\pdfoutput=1
\RequirePackage[T1]{fontenc}
\documentclass[12pt]{article}
\usepackage[height=8.85in,width=6.45in]{geometry}

\usepackage[utf8]{inputenc}
\usepackage{amsmath}
\usepackage{amssymb}
\usepackage{mathtools}
\numberwithin{equation}{section}
\usepackage{slashed}
\usepackage{braket}
\usepackage{enumitem}
\usepackage{dsfont}
\usepackage{bbm}
\usepackage[svgnames]{xcolor}
\usepackage[colorlinks,citecolor=DarkGreen,linkcolor=FireBrick,urlcolor=FireBrick,linktocpage,unicode]{hyperref}
\usepackage{tikz}
\usetikzlibrary{positioning, shapes.geometric, calc, arrows.meta}
\usepackage{pgfplots}

\usepackage{tocloft}

\usepackage[bottom]{footmisc}
\allowdisplaybreaks
\usepackage{graphicx}
\usepackage{tikz}
\usepackage{tikz-cd}
\usepackage{times}
\usepackage{courier}
\usepackage{bm}
\usepackage{physics}
\usepackage{xcolor}
\usepackage{natbib}
\usepackage{mdframed}
\usepackage{nicefrac}
\usepackage{booktabs}
\usepackage{lipsum}
\usepackage{titlesec}
\usepackage{wrapfig,lipsum,booktabs}
\usepackage{blindtext}
\usepackage{algorithm}
\usepackage{algorithmicx}
\usepackage{algpseudocode}
\usepackage{pifont}
\usepackage{fontawesome}
\usepackage{titletoc}
\usepackage{todonotes}
\usepackage{titletoc}
\usepackage{setspace}
\usepackage{dsfont} %
\newcommand{\one}{\mathds{1}}

\usepackage{tikz}
\usetikzlibrary{positioning, shapes.geometric, calc, arrows.meta}
\usepackage{pgfplots}
\pgfplotsset{compat=1.18}

\usepackage{dsfont}
\usepackage{bbm}

\usepackage{verbatim}

\usepackage[nameinlink,capitalize,noabbrev]{cleveref}

\usepackage{array}
\usepackage{makecell}

\usepackage{slashed}

\def\({\left(}
\def\){\right)}
\def\[{\left[}
\def\]{\right]}

\usepackage{dashbox}
\usepackage{xcolor}
\usepackage{colortbl}
\definecolor{lightyellow}{rgb}{1.0, 0.95, 0.7}
\definecolor{Blue}{rgb}{0, 0, 0.8}
\definecolor{blue}{rgb}{0,0,1}
\definecolor{darkgreen}{rgb}{0,0.40,0}
\definecolor{firebrick}{rgb}{0.698,0.133,0.133}
\definecolor{mypurple}{rgb}{0.5,0,0.5}

\newcommand*{\blue}[1]{\textcolor{blue}{#1}}

\definecolor{lesslightgray}{rgb}{0.5,0.5,0.5}
\definecolor{light-gray}{gray}{0.95}

\let\tilde\widetilde
\let\hat\widehat

\newcommand{\calA}{\mathcal{A}}

\newcommand{\calC}{\mathcal{C}}
\newcommand{\calD}{\mathcal{D}}

\newcommand{\calF}{\mathcal{F}}

\newcommand{\calH}{\mathcal{H}}

\newcommand{\calL}{\mathcal{L}}

\newcommand{\calN}{\mathcal{N}}
\newcommand{\calO}{\mathcal{O}}
\newcommand{\calP}{\mathcal{P}}

\newcommand{\calR}{\mathcal{R}}
\newcommand{\calS}{\mathcal{S}}
\newcommand{\calT}{\mathcal{T}}

\newcommand{\calV}{\mathcal{V}}

\newcommand{\EE}{\mathbb{E}}
\newcommand{\RR}{\mathbb{R}}

\newcommand{\poly}{\mathrm{poly}}

\newcommand{\Softmax}{\mathop{\rm{Softmax}}}

\def\R{\mathbb{R}}

\def\E{\mathbb{E}}

\let\cite\citep

\newcommand{\cov}{\mathop{\rm{Cov}}}

\def\tilded{\tilde{d}}
\def\tildeL{\tilde{L}}
\def\tildeR{\tilde{R}}

\def\barh{\bar{h}}
\def\holder{H\"{o}lder }
\def\lip{Lipschitz }
\def\relu{$\mathrm{ReLU}$ }

\DeclareDocumentCommand{\al}{o m}{\IfNoValueTF{#1}{=}{#1} & ~ #2}

\usepackage{amsthm}
\makeatletter
\def\th@remark{%
  \thm@headfont{\bfseries}%
  \normalfont %
  \thm@preskip\topsep \divide\thm@preskip\tw@
  \thm@postskip\thm@preskip
}
\makeatother
\usepackage[many]{tcolorbox}
\theoremstyle{definition}
\newtheorem{theorem}{Theorem}[section]
\tcolorboxenvironment{theorem}{
  breakable,
  colback=black!10,
  colframe=white,%
  width=\dimexpr\linewidth+10pt\relax,%
  enlarge left by=-5pt,%
  enlarge right by=-5pt,%
  boxsep=5pt,%
  boxrule=0pt,
  left=0pt,right=0pt,top=0pt,bottom=0pt,
  sharp corners,
  before skip=\topsep,
  after skip=\topsep
}
\newtheorem{lemma}{Lemma}[section]
\tcolorboxenvironment{lemma}{
  breakable,
  colback=black!10,
  colframe=white,%
  width=\dimexpr\linewidth+10pt\relax,%
  enlarge left by=-5pt,%
  enlarge right by=-5pt,%
  boxsep=5pt,%
  boxrule=0pt,
  left=0pt,right=0pt,top=0pt,bottom=0pt,
  sharp corners,
  before skip=\topsep,
  after skip=\topsep
}
\newtheorem{corollary}{Corollary}[theorem]
\tcolorboxenvironment{corollary}{
  breakable,
  colback=black!10,
  colframe=white,%
  width=\dimexpr\linewidth+10pt\relax,%
  enlarge left by=-5pt,%
  enlarge right by=-5pt,%
  boxsep=5pt,%
  boxrule=0pt,
  left=0pt,right=0pt,top=0pt,bottom=0pt,
  sharp corners,
  before skip=\topsep,
  after skip=\topsep
}

\tcolorboxenvironment{proposition}{
  breakable,
  colback=black!10,
  colframe=white,%
  width=\dimexpr\linewidth+10pt\relax,%
  enlarge left by=-5pt,%
  enlarge right by=-5pt,%
  boxsep=5pt,%
  boxrule=0pt,
  left=0pt,right=0pt,top=0pt,bottom=0pt,
  sharp corners,
  before skip=\topsep,
  after skip=\topsep
}
\theoremstyle{definition}
\newtheorem{definition}{Definition}[section]
\tcolorboxenvironment{definition}{
  breakable,
  colback=black!10,
  colframe=white,%
  width=\dimexpr\linewidth+10pt\relax,%
  enlarge left by=-5pt,%
  enlarge right by=-5pt,%
  boxsep=5pt,%
  boxrule=0pt,
  left=0pt,right=0pt,top=0pt,bottom=0pt,
  sharp corners,
  before skip=\topsep,
  after skip=\topsep
}
\newtheorem{remark}{Remark}[section]
\newtheorem{assumption}{Assumption}[section]
\tcolorboxenvironment{assumption}{
  breakable,
  colback=black!10,
  colframe=white,%
  width=\dimexpr\linewidth+10pt\relax,%
  enlarge left by=-5pt,%
  enlarge right by=-5pt,%
  boxsep=5pt,%
  boxrule=0pt,
  left=0pt,right=0pt,top=0pt,bottom=0pt,
  sharp corners,
  before skip=\topsep,
  after skip=\topsep
}

\tcolorboxenvironment{problem}{
  breakable,
  colback=black!10,
  colframe=white,%
  width=\dimexpr\linewidth+10pt\relax,%
  enlarge left by=-5pt,%
  enlarge right by=-5pt,%
  boxsep=5pt,%
  boxrule=0pt,
  left=0pt,right=0pt,top=0pt,bottom=0pt,
  sharp corners,
  before skip=\topsep,
  after skip=\topsep
}

\tcolorboxenvironment{question}{
  breakable,
  colback=black!10,
  colframe=white,%
  width=\dimexpr\linewidth+10pt\relax,%
  enlarge left by=0pt,%
  enlarge right by=0pt,%
  boxsep=5pt,%
  boxrule=0pt,
  left=0pt,right=0pt,top=0pt,bottom=0pt,
  sharp corners,
  before skip=\topsep,
  after skip=\topsep
}
\crefname{question}{Question}{Questions}

\crefname{theorem}{Theorem}{Theorems}
\crefname{proposition}{Proposition}{Propositions}
\crefname{lemma}{Lemma}{Lemmas}
\crefname{corollary}{Corollary}{Corollaries}
\crefname{definition}{Definition}{Definitions}
\crefname{assumption}{Assumption}{Assumptions}
\crefname{remark}{Remark}{Remarks}
\crefname{problem}{Problem}{Problems}
\crefname{property}{Property}{property}

\numberwithin{equation}{section}
\numberwithin{theorem}{section}
\numberwithin{proposition}{section}
\numberwithin{definition}{section}
\numberwithin{lemma}{section}
\numberwithin{assumption}{section}
\numberwithin{remark}{section}

\newcommand*{\annot}[1]{\tag*{\footnotesize{\textcolor{black!50}{\big(#1\big)}}}}

\makeatletter
\let\save@mathaccent\mathaccent
\newcommand*\if@single[3]{%
    \setbox0\hbox{${\mathaccent"0362{#1}}^H$}%
    \setbox2\hbox{${\mathaccent"0362{\kern0pt#1}}^H$}%
    \ifdim\ht0=\ht2 #3\else #2\fi
}
\newcommand*\rel@kern[1]{\kern#1\dimexpr\macc@kerna}
\newcommand*\widebar[1]{\@ifnextchar^{{\wide@bar{#1}{0}}}{\wide@bar{#1}{1}}}
\newcommand*\wide@bar[2]{\if@single{#1}{\wide@bar@{#1}{#2}{1}}{\wide@bar@{#1}{#2}{2}}}
\newcommand*\wide@bar@[3]{%
    \begingroup
    \def\mathaccent##1##2{%
        \let\mathaccent\save@mathaccent
        \if#32 \let\macc@nucleus\first@char \fi
        \setbox\z@\hbox{$\macc@style{\macc@nucleus}_{}$}%
        \setbox\tw@\hbox{$\macc@style{\macc@nucleus}{}_{}$}%
        \dimen@\wd\tw@
        \advance\dimen@-\wd\z@
        \divide\dimen@ 3
        \@tempdima\wd\tw@
        \advance\@tempdima-\scriptspace
        \divide\@tempdima 10
        \advance\dimen@-\@tempdima
        \ifdim\dimen@>\z@ \dimen@0pt\fi
        \rel@kern{0.6}\kern-\dimen@
        \if#31
        \overline{\rel@kern{-0.6}\kern\dimen@\macc@nucleus\rel@kern{0.4}\kern\dimen@}%
        \advance\dimen@0.4\dimexpr\macc@kerna
        \let\final@kern#2%
        \ifdim\dimen@<\z@ \let\final@kern1\fi
        \if\final@kern1 \kern-\dimen@\fi
        \else
        \overline{\rel@kern{-0.6}\kern\dimen@#1}%
        \fi
    }%
    \macc@depth\@ne
    \let\math@bgroup\@empty \let\math@egroup\macc@set@skewchar
    \mathsurround\z@ \frozen@everymath{\mathgroup\macc@group\relax}%
    \macc@set@skewchar\relax
    \let\mathaccentV\macc@nested@a
    \if#31
    \macc@nested@a\relax111{#1}%
    \else
    \def\gobble@till@marker##1\endmarker{}%
    \futurelet\first@char\gobble@till@marker#1\endmarker
    \ifcat\noexpand\first@char A\else
    \def\first@char{}%
    \fi
    \macc@nested@a\relax111{\first@char}%
    \fi
    \endgroup
    }
\makeatother
\let\bar\widebar

\makeatletter
\newcommand*{\redefinesymbolwitharg}[1]{%
  \expandafter\let\csname ltx#1\expandafter\endcsname\csname #1\endcsname
  \@namedef{#1}{\@ifnextchar{^}{\@nameuse{#1@}}{\@nameuse{#1@}^{}}}%
  \expandafter\def\csname #1@\endcsname^##1##2{%
     \csname ltx#1\endcsname\ifx!##1!\else^{##1}\fi\mathopen{}\mathclose\bgroup\left(##2\aftergroup\egroup\right)
     }%
}
\makeatother
\redefinesymbolwitharg{sin}
\redefinesymbolwitharg{cos}

\usepackage{multirow}

\newcommand*{\email}[1]{\footnote{\href{mailto:#1}{\texttt{#1}}}}

\titlespacing{\section}{0pt}{*1}{*-1}
\titlespacing{\subsection}{0pt}{*1}{*-1}
\titlespacing{\subsubsection}{0pt}{*1}{*-1}
\titlespacing{\paragraph}{0pt}{*1}{1em} %

\setlist[itemize]{leftmargin=1em,
topsep=0em, partopsep=0em, itemsep=0.08em}

\setlist[enumerate]{leftmargin=1.4em,
topsep=0em, partopsep=0em, itemsep=0.08em}

\definecolor{blue}{named}{black}

\begin{document}

\begin{titlepage}

\begin{flushright}
Last Update: \today
\end{flushright}

\begin{center}

{
\Large \bfseries %
\begin{spacing}{1.15} %
On Statistical Rates of Conditional Diffusion Transformers: Approximation, Estimation and Minimax Optimality

\end{spacing}
}

Jerry Yao-Chieh Hu$^{\dagger\ddag*}$\email{jhu@u.northwestern.edu}\quad
Weimin Wu$^{\dagger\ddag*}$\email{wwm@u.northwestern.edu}\quad
Yi-Chen Lee$^{\natural*}$\email{b10202055@ntu.edu.tw}\quad
Yu-Chao Huang$^{\sharp\natural*}$\email{yuchaohuang@g.ntu.edu.tw}\\
Minshuo Chen$^{\flat}$\email{minshuo.chen@northwestern.edu}\quad
Han Liu$^{\dagger\ddag\S}$\email{hanliu@northwestern.edu}

\def\thefootnote{*}
\footnotetext{These authors contributed equally to this work.}

{\footnotesize
\begin{tabular}{ll}
 $^\dagger\;$Center for Foundation Models and Generative AI, Northwestern University, Evanston, IL 60208, USA\\
 $^\ddag\;$Department of Computer Science, Northwestern University, Evanston, IL 60208 USA\\
 $^\sharp\;$Department of Physics, National Taiwan University, Taipei 106319, Taiwan\\
 $^\natural\;$Physics Division, National Center for Theoretical Sciences, Taipei 106319, Taiwan\\
 $^\flat\;$Department of Industrial Engineering \& Management Sciences, Northwestern University, Evanston, IL 60208 USA\\
 $^\S\;$Department of Statistics and Data Science, Northwestern University, Evanston, IL 60208 USA
\end{tabular}}

\end{center}

\noindent

We investigate the approximation and estimation rates of conditional diffusion transformers (DiTs) with classifier-free guidance.
We present a comprehensive analysis for ``in-context'' conditional DiTs under four common data assumptions.
We show that both conditional DiTs and their latent variants lead to the minimax optimality of unconditional DiTs under identified settings.
Specifically, we discretize the input domains into infinitesimal grids and then perform a term-by-term Taylor expansion on the conditional diffusion score function under \holder smooth data assumption.
This enables fine-grained use of transformers' universal approximation through a more detailed piecewise constant approximation and hence obtains tighter bounds.
Additionally, we extend our analysis to the latent setting under the linear latent subspace assumption.
We not only show that latent conditional DiTs achieve lower bounds than conditional DiTs both in approximation and estimation, but also show the minimax optimality of latent unconditional DiTs.
Our findings establish statistical limits for conditional and unconditional DiTs, and offer
practical guidance toward developing more efficient and accurate DiT models.

\end{titlepage}

{
\setlength{\parskip}{0em}
\setcounter{tocdepth}{2}
\tableofcontents
}
\setcounter{footnote}{0}

\section{Introduction}
\label{sec:intro}
We investigate the approximation and estimation rates of conditional diffusion transformers (DiTs) with classifier-free guidance.
Specifically, we derive score approximation, score estimation, and distribution estimation guarantees for both conditional DiTs and their latent variants. 
We provide a comprehensive analysis under various data conditions.
Moreover, we show that both conditional DiTs and their latent variants lead to the minimax optimality of unconditional DiTs under identified settings.
This analysis is not only practical but also timely.
Transformer-based conditional diffusion models are at the forefront of generative AI due to their success as scalable and flexible backbones for image \citep{wu2024med, bao2023all, batzolis2021conditional} and video generation \citep{liu2024sora, ni2023conditional, saharia2022photorealistic, voleti2022mcvd}. 
However, the theoretical understanding of conditional DiTs remains limited. 
\blue{
On the one hand, while prior work by \citet{hu2024statistical} reports approximation and estimation rates of DiTs using the established universality of transformers \cite{yun2019transformers}, 
their results are not tight and are limited to unconditional diffusion.
}
On the other hand, existing theoretical works on conditional diffusion models only focus on \relu networks \cite{fu2024diffusion,yuan2023reward}, model-free settings 
\cite{ye2024tfg,guo2024gradient} or  generative sampling process \cite{dinh2023rethinking}, without considering the transformer architectures.
This work addresses this gap by providing a timely analysis of the statistical limits of conditional DiTs.

In this work, 
we present a comprehensive analysis of conditional DiT and its latent setting under four common data assumptions.
We also establish the minimax optimality of unconditional DiT and its latent version by deriving the tight distribution estimation error bounds.
Our techniques include two key parts:
(i) Discretizing the input domains into infinitesimal grids.
(ii) On each grid, performing a term-by-term Taylor expansion on the conditional diffusion score function under generic and stronger H\"{o}lder smooth data assumptions, motivated by the local diffused polynomial analysis \cite{fu2024diffusion,oko2023diffusion}.
These techniques leverage the nice regularity of the score function imposed by the H\"{o}lder smoothness data assumptions and hence enable fine-grained use of transformers' universal approximation \cite{kajitsuka2023transformers, yun2019transformers} through a more detailed piecewise constant approximation.
Consequently, we obtain tighter bounds.

\begin{table}[!ht]
\centering
\caption{\small \textbf{Summary of Theoretical Results.}  
\blue{The initial data is $d_x$-dimensional, and the condition is $d_y$-dimensional. For latent DiT, the latent variable is $d_0$-dimensional. 
$\sigma_{t}^{2} = 1 - e^{-t}$ is the denoising scheduler. 
The sample size is $n$, and $0 < \epsilon < 1$ represents the score approximation error. 
While we report asymptotics for large $d_x,d_0$, we reintroduce the $n$ dependence in the estimation results to emphasize sample complexity convergence.}
}

\renewcommand{\arraystretch}{0.7}
\setlength\extrarowheight{0pt}
\resizebox{\textwidth}{!}{%
\begin{tabular}{m{4.2cm}|m{3cm}|m{3.7cm}|m{3.7cm}|m{1.5cm}}

\toprule
\makecell{Assumption} & \makecell{Score \\Approximation} & \makecell{Score \\Estimation} & \makecell{Dist. Estimation \\(Total Variation)} & \makecell{Minimax \\Optimality} \\

\midrule 

\multirow{1}{*}
\makecell{Generic H\"{o}lder Smooth Data Dist. (\cref{sec:generic_holder,sec:score_est_dist_est})} & \makecell{\blue{$\calO((\log(\frac{1}{\epsilon}))^{d_x}/\sigma_t^4)$}} & \makecell{\blue{$ n^{-o(1/d_x)}\cdot(\log n)^{\calO(d_x)}$}} & \makecell{\blue{$n^{-o(1/d_x)}\cdot(\log n)^{\calO(d_x)}$}} & \makecell{\faTimes} \\ 

\midrule 

\multirow{1}{*}
\makecell{Stronger H\"{o}lder Smooth Data Dist. (\cref{sec:bounded_holder,sec:score_est_dist_est})} & \makecell{ \blue{$(\log (\frac{1}{\epsilon}))^{\calO(1)}/\sigma_t^2$}} & \makecell{\blue{$n^{-o(1/d_x^2)}\cdot(\log n)^{\calO(1)}$}} & \makecell{\blue{$n^{-o(1/d_x)}\cdot(\log n)^{\calO(1)}$}} & \makecell{\faCheck} \\

\midrule 

\multirow{1}{*}
\makecell{Latent Subspace + Generic H\"{o}lder Smooth Data Dist. (\cref{sec:latent_con_dit})}  & \makecell{\blue{$\calO((\log(\frac{1}{\epsilon}))^{d_0}/\sigma_t^4)$}} & \makecell{$\blue{n^{-o(1/d_0)}\cdot(\log n)^{\calO(d_0)}}$} & \makecell{\blue{$n^{-o(1/d_0)}\cdot(\log n)^{\calO(d_0)}$}} &  \makecell{\faTimes} \\ 

\midrule 

\multirow{1}{*}
\makecell{Latent Subspace + Stronger H\"{o}lder Smooth Data Dist. (\cref{sec:latent_con_dit})}  & \makecell{\blue{$(\log (\frac{1}{\epsilon}))^{\calO(1)}/\sigma_t^2$}} & \makecell{\blue{$n^{-o(1/d_0^2)}\cdot(\log n)^{\calO(1)}$}} & \makecell{\blue{$n^{-o(1/d_0)}\cdot(\log n)^{\calO(1)}$}} &  \makecell{\faCheck} \\

\bottomrule

\end{tabular}
}
\hfill

\label{table:summary_results}
\end{table}

\textbf{Contributions.}
We summarize the theoretical results in \cref{table:summary_results}.
Our contributions are threefold:
\begin{itemize}
    \item \textbf{Score Approximation.}
    We characterize the approximation limit of matching the conditional DiT score function with a transformer-based score estimator.
    The approximation results explain the expressiveness of conditional DiT and its latent version, and guide the score network's structural configuration for practical implementations (\cref{thm:Main_1,thm:Main_2_informal,theorem:latent_main1_infor}).
    The results also show that the latent version achieves a better approximation for the score function.  
    
    \item \textbf{Score and Distribution Estimation.}
    We study the score and distribution estimation of conditional DiTs in practical training scenarios. 
    Specifically, we provide a sample complexity bound for score estimation (\cref{thm:main_risk_bounds,thm:latent_holder_est}), using norm-based covering number bound of transformer architecture. 
    Additionally, we show that the learned score estimator can recover the initial data distribution in both conditional DiT and its latent setting (\cref{thm:distribution_TV_bound,thm:lantent_distribution_TV_bound_main}).

    \item \textbf{Minimax Optimal Estimator.}
    We extend our analysis to unconditional DiT and investigate whether the generated data distribution achieves the minimax optimality in the total variation distance. 
    Specifically, we show that the upper bounds on the distribution estimation error match the lower bounds under stronger H\"{o}lder smooth data distribution (\cref{cor:holder_optimal} and \cref{remark:latent_minimax}).
    
\end{itemize}

\textbf{Organization.}
\cref{sec:background} presents preliminaries and the problem setup.
\cref{sec:con_dit} presents the results of conditional DiTs.
\cref{sec:latent_con_dit} presents the results of latent conditional DiTs.
\cref{sec:related} presents related works' discussions.
The appendix contains an extended and improved version of \cite{hu2024statistical} on conditional DiTs (\cref{sec:appendix_latent_dit_lipschitz}), additional results, and detailed proofs. 

\textbf{Notations.} 
The index set $\{ 1, ..., I \}$ is denoted by $[ I ]$, where $I \in \mathbb{N}^+$. 
We denote (column) vectors by lower case letters, and matrices by upper case letters.
Let $a[i]$ denote the $i$-th component of vector $a$.
Let $A_{ij}$ denotes the $(i,j)$-th entry of matrix $A$.
$\norm{x}$, $\norm{x}_{1}$ and $\norm{x}_{\infty}$ denote the Euclidean norm, 1-norm, and infinite norm.
$\norm{W}_{2}$ and $\norm{W}_{F}$ denote the spectral norm and Frobenius norm, and $\norm{W}_{p,q}$ denotes the $(p,q)$-norm where $p$-norm is over
columns and $q$-norm is over rows.

\section{Background and Preliminaries}
\label{sec:background}
In this section, we provide a high-level overview of the conditional diffusion model with classifier-free guidance in \cref{sec:CDM_intro} and conditional Diffusion Transformer (DiT) networks in \cref{sec:con_DiT_intro}.

\subsection{Conditional Diffusion Model with Classifier-free Guidance}
\label{sec:CDM_intro}

\textbf{Forward and Backward Conditional Diffusion Process.}
In the \textit{forward} process, conditional diffusion models gradually add noise to the original data $x_0\in \mathbb{R}^{d_x}$.
Give a condition $y \in \R^{d_y}$, and $x_0 \sim P_{0}(\cdot | y)$.
Let $x_t$ denote the noisy data at the timestamp $t$, with marginal distribution and density as $P_t(\cdot | y)$ and $p_t(\cdot | y)$.
The conditional distribution $P_t(x_t| y)$ follows $N(\alpha_t x_0, \sigma_t^{\textcolor{blue}{2}} I_{d_x})$, where $\alpha_{t}=e^{-t/2}$, $\sigma_{t}^{2}=1-e^{-t}$, and $w(t)>0$ is a nondecreasing weighting function. 
In practice, the forward process terminates at a large enough $T$ such that $P_T$ is close to $N(0,I_{d_x})$.
In the \textit{backward} process, we obtain $x_t^{\leftarrow}$ by reversing the forward process.
The generation of $x_t^{\leftarrow}$ depends on the score function $\nabla \log p_t(\cdot | y)$.
See \cref{app:back_con_diff} for the details.
In below, when the context is clear, we suppress the notation dependence of $x_t$ on the time step $t$.

\textbf{Classifier-Free Guidance.}
Classifier-free guidance \cite{ho2022classifier} is the standard workhorse for training condition diffusion models.
It approximates both conditional and unconditional score functions using neural networks $s_W$ with parameters $W$.
It uses the following loss function:
\begin{align*}
    \ell(x_0, y ; s_W)=\int_{t_0}^T \frac{1}{T-t_0} \EE_{\tau, x_t \sim N(\alpha_t x_0, \sigma_t^2 I_{d_x})} \left[\left\|s_W (x_t, \tau y, t)-\nabla_{x_t} \log \phi_t\left(x_t | x_0 \right)\right\|_2^2\right] \dd t,
\end{align*}
where $\nabla_{x_t} \log \phi_t\left(x_t | x_0 \right) = -(x_t-\alpha_t x_0)/\sigma_t^2$, 
$t_0$ is a small cutoff to stabilize training \footnote{$t_0$ is the early stopping time to prevent the score function from blowing up \cite{fu2024diffusion, chen2023score, dhariwal2021diffusion, song2021score}.}.
$\tau = \emptyset$ denotes the unconditional version, $\tau = \mathrm{id}$ denotes the conditional version, and $P(\tau=\emptyset)=P(\tau=\mathrm{id})=0.5$.
To train $s_W$, we select $n$ i.i.d. samples $\{x_{0,i}, y_i \}_{i=1}^{n}$, where $x_{0,i} \sim P_0(\cdot | y_i)$.
We use 
\begin{align}
\label{eqn:empirical_loss}
    \hat{\calL}(s_W)  \coloneqq \frac{1}{n} \sum_{i=1}^n \ell(x_{0,i}, y_i ; s_{W}),
\end{align}
as \blue{the} empirical loss.
In addition, we denote population loss as ${\calL}(s_W)$.
See \cref{app:class_free_guide} for details.

\subsection{Conditional Diffusion Transformer Networks}\label{sec:con_DiT_intro}

\begin{figure}[t!]
\centering
\definecolor{encodercolor}{RGB}{255,99,71}
\definecolor{decodercolor}{RGB}{135,206,250}
\definecolor{networkcolor}{RGB}{211,211,211}
\definecolor{sumcolor}{RGB}{139,138,123}
\definecolor{greyblockcolor}{RGB}{169,169,169}
\definecolor{reshapecolor}{RGB}{169,169,169}
\definecolor{attncolor}{rgb}{0.698,0.133,0.133}
\definecolor{ffcolor}{rgb}{0,0.40,0}
\definecolor{normcolor}{RGB}{173,216,230}
\definecolor{concatcolor}{RGB}{169,169,169}

\resizebox{\textwidth}{!}{
\begin{tikzpicture}[
    encoder/.style={trapezium, trapezium angle=60, draw=encodercolor, fill=encodercolor!50, thick, minimum height=1.8cm, minimum width=2cm, align=center, rotate=270},
    decoder/.style={trapezium, trapezium angle=60, draw=decodercolor, fill=decodercolor!50, thick, minimum height=1.8cm, minimum width=2cm, align=center, rotate=90},
    network/.style={rectangle, draw=attncolor!70, fill=attncolor!30, thick, minimum height=2cm, minimum width=3cm, align=center},
    greyblock/.style={rectangle, draw=greyblockcolor, fill=greyblockcolor!50, thick, minimum height=2cm, minimum width=1cm, align=center},
    yellowblock/.style={rectangle, draw=reshapecolor, fill=reshapecolor!50, thick, minimum height=1.5cm, minimum width=1.2cm, align=center},
    concatblock/.style={rectangle, draw=reshapecolor, fill=reshapecolor!50, thick, minimum height=2.0cm, minimum width=1.2cm, align=center},
    ffblock/.style={rectangle, draw=ffcolor!70, fill=ffcolor!30, thick, minimum height=1.5cm, minimum width=1cm, align=center},
    attnblock/.style={rectangle, draw=attncolor!70, fill=attncolor!30, thick, minimum height=1.5cm, minimum width=1cm, align=center},
    sum/.style={circle, draw=sumcolor, fill=sumcolor!50, thick, minimum size=0.4cm},
    shortcut/.style={dashed,  -{Latex[scale=1.5]}},
    myarrow/.style={thick, ->, -{Latex[scale=1.5]}},
    node distance=0.8cm, auto, scale=0.75, transform shape,
    decorate,decoration={brace,amplitude=10pt,raise=3pt},
    smallrect/.style={rectangle, draw=black, fill=gray!30, minimum width=0.01cm, minimum height=0.01cm},
    highlightrect/.style={rectangle, draw=black, fill=ffcolor!30, minimum width=0.01cm, minimum height=0.01cm},
]

\node (input) at (-2.3,1) {};
\node (y) at (-2.3,-1) {};

\node[yellowblock] at (0.5,1) (reshape) {{$ R(\cdot) $}};
\node at (reshape.north) [above, yshift=-0.0cm] {\shortstack{Reshape Layer}};

\node[yellowblock] at (0.5,-1) (embed) {{Embed}};

\node[concatblock] at (3.0,0) (concat) {{Concat}};

\node[network] at (7.3,0) (network) {$f_{\calT} \in \calT^{\textcolor{blue}{h,s,r}}$};
\node at (network.south) [below, yshift=-0.0cm] {\shortstack{Transformer Network}};

\node[yellowblock] at (13.7,0) (reshapei) {{$ R^{-1}(\cdot) $}};
\node at (reshapei.north) [above, yshift=-0.0cm] {\shortstack{Reversed\\ Reshape Layer}};

\node at (16.4,0) (output) {};

\draw[myarrow] (input) -- (reshape) node[midway, below, yshift=-0.2cm] {$x \in \R^{d_x}$};
\draw[myarrow] (y) -- (embed) node[midway, below, yshift=-0.2cm] {\parbox{1.8cm}{Label $y$ \\Timestep $t$}};

\draw[myarrow] (reshape) -- (concat) node[midway, above, xshift=0.0cm, yshift=0.2cm] {$ \R^{d \times L} $};
\draw[myarrow] (embed) -- (concat) node[midway, below, xshift=0.0cm, yshift=-0.2cm] {$ \R^{d \times 2} $};

\draw[myarrow] (concat) -- (network) node[midway, below, yshift=-0.2cm] {$ \R^{d \times (L+2)} $};

\draw[myarrow] (reshapei) -- (output) node[midway, below, yshift=-0.2cm] {$ \R^{d_x} $};

\draw[myarrow] (network) -- ++(3.4,0) node[midway, below, yshift=-0.2cm] {$ \R^{d \times (L+2)} $};

\node[highlightrect] at (11.0,0.7) (token1) {};
\node[highlightrect] at (11.0,0.525) (token2) {};
\node[highlightrect] at (11.0,0.35) (token3) {};
\node[highlightrect] at (11.0,0.175) (token4) {};
\node[highlightrect] at (11.0,0) (token5) {};
\node[highlightrect] at (11.0,-0.175) (token6) {};
\node[highlightrect] at (11.0,-0.35) (token7) {};
\node[smallrect] at (11.0,-0.525) (cls_token) {}; 
\node[smallrect] at (11.0,-0.7) (cls_token) {}; 

\draw[myarrow] (11.30,0) -- (reshapei) node[midway, below,yshift=-0.3cm] {$\R^{d \times L}$};

\draw[dashed, thick] (10.78, 0.88) rectangle (11.22, -0.47);

\end{tikzpicture}
}
\caption{\small
\textbf{Conditional DiT Network Architecture.} 
The architecture consists of a reshape layer $R(\cdot)$, a reversed reshape layer $R^{-1}(\cdot)$, and the embedding layers for label $y$ and timestep $t$. 
The embeddings of $y$ and $t$ are concatenated with input sequences and then processed by a transformer network $ f_{\calT} \in \calT^{\textcolor{blue}{h,s,r}}$.
}
\label{fig:condition_DiT}
\end{figure}
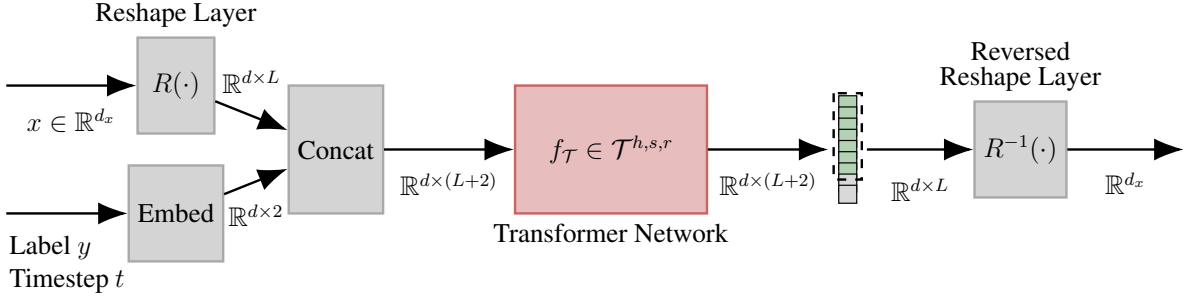

We use a transformer network as a score estimator $s_W$.
Our notation follows \cite{hu2024statistical}.

\textbf{Transformer Block.}
Let $f^{(\mathrm{SA})}:\R^{d\times L}\to \R^{d\times L}$ denote the self-attention layer.
Let $h$ and $s$ denote the number of heads and hidden dimension in the self-attention layer, and then we have
\begin{align}
\label{eq:self_attn}
    f^{(\text{SA})} \left( Z \right) 
    \coloneqq 
    Z +
    \sum_{i=1}^{h} W_O^i
    (W_V^i Z) 
    \Softmax
    \big[ (W_K^i Z)^{\top}(W_Q^i Z )\big],
\end{align} 
where $W_V^{i}, W_K^i, W_Q^i \in \mathbb{R}^{s \times d}$, and $W_{O}^i\in \mathbb{R}^{d \times s}$ are the weight matrices. 
Next, we define the feed-forward layer with MLP dimension $r$:
\begin{equation} 
\label{eq:ff_layer}
    f^{(\text{FF})}(Z) 
    \coloneqq
    Z +
     W_2 {\rm ReLU}(W_1 Z
     +
     b_1)+b_2,
\end{equation}
where $W^{(1)} \in \mathbb{R}^{r \times d}$ and $W^{(2)} \in \mathbb{R}^{d \times r}$ are weight matrices, and $b^{(1)} \in \mathbb{R}^{r}$, and $b^{(2)} \in \mathbb{R}^{d}$ are bias.

\begin{definition}[Transformer Block]\label{def:transformer_block}
We define a transformer block of $h$-head, $s$-hidden dimension, $r$-MLP dimension, and with positional encoding $E \in \mathbb{R}^{d\times L}$ as
\begin{align*}
    f^{\textcolor{blue}{h,s,r}} \left( Z \right)
    \coloneqq 
    f^{(\text{FF})} \left( f^{(\text{SA})} \left( Z + E \right) \right) : \mathbb{R}^{d\times L} \mapsto \mathbb{R}^{d\times L}.
\end{align*}
\end{definition}

Now, we define the transformer networks as compositions of transformer blocks. 
\begin{definition}[Transformer Network Function Class]\label{def:transformer_class}
    Let $\mathcal{T}^{\textcolor{blue}{h,s,r}}$ denote the transformer network function class where each function $\tau \in \mathcal{T}^{\textcolor{blue}{h,s,r}}$ is a composition of transformer blocks $f^{\textcolor{blue}{h,s,r}}$, i.e., 
\begin{align*}
    \mathcal{T}^{\textcolor{blue}{h,s,r}} 
    \coloneqq 
    \{ 
    \tau: \mathbb{R}^{d\times L} \mapsto \mathbb{R}^{d\times L} \mid 
    \tau = 
    f^{\textcolor{blue}{h,s,r}} \circ \cdots \circ f^{\textcolor{blue}{h,s,r}} \}.
\end{align*}
\end{definition}

\textbf{Conditional Diffusion Transformer (DiT).}
Let $f \in \mathcal{T}^{\textcolor{blue}{h,s,r}}$ be a transformer network, and $(x,y,t) \in \R^{d_x} \times \R^{d_y} \times [t_0, T]$ be the input data.
We follow the ``in-context conditioning''  conditional DiT network in \cite{peebles2023scalable} as  in \cref{fig:condition_DiT}. 
The following reshape layer converts a vector input $x\in\R^{d_x}$ into the sequential matrix input format $Z\in\R^{d\times L}$ for transformer with $d_x=d\cdot L$.
\begin{definition}[DiT Reshape Layer $R(\cdot)$]
\label{def:reshape_layer}
     Let $R(\cdot): \mathbb{R}^{d_x} \to \mathbb{R}^{d \times L}$ be a reshape layer that transforms the $d_x$-dimensional input into a $d \times L$ matrix. 
     Specifically, for any $d_x=i\times i$ image input, $R(\cdot)$ converts it into a sequence representation with feature dimension $d \coloneqq p^2$ (where $p \geq 2$) and sequence length $L \coloneqq \left(i/p\right)^2$.
     Besides, we define the corresponding reverse reshape (flatten) layer $R^{-1} (\cdot): \mathbb{R}^{d \times L} \to \mathbb{R}^{d_x}$ as the inverse of $R(\cdot)$.
     By $d_x= dL$, $R,R^{-1}$ are associative w.r.t. their input. 
\end{definition}

We define the following transformer network function class with the reshape layer.

\begin{definition}[Transformer Network Function Class with Reshape Layer $\calT_R^{\textcolor{blue}{h,s,r}}$]
\label{def:Tran_reshape_class}
    \begin{align*}
        \calT_R^{\textcolor{blue}{h,s,r}}  & (C_{\calT}, C_{Q}^{2,\infty}, C_{Q}, 
        C_{K}^{2,\infty}, C_{K},
        C_{V}^{2,\infty}, C_{V},
        C_{O}^{2,\infty}, C_{O},
        C_E, 
        C_{f_1}^{2,\infty}, C_{f_1}, 
        C_{f_2}^{2,\infty}, C_{f_2}, 
        L_{\calT})~
        \text{satisfies}
    \end{align*}

    \begin{itemize}[after=\vspace{-.2em}]
        \item $\calT_R^{\textcolor{blue}{h,s,r}} 
        \coloneqq 
        \{ 
        R^{-1}\circ f_{\calT} \circ R:  \R^{d_x} \mapsto \mathbb{R}^{d_x} \mid  f_{\calT} \in \calT^{\textcolor{blue}{h,s,r}} 
        \}$;
        \item Model output bound: $\sup_{Z}\norm{f_{\calT}(Z)}_2\leq C_{\calT}$;
        \item Parameter bound in ${f^{({\rm SA})}}$:  $\norm{(W_{Q})^\top}_{2,\infty}\leq C_{Q}^{2,\infty}$, $\norm{(W_{Q})^\top}_2 \leq C_{Q}$, $\norm{W_{K}}_{2,\infty}\leq C_{K}^{2,\infty}$, $\norm{W_{K}}_{2}\leq C_{K}$, 
        $\norm{W_{V}}_{2,\infty}\leq C_{V}^{2,\infty}$, $\norm{W_{V}}_{2}\leq C_{V}$,
        $\norm{W_O}_{2,\infty}\leq C_{O}^{2,\infty}$, $\norm{W_O}_{2}\leq C_{O}$,
        $\norm{E^\top}_{2,\infty}\leq C_E$;
        \item Parameter bound in $f^{\text{(FF)}}$:      $\norm{W_{1}}_{2,\infty} \leq C_{f_1}^{2,\infty}$,
        $\norm{W_{1}}_{2} \leq C_{f_1}$,
        $\norm{W_{2}}_{2,\infty} \leq C_{f_2}^{2,\infty}$,
        $\norm{W_{2}}_{2} \leq C_{f_2}$;
        \item Lipschitz of $f_{\calT} \in \calT^{\textcolor{blue}{h,s,r}}$: $\norm{f_{\calT}(Z_1)-f_{\calT}(Z_2)}_F \leq L_{\calT}\norm{Z_1-Z_2}_F$, for any $Z_1,Z_2 \in \R^{d\times L}$.
    \end{itemize}
\end{definition}
These norm bounds are critical to quantify the interplay between model, performance and data.

\section{Statistical Limits of Conditional DiTs}
\label{sec:con_dit}

In this section, we present a refined decomposition scheme for the fine-grained analysis of score approximation, score estimation, and distribution estimation in conditional DiT.
Our analysis considers two assumptions on initial data distributions:
\begin{itemize}
    \item
    A generic H\"{o}lder smooth data assumption (\cref{sec:generic_holder} for approximation, and \cref{sec:score_est_dist_est} for estimation),

    \item 
    A stronger H\"{o}lder smooth data  assumption (\cref{sec:bounded_holder} for approximation, and \cref{sec:score_est_dist_est} for estimation).
\end{itemize}
This new scheme leads to tighter bounds, including the minimax optimality of the unconditional DiT score estimator.

\subsection{Score Approximation: Generic H\"{o}lder Smooth Data Distributions}\label{sec:generic_holder}

We present a fine-grained piecewise approximation using transformers to approximate the conditional score function under the H\"{o}lder smoothness assumption on the initial data \cite{fu2024unveil}. 
At its core, we introduce a score function decomposition scheme with term-by-term tractability.

We first introduce the definition of H\"{o}lder space and H\"{o}lder ball following \cite{fu2024unveil}.
\begin{definition}[H\"{o}lder Space]
\label{def:holder_norm_space}
Let $\alpha \in \mathbb{Z}_{+}^{d}$, and let $\beta = \textcolor{blue}{k_1} + \gamma$ denote the smoothness parameter, where $\textcolor{blue}{k_1} = \lfloor \beta \rfloor$ and $\gamma \in [0,1)$. 
For a function $f: \R^{d} \rightarrow \R$, the H\"{o}lder space $\calH^{\beta}(\R^{d})$ is defined as the set of $\alpha$-differentiable functions satisfying:
$\calH^{\beta}(\R^{d}) \coloneqq \left\{ f: \R^{d} \rightarrow \R \mid \|f\|_{\calH^{\beta}(\R^{d})} < \infty \right\}$,
where the H\"{o}lder norm $\| f \|_{\calH^{\beta}(\R^{d})}$ satisfies:
\begin{align*}
\| f \|_{\calH^{\beta}(\R^{d})} \coloneqq \max_{\alpha : \|\alpha\|_{1} < \textcolor{blue}{k_1}} \sup_{x} \left| \partial^{\alpha} f(x) \right| + \max_{\alpha : \|\alpha\|_{1} = \textcolor{blue}{k_1}} \sup_{x \neq x'} \frac{\left| \partial^{\alpha} f(x) - \partial^{\alpha} f(x') \right|}{\|x - x'\|_{\infty}^{\gamma}}.
\end{align*}
We also define the H\"{o}lder ball of radius $B$:
$
\calH^{\beta}(\R^{d}, B) \coloneqq \left\{ f : \R^{d} \rightarrow \R \mid \| f \|_{\calH^{\beta}(\R^{d})} < B \right\}.$
\end{definition}

Let $x_0 \in \R^{d_x}$ denote the initial data, and $y \in [0,1]^{d_y}$ the conditional label.
With \cref{def:holder_norm_space},
we state the first assumption on the conditional distribution of initial data $x_0$.
\begin{assumption}[Generic H\"{o}lder Smooth Data]
\label{assumption:conditional_density_function_assumption_1}
The conditional density function $p_0(x_0|y)$ is defined on the domain $\R^{d_{x}}\times[0,1]^{d_{y}}$ and belongs to H\"{o}lder ball of radius $B>0$ for H\"{o}lder index $\beta>0,$
denoted by $p_0(x_0|y)\in\calH^{\beta}(\R^{d_{x}}\times[0,1]^{d_{y}},B)$ (see \cref{def:holder_norm_space} for precise definition.)
Also, for any $y\in[0,1]^{d_{y}}$, there exist positive constants $C_{1}, C_{2}$ such that $p_0(x_0|y)\leq C_{1}\exp(-C_{2}\norm{x_0}_{2}^{2}/2)$. 
\end{assumption}

\begin{remark}
The H\"{o}lder continuity assumption captures various smoothness levels in the conditional density function.
The light-tail condition relaxes the bounded support assumption in \cite{oko2023diffusion}.
Moreover,
\cref{assumption:conditional_density_function_assumption_1} only applies to the initial conditional distribution and imposes no constraints on the induced conditional score function.
This is far less restrictive than the Lipschitz score condition in prior works \cite{yuan2024reward,lee2023convergence,chen2022sampling}.
\end{remark}

In our work, we aim to approximate the conditional score function $\nabla \log p_t(x_t|y)$ using transformer architectures. 
\citet{hu2024statistical} analyze the unconditional DiTs based on the established universality of transformers \cite{yun2019transformers}. 
These theories discretize the input and output domains into infinitesimal grids and employ piecewise constant approximations to construct universal approximators with controllable errors.
However, such methods do not yield tight bounds for DiT architectures \cite{hu2024statistical}.
To combat this, we build on the key observation by \citet{fu2024diffusion}\footnote{Recall that $p_{t}(x_t|y) = \int_{\R^{d_{x}}} p(x_{0}|y) p_{t}(x_t|x_{0}) \, \dd x_{0}$ with $P_t(\cdot| y)\sim N(\alpha_t x_0, \sigma_t I_{d_x})$.
In below, when the context is clear, we suppress the notation dependence of $x_t$ on the time step $t$.}:
\begin{align}\label{eqn:decomposed_cond}
p_{t}(x_t|y) 
=\int_{\R^{d_{x}}} \frac{\dd x_{0}}{\sigma_{t}^{d_x}(2\pi)^{d_x/2}}
\cdot 
\underbrace{p_0(x_{0}|y)}_{\approx \textcolor{blue}{k_1}\text{-order Taylor polynomial}}
\cdot \underbrace{\exp\left(-\frac{\norm{\alpha_{t}x_{0}-x_t}^{2}}{2\sigma_{t}^{2}}\right)}_{\approx \textcolor{blue}{k_2}\text{-order Taylor polynomial}} .
\end{align}
A term-by-term Taylor expansion of the above conditional distribution under \cref{assumption:conditional_density_function_assumption_1} enables a more fine-grained analysis (e.g., \cref{lem:f_1_expression}). 
As a result, we propose a \textit{fine-grained version} of \textit{piecewise constant approximation} for conditional DiTs, allowing transformers to approximate the conditional score function with tighter error bounds. 
In particular, we utilize a refined transformer universal approximation modified from \cite{kajitsuka2023transformers} (see \cref{sec:trans_univeral_approx} for details).

Our score approximation procedure has two stages: first, we approximate $p_t$ and $\nabla p_t$ using a Taylor expansion, then use transformers to approximate $p_t$, $\nabla p_t$, and the required algebraic operators to construct $\nabla \log{p_{t}}(x|y) = \frac{\nabla p_{t}(x|y)}{p_{t}(x|y)}$.
These lead to provably tight estimation results in \cref{sec:score_est_dist_est}.

We state our main result of score approximation using transformers under \cref{assumption:conditional_density_function_assumption_1} as follows:

\begin{theorem}[Conditional Score Approximation under \cref{assumption:conditional_density_function_assumption_1}]
\label{thm:Main_1}
Assume \cref{assumption:conditional_density_function_assumption_1} \blue{and $d_x=\Omega( \frac{\log N}{\log \log N})$}. 
For any precision parameter $0 < \epsilon < 1$ and smoothness parameter $\beta > 0$, let $\epsilon \le \calO(N^{-\beta})$ for some $N \in \mathbb{N}$.
For some positive constants $C_{\alpha},C_{\sigma}>0$, for any $y \in [0,1]^{d_{y}}$ and $t \in [N^{-C_{\sigma}}, C_{\alpha} \log N]$, there exists a $\calT_{\text{score}}(x,y,t)\in\calT_R^{\textcolor{blue}{h,s,r}}$ such that
\begin{align*}
\int_{\R^{d_{x}}} \| \calT_{\text{score}}(x,y,t) - \nabla \log p_{t}(x|y) \|_{2}^{2} \cdot p_{t}(x|y) \, \dd x = \calO\left( \frac{B^{2}}{\blue{\sigma_{t}^{4}}} \cdot N^{-\frac{\beta}{d_{x} + d_{y}}} \cdot (\log N)^{d_{x} + \frac{\beta}{2} + 1} \right).
\end{align*}
Notably, for $\epsilon=\calO(N^{-\beta})$, the approximation error has the upper bound \blue{$\calO((\log(\frac{1}{\epsilon}))^{d_x}/\sigma_t^4)$}.

The parameter bounds for the transformer network class are as follows:
\begin{align*}
& \norm{W_{Q}}_{2}, \norm{W_{K}}_{2},
\norm{W_{Q}}_{2,\infty}, \norm{W_{K}}_{2,\infty}
=
\calO\left(N^{\frac{7\beta}{d_x+d_y}+6C_{\sigma}}\right); \\
&
\norm{W_{O}}_{2},\norm{W_{O}}_{2,\infty}=\calO\left(N^{-\frac{3\beta}{d_x+d_y}+6C_{\sigma}}(\log{N})^{3(d_x+\beta)}\right);\\
&
\norm{W_{V}}_{2}=\calO(\sqrt{d});
\quad \norm{W_{V}}_{2,\infty}=\calO (d);\\ 
& \norm{W_{1}}_{2}, \norm{W_{1}}_{2,\infty}
=
\calO\left(N^{\frac{2\beta}{d_x+d_y}+4C_{\sigma}}\right);
\norm{E^{\top}}_{2,\infty}=\calO\left(d^{\frac{1}{2}}L^{\frac{3}{2}}\right);
\\
& \norm{W_{2}}_{2}, \norm{W_{2}}_{2,\infty}
=
\calO\left(N^{\frac{3\beta}{d_x+d_y}+2C_{\sigma}}\right);
C_\calT=\calO\left(\sqrt{\log{N}}/\sigma_{t}^{2}\right).
\end{align*}
\end{theorem}
\begin{remark}
   $N$ is the resolution of the input domain discretization (see \cref{lem:f_1_expression}).
   \blue{We remark that domain discretization is essential for utilizing the local smoothness of functions under H\"{o}lder  assumptions.}
   $C_\sigma$ and $C_\alpha$ control the stability cutoff and early stopping time, respectively.   
\end{remark}

\begin{proof}[Proof Sketch]
Recall that $\nabla \log{p_{t}}(x|y) = \frac{\nabla p_{t}(x|y)}{p_{t}(x|y)}$. 
We employ  the following strategy: 
discretize the domains, apply a term-by-term Taylor approximation to the decomposed conditional distribution \eqref{eqn:decomposed_cond}, decompose the conditional score function $\nabla \log{p_{t}}(x|y) = \frac{\nabla p_{t}(x|y)}{p_{t}(x|y)}$ into two fundamental functions and a parsimonious set of algebraic operators, and then approximate the fundamental functions and operators with transformer networks.
The resulting joint error of this strategy is controllable under \cref{assumption:conditional_density_function_assumption_1}.
Our proof follows three steps:

\textbf{Step 1. Input Domains Discretization.} 
For any $x \in \R^{d_x}$, we construct a bounded domain $B_{x,N}$ to approximate polynomial functions evaluated at $x$ on $\R^{d_x}$ with the same functions on $B_{x,N}$ to arbitrary precision $1/N$ (\cref{clipping_integral}).
Then, we discretize $B_{x,N} \times [0,1]^{d_{y}}$ into $N^{d_{x}+d_{y}}$ hypercubes (\cref{lem:f_1_expression}).
This technique confines the approximation to a compact domain by controlling error outside this domain under \cref{assumption:conditional_density_function_assumption_1}.
Each hypercube is now compact and local, enabling a well-behaved Taylor expansion at $x$. 
This confinement reduces approximation error in \textbf{Step 2}.
    
\textbf{Step 2. Local, Term-by-Term Taylor Expansion for $\nabla \log{p_{t}}$.} 
    To approximate $\nabla \log{p_{t}}$, 
    we expand $p_{t}(x|y)$ and $\nabla p_{t}(x|y)$ with Taylor polynomials on each local grid on $B_{x,N}$, following the term-by-term expansion \eqref{eqn:decomposed_cond}. 
    Specifically, we approximate $p_{t}(x|y)$ with a scalar polynomial function $f_{1}(x,y,t) \in \R$ (\cref{lemma:diffused_local_polynomials}) and $\nabla p_{t}(x|y)$ with a vector-valued polynomial function $f_{2}(x,y,t) \in \R^{d_{x}}$ (\cref{lemma:diffused_local_polynomials_grad}).
    Together with a parsimonious set of algebraic operators (inverse, product), the obtained $f_1,f_2$ resemble $\nabla \log{p_{t}}$ with a bounded error ${\rm Error}_{\rm Taylor}$.
    
\textbf{Step 3. Term-by-Term Approximations with Transformers.} 
We utilize a refined universal approximation theorem for transformers (\cref{sec:trans_univeral_approx}) to approximate all Taylor-expanded terms: $f_1$, $f_2$, and the set of algebraic operators.
Specifically,  we approximate $f_{1}(x,y,t)$ and $f_{2}(x,y,t)$ with transformer models $\calT_{f_{1}}$ (\cref{lemma:Trans_Approx_Poly}) and $\calT_{f_{2}}$ (\cref{lemma:Trans_Approx_Poly_Gradient}).
For the operators, we also approximate each of them with a corresponding transformer $\calT_\mu$ with $\mu=\{{\rm inverse},{\rm square}\ldots\}$ (\cref{lemma:inverse_trans,lemma:trans_approx_variance,lemma:inverse_trans,lemma:approx_prod_with_trans}).
All approximations have precision guarantees.
Finally, we combine the transformer approximations $\calT_{f_{1}}$, $\calT_{f_{2}}$ and  ${\calT_\mu}$ for the set of algebraic operators, resulting in a joint approximation for $\nabla \log p_{t}$ (see \cref{fig:error_main1}) with arbitrary small error ${\rm Error_{\calT}}$.

\textbf{Error Matching.}
The overall error includes ${\rm Error}_{\rm Taylor}$ and ${\rm Error_{\calT}}$.
Given a fixed discretization resolution $N$, ${\rm Error}_{\rm Taylor}$ remains fixed.
However, the approximation error bound of the transformer can be an arbitrary value.
We align ${\rm Error_{\calT}}$ and ${\rm Error}_{\rm Taylor}$ to optimize the final results.

\begin{figure}[t!]
\centering
\definecolor{red}{rgb}{0.698,0.133,0.133}
\definecolor{green}{rgb}{0,0.40,0}
\definecolor{gray}{RGB}{169,169,169}

\resizebox{\textwidth}{!}{
\begin{tikzpicture}[node distance=3.5cm and 2cm, auto, font=\footnotesize, scale=0.15]

\node (main1) [] {$\calT_{\text{score}}$};

\node (min1) [right of=main1, xshift=-1.0cm, yshift=0cm, draw=green, circle, fill=green!50, inner sep=0pt, minimum size=8pt] {};
\node at (min1.north) [above, xshift=-0.20cm, yshift=0.05cm, text=green] {entrywise min};

\node (square1) [right of=min1, xshift=-1.0cm, yshift=-0.5cm, draw=green, circle, fill=green!50, inner sep=0pt, minimum size=8pt] {};
\node at (square1.north) [below, yshift=-0.3cm, text=green] {\shortstack{square \\\cref{lemma:approx_prod_with_trans}}};

\node (inv1) [right of=square1, xshift=-1.0cm, yshift=0.0cm, draw=green, circle, fill=green!50, inner sep=0pt, minimum size=8pt] {};
\node at (inv1.north) [below, yshift=-0.3cm, text=green] {\shortstack{inverse \\\cref{lemma:inverse_trans}}};

\node (sig1) [right of=inv1,xshift=-1.0cm,yshift=0.0cm] {\shortstack{$\sigma_t$\\ \cref{lemma:trans_approx_variance}}};

\node (prod1) [right of=min1, xshift=-1.0cm, yshift=1.0cm, draw=green, circle, fill=green!50, inner sep=0pt, minimum size=8pt] {};
\node at (prod1.north) [above, xshift=-1cm, yshift=0.05cm, text=green] {\shortstack{product \\\cref{lemma:approx_prod_with_trans}}};

\node (f2) [right of=prod1, xshift=-1.0cm, yshift=-0.8cm] {\shortstack{$f_2$ \\ \cref{lemma:Trans_Approx_Poly_Gradient}}};

\node (inv2) [right of=prod1, xshift=-1.0cm, yshift=0.0cm, color=green] { \shortstack{inverse \\ \cref{lemma:inverse_trans} }};

\node (max1) [right of=inv2, xshift=-1.0cm, yshift=0.0cm, draw=green, circle, fill=green!50, inner sep=0pt, minimum size=8pt] {};
\node at (max1.north) [above, yshift=0.05cm, text=green] {\shortstack{max}};

\node (f1) [right of=max1, xshift=-1.0cm, yshift=0.0cm] {\shortstack{$f_1$ \\ \cref{lemma:Trans_Approx_Poly}}};

\node (eps_low) [below of=f1, xshift=0.0cm, yshift=+2.7cm] {\shortstack{$\epsilon_{\text{low}}$ }};

\node (inv3) [above of=inv2, xshift=0.0cm, yshift=-2.6cm, color=green] { \shortstack{inverse \\ \cref{lemma:inverse_trans} }};

\node (sig2) [right of=inv3,xshift=-1.0cm,yshift=0.0cm] {\shortstack{$\sigma_t$\\ \cref{lemma:trans_approx_variance}}};

\draw[->, line width=0.3mm] (min1) -- (main1) node[midway, above, yshift=0.0cm,xshift=0.0cm] {};

\draw[->, line width=0.3mm] (square1) -- (min1) node[midway, below, yshift=0.0cm,xshift=-0.4cm, sloped] {\scriptsize \shortstack{Rescale \\ \cref{lemma:Score_Approx_Trans}}};

\draw[->, line width=0.3mm] (inv1) -- (square1) node[midway, above, yshift=0.0cm,xshift=0.1cm, sloped] {\shortstack{}};

\draw[->, line width=0.3mm] (sig1) -- (inv1) node[midway, above, yshift=0.0cm,xshift=0.1cm, sloped] {\shortstack{}};

\draw[->, line width=0.3mm] (prod1) -- (min1) node[midway, above, yshift=0.0cm,xshift=0.0cm, sloped] {\shortstack{}};

\draw[->, line width=0.3mm] (f2) -| (prod1) node[midway, above, yshift=0.0cm, xshift=0.0cm, sloped] {\shortstack{}};

\draw[->, line width=0.3mm] (inv2) -- (prod1) node[midway, above, yshift=0.0cm, xshift=0.0cm, sloped] {\shortstack{}};

\draw[->, line width=0.3mm] (max1) -- (inv2) node[midway, above, yshift=0.0cm, xshift=0.0cm, sloped] {\shortstack{}};

\draw[->, line width=0.3mm] (f1) -- (max1) node[midway, above, yshift=0.0cm, xshift=0.0cm, sloped] {\shortstack{}};

\draw[->, line width=0.3mm] (eps_low) -| (max1) node[midway, above, yshift=0.0cm, xshift=0.0cm, sloped] {\shortstack{}};

\draw[->, line width=0.3mm] (inv3) -| (prod1) node[midway, above, yshift=0.0cm, xshift=0.0cm, sloped] {\shortstack{}};

\draw[->, line width=0.3mm] (sig2) -- (inv3) node[midway, above, yshift=0.0cm, xshift=0.0cm, sloped] {\shortstack{}};

\end{tikzpicture}
}
\vspace{-.5em}
\caption{\small
\textbf{Approximate Score Function with Transformer $\calT_{\text{score}}$ under \cref{assumption:conditional_density_function_assumption_1}.} 
The construction consists of the transformers to approximate local polynomials $f_1$ and $f_2$, and the algebraic operators.
We highlight the overall term-by-term approximations and their corresponding lemmas to ensemble the transformers.
}
\label{fig:error_main1}
\end{figure}
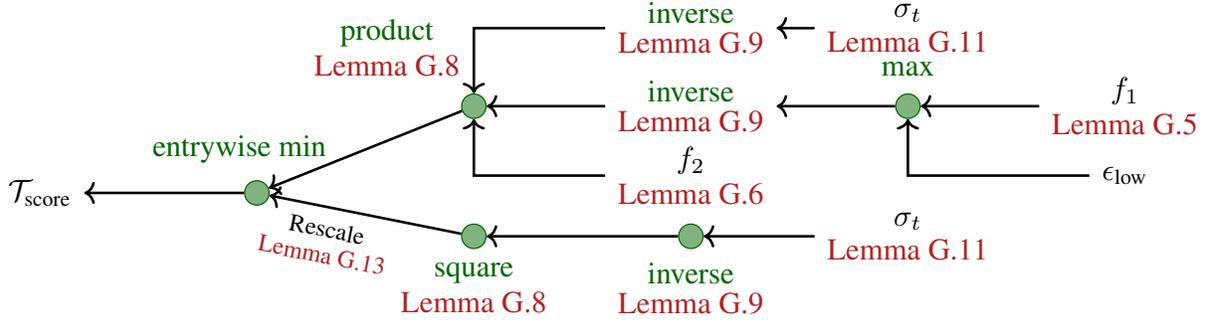

Please see \cref{sec:appendix_proof_main1} for a detailed proof.
\end{proof}

\begin{remark}[Approximation Rate]
    Given a fixed resolution $N$, the approximation error scales inversely with the smoothness  $\beta$.
    As the smoothness increases, we get a tighter approximation error.
\end{remark}

\begin{remark}[Comparing with Existing Works]
\citet{fu2024diffusion} provide approximation rates for conditional diffusion models using ReLU networks. 
We are the first to establish approximation error bounds with transformer networks. 
Additionally, \citet{oko2023diffusion} establish approximation rates under a compactness condition on the input data. 
We mitigate this compactness requirement by applying a H\"{o}lder smoothness assumption to control approximation error outside a compact domain.
\end{remark}

\subsection{Score Approximation: Stronger H\"{o}lder Smooth Data Distributions}\label{sec:bounded_holder}

Next, we study the conditional DiT score approximation problem using our score decomposition scheme under the stronger H\"{o}lder smoothness assumption from \citet[Assumption 3.3]{fu2024unveil}.

\begin{assumption}[Stronger H\"{o}lder Smooth Data]
\label{assumption:conditional_density_function_assumption_2}
Let function $f\in\calH^{\beta}(\R^{d_{x}}\times[0,1]^{d_{y}},B)$.
Given a constant radius $B$, positive constants $C$ and $C_{2}$, we assume the conditional density function $p(x_{0}|y)=\exp(-C_{2}\norm{x_{0}}_{2}^{2}/2)\cdot f(x_{0},y)$ and $f(x_{0},y)\geq C$ for all $(x_{0},y) \in \R^{d_{x}}\times[0,1]^{d_{y}}$.
\end{assumption}

\cref{assumption:conditional_density_function_assumption_2} imposes stronger assumption than \cref{assumption:conditional_density_function_assumption_1} and induces a refined conditional score function decomposition.
Explicitly, by \cref{lemma:score_decomp}, $\nabla\log{p_{t}(x|y)}$ becomes:
\begin{align}\label{eqn:stronger_holder_score_decom}
\nabla\log{p_{t}(x|y)} = \frac{-C_{2}x}{\alpha_{t}^{2}+C_{2}\sigma_{t}^{2}} + \frac{\nabla h(x,y,t)}{h(x,y,t)},
\end{align}
where $h(x,y,t) \coloneqq \int_{\R^{d_x}}\frac{f(x_{0},y)}{\hat{\sigma}_{t}^{d_x}(2\pi)^{d_x/2}}\exp(-\frac{\norm{x_{0}-\hat{\alpha}_{t}x}^{2}}{2\hat{\sigma}_{t}^{2}})\dd x_{0}$, $\hat{\sigma}_{t} = \frac{\sigma_{t}}{\sqrt{\alpha_{t}^{2}+C_{2}\sigma_{t}^{2}}}$, and
$\hat{\alpha}_{t}=\frac{\alpha_{t}}{\alpha_{t}^{2}+C_{2}\sigma_{t}^{2}}$.

We highlight that \eqref{eqn:stronger_holder_score_decom} leads to a tighter approximation error compared with \cref{thm:Main_1}.
Intuitively, \cref{assumption:conditional_density_function_assumption_2} imposes a lower bound on the conditional density function and hence implies in better regularity of the score function.
In contrast, under \cref{assumption:conditional_density_function_assumption_1},
the score function lacks such regularity and may explode when $p_t$ is small.
These low-density regions act as holes in the data support.
They cause the score function to diverge near the boundary of these holes. 
To combat this, an implication of  \eqref{eqn:stronger_holder_score_decom} is handy --- $h$ is bounded from zero, ensuring that the score function remains well-behaved across the entire data domain.
To elaborate more, two technical remarks are in order.
\begin{remark}[Linearity]
The first term on the RHS of \eqref{eqn:stronger_holder_score_decom} is linear in $x$.
This makes part of $\nabla \log p_{t}(x|y)$ a \textit{linear} function of $x$, enabling easy approximation with a tighter bound.
\end{remark}
\begin{remark}[Tightened Approximation Induced by $h$'s Lower Bound]
Moreover, the introduction of $h$ tightens the approximation error due to the lower bound imposed by \cref{assumption:conditional_density_function_assumption_2} (i.e., $f(x, y) \geq C$). 
The second term on the RHS of \eqref{eqn:stronger_holder_score_decom} mirrors the form $\nabla \log p_t(x|y) = \frac{\nabla p_t(x|y)}{p_t(x|y)}$ by replacing $p$ with $h$. 
In the analysis of \cref{sec:generic_holder}, especially in \textbf{Step 2} of the proof (resembling $f_1,f_2$ to approximate $\nabla p_t(x|y)$), 
we have to impose
a threshold on the denominator of $\frac{\nabla p_t(x|y)}{p_t(x|y)}$ to prevent score explosion under \cref{assumption:conditional_density_function_assumption_1}.
This threshold introduces additional approximation error (\cref{lemma:Score_Approx_Trans}).
\cref{assumption:conditional_density_function_assumption_2} remedies this by ensuring a lower bound on $p_{t}(x|y)$ through the minimum values of $f(x,y)$ and ${\rm exp} (-C_{2}\norm{x}_{2}^{2}/2)$ within the compact domain after discretization.
Setting this lower bound eliminates the need for a threshold and improves the approximation.
\end{remark}

Consequently, decomposition \eqref{eqn:stronger_holder_score_decom} improves our approximation result from \cref{sec:generic_holder}.
We state our main result of score approximation using transformers under \cref{assumption:conditional_density_function_assumption_2} as follows:

\begin{theorem}[Conditional Score Approximation under \cref{assumption:conditional_density_function_assumption_2} (Informal Version of \cref{thm:Main_2})]
\label{thm:Main_2_informal}
Assume \cref{assumption:conditional_density_function_assumption_2}.
For any precision parameter $0 < \epsilon < 1$ and smoothness parameter $\beta > 0$, let $\epsilon \le \calO(N^{-\beta})$ for some $N \in \mathbb{N}$.
For some positive constants $C_{\alpha},C_{\sigma}>0$, for any $y \in [0,1]^{d_{y}}$ and $t \in [N^{-C_{\sigma}}, C_{\alpha} \log N]$, there exists a $\calT_{\text{score}}(x,y,t)\in\calT_R^{\textcolor{blue}{h,s,r}}$ such that
\begin{align*}
\int_{\R^{d_x}}\norm{\calT_{\text{score}}(x,y,t)-\nabla\log{p_{t}(x|y)}}_{2}^{2}\cdot p_{t}(x|y) \dd x = \calO\left(\frac{B^{2}}{\sigma_{t}^{2}}\cdot N^{-\frac{2\beta}{d_{x}+d_{y}}}\cdot(\log{N})^{{\beta}+1}\right).
\end{align*}
Notably, for $\epsilon=\calO(N^{-\beta})$, the approximation error has the upper bound \blue{$(\log (\frac{1}{\epsilon}))^{\calO(1)}/\sigma_t^2$}.
\end{theorem}

\begin{proof}[Proof Sketch]
Our proof follows \cref{thm:Main_1},
but uses a different conditional score function decomposition in the form of \eqref{eqn:stronger_holder_score_decom}.
We highlight key differences in each corresponding step:

\textbf{Step 0: Score Decomposition and Bounds on $h$ and $\nabla h$.} 
    We decompose $\nabla p_{t}$ to the form of \eqref{eqn:stronger_holder_score_decom} by \cref{lemma:score_decomp}.
    Different from \cref{sec:bounded_holder}, we derive a lower bound on $h$  in \cref{lemma:bounds_on_h}.

\textbf{Step 1: Input Domains Discretization.}
This step remains the same as \cref{sec:generic_holder}, except the approximation target changes from $p, \nabla p$ to $h, \nabla h$.
We confine and discretize input domains $\R^{d_x}\times [0,1]^{d_y}$ into $N^{d_{x}+d_{y}}$ hypercubes (\cref{lem:f_1_expression}), each supporting well-behaved Taylor expansions.

\textbf{Step 2: Local, Term-by-Term Taylor Expansion for $h$ and $\nabla h$.} 
Similar to \cref{sec:generic_holder}, we utilize Taylor polynomials $f_{1}$ and $f_{2}$ to approximate $h$ and $\nabla h$ on obtained hypercubes.
The approximation on $h$ and $\nabla h$ differs from approximation on $p_{t}$ and $\nabla p_{t}$, 
as their boundedness eliminates the need for a threshold to prevent score function blow-up.
This leads to a faster approximation rate.
    
 \textbf{Step 3: Transformer Network Approximation.} 
    Similar to \cref{sec:generic_holder},
    we approximate polynomial functions $f_1, f_2$ and all necessary algebraic operators to construct an approximator $f_3$ for $\nabla p_t$:
    \begin{align}\label{eqn:thm32_step3}
    f_{3}(x,y,t) = - \frac{C_{2}x}{\alpha_{t}^{2} + C_{2}\sigma_{t}^{2}}+\frac{\hat{\alpha}_{t}}{\hat{\sigma}_{t}} \cdot \frac{f_{2}(x,y,t)}{f_{1}(x,y,t)}, 
    \end{align}
    following \eqref{eqn:stronger_holder_score_decom}.
    Differed from \cref{sec:generic_holder}, \eqref{eqn:stronger_holder_score_decom} requires transformers to approximate two additional operators, $\hat{\sigma}_t$ and $\hat{\alpha}_t$.
    All approximations have precision guarantees. 
    Finally, we combine all transformer approximations required in \eqref{eqn:thm32_step3} 
    and obtain a joint approximation error for $\nabla \log p_t$  (see \cref{fig:error_main2}) with arbitrary precision.
    We complete the proof by matching the approximation errors of the Taylor polynomial and transformer.
    Importantly, second term on the RHS of \eqref{eqn:thm32_step3} manifests a tighter bound than that of $\frac{\nabla p_{t}(x|y)}{p_{t}(x|y)}$. 
    The first linear-in-$x$ term achieves a even tighter bound due to its linearity.
    Combined, we obtain a smaller overall joint approximation error than \cref{thm:Main_1}.

Please see \cref{sec:appendix_proof_main2} for a detailed proof, and see \cref{thm:Main_2} for the formal version.
\end{proof}

\begin{remark}[Comparing with \cref{thm:Main_1}]
Let $\tilde{\calO}(\cdot)$ hide the terms about $t_0$, $\log t_0$, $\log n$.
    In \cref{thm:Main_2_informal}, the approximation rate $\tilde{\calO}(N^{-\frac{2\beta}{d_{x}+d_{y}}})$ is faster than that of \cref{thm:Main_1}, i.e.,  $\tilde{\calO}(N^{-\frac{\beta}{d_{x}+d_{y}}})$.
\end{remark}

\subsection{Score Estimation and Distribution Estimation of Conditional DiTs}\label{sec:score_est_dist_est}

Next, we study score and distribution estimations based on the two score approximation results for two different data assumptions: \cref{thm:Main_1,thm:Main_2_informal}.
Let $\hat{s}$ denote the trained score estimator.

\textbf{Score Estimation.}
Building on our approximation results from \cref{sec:generic_holder,sec:bounded_holder}, 
the next objective is to evaluate the performance of the score estimator $\hat{s}$ trained with a set of finite samples by optimizing the empirical loss \eqref{eqn:empirical_loss}.
To quantify this, we introduce the notion of score estimation risk and characterize its upper bound.
\begin{definition}[Conditional Score Risk]\label{def:risk}
Given a score estimator $\hat{s}$, 
we define risk as the expectation of the squared $\ell_{2}$ difference between the score estimator and the ground truth with respect to $(x_{t},y,t)$:
\begin{align*}
\calR(\hat{s})\coloneqq\int_{t_{0}}^{T}\frac{1}{T-t_{0}}\E_{x_{t},y}\norm{\hat{s}(x_{t},y,t)-\nabla\log{p_{t}(x_t|y)}}_{2}^{2}\dd t.
\end{align*}
\end{definition}
Given a set of i.i.d sample $\{x_{i},y_{i}\}_{i\in[n]}$, direct computation of $\E_{\{x_{i},y_{i}\}_{i\in[n]}}[\calR(\hat{s})]$ is infeasible due to the absence of access to the joint distribution $P(x_{t},y)$.
To address this, we:
(i) Decompose the risk into estimation and approximation errors,
(ii) Bound the estimation error using the covering number of transformers, and
(iii) Bound the approximation error using \cref{thm:Main_1} and \cref{thm:Main_2_informal}.

\begin{theorem}[Conditional Score Estimation with Transformer]
\label{thm:main_risk_bounds}
\blue{Assume $d_x=\Omega( \frac{\log N}{\log \log N})$.}
\begin{itemize}[leftmargin=.8em]
    \item 
Under \cref{assumption:conditional_density_function_assumption_1}, by taking $N=n^{\frac{1}{\nu_{1}}\cdot \frac{d_{x}+d_{y}}{\beta+d_{x}+d_{y}}}$, $t_0=N^{-C_{\sigma}}<1$ and $T=C_{\alpha}\log{n}$, it holds
    \begin{align*}
    \E_{\{x_{i},y_{i}\}_{i=1}^{n}}\left[\calR(\hat{s})\right]=
    \calO\left(\frac{1}{t_{0}}n^{-\frac{1}{\nu_{1}}\cdot \frac{\beta}{d_{x}+d_{y}+\beta}}(\log{n})^{\nu_{2}+2}\right),
    \end{align*}
where \blue{$\nu_{1}=\frac{68\beta}{(d_x+d_y)}+104C_{\sigma}$
and
$\nu_{2}=12d_x+12\beta+2$.}

\item 
Under \cref{assumption:conditional_density_function_assumption_2},
    by taking $N=n^{\frac{1}{\nu_{3}}\cdot\frac{d_{x}+d_{y}}{2\beta+d_{x}+d_{y}}},$ $t_0=N^{-C_{\sigma}}<1$ and $T=C_{\alpha}\log{n}$, it holds
    \begin{align*}
    \E_{\{x_{i},y_{i}\}_{i=1}^{n}}\left[\calR(\hat{s})\right]=
    \calO\left(\log{\frac{1}{t_{0}}}n^{-\frac{1}{\nu_{3}}\cdot \frac{\blue{2\beta}}{d_{x}+d_{y}+2\beta}}(\log{n})^{{\max(10,\beta+1})}\right),
    \end{align*}
where 
\blue{$\nu_{3}={\frac{4(12\beta d_{x}+31\beta d+ 6\beta)}{d(d_x+d_y)} + \frac{12(12C_{\alpha}d_x + 25C_{\alpha}\cdot d + 6C_{\alpha})}{d}}
+72C_{\sigma}$}.
\end{itemize}

\end{theorem}
\begin{corollary}[Low-Dimensional Input Region]\label{cor:low_dim_score_est}
\blue{
Assume $d_x = o\left(\frac{\log N}{\log \log N}\right)$ , i.e., $d_x\ll n$. 
Under \cref{assumption:conditional_density_function_assumption_1}, by setting $N, t_0, T$ as specified in \cref{thm:main_risk_bounds}, we have
$\E_{\{x_i, y_i\}_{i=1}^{n}}\left[\calR(\hat{s})\right] = \calO\left(\frac{1}{t_0} n^{-\frac{1}{\nu_4} \cdot \frac{\beta}{d_x + d_y + \beta}}\right)$,
where
$\nu_4=\frac{72\beta(2d_x+5d+1)}{d(d_x+d_y)} + \frac{48C_{\sigma}(2d_x+5d+1)}{d} - 4\beta$.
}
\end{corollary}

\begin{proof}
    Please see \cref{app:proof_main_risk_bounds} and  \cref{proof:cor:low_dim_score_est} for detailed proofs.
\end{proof}

\begin{remark}[Sample Complexity Bounds]
To obtain $\epsilon$-error in terms of score estimation,
we have the sample complexity $\tilde{\calO}\left( \epsilon^{-{\nu_{1}(d_{x}+d_{y}+\beta)}/{\beta}}\right)$ under \cref{assumption:conditional_density_function_assumption_1} and $\tilde{\calO}\left( \epsilon^{-{\nu_{3}(d_{x}+d_{y}+2\beta)}/{\textcolor{blue}{2\beta}}} \right)$ under \cref{assumption:conditional_density_function_assumption_2}.
Here $\tilde{\calO}(\cdot)$ ignores the terms about $t_0$, $\log t_0$ and $\log n$.
The H\"older data smoothness degree $\beta$ affects the sample complexity.
This indicates that the regularity of the initial data distribution determines the complexity of score estimation.
\end{remark}

\textbf{Distribution Estimation.} 
Next, we study the distributional estimation capability of the trained conditional score network $s(x, y, t)$ by analyzing the total variation distance between the estimated and true distributions.
Our strategy uses a three-part decomposition:
(i) the total variation between the true distributions at timestamps $0$ and $t_0$,
(ii) the total variation between the true distribution at $t_0$ and the reverse process distribution using the true score function, and
(iii) the total variation between the reverse process distributions using the true and estimated score functions at $t_0$.

\begin{theorem}[Conditional Distribution Estimation]
\label{thm:distribution_TV_bound}
\blue{Assume $d_x=\Omega( \frac{\log N}{\log \log N})$}.
\blue{For $y\in[0,1]^{d_y}$, 
let $\hat{P}_{t_0}(\cdot | y)$ denote \textit{estimated} conditional distributions at $t_0$.
Recall that $P_0(\cdot|y)$ is the conditional distribution of initial data $x_0$ given $y$.
Assume $\mathrm{KL}\left(P_0(\cdot|y)\mid N(0,I)\right) \le c$ for some constant $c<\infty$.}

\begin{itemize}[leftmargin=.8em]
\item Under \cref{assumption:conditional_density_function_assumption_1}, by
taking the early-stopping time $t_0=n^{-\frac{\beta}{\blue{d_{x}+d_{y}+\beta}}}$ and terminal time $T=\frac{2\beta}{d_{x}+d_{y}+2\beta}\log{n}$, it holds
\begin{align*}
    \mathbb{E}_{\{x_i, y_i\}_{i=1}^n} \left[ \mathbb{E}_{y} \left[ {\rm TV} \left( \hat{P}_{t_0}(\cdot | y), P_0(\cdot | y) \right) \right] \right]
    = \calO \left(n^{-\frac{\beta}{2(\nu_{1}\blue{-}1)(d_{x}+d_{y}+\beta)}}(\log{n})^{\frac{\nu_{2}}{2}+\frac{3}{2}}\right),
\end{align*}
where \blue{$\nu_{1}=\frac{68\beta}{(d_x+d_y)}+104C_{\sigma}$,
$\nu_{2}=12d_x+12\beta+2$
and $C_{\sigma}=\frac{\beta}{d_x+d_y+\beta}$.}

\item
Under \cref{assumption:conditional_density_function_assumption_2},
by taking $t_{0}=n^{-\frac{4\beta}{d_{x}+d_{y}+2\beta}-1}$,
it holds
\begin{align*}
    \mathbb{E}_{\{x_i, y_i\}_{i=1}^n} \left[ \mathbb{E}_{y} \left[ {\rm TV} \left( \hat{P}_{t_0}(\cdot | y), P_0(\cdot | y) \right) \right] \right]=
    \calO\left(n^{-\frac{1}{2\nu_{3}}\frac{\beta}{d_{x}+d_{y}+2\beta}}(\log{n})^{{\max(6,\frac{\beta}{2}+\frac{3}{2}})}\right),
\end{align*}
where 
\blue{$\nu_{3}={\frac{4(12\beta d_{x}+31\beta d+ 6\beta)}{d(d_x+d_y)} + \frac{12(12C_{\alpha}d_x + 25C_{\alpha}\cdot d + 6C_{\alpha})}{d}}
+72C_{\sigma}$ and $C_{\alpha}=\frac{2\beta}{d_x+d_y+2\beta}$.}

\end{itemize}
We remark that the choice of $t_0,T$ (i.e., $C_\sigma,C_\alpha$) leads to the tightest rates in our analysis.
\end{theorem}
\begin{corollary}[Low-Dimensional Input Region]\label{cor:low_dim_distribution_est}
\blue{Assume $d_x = o\left(\frac{\log N}{\log \log N}\right)$ , i.e., $d_x\ll n$. 
Under \cref{assumption:conditional_density_function_assumption_1}, by setting $t_0, T$ as specified in \cref{thm:distribution_TV_bound}, we have
\begin{align*}
\mathbb{E}_{\{x_i, y_i\}_{i=1}^n} \left[ \mathbb{E}_{y} \left[ {\rm TV} \left( \hat{P}_{t_0}(\cdot | y), P_0(\cdot | y) \right) \right] \right]
    = \calO \left(n^{-\frac{\beta}{2(\nu_{4}+1)(d_{x}+d_{y}+\beta)}}\right),    
\end{align*}
where
$\nu_4=\frac{72\beta(2d_x+5d+1)}{d(d_x+d_y)} + \frac{48C_{\sigma}(2d_x+5d+1)}{d} - 4\beta$.}
\end{corollary}

\begin{proof}
    Please see \cref{app:proof_distribution_TV_bound} for a detailed proof.
\end{proof}

\subsection{Minimax Optimal Estimation of Unconditional DiTs}
\label{subsec:mini_op_con_dit}

In this section, we show the minimax optimality of the unconditional DiT architecture under \cref{assumption:conditional_density_function_assumption_2}.
Specifically, we obtain the distribution estimation error of unconditional DiTs by removing the condition $y$ and let $d_y=0$ in \cref{thm:distribution_TV_bound}.
Then the distribution estimation error becomes $\tilde{\calO}( \epsilon^{-\frac{1}{2\nu_{3}} \frac{\beta}{d_x+2\beta}} )$ under \cref{assumption:conditional_density_function_assumption_2}.
Here $\tilde{\calO}(\cdot)$ ignores the term about $\log n$.
By setting $2\nu_{3}=1$, we show that the unconditional DiT is the minimax optimal distribution estimator.

\begin{corollary}[Proposition 4.3 of \citet{fu2024unveil}]
\label{cor:holder_optimal}
For a fixed constant $C_{2}$ and a H\"{o}lder index $\beta>0$.
We consider the task of estimating a probability distribution $P(x)$ with its density function defined within the following function space
\begin{align*}
\calP=\left\{p(x)=f(x)\exp(-C_{2}\norm{x}_{2}^{2}):f(x)\in\calH^{\beta}(\R^{d_{x}},B),f(x)\geq C\geq0\right\},
\end{align*}
Given $n$ i.i.d data $\{x_{i}\}_{i=1}^{n}$,
we have $\inf_{\hat{\mu}}\sup_{p\in\calP}\E_{\{x_{i}\}_{i=1}^{n}}\left[\rm{TV}(\hat{\mu},P)\right] \geq \Omega(n^{-\frac{\beta}{d_{x}+2\beta}})$.
Here, the estimator $\hat{\mu}$ ranges over all possible estimators constructed from the data.
\end{corollary}
\begin{remark}
[Comparing with Existing Works] 
\citet{oko2023diffusion} analyze the \relu network and provide the near minimax optimal estimation rates in both the total variation distance and Wasserstein distance of order one.
\citet{fu2024unveil} also uses the \relu network and provides the minimax optimality for distribution in total variation.
Our results offer the first and exact minimax optimal guarantee for unconditional DiTs in distribution estimation.
\end{remark}

\section{Latent Conditional DiTs}
\label{sec:latent_con_dit}In this section, we extend the results from \cref{sec:con_dit} by considering the latent conditional DiTs. 
Specifically, we assume the raw input $x \in \R^{d_x}$ has an intrinsic lower-dimensional representation.

\begin{assumption}[Low-Dimensional Linear Latent Space]
\label{assumption:low_dim_linear_latent_space}
\blue{Initial data $x$ has a latent representation via  $x = U h$, where $U \in \mathbb{R}^{d_x \times d_0}$ is an unknown matrix with orthonormal columns. 
The latent variable $h \in \mathbb{R}^{d_0}$ follows the distribution $P_h$ with a density function $p_h$.}
\end{assumption}

\begin{remark}
    ``Linear Latent Space'' means that each entry of a given latent vector is a linear combination of the corresponding input, i.e., $x=Uh$. 
    This is also known as the ``low-dimensional data'' assumption in literature~\citep{hu2024statistical,chen2023score}.
    This assumption is fundamental in dimensionality reduction techniques for capturing the intrinsic lower-dimensional structure of data.
\end{remark}

\begin{figure}[t!]
\centering
\definecolor{encodercolor}{RGB}{255,99,71}
\definecolor{decodercolor}{RGB}{135,206,250}
\definecolor{networkcolor}{RGB}{211,211,211}
\definecolor{sumcolor}{RGB}{139,138,123}
\definecolor{greyblockcolor}{RGB}{169,169,169}
\definecolor{reshapecolor}{RGB}{169,169,169}
\definecolor{attncolor}{rgb}{0.698,0.133,0.133}
\definecolor{ffcolor}{rgb}{0,0.40,0}
\definecolor{normcolor}{RGB}{173,216,230}
\definecolor{concatcolor}{RGB}{169,169,169}
\resizebox{\textwidth}{!}{%
\begin{tikzpicture}[
    encoder/.style={trapezium, trapezium angle=60, draw=attncolor!70, fill=attncolor!30, thick, minimum height=1.2cm, minimum width=1.0cm, align=center, rotate=270},
    decoder/.style={trapezium, trapezium angle=60, draw=attncolor!70, fill=attncolor!30, thick, minimum height=1.2cm, minimum width=1.0cm, align=center, rotate=90},
    network/.style={rectangle, draw=ffcolor!70, fill=ffcolor!30, thick, minimum height=2cm, minimum width=3cm, align=center},
    greyblock/.style={rectangle, draw=greyblockcolor, fill=greyblockcolor!50, thick, minimum height=2cm, minimum width=1cm, align=center},
    yellowblock/.style={rectangle, draw=reshapecolor, fill=reshapecolor!50, thick, minimum height=1.5cm, minimum width=1.2cm, align=center},
    concatblock/.style={rectangle, draw=reshapecolor, fill=reshapecolor!50, thick, minimum height=2.0cm, minimum width=1.2cm, align=center},
    ffblock/.style={rectangle, draw=ffcolor!70, fill=ffcolor!30, thick, minimum height=1.5cm, minimum width=1cm, align=center},
    attnblock/.style={rectangle, draw=attncolor!70, fill=attncolor!30, thick, minimum height=1.5cm, minimum width=1cm, align=center},
    sum/.style={circle, draw=sumcolor, fill=sumcolor!50, thick, minimum size=0.4cm},
    shortcut/.style={dashed,  -{Latex[scale=1.5]}},
    myarrow/.style={thick, ->, -{Latex[scale=1.5]}},
    node distance=0.8cm, auto, scale=0.75, transform shape,
    decorate,decoration={brace,amplitude=10pt,raise=3pt},
    smallrect/.style={rectangle, draw=black, fill=gray!30, minimum width=0.01cm, minimum height=0.01cm},
    highlightrect/.style={rectangle, draw=black, fill=ffcolor!30, minimum width=0.01cm, minimum height=0.01cm},
    scale=0.85
]

\node (input) at (-4.1,1) {};
\node (y) at (-1.8,-1) {};

\node[encoder] at (-1.8,1) (encoder) {\rotatebox{90}{$ W_U^{\top} $}};
\node at (encoder.north) [above, xshift=-0.5cm, yshift=-1.8cm, text=black] {\shortstack{Latent\\ Encoder}};

\node[decoder] at (16.0,0) (decoder) {\rotatebox{-90}{$W_U$}};
\node at (decoder.north) [above, xshift=0.5cm, yshift=-1.8cm, text=black] {\shortstack{Latent\\ Decoder}};

\node[yellowblock] at (1.1,1) (reshape) {{$ \tilde{R}(\cdot) $}};
\node at (reshape.north) [above, xshift=0.0cm, yshift=-0.0cm, text=black] {\shortstack{Reshape Layer}};

\node[yellowblock] at (1.1,-1) (embed) {{Embed}};

\node[concatblock] at (3.5,0) (concat) {{Concat}};

\node[network] at (7.5,0) (network) {$g_{\calT} \in \calT^{\textcolor{blue}{h,s,r}}$};
\node at (network.south) [above, xshift=0.0cm, yshift=-0.5cm] {\shortstack{Transformer Network}};

\node[yellowblock] at (13.5,0) (reshapei) {{$ \tilde{R}^{-1}(\cdot) $}};
\node at (reshapei.north) [above, xshift=0.0cm, yshift=-0.0cm, text=black] {\shortstack{Reversed\\ Reshape Layer}};

\node (sum) at (18.2,0) {{\LARGE $\oplus$}};
\node at (19.5,0) (output) {};

\draw[myarrow] (input) -- (encoder) node[midway, below, yshift=-0.3cm, xshift=-0.1cm] {$x \in \R^{d_x}$};
\draw[myarrow] (encoder) -- (reshape) node[midway, below, yshift=-0.3cm,xshift=-0.1cm] {$x \in \R^{d_0}$};
\draw[myarrow] (y) -- (embed) node[midway, below, yshift=-0.3cm] {\parbox{1.8cm}{Label $y$ \\Timestep $t$}};

\draw[myarrow] (reshape) -- (concat) node[midway, above, xshift=0.0cm, yshift=0.1cm] {$ \R^{\tilde{d} \times \tilde{L}} $};
\draw[myarrow] (embed) -- (concat) node[midway, below, xshift=0.0cm, yshift=-0.0cm] {$ \R^{\tilde{d} \times 2} $};

\draw[myarrow] (concat) -- (network) node[midway, below, yshift=-0.3cm] {$ \R^{\tilde{d} \times (\tilde{L}+2)} $};
\draw[myarrow] (reshapei) -- (decoder) node[midway, below, yshift=-0.3cm] {$ \R^{d_0} $};
\draw[myarrow] (decoder) -- (sum) node[midway, below, yshift=-0.3cm] {$ \R^{d_x} $};

\draw[myarrow] (network) -- ++(3.4,0) node[midway, below, yshift=-0.3cm] {$ \R^{d \times (\tilde{L}+2)} $};

\node[highlightrect] at (11.2,0.7) (token1) {};
\node[highlightrect] at (11.2,0.525) (token2) {};
\node[highlightrect] at (11.2,0.35) (token3) {};
\node[highlightrect] at (11.2,0.175) (token4) {};
\node[highlightrect] at (11.2,0) (token5) {};
\node[highlightrect] at (11.2,-0.175) (token6) {};
\node[highlightrect] at (11.2,-0.35) (token7) {};
\node[smallrect] at (11.2,-0.525) (cls_token) {}; 
\node[smallrect] at (11.2,-0.7) (cls_token) {}; 

\draw[myarrow] (11.5,0) -- (reshapei) node[midway, below,yshift=-0.3cm] {$\R^{\tilde{d} \times \tilde{L}}$};

\draw[dashed, thick] (10.98, 0.88) rectangle (11.42, -0.47);

\draw[shortcut] (-3.6,1.1) -- ++(0,1.3) -- ++(21.8,0) node[midway, below] {$-{1}/{\sigma_{t}^{2}}$} -- (sum.north);

\draw[myarrow] (sum) -- (output) node[midway, below, yshift=-0.3cm] {$s_{{W}}$};

\end{tikzpicture}
}
\vspace{-1em}
\caption{\small
\textbf{Network Architecture of Latent Conditional DiT.} 
The overall architecture \blue{consists} of linear layer of encoder and decoder $W_U^{\top}$ and $W_U$ that transform input $x\in \R^{d_x}$ into linear latent space $\R^{d_0}$, reshaping layer $\tilde{R}(\cdot)$ and $\tilde{R}^{-1}(\cdot)$, embedding layer for label $y$ and timestep $t$. The embedding concatenates with input sequences and processes by the adapted transformer network ${\calT}^{\textcolor{blue}{h,s,r}}_{\tilde{R}} = \tilde{R}^{-1} \circ g_{\calT} \circ f^{({\rm FF})} \circ \tilde{R}$.
}
\label{fig:Latent_DiT}
\end{figure}
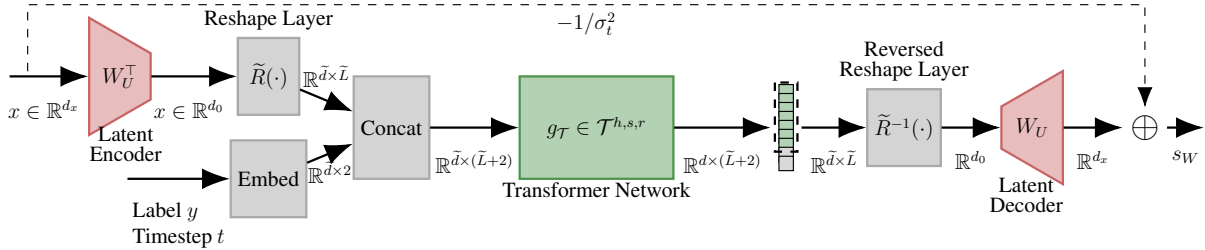

\textbf{Score Decomposition and Model Architecture.} 
To derive approximation and estimation results, we extend the techniques and network architecture presented in \cref{sec:con_dit} to latent diffusion by considering the ``low-dimensional linear subpace''.
Under \cref{assumption:low_dim_linear_latent_space}, we decompose the score:
\begin{align}
\label{eqn:score_docom_rearange}
    \nabla \log p_t(x | y) = U ( \underbrace{\sigma_{t}^2\nabla \log p_t^{h} ( U^\top x | y)
    +U^{\top}x}_{\coloneqq q(U^\top x, y,t):\;\R^{d_0} \times \R^{d_y} \times [t_0, T] \;\to\; \R^{d_0}})/\sigma_{t}^2 -  \underbrace{x/\sigma_{t}^2}_{\text{residual connection}},
\end{align}
following \citet{hu2024statistical,chen2023score}  (see \cref{lemma:subspace_score}).
Based on this decomposition, we construct the model architecture in \cref{fig:Latent_DiT}.
The network detail for approximate \eqref{eqn:score_docom_rearange} are as follow: a transformer $g_{\calT}(W_U^\top x, y, t) \in \calT_{\tildeR}^{h,s,r}$ to approximate $q(U^\top x, y,t)$, a latent encoder $W_U^{\top}\in \R^{d_0\times d_x}$ and decoder $W_U\in \R^{d_x\times d_0}$ to approximate $U^{\top} \in \R^{d_0\times d_x}$ and $U\in \R^{d_x\times d_0}$, and a residual connection to approximate $-x/\sigma_{t}^2$.
Importantly, $d_0$ is the latent dimension.

For latent diffusion, we follow the standard setting by \citet{peebles2023scalable}.
For each input $x \in \R^{d_x}$ and corresponding label $y \in \R^{d_y}$, we use a transformer network to obtain a score estimator $s_W \in \R^{d_x}$.
The key differences from \cref{sec:con_dit} are as follows:
First, we apply a latent encoder $W_U^\top \in \R^{d_0 \times d_x}$ to map the raw data $x \in \mathbb{R}^{d_x}$ into a low-dimensional representation $h\coloneqq W_U^\top x \in \mathbb{R}^{d_0}$, where $d_0 \le d_x$.
Second, we reshape $h \in \R^{d_0}$ into a sequence $H \in \R^{\tilde{d} \times \tilde{L}}$ using a layer $\tilde{R}(\cdot): \mathbb{R}^{d_0} \to \mathbb{R}^{\tilde{d} \times \tilde{L}}$, with $d_0 = \tilde{d} \cdot \tilde{L}$.
Note that, by $d_0 \leq d_x$, $\tilde{d} \leq d$, and $\tilde{L} \leq L$.
Third, we pass $H \in \R^{\tilde{d} \times \tilde{L}}$ through the transformer $g_{\calT}$. 
Lastly, We then obtain the predicted score $s_W \in \R^{d_x}$ by applying the inverse reshape layer $\tilde{R}^{-1}(\cdot): \mathbb{R}^{\tilde{d} \times \tilde{L}} \to \mathbb{R}^{d_0}$, followed by the latent decoder $W_U: \R^{d_0} \to \R^{d_x}$.

For our analysis, we study the cases under both the generic and strong  H\"older smoothness assumptions on latent representation $z \in \R^{d_0}$. 
Specifically, we assume the ``latent'' data is $\beta_0$-H\"older smooth with radius $B_0$ following \cref{assumption:conditional_density_function_assumption_1,assumption:conditional_density_function_assumption_2}.
We extend both approximation and estimation results from \cref{sec:con_dit} to latent diffusion and establish the minimax optimality of latent conditional DiTs.

\textbf{Score Approximation.}
We now present the approximation rates for  latent score function under both generic and stronger H\"older data assumptions.
Let $h\coloneqq W_U^\top x \in\R^{d_0}$ and $\bar{h}\coloneqq U^\top x\in \R^{d_0}$ be the estimated and ground truth (according to \cref{assumption:low_dim_linear_latent_space}) latent representations, respectively.

\begin{theorem}[Score Approximation of Latent Conditional DiTs (Informal Version of \cref{theorem:latent_main1_formal,theorem:latent_main2_formal})]
\label{theorem:latent_main1_infor}
\blue{Assume $d_x=\Omega( \frac{\log N}{\log \log N})$}.
For any precision  $0 < \epsilon < 1$ and smoothness  $\beta_0 > 0$, let $\epsilon \le \calO(N^{-\beta_0})$ for some $N \in \mathbb{N}$.
For some positive constants $C_{\alpha},C_{\sigma}>0$, for any $y \in [0,1]^{d_{y}}$ and $t \in [N^{-C_{\sigma}}, C_{\alpha} \log N]$, there exists a $\calT_{\text{score}}(x,y,t)\in\calT_{\textcolor{blue}{\tildeR}}^{\textcolor{blue}{h,s,r}}$ 
such that 

\begin{itemize}
    \item Under \cref{assumption:conditional_density_function_assumption_1}, we have 
\begin{align*}
\int_{\R^{d_0}}\norm{\calT_{\text{score}}(\bar{h},y,t)-\nabla\log{p_{t}^h(\bar{h}|y)}}_{2}^{2}\cdot p_{t}^h(\bar{h}|y) \dd \barh = \calO\left(\frac{B_0^{2}}{\sigma_{t}^{4}}\cdot N^{-\frac{\beta_0}{d_0+d_{y}}}\cdot(\log{N})^{d_0+\frac{\beta_0}{2}+1}\right).
\end{align*}
Notably, for $\epsilon=\calO(N^{-\beta_0})$, the approximation error has the upper bound \blue{$\calO((\log (\frac{1}{\epsilon}))^{d_0}/\sigma_t^4$)}.

    \item Under \cref{assumption:conditional_density_function_assumption_2}, we have
\begin{align*}
\int_{\R^{d_0}}\norm{\calT_{\text{score}}(x,y,t)(\bar{h},y,t)-\nabla\log{p_{t}^h(\bar{h}|y)}}_{2}^{2}\cdot p_{t}^h(\bar{h}|y) \dd \barh = \calO\left(\frac{B_0^{2}}{\sigma_{t}^{2}}\cdot N^{-\frac{2\beta_0}{d_0+d_{y}}}\cdot(\log{N})^{\beta_0+1}\right).
\end{align*}
Notably, for $\epsilon=\calO(N^{-\beta_0})$, the approximation error has the upper bound \blue{$(\log (\frac{1}{\epsilon}))^{\calO(1)}/\sigma_t^2$}.
\end{itemize}

\end{theorem}

\begin{proof}
See \cref{theorem:latent_main1_formal,theorem:latent_main2_formal} for the formal versions \blue{and \cref{sec:appendix_proof_main1,sec:appendix_proof_main2} for proofs.}
\end{proof}

\begin{remark}[Comparing with \cref{thm:Main_1,thm:Main_2_informal}]
Recall $d_x \ge d_0$, and the approximation error bounds are $\tilde{\calO}( \epsilon^{1/(d_{x} + d_{y})}/\sigma_{t}^{2} )$ in \cref{thm:Main_1} and $\tilde{\calO}( \epsilon^{2/(d_{x} + d_{y})}/\sigma_{t}^{2} )$ in \cref{thm:Main_2_informal}.
These results show that the latent conditional DiT achieves better approximation and has the potential to bypass the challenges associated with the high dimensionality of initial data.

\end{remark}

\textbf{Score and Distribution Estimation.}
Based on \cref{theorem:latent_main1_infor}, we derive the score estimation bounds in \cref{thm:latent_holder_est}, and report 
the results for distribution estimation in next theorem.

\begin{theorem}[Distribution Estimation of Latent Conditional DiTs]
\label{thm:lantent_distribution_TV_bound_main}
\blue{Assume $d_0=\Omega( \frac{\log N}{\log \log N})$}.
\blue{For $y\in[0,1]^{d_y}$, 
let $\hat{P}_{t_0}(\cdot | y)$ denote \textit{estimated} conditional distributions at $t_0$.
Recall that $P_0(\cdot|y)$ is the conditional distribution of initial data $x_0$ given $y$.
Assume $\mathrm{KL}\left(P_0(\cdot|y)\mid N(0,I)\right) \le c$ for some constant $c<\infty$.}
\begin{itemize}
    \item Under \cref{assumption:conditional_density_function_assumption_1}, 
    taking $t_0=n^{-\frac{\beta_0}{\blue{(d_{0}+d_{y}+\beta_0)}}}$ and $T=\frac{2\beta_0}{d_{0}+d_{y}+2\beta_0}\log{n}$,it holds
\begin{align*}
    \mathbb{E}_{\{x_i, y_i\}_{i=1}^n} \left[ \mathbb{E}_{y} \left[ {\rm TV} \left( \hat{P}_{t_0}(\cdot | y), P_0(\cdot | y) \right) \right] \right]
    = \calO \left(n^{-\frac{\beta_0}{2(\tilde{\nu}_{1}\blue{-}1)(d_{0}+d_{y}+\beta_0)}}(\log{n})^{\frac{\tilde{\nu}_{2}}{2}+\frac{3}{2}}\right),
\end{align*}
where \blue{$\tilde\nu_{1}=\frac{68\beta_0}{(d_0+d_y)}+104C_{\sigma}$,
$\tilde{\nu}_{2}=12d_0+12\beta_0+2$ 
and $C_{\sigma}=\frac{\beta_0}{d_0+d_y+\beta_0}$.}

    \item Under \cref{assumption:conditional_density_function_assumption_2},
    taking $t_0=n^{-\frac{\beta_0}{4(d_{0}+d_{y}+\beta_0)}}$ and $T=\frac{2\beta_0}{d_{0}+d_{y}+2\beta_0}\log{n}$,it holds
\begin{align*}
    \mathbb{E}_{\{x_i, y_i\}_{i=1}^n} \left[ \mathbb{E}_{y} \left[ {\rm TV} \left( \hat{P}_{t_0}(\cdot | y), P_0(\cdot | y) \right) \right] \right]=
    \calO\left(n^{-\frac{1}{2\tilde{\nu}_{3}}\frac{\beta_0}{d_{0}+d_{y}+2\beta_0}}(\log{n})^{{\max(6,\frac{\beta_0}{2}+\frac{3}{2})}}\right),
\end{align*}
where 
\blue{$\tilde\nu_{3}={\frac{4(12\beta_0 d_{0}+31\beta_0 \tilded+ 6\beta_0)}{\tilde{d}(d_0+d_y)} + \frac{12(12C_{\alpha}d_0 + 25C_{\alpha}\cdot\tilde{d} + 6C_{\alpha})}{\tilde{d}}}
+72C_{\sigma}$ and $C_{\alpha}=\frac{2\beta_0}{d_0+d_y+2\beta_0}$.
}
\end{itemize}
\end{theorem}
\begin{proof}
    \blue{Please see \cref{app:proof_distribution_TV_bound} for a detailed proof.}
\end{proof}

\begin{remark}[Minimax Optimal Estimation]\label{remark:latent_minimax}
Following the same idea in \cref{subsec:mini_op_con_dit}, we show that the estimation error bound in \cref{thm:lantent_distribution_TV_bound_main} is the optimal tight bound for the latent unconditional DiT.
Specifically, by applying \cref{cor:holder_optimal} and substituting $p(x|y)$ and $d_x$ by ${p_{t}^h(\bar{h}|y)}$ and $d_0$ respectively in \cref{assumption:conditional_density_function_assumption_2}, we establish a distribution estimation lower bound of  $\calO(n^{-\beta_0/(d_{0}+2\beta_0)})$. 
Setting $2\tilde{\nu}_{3}=1$, we obtain the minimax optimality of latent unconditional DiT.
\end{remark}

\section{Discussion and Conclusion}
\label{sec:com_ass}

We investigate the approximation and estimation rates of conditional DiT and its latent setting.
We focus on the ``in-context'' conditional DiT setting presented by \citet{peebles2023scalable}, and conduct a comprehensive analysis under various common data conditions (\cref{sec:con_dit} for generic and strong H\"older smooth data, \cref{sec:latent_con_dit} for data with intrinsic latent subspace).

Interestingly, we establish the minimax optimality of the unconditional DiTs' estimation by reducing our analysis of conditional DiTs to the unconditional setting (\cref{subsec:mini_op_con_dit} and \cref{remark:latent_minimax}).
Our key techniques include a well-designed score decomposition scheme (\cref{sec:generic_holder}).
These enable a finer use of transformers’ universal approximation, compared to the prior statistical rates of DiTs derived from the universal approximation results in \cite{yun2019transformers} by \citet{hu2024statistical}.

Consequently, we provide two extensions in the appendix:
\begin{itemize}
    \item 
    In \cref{sec:appendix_latent_dit_holder}, we    expand  \cref{sec:latent_con_dit} and extend our well-designed score decomposition scheme from \cref{sec:con_dit} to the latent conditional DiT.
    Notably, we also obtain provably tight rate, i.e., for distribution estimation under \cref{assumption:conditional_density_function_assumption_2} (\cref{remark:latent_minimax}).

    \item 
    In \cref{sec:appendix_latent_dit_lipschitz}, we extend the analysis of \cite{hu2024statistical} to the conditional DiT setting and provide an improved version.     
    In particular, we analyze conditional latent DiTs under the following three assumptions from \cite{hu2024statistical} and obtained sharper rates:
    \begin{itemize}
        \item Low-Dimensional Linear Latent Space Data  (\cref{assumption:low_dim_linear_latent_space})

        \item Lipschitz Score Function (\cref{assumption:latent_on_support_lipschitz})

        \item Light Tail Data Distribution (\cref{assumption:latent_tail_behavior})
        
    \end{itemize}
    
    \blue{In detail, we use a modified universal approximation of the single-layer self-attention transformers (modified from \cite{kajitsuka2023transformers}) to avoid the need for dense layers required in \cite{yun2019transformers}.}
    This refinement results in tighter error bounds for both score and distribution estimation.
    Consequently, our sample complexity error bounds avoid the gigantic double exponential term $2^{(1/\epsilon)^{2L}}$ reported by \citet{hu2024statistical}, and obtain   sharper rates than those of \cite{hu2024statistical}.
\end{itemize}

\section*{Acknowledgments}
The authors thank the authors of \cite{fu2024diffusion} for clarifications; Jialong Li, Stephen Cheng, Dino Feng, and Andrew Chen for insightful discussions; Yibo Wen, Zhenyu Pan, Damian Jiang, Hong-Yu Chen, Zhao Song, and Sophia Pi for collaborations on related topics; and Jiayi Wang for assisting with experimental deployments. JH also extends gratitude to the Red Maple Family for their support. The authors thank the anonymous reviewers and program chairs for their constructive feedback.

JH is partially supported by the Walter P. Murphy Fellowship.
HL is partially supported by NIH R01LM1372201, AbbVie and Dolby.
This research was supported in part through the computational resources and staff contributions provided for the Quest high performance computing facility at Northwestern University which is jointly supported by the Office of the Provost, the Office for Research, and Northwestern University Information Technology.
The content is solely the responsibility of the authors and does not necessarily represent the official
views of the funding agencies.

\newpage
\appendix

\normalsize

\setlist[itemize]{leftmargin=1em, itemsep=0.5em, parsep=\parskip, before=\vspace{0em}, after=\vspace{0em}}
\setlist[enumerate]{leftmargin=1.4em, itemsep=0.5em, parsep=\parskip, before=\vspace{0em}, after=\vspace{0em}}

\setlength{\abovedisplayskip}{\baselineskip} %
\setlength{\abovedisplayshortskip}{0.5\baselineskip} %
\setlength{\belowdisplayskip}{\baselineskip}
\setlength{\belowdisplayshortskip}{0.5\baselineskip}

\label{sec:append}
\part*{Appendix}
{
\setlength{\parskip}{-0em}
\startcontents[sections]
\printcontents[sections]{ }{1}{}
}

{
\setlength{\parskip}{-0em}
\startcontents[sections]
\printcontents[sections]{ }{1}{}
}

\clearpage

\clearpage
\section{Notation Table}
\begin{table}[h!]
    \vspace{-1.5em}
    \caption{Mathematical Notations and Symbols}
    \centering
    \resizebox{.88 \textwidth}{!}{ 
    \begin{tabular}{ll}
    \toprule
        Symbol & Description \\
    \midrule
        $[I]$ & The index set $\{ 1, ..., I \}$, where $I \in \mathbb{N}^+$ \\
        $a[i]$ & The $i$-th component of vector $a$ \\
        $A_{ij}$ & The $(i,j)$-th entry of matrix $A$ \\
        $\norm{x}$ & Euclidean norm of vector $x$ \\
        $\norm{x}_1$ & 1-norm of vector $x$ \\
        $\norm{x}_2$ & 2-norm of vector $x$\\
        $\norm{x}_{\infty}$ & Infinite norm of vector $x$ \\
        $\norm{W}_2$ & Spectral norm of matrix $W$ \\
        $\norm{W}_F$ & Frobenius norm of matrix $W$ \\
        $\norm{W}_{p,q}$ & $(p,q)$-norm of matrix $W$, where $p$-norm is over columns and $q$-norm is over rows \\
        $\norm{f(x)}_{L^2}$ & $L^2$-norm, where $f$ is a function\\
        $\norm{f(x)}_{L^2(P)}$ & $L^2(P)$-norm, where $f$ is a function and $P$ is a distribution\\
        $\norm{f(\cdot)}_{Lip}$ & Lipschitz-norm, where $f$ is a function\\
        \blue{$d_p(f,g)$} & \blue{$p$-norm of the difference between functions $f$ and $g$ defined as $d_p(f,g) = \qty (\int |f(x) - g(x)|^p \, dx ) ^{{1}/{p}}$} \\
        $f_{\sharp} P$ & Pushforward measure, where $f$ is a function and $P$ is a distribution\\
        ${\rm KL}(P , Q)$ & Kullback-Leibler (KL) divergence between distributions $P$ and $Q$ \\
        ${\rm TV}(P, Q)$ & Total variation (TV) distance between distributions $P$ and $Q$ \\
        $N(\mu, \sigma^2)$ & Normal distribution with mean $\mu$ and variance $\sigma^2$ \\
        \blue{$a \lesssim b$} & \blue{There exist constants $C > 0$ such that $a \le Cb$} \\

    \midrule
        $n$ & Sample size\\
        $x$ & Data point in original data space, $x\in\mathbb{R}^{d_x}$\\
        $y$ & Conditioning Label, $x\in\mathbb{R}^{d_y}$\\
        $h$ & Latent variable in low-dimensional subspace, $h\in\mathbb{R}^{d_0}$\\
        $\Bar{h}$ & $\Bar{h}=U^\top  x $\\
        $p_h$ & The density function of $h$\\
        $U$ & The matrix with orthonormal columns to transform $h$ to $x$, where $U \in \RR^{d\times d_0}$\\
    \midrule
        $B$ & Radius of H\"older ball for conditional density function $p(x|y)$\\
        $B_0$ & Radius of H\"older ball for latent conditional density function $p(\barh|y)$\\
        $\beta$ & H\"older index for conditional density function $p(x|y)$\\
        $\beta_0$ & H\"older index for latent conditional density function $p(\barh|y)$\\
        $D$ & Granularity in the construction of the transformer universal approximation \\
        $N$ & Resolution of the discretization of the input domain\\
        $\calR$ & Score risk (expectation of squared $\ell^2$ difference between score estimator and ground truth)\\
        $\calN(\epsilon,\calF,\norm{\cdot})$ & Covering number of collection $\calF$ (see \cref{def:covering_number})\\
    \midrule
        $T$ & Stopping time in the forward process of diffusion model\\
        $t_0$ & Stopping time in the backward process of diffusion model\\
        $\mu$ & Discretized step size in backward process\\
        $p_t(\cdot)$ & The density function of $x$ at time $t$\\
        $p_t^{h}(\cdot)$ & The density function of $\Bar{h}$ at time $t$\\
        $\psi$ & (Conditional) Gaussian density function\\
    \midrule
        $\calT^{\textcolor{blue}{h,s,r}}$ & Transformer network function class (see \cref{def:transformer_class})\\
        $f^{\textcolor{blue}{h,s,r}}$ & Transformer block of $h$-head, $s$-hidden size, $r$-MLP dimension (see \cref{def:transformer_block})\\
        $d$ & Input dimension of each token in the transformer network of DiT\\
        $L$ & Token length in the transformer network of DiT\\
        $\tilded$ & Latent data input dimension of each token in the transformer network of DiT\\
        $\tildeL$ & Latent data token length in the transformer network of DiT\\
        $X$ & Sequence input of transformer network in DiT, where $X \in \mathbb{R}^{d \times L}$\\
        $H$ & Sequence latent data input of transformer network in DiT, where $X \in \mathbb{R}^{d \times L}$\\
        $E$ & Position encoding, where $E \in \mathbb{R}^{d \times L}$\\
        $R(\cdot)$ & Reshape layer in DiT, $R(\cdot): \mathbb{R}^{d_x} \to \mathbb{R}^{d \times L}$\\
        $\tildeR(\cdot)$ & Reshape layer in DiT, $\tildeR(\cdot): \mathbb{R}^{d_0} \to \mathbb{R}^{\tilded \times \tildeL}$\\
        $R^{-1}(\cdot)$ & Reverse reshape layer in DiT, $R^{-1}(\cdot): \mathbb{R}^{d \times L} \to \mathbb{R}^{d_x}$\\
        $\tildeR^{-1}(\cdot)$ & Reverse reshape layer in DiT, $\tildeR^{-1}(\cdot):  \mathbb{R}^{\tilded \times \tildeL}  \to \mathbb{R}^{d_0}$\\
        $W_U$ & The orthonormal matrix to approximate $U$, where $W_U \in \RR^{d_x\times d_0}$\\
    \bottomrule
    \end{tabular}
    }
    \label{tab:nomenclature}
    \vspace{-4em}
\end{table}

\clearpage
\section{Related Works, Broader Impact and Limitations}
\label{sec:related_broad_limit}
\subsection{Related Works}
\label{sec:related}

In the following, we discuss the recent success of the techniques used in our work.
We first give the universality (universal approximation) of the transformer.
Then, we discuss recent theoretical developments (approximation and estimation) in diffusion generative models.

\paragraph{Universality of Transformers.}
The universality of transformers refers to their ability to approximate any sequence-to-sequence function with arbitrary precision. \citet{yun2019transformers} establish this by showing that transformers is capable of universally approximate sequence-to-sequence functions using deep stacks of feed-forward and self-attention layers. 
Additionally, \citet{alberti2023sumformer} demonstrate universal approximation for architectures employing non-standard attention mechanisms. 
Recently, \citet{kajitsuka2023transformers} show that even a single-layer transformer with self-attention suffices for universal approximation assuming all attention weights are rank-$1$. 
Moreover, \citet{hu2024statistical} leverage \citet{yun2019transformers} universality results to analyze the approximation and estimation capabilities of DiT.

Our paper is motivated by and builds upon the works of \citet{hu2024statistical,kajitsuka2023transformers,yun2019transformers}. 
Specifically, we utilize and extend the transformer universality result from \citet{kajitsuka2023transformers}. 
We employ a relaxed contextual mapping property in \citet{kajitsuka2023transformers} (see \cref{sec:trans_univeral_approx}). 
This generalization allows us to avoid the ``double exponential'' sample complexity bounds in previous DiT analyses \cite[Remark~3.4]{hu2024statistical} and establish transformer approximation in the simplest configuration — a single-layer, single-head attention model.

\paragraph{Approximation and Estimation Theories of Diffusion Models.}
The theories of DiTs revolve around two main frontiers: score function approximation and statistical estimation \cite{chen2024overview, tang2024score}.
Score function approximation refers to the ability of the score network to approximate the score function.
It leverages the universal approximation ability of the neural network in $L^p$ norms \citep{Hayakawa2020minimax}, the approximation characterized as Taylor polynomial \citep{fu2024diffusion} or B-Spline \citep{oko2023diffusion}.
\citet{chen2023score} and \citet{fu2024diffusion} investigate score approximation under specific conditions, such as low-dimensional linear subspaces and H\"{o}lder smooth data assumptions, using ReLU-based models. 
Furthermore, \citet{hu2024statistical} presents the first characterization of score approximation in diffusion transformers (DiTs).

The statistical estimation includes score function and distribution estimation \cite{wu2024theoretical,dou2024theory, guo2024gradient, chen2023score}.
Under a $L_2$ accurate score estimation, several works have provided the convergence bounds
under either smoothness assumptions \citep{benton2024nearly,chen2022sampling} or bounded second-order moment assumptions \citep{chen2023improved,lee2023convergence}.
\citet{chen2023score} provide the first complete estimation theory using ReLU networks without precise estimators.
\citet{oko2023diffusion} achieve nearly minimax optimal estimation rates for total variation and Wasserstein distances. 
Meanwhile, \citet{dou2024optimal} define exact minimax optimality using kernel functions without characterizing the network architectures. 
In the realm of diffusion transformers, \citet{hu2024statistical} introduces the first complete estimation theory.
\citet{jiao2024latent,jiao2024convergence} demonstrate theoretical convergence for latent DiTs using ODE-based and Schr\"{o}dinger bridge diffusion models.\footnote{Of independent interest, many works investigate the convergence rates of diffusion models under various score and data smoothness assumptions or with different samplers.
Please see \cite{li2024accelerating,li2024towards,li2024sharp,potaptchik2024linear,wu2024stochastic,liang2024non,liang2024unraveling,gatmiry2024learning,gu2024exploring,guo2024gaussian,chen2024probability,chen2023improved,chen2022sampling,lee2023convergence,lee2022convergence} and references therein.}

Our paper advances the foundational works of \citet{fu2024unveil, oko2023diffusion, hu2024statistical}. 
We adopt the H\"{o}lder smooth data distribution assumption\footnote{Recent work by \citet{havrilla2024understanding} examines the generalization and approximation of transformers under Hölder smoothness and low-dimensional subspace assumptions.}, a more practical approach than the bounded support assumption in \citet{oko2023diffusion}. 
Unlike the simple ReLU networks in \citet{fu2024unveil}, we provide a complete approximation and estimation analysis for conditional DiTs and establish their exact minimax optimality. 
Furthermore, while \citet{hu2024statistical} analyze DiTs, their estimation upper bounds are suboptimal. 
We refine this by avoiding the substantial double exponential term $2^{(1/\epsilon)^{2L}}$ reported by \citet[Remark 3.4]{hu2024statistical} and present a provably tight, minimax optimal estimation.

\subsection{Broader Impact}
This theoretical work aims to shed light on the foundations of generative diffusion models and is not expected to have negative social impacts.

\subsection{Limitations}
\label{subsec:limit}
Although our study provides a complete theoretical analysis of the conditional DiTs and establishes the minimax optimality of the unconditional DiT, we acknowledge three main limitations:
\begin{itemize}
    \item 
    The minimax optimality of conditional DiT remains not clear.

    \item 
    We did not explore other architectures such as ``adaptive layer norm'' and ``cross-attention'' DiT. 
    A potential direction is by establishing the universal approximation capacity of the transformer with cross-attention mechanisms. 

    \item 
    Although we achieve a better bound for the latent conditional DiT under the Lipschitz assumption than under the H\"{o}lder assumption, we do not show the minimax optimality under the Lipschitz assumption.
\end{itemize}
We leave these for future work. 

\blue{Furthermore, there are limitations regarding the H\"{o}lder smooth data assumptions in \cref{assumption:conditional_density_function_assumption_1} and \cref{assumption:conditional_density_function_assumption_2}.
Our results in \cref{sec:con_dit} and \cref{sec:latent_con_dit} depend on the H\"{o}lder smooth data assumptions.
However, it is challenging to measure the smoothness of a given dataset (e.g., CIFAR10), because it requires knowledge of the dataset's exact distribution. 
Conversely, it is feasible to create a dataset with a predefined level of smoothness. 
To illustrate this, we provide two examples:}
\begin{itemize}
    \item \blue{Diffusion Models in Image Generation: When modeling conditional distributions of images given attributes (e.g., generating images based on class labels), these assumptions hold if the data distribution around these attributes is smooth and decays. 
    In diffusion-based generative models, the data distribution often decays smoothly in high-dimensional space. The assumption that the density function decays exponentially reflects the natural behavior of image data, where pixels or features far from a central region or manifold are less likely. This is commonly observed in images with blank boundaries.}

    \item \blue{Physical Systems with Gaussian-Like Decay: This applies to cases where the spatial distribution of a physical quantity, such as temperature, is smooth and governed by diffusion equations with exponential decay. 
    In physics-based diffusion models, like those simulating the spread of particles or heat in a medium (e.g., stars in galaxies for astrophysics applications), the conditional density typically decays exponentially with distance from a central region.}

\end{itemize}

\clearpage
\section{Latent Conditional DiT with H\"{o}lder Assumption}\label{sec:appendix_latent_dit_holder}
In this section, we extend the results \blue{on approximation and estimation of DiT} from \cref{sec:con_dit} by considering the latent conditional DiTs. 
\blue{Latent DiTs enables efficient data generation from latent space 
and therefore scales better in terms of spatial dimensionality \cite{rombach2022high}.}
Specifically, we assume the raw input $x \in \R^{d_x}$ has an intrinsic lower-dimensional representation in a $d_0$-dimensional subspace, where $d_0 \leq d_x$.
This setting is common in both empirical \cite{peebles2022scalable,rombach2022high} and theoretical studies \cite{hu2024statistical,chen2023score}.

\textbf{Organization.}
We present the statistical results under H\"older data smooth \cref{assumption:conditional_density_function_assumption_1,assumption:conditional_density_function_assumption_2} and state the results in \cref{theorem:latent_main1_formal}, \cref{theorem:latent_main2_formal}, \cref{thm:latent_holder_est}, and \cref{thm:lantent_distribution_TV_bound}, respectively.
\cref{sec:appendix_latent_score_approx_dit_holder} discusses score approximation.
\cref{sec:appendix_latent_score_est_dit_holder}
discusses score estimation.
\cref{sec:appendix_latent_dist_est_dit_holder} discusses distribution estimation.
The proofs in this section primarily follow \cref{sec:appendix_proof_main1,sec:appendix_proof_main2}.

Let $d_0$ denote the latent dimension.
We summarize the  key points of this section \blue{as follows}:

\begin{itemize}[leftmargin=2.0em]
    \item [\textbf{K1.}]\label{item:K1} \textbf{Low-Dimensional Subspace Space Data Assumption.} 
    We consider the setting that latent representation lives in a ``Low-Dimensional Subspace'' under \cref{assumption:low_dim_linear_latent_space}, following \cite{hu2024statistical,chen2023score}.
    \begin{assumption}[Low-Dimensional Linear Latent Space (\cref{assumption:low_dim_linear_latent_space} Restated)]
    Data point $x = U h$, where $U \in \mathbb{R}^{d_x \times d_0}$ is an unknown matrix with orthonormal columns. 
    The latent variable $h \in \mathbb{R}^{d_0}$ follows a distribution $P_h$ with a density function $p_h$.
    \end{assumption}
    
    For raw data $x\in\R^{d_x}$, we utilize linear encoder $W_U^\top\in\R^{d_0\times d_x}$ and decoder $W_U\in \R^{d_x\times d_0}$ to convert the raw $x\in\R^{d_x}$ and latent $h\in\R^{d_0}$ data representations.
    Importantly, $x=Uh$ with $U\in \R^{d_x\times d_0}$ by \cref{assumption:low_dim_linear_latent_space}.
    
    For each input $x \in \R^{d_x}$ and corresponding label $y \in \R^{d_y}$, we use a transformer network to obtain a score estimator $s_W \in \R^{d_x}$.
    To utilize the transformer network as the score estimator, 
    we introduce reshape layer to convert vector input $h\in\R^{d_0}$ to matrix (sequence) input $H\in\R^{\tilde{d}\times \tilde{L}}$.
    Specifically,    
    the reshape layer in the network \cref{fig:Latent_DiT} is defined as $\tilde{R}(\cdot): \mathbb{R}^{d_0} \to \mathbb{R}^{\tilde{d} \times \tilde{L}}$ and its reverse $\tilde{R}^{-1} (\cdot): \mathbb{R}^{\tilde{d} \times \tilde{L}} \to \mathbb{R}^{d_0}$, where $d_0 \le d_x$, $\tilde{d}\le d$, and $\tilde{L} \le L$.
    
    We remark that the ``low-dimensional data'' assumption leads to tighter approximation rates than those of \cref{sec:generic_holder,sec:bounded_holder} and estimation errors due to  $d_0\le d_x$ (\cref{theorem:latent_main1_formal,theorem:latent_main2_formal}).
    
    \item [\textbf{K2.}]\label{item:K2} \textbf{H\"older Smooth Assumption.} 
    For approximation and estimation results for latent conditional DiTs (\cref{theorem:latent_main1_formal,theorem:latent_main2_formal,thm:latent_holder_est,thm:lantent_distribution_TV_bound}), 
    we study the cases under both the generic and strong  H\"older smoothness assumptions on latent representation $h \in \R^{d_0}$. 
    Specifically, we assume the ``latent'' data is $\beta_0$-H\"older smooth with radius $B_0$ following \cref{assumption:conditional_density_function_assumption_1,assumption:conditional_density_function_assumption_2}.
    We extend both approximation and estimation results from \cref{sec:con_dit} to latent diffusion and establish the minimax optimality of latent conditional DiTs.

    \begin{assumption}[Generic H\"{o}lder Smooth Data (\cref{assumption:conditional_density_function_assumption_1} Restated)]
    The conditional density function $p_0^h(h_0|y)$ is defined on the domain $\R^{d_{0}}\times[0,1]^{d_{y}}$ and belongs to H\"{o}lder ball of radius $B_0>0$ for H\"{o}lder index $\beta_0>0,$
    denoted by $p_0^h(h_0|y)\in\calH^{\beta_0}(\R^{d_{0}}\times[0,1]^{d_{y}},B_0)$ (see \cref{def:holder_norm_space} for precise definition.)
    Also, for any $y\in[0,1]^{d_{y}}$, there exist positive constants $C_{1}, C_{2}$ such that $p_0^h(h_0|y)\leq C_{1}\exp(-C_{2}\norm{h_0}_{2}^{2}/2)$. 
    \end{assumption}

    \begin{assumption}[Stronger H\"{o}lder Smooth Data  (\cref{assumption:conditional_density_function_assumption_2} Restated)]
    Let function $f\in\calH^{\beta_0}(\R^{d_{0}}\times[0,1]^{d_{y}},B_0)$.
    Given a constant radius $B_0$, positive constants $C$ and $C_{2}$, we assume the conditional density function $p(h_{0}|y)=\exp(-C_{2}\norm{h_{0}}_{2}^{2}/2)\cdot f(h_{0},y)$ and $f(h_{0},y)\geq C$ for all $(h_{0},y) \in \R^{d_{0}}\times[0,1]^{d_{y}}$.
    \end{assumption}

    \item [\textbf{K3.}]\label{item:K3} \textbf{Latent Score Network.} Under low-dimensional data assumption, we decompose the score function following \cite{hu2024statistical,chen2023score} (see \cref{lemma:subspace_score}):
    \begin{align}
    \label{eqn:score_docom_rearange_appendix}
    \nabla \log p_t(x | y) = U ( \underbrace{\sigma_{t}^2\nabla \log p_t^{h} ( U^\top x | y)
    +U^{\top}x}_{\coloneqq q(U^\top x, y,t):\;\R^{d_0} \times [t_0, T] \;\to\; \R^{d_0}})/\sigma_{t}^2 -  \underbrace{x/\sigma_{t}^2}_{\text{residual connection}}.
    \end{align}
    Based on this decomposition, we construct the model architecture in \cref{fig:Latent_DiT}.
    The network detail for approximate \eqref{eqn:score_docom_rearange_appendix} are as follow: a transformer $g_{\calT}(W_U^\top x, y, t) \in \calT^{\textcolor{blue}{h,s,r}}$ to approximate $q(U^\top x, y,t)$, a latent encoder $W_U^{\top}\in \R^{d_0\times d_x}$ and decoder $W_U\in \R^{d_x\times d_0}$ to approximate $U^{\top} \in \R^{d_0\times d_x}$ and $U\in \R^{d_x\times d_0}$, and a residual connection to approximate $-x/\sigma_{t}^2$.

    \end{itemize}

We adopt the following transformer network class of one-layer single-head self-attention
\begin{align}\label{eqn:score_network_S_formal}
    \textcolor{blue}{{\calT^{h,s,r}_{\tildeR}}} = \bigg\{ s_{W} \left( x,y,t \right) = 
    \dfrac{1}{\sigma_{t}^2} {W_U} 
    {g_{\calT} \left( W_U^{\top} x, y, t \right)}  - 
    \underbrace{\dfrac{1}{\sigma_{t}^2}x}_{\text{residual connection}} \bigg\},
\end{align}
where $g_{\calT}\in\calT^{\textcolor{blue}{h,s,r}} = 
\{f^{\text{FF}}_{2}
\circ f^{\textcolor{blue}{h,s,r}}:
\R^{\tilde{d}\times \tilde{L}}\rightarrow\R^{\tilde{d}\times \tilde{L}}\}$.

Let $h\coloneqq W_U^\top x \in\R^{d_0}$ and $\bar{h}\coloneqq U^\top x\in \R^{d_0}$ be the estimated and ground truth (according to \cref{assumption:low_dim_linear_latent_space}) latent representations, respectively.
Here we construct a network $s_{W} \left( x, y, t \right)$ to approximate the score function in \eqref{eqn:score_docom_rearange_appendix} (see \cref{fig:Latent_DiT} for network illustration).

In \cref{sec:con_dit}, we derive the approximation theory of conditional DiTs using a one-layer, single-head self-attention transformer to approximate the score function $\nabla \log p_t (x | y)$. 
Here, we use the similar transformer architecture to approximate latent score function $\nabla \log p_t^{h} (\Bar{h} | y)$, where $p_t^{h}(\Bar{h} | y) = \int \psi_t (\Bar{h} | h) p_h (h | y) \dd h$, $\psi_t( \cdot | h)$ is the Gaussian density function of $N (\beta_t h, \sigma_t^2 I_{d_0})$, $\beta_t = e^{-t/2}$, and $\sigma_t^2 = 1 - e^{-t}$.

Base on the latent network construction in \hyperref[item:K3]{(K3)}, we employ the same techniques presented in \cref{sec:con_dit} for score function approximation and estimation.
We restate for completeness.
First, we decompose the conditional score function $\nabla \log p_t^h(\Bar{h} | y)$ as following:
\begin{align}\label{eqn:score_sbb}
\nabla \log p_t^h\left(\Bar{h} | y \right) = \frac{\nabla  p_t^h\left(\Bar{h} | y \right)}{p_t^h\left(\Bar{h} | y \right)}. 
\end{align}
By the definition of Gaussian kernel, we have
\begin{align*}
   p_t^h \left( \Bar{h} | y \right) = \int_{\R^{d_0}} 
   ( 2 \pi \sigma_{t}^2 )^{-d_x/2} 
   \underbrace{
   p_h \left( h | y \right)
   }_{\approx \textcolor{blue}{k_1}\text{-order Taylor polynomial}}
   \underbrace{
   \exp\left(-\frac{\norm{\beta_{t}h - \Bar{h}}_2^2}{2\sigma_{t}^2} \right)
   }_{\approx \textcolor{blue}{k_2}\text{-order Taylor polynomial}}
   \dd h. 
\end{align*}

Similar to \cref{sec:con_dit}, our strategy is to expand above term-by-term with \blue{$k_1$- and $k_2$-order} Taylor polynomials for fine-grained characterizations.

\begin{remark}
    Here in the latent density function, we have $ ( 2 \pi \sigma_{t}^2 )^{-d_x/2}$ instead of $ ( 2 \pi \sigma_{t}^2 )^{-d_0/2}$.
    However, the additional $( 2 \pi \sigma_{t}^2 )^{-(d_x-d_0)/2}$ term does not affect the application of \cref{sec:con_dit} into latent diffusion approximation.
\end{remark}

Based on the low-dimensional data structure assumption, we have the following score decomposition \blue{terms}: on-support score $s_+(U^\top x, y, t)$ and orthogonal score $s_-(x, y, t)$.

\begin{lemma}[Score Decomposition, Lemma 1 of \cite{chen2023score}]
\label{lemma:subspace_score}
Let data $x = Uh$ follow \cref{assumption:low_dim_linear_latent_space}. 
The decomposition of score function $\nabla \log p_t(x)$ is
\begin{align}\label{eqn:score_decom}
~\nabla \log p_t(x) = \underbrace{U\nabla \log p_t^{h}(\Bar{h} | y)}_{s_+(\Bar{h}, y, t)}   \underbrace{-  \left(I_D - UU^\top \right) x/\sigma_{t}^2}_{s_-(x, t)}, ~\Bar{h}=U^\top x,
\end{align}
where
$p_t^{h}\left(\Bar{h} | y \right) \coloneqq  \int  \psi_t(\Bar{h}|h)p_h\left(h | y \right) \dd h$, 
$\psi_t( \cdot | h)$ is the Gaussian density function of $N(\beta_{t}h, \sigma_{t}^2I_{d_0})$, $\beta_{t} = e^{-t/2}$ and $\sigma_{t}^2 = 1 - e^{-t}$.
\end{lemma}

Following the proof strategy of conditional DiTs in \cref{sec:appendix_proof_main1,sec:appendix_proof_main2} with differences highlighted in \hyperref[item:K1]{(K1)}, \hyperref[item:K2]{(K2)}, and the latent network in \hyperref[item:K3]{(K3)}.
To derive the approximation and estimation under generic and stronger \holder assumptions results in \cref{thm:Main_1,thm:Main_2_informal,thm:main_risk_bounds,thm:distribution_TV_bound} for data under low-dimensional data assumption, we
just need to replace the input dimension $d$, $L$ to $\tilded$ and $\tildeL$, and the input dimension $d_x$ with $d_0$, and consider the $\beta_0$-\holder smoothness assumption on latent data.

To begin, we clarify the relation between initial data admits to $p(x|y)\in\calH^{\beta}(\R^{d_{x}}\times[0,1]^{d_{y}},B)$, and under linear transformed data \cref{assumption:low_dim_linear_latent_space} admits to $p(\bar{h}|y)\in\calH^{\beta_0}(\R^{d_{0}}\times[0,1]^{d_{y}},B_0)$ where $\beta_0=\beta$ and $B_0\le \tilde{C}B$ by \cref{lemma:holder_smooth_transformation}.

\begin{lemma}[Transformation of Stronger \holder Smooth Data Distribution under Linear Mapping]
\label{lemma:holder_smooth_transformation}
Let $f \in H^\beta (\mathbb{R}^{d_x} \times [0,1]^{d_y}, B)$ satisfy $f(x,y) \geq C > 0$ for all $(x,y) \in \mathbb{R}^{d_x} \times [0,1]^{d_y}$. Consider the conditional density function:
\begin{align*}
p(x|y) = f(x,y) \exp\left( - \frac{C_2}{2} \| x \|_2^2 \right).
\end{align*}
Suppose the data undergo the linear transformation $x = Uh$, where $U \in \mathbb{R}^{d_x \times d_0}$ has orthonormal columns ($U^\top U = I_{d_0}$) and $f_0(h|y)=f(Uh|y)$. The transformed density $p(h|y)$ becomes:
\begin{align*}
p(h|y) = f(Uh, y) \exp\left( - \frac{C_2}{2} \| h \|_2^2 \right).
\end{align*}
The following condition holds for H\"{o}lder smooth data undergo linear transformation: $f_0 \in H^\beta (\mathbb{R}^{d_x} \times [0,1]^{d_y}, B_0)$ with $B_0 \le \tilde{C}B$, where $\tilde{C}=\max \{ C^{\prime}, C^{\prime\prime} \}$.

\end{lemma}
\begin{proof}
First, we compute the partial derivative of the transformed function $f_0(h|y) \coloneqq f(Uh|y)$.
From the definition of H\"{o}lder space \cref{def:holder_norm_space}, and let $\alpha=(\alpha_h,\alpha_y)$ where $\alpha_h+\alpha_y \le \textcolor{blue}{k_1}$.
We compute the partial derivative up to the order of \blue{$k_1$} and show that it is bounded by some $C^{\prime}$, that is
\begin{align*}
    \partial^{\alpha_h}_{h} \partial^{\alpha_y}_{y}p(h|y)
    = & ~ 
    \partial^{\alpha_h}_{h} \partial^{\alpha_y}_{y} \left[ f(Uh,y) \exp\left( - \frac{C_2}{2} \| h \|_2^2 \right) \right]\\
    = & ~  
    \sum_{\alpha \le \nu} \binom{\alpha}{\mu} \left(\partial^{\alpha_\mu}_{h}f(Uh,y) \right) \left(\partial^{(\alpha-\nu)}_h \exp\left( - \frac{C_2}{2} \| h \|_2^2 \right) \right). \annot{By product rule}
\end{align*}
From the relation $\partial^{\alpha_h}_{h}f(Uh,y)=U^{\alpha_h}\partial^{\alpha_h}_xf(Uh,y)$ where $U^{\alpha_h}$ is the product of $U$ entries correspond to $\alpha_h$.
Therefore, $\norm{\partial^{\alpha_h}_{h} \partial^{\alpha_y}_{y}f_0(h|y)} \le C^{\prime}B$ for some $C^{\prime}$ depends on $U$ and $\alpha_h$.
Since $f$ satisfied H\"{o}lder condition and the mapping $h \mapsto Uh$ is linear, for H\"{o}lder condition $|\alpha_h|+|\alpha_y|=\textcolor{blue}{k_1}$ there exist $C^{\prime\prime}$ such that
\begin{align*}
\frac{\left| \partial^{\alpha_h}_{h} \partial^{\alpha_y}_{y}f_0(h|y) - \partial^{\alpha_h}_{h} \partial^{\alpha_y}_{y}f_0(h^\prime|y^\prime) \right|}{\|(h,y) - (h^\prime,y^\prime)\|_{\infty}^{\gamma}}
\le C^{\prime\prime}B.
\end{align*}
The bounded partial derivate up to order $\textcolor{blue}{k_1}$ satisfied H\"{o}lder condition.

This completes the proof.
\end{proof}

\subsection{Score Approximation}\label{sec:appendix_latent_score_approx_dit_holder}

We present the approximation rate of latent score function under generic \holder and stronger \holder data assumption in \cref{theorem:latent_main1_formal,theorem:latent_main2_formal}, respectively. 

\begin{theorem}[Latent Conditional DiT Score Approximation (Formal Version of \cref{theorem:latent_main1_infor})]
\label{theorem:latent_main1_formal}
Assume \cref{assumption:conditional_density_function_assumption_1} and \blue{Assume $d_x=\Omega( \frac{\log N}{\log \log N})$}.
For any precision  $0 < \epsilon < 1$ and smoothness  $\beta_0 > 0$, let $\epsilon \le \calO(N^{-\beta_0})$ for some $N \in \mathbb{N}$.
For some positive constants $C_{\alpha},C_{\sigma}>0$, for any $y \in [0,1]^{d_{y}}$ and $t \in [N^{-C_{\sigma}}, C_{\alpha} \log N]$, there exists a $\calT_{\text{score}}(x,y,t)\in\calT_{\tildeR}^{\textcolor{blue}{h,s,r}}$ 
such that
\begin{align*}
\int_{\R^{d_0}}\norm{\calT_{\text{score}}(\bar{h},y,t)-\nabla\log{p_{t}^h(\bar{h}|y)}}_{2}^{2}\cdot p_{t}^h(\bar{h}|y) \dd \barh = \calO\left(\frac{B_0^{2}}{\sigma_{t}^{4}}\cdot N^{-\frac{\beta_0}{d_0+d_{y}}}\cdot(\log{N})^{d_0+\frac{\beta_0}{2}+1}\right).
\end{align*}
Notably, for $\epsilon=\calO(N^{-\beta_0})$, the approximation error has the upper bound \blue{$\calO((\log(\frac{1}{\epsilon}))^{d_0}/\sigma_t^4)$}.

The parameter bounds for the transformer network class are as follows:
\begin{align*}
& \norm{W_{Q}}_{2}, \norm{W_{K}}_{2},
\norm{W_{Q}}_{2,\infty}, \norm{W_{K}}_{2,\infty}
=
\calO\left(N^{\frac{7\beta_0}{d_0+d_y}+6C_{\sigma}}\right); \\
&
\norm{W_{O}}_{2},\norm{W_{O}}_{2,\infty}=\calO\left(N^{-\frac{3\beta_0}{d_0+d_y}+6C_{\sigma}}(\log{N})^{3(d_0+\beta_0)}\right);\\
&
\norm{W_{V}}_{2}=\calO(\sqrt{\tilded});
\quad \norm{W_{V}}_{2,\infty}=\calO (\tilded);\\ 
& \norm{W_{1}}_{2}, \norm{W_{1}}_{2,\infty}
=
\calO\left(N^{\frac{2\beta_0}{d_0+d_y}+4C_{\sigma}}\right);
\norm{E^{\top}}_{2,\infty}=\calO\left(\tilded^{\frac{1}{2}}\tildeL^{\frac{3}{2}}\right);
\\
& \norm{W_{2}}_{2}, \norm{W_{2}}_{2,\infty}
=
\calO\left(N^{\frac{3\beta_0}{d_0+d_y}+2C_{\sigma}}\right);
C_\calT=\calO\left(\sqrt{\log{N}}/\sigma_{t}^{2}\right).
\end{align*}
\end{theorem}
\begin{proof}[Proof Sketch]
The proof closely follows \cref{thm:Main_1}, with differences highlighted in \hyperref[item:K1]{(K1)} and \hyperref[item:K2]{(K2)}. 
By replacing the input dimension $d$, $L$ to $\tilded$ and $\tildeL$, and the input dimension $d_x$ with $d_0$ in \cref{thm:Main_1}, and under the the $\beta_0$-\holder smoothness assumption on latent data detailed in \hyperref[item:K2]{(K2)}, the proof is complete. 
Please see \cref{sec:appendix_proof_main1} for a detailed proof.
\end{proof}

\begin{theorem}[Latent Conditional DiT Score Approximation under Stronger H\"older Assumption under Generic H\"older Assumption 
 (Formal Version of \cref{theorem:latent_main1_infor})]
\label{theorem:latent_main2_formal}
Assume \cref{assumption:conditional_density_function_assumption_2} and \blue{Assume $d_x=\Omega( \frac{\log N}{\log \log N})$}.
For any precision  $0 < \epsilon < 1$ and smoothness  $\beta_0 > 0$, let $\epsilon \le \calO(N^{-\beta_0})$ for some $N \in \mathbb{N}$.
For some positive constants $C_{\alpha},C_{\sigma}>0$, for any $y \in [0,1]^{d_{y}}$ and $t \in [N^{-C_{\sigma}}, C_{\alpha} \log N]$, there exists a $\calT_{\text{score}}(x,y,t)\in\calT_{\tildeR}^{\textcolor{blue}{h,s,r}}$ 
such that
\begin{align*}
\int_{\R^{d_0}}\norm{\calT_{\text{score}}(x,y,t)(\bar{h},y,t)-\nabla\log{p_{t}^h(\bar{h}|y)}}_{2}^{2}\cdot p_{t}^h(\bar{h}|y) \dd \barh = \calO\left(\frac{B_0^{2}}{\sigma_{t}^{2}}\cdot N^{-\frac{2\beta_0}{d_0+d_{y}}}\cdot(\log{N})^{\beta_0+1}\right).
\end{align*}
Notably, for $\epsilon=\calO(N^{-\beta_0})$, the approximation error has the upper bound 
\blue{$(\log (\frac{1}{\epsilon}))^{\calO(1)}/\sigma_t^2$}.

The parameter bounds in the transformer network class satisfy
\begin{align*}
& \norm{W_{Q}}_{2},\norm{W_{K}}_{2}, \norm{W_{Q}}_{2,\infty}, \norm{W_{K}}_{2,\infty}
=\calO\left(N^{\frac{3\beta_0(2d_0+4\tilded+1)}{\tilded(d_0+d_y)}+\frac{9C_{\alpha}(2d_0+4\tilded+1)}{\tilded}}\right);\\
&
\norm{W_{V}}_{2}=\calO(\sqrt{\tilded}); \norm{W_{V}}_{2,\infty}=\calO(\tilded); \norm{W_{O}}_{2},\norm{W_{O}}_{2,\infty}=\calO\left(N^{-\frac{\beta_0}{d_0+d_y}}\right);\\  
& \norm{W_{1}}_{2}, \norm{W_{1}}_{2,\infty}
=
\calO\left(N^{\frac{4\beta_0}{d_0+d_y}+9C_{\sigma}+\frac{3C_{\alpha}}{2}}\cdot\log{N}\right);
\norm{E^{\top}}_{2,\infty}=\calO\left(\tilded^{\frac{1}{2}}\tildeL^{\frac{3}{2}}\right);\\
& 
\norm{W_2}_{2}, \norm{W_{2}}_{2,\infty}
=
\calO\left(N^{\frac{4\beta_0}{d_0+d_y}+9C_{\sigma}+\frac{3C_{\alpha}}{2}}\right);
C_{\calT}=\calO\left(\sqrt{\log{N}}/\sigma_{t}\right).
\end{align*}
\end{theorem}
\begin{proof}[Proof Sketch]
The proof closely follows \cref{thm:Main_2}, with differences highlighted in \hyperref[item:K1]{(K1)} and \hyperref[item:K2]{(K2)}. 
By replacing the input dimension $d$, $L$ to $\tilded$ and $\tildeL$, and the input dimension $d_x$ with $d_0$ in \cref{thm:Main_2}, and under the the $\beta_0$-\holder smoothness assumption on latent data detailed in \hyperref[item:K2]{(K2)}, the proof is complete. 
Please see \cref{sec:appendix_proof_main2} for a detailed proof.
\end{proof}

\begin{remark}[Score Approximation for Low-Dimensional Linear Latent Space]
With the assumption of low-dimensional latent space \cref{assumption:low_dim_linear_latent_space}, \cref{theorem:latent_main1_formal,theorem:latent_main2_formal} provide better approximation rates than \cref{thm:Main_1,thm:Main_2_informal} under H\"{o}lder smooth assumptions in \cref{assumption:conditional_density_function_assumption_1,assumption:conditional_density_function_assumption_2}, respectively.
Specifically, from \cref{lemma:holder_smooth_transformation} we have $\beta_0=\beta$ and $B_0 \lesssim B$. 
Therefore, \cref{theorem:latent_main1_formal,theorem:latent_main2_formal} deliver $\calO \left( N^{2\beta \left( \frac{d_x-d_0}{(d_0+d_y)(d_x+d_y)} \right)} \right)$ better approximation error over \cref{thm:Main_1,thm:Main_1}, where $d_0 \leq d_x$.
\end{remark}

\subsection{Score Estimation}\label{sec:appendix_latent_score_est_dit_holder}

In this section, we provide the extended results for \cref{sec:score_est_dist_est} on score estimation with the estimator $\calT_{\text{score}}$.
We state the main results under H\"older data assumptions in \cref{thm:latent_holder_est}.

\begin{theorem}[Conditional Score Estimation of Latent DiT]
\label{thm:latent_holder_est}
\blue{Assume $d_x=\Omega( \frac{\log N}{\log \log N})$.}
Let $\hat{s}$ denote the score estimator trained with a set of finite samples $\{x_i,y_i\}_{i\in[n]}$ by optimizing the empirical loss \eqref{eqn:empirical_loss}, and $\calR$ denote the conditional score risk defined in \cref{def:risk}.
\begin{itemize}[leftmargin=.8em]
    \item Under \cref{assumption:conditional_density_function_assumption_1}, by taking $N=n^{\frac{1}{\tilde{\nu}_{1}}\cdot \frac{d_{0}+d_{y}}{\beta_0+d_{0}+d_{y}}},$ $t_0=N^{-C_{\sigma}}<1$ and $T=C_{\alpha}\log{n}$, it holds
    \begin{align*}
    \E_{\{x_{i},y_{i}\}_{i=1}^{n}}\left[\calR(\hat{s})\right]=
    \calO\left(\frac{1}{t_{0}}n^{-\frac{\beta_0}{\tilde{\nu}_{1}(d_{0}+d_{y}+\beta_0)}}(\log{n})^{\tilde{\nu}_{2}+2}\right),
    \end{align*}
    where \blue{$\tilde{\nu}_{1}=68\beta_0/(d_0+d_y)+104C_{\sigma}$
    and
    $\tilde{\nu}_{2}=12d_0+12\beta_0+2$.}

    \item Under \cref{assumption:conditional_density_function_assumption_2},
    by taking $N=n^{\frac{1}{\tilde{\nu}_{3}}\cdot\frac{d_{0}+d_{y}}{2\beta_0+d_{0}+d_{y}}},$ $t_0=N^{-C_{\sigma}}<1$ and $T=C_{\alpha}\log{n}$, it holds
    \begin{align*}
    \E_{\{x_{i},y_{i}\}_{i=1}^{n}}\left[\calR(\hat{s})\right]=
    \calO\left(\log{\frac{1}{t_{0}}}n^{-\frac{1}{\tilde{\nu}_{3}}\frac{2\beta_0}{d_{0}+d_{y}+2\beta_0}}(\log{n})^{{\max(10,\beta_0+1})}\right),
    \end{align*}
    where 
    \blue{$\tilde{\nu}_{3}={\frac{4(12\beta_0 d_{0}+31\beta_0 \tilded+ 6\beta_0)}{\tilded(d_0+d_y)} + \frac{12(12C_{\alpha}d_0 + 25C_{\alpha}\cdot \tilded + 6C_{\alpha})}{\tilded}}
    +72C_{\sigma}$}.

\end{itemize}
\end{theorem}
\begin{proof}[Proof Sketch]
    The proof closely follows \cref{thm:main_risk_bounds}, with differences highlighted in \hyperref[item:K1]{(K1)} and \hyperref[item:K2]{(K2)}. 
    By replacing the input dimension $d$, $L$ to $\tilded$ and $\tildeL$, and the input dimension $d_x$ with $d_0$ in \cref{thm:main_risk_bounds}, and under the the $\beta_0$-\holder smoothness assumption on latent data detailed in \hyperref[item:K2]{(K2)}, the proof is complete. 
    Please see \cref{app:proof_main_risk_bounds} for a detailed proof.
\end{proof}

Next,
we present the score estimation result for low-dimensional input data.

\begin{corollary}[Low-Dimensional Input Region]\label{cor:low_dim_latent_score_est}
\blue{
Assume $d_0 = o\left(\frac{\log N}{\log \log N}\right)$ , i.e., $d_0\ll n$. 
Under \cref{assumption:conditional_density_function_assumption_1}, by setting $N, t_0, T$ as specified in \cref{thm:latent_holder_est}, we have
$\E_{\{x_i, y_i\}_{i=1}^{n}}\left[\calR(\hat{s})\right] = \calO\left(\frac{1}{t_0} n^{-\frac{1}{\tilde{\nu}_4} \cdot \frac{\beta_0}{d_0 + d_y + \beta_0}}\right)$,
where
$\tilde{\nu}_4=\frac{72\beta_0(2d_0+5\tilded+1)}{\tilded(d_0+d_y)} + \frac{48C_{\sigma}(2d_0+5\tilded+1)}{\tilded} - 4\beta_0$.
}
\end{corollary}

\begin{proof}
    The proof closely follows \cref{cor:low_dim_score_est}, with differences highlighted in \hyperref[item:K1]{(K1)} and \hyperref[item:K2]{(K2)}. 
    By replacing the input dimension $d$, $L$ to $\tilded$ and $\tildeL$, and the input dimension $d_x$ with $d_0$ in \cref{cor:low_dim_score_est}, and under the the $\beta_0$-\holder smoothness assumption on latent data detailed in \hyperref[item:K2]{(K2)}, the proof is complete. 
    Please see \cref{app:proof_main_risk_bounds} and  \cref{proof:cor:low_dim_score_est} for detailed proofs.
\end{proof}

\begin{remark}[Comparing Score Estimation in \cref{thm:main_risk_bounds,thm:latent_holder_est}]
Under H\"older data assumption, the sample complexity of $L_2$ estimator for achieving $\epsilon$-error are bound by $\tilde{\calO}\left(\epsilon^{-{\tilde{\nu}_{1}(d_{0}+d_{y}+\beta_0)}/{\beta_0}}\right)$ and $\tilde{\calO}\left( \epsilon^{-{\tilde{\nu}_3(d_{0}+d_{y}+2\beta_0)}/{\beta_0}} \right)$ where $\tilde{\calO}$ ignores $\tilded$, $\tildeL$, $\log {\tildeL}$, $\log 1/t_0$, $1/t_0$, and $\log n$.
Invoking \cref{lemma:holder_smooth_transformation} where $\beta_0=\beta$ and $B_0 \lesssim B$ the sample complexity in \cref{thm:latent_holder_est} improves \cref{thm:main_risk_bounds} by $\calO \left( \epsilon^{-\zeta\qty(d_x-d_0)} \right)$ where $\zeta$ is a positive constant defined by $\zeta=104C_\sigma/\beta-68\beta\qty(1/\qty(\qty(d_x+d_y)\qty(d_0+d_y)))$ and $d_0 \le d_x$.
\end{remark}

\subsection{Distribution Estimation}\label{sec:appendix_latent_dist_est_dit_holder}

In this section, we provide the extended results for \cref{sec:score_est_dist_est} on distribution estimation with the estimator $\calT_{\text{score}}$.
We state the main results under H\"older data assumptions in \cref{thm:latent_holder_est}.

\begin{theorem}[Conditional Distribution Estimation of Latent DiT]
\label{thm:lantent_distribution_TV_bound}
\blue{Assume $d_x=\Omega( \frac{\log N}{\log \log N})$.}
For all $y\in[0,1]^{d_y}$, let $\mathrm{KL}\left(P(\cdot|y)| N(0,I)\right) \le c$ for some constant $c<\infty$.
Taking the early-stopping time $t_0=n^{-\frac{\beta_0}{\blue{(d_{0}+d_{y}+\beta_0)}}}$ and terminal time $T=\frac{2\beta_0}{d_{0}+d_{y}+2\beta_0}\log{n}$.
\begin{itemize}
    \item Under \cref{assumption:conditional_density_function_assumption_1}, we have 
    \begin{align*}
    \mathbb{E}_{\{x_i, y_i\}_{i=1}^n} \left[ \mathbb{E}_{y} \left[ {\rm TV} \left( \hat{P}_{t_0}(\cdot | y), P(\cdot | y) \right) \right] \right]
    = \calO \left(n^{-\frac{\beta}{2(\tilde{\nu}_{1}-1)(d_{0}+d_{y}+\beta_0)}}(\log{n})^{\frac{\tilde{\nu}_{2}}{2}+\frac{3}{2}}\right),
\end{align*}
where \blue{$\tilde{\nu}_{1}=68\beta_0/(d_0+d_y)+104C_{\sigma}$
and
$\tilde{\nu}_{2}=12d_0+12\beta_0+2$.}

    \item Under \cref{assumption:conditional_density_function_assumption_2}.
we have
\begin{align*}
    \mathbb{E}_{\{x_i, y_i\}_{i=1}^n} \left[ \mathbb{E}_{y} \left[ {\rm TV} \left( \hat{P}_{t_0}(\cdot | y), P(\cdot | y) \right) \right] \right]=
    \calO\left(n^{-\frac{1}{2\tilde{\nu}_{3}}\frac{\beta_0}{d_{0}+d_{y}+2\beta_0}}(\log{n})^{{\max(6,\frac{\beta_0}{2}+\frac{3}{2}})}\right),
\end{align*}
where 
\blue{$\tilde{\nu}_{3}={\frac{4(12\beta_0 d_{0}+31\beta_0 \tilded+ 6\beta_0)}{\tilded(d_0+d_y)} + \frac{12(12C_{\alpha}d_0 + 25C_{\alpha}\cdot \tilded + 6C_{\alpha})}{\tilded}}
+72C_{\sigma}$}.

\end{itemize}
\end{theorem}

\begin{proof}
The proof closely follows \cref{thm:distribution_TV_bound}, with differences highlighted in \hyperref[item:K1]{(K1)} and \hyperref[item:K2]{(K2)}. 
By replacing the input dimension $d$, $L$ to $\tilded$ and $\tildeL$, and the input dimension $d_x$ with $d_0$ in \cref{thm:distribution_TV_bound}, and under the the $\beta_0$-\holder smoothness assumption on latent data detailed in \hyperref[item:K2]{(K2)}, the proof is complete. 
Please see \cref{app:proof_distribution_TV_bound} for a detailed proof.
\end{proof}

Next,
we present the distribution estimation result for low-dimensional input data.

\begin{corollary}[Low-Dimensional Input Region]\label{cor:low_dim_latent_distribution_est}
Assume $d_0 = o\left(\frac{\log N}{\log \log N}\right)$ , i.e., $d_0\ll n$. 
Under \cref{assumption:conditional_density_function_assumption_1}, by setting $t_0, T$ as specified in \cref{thm:lantent_distribution_TV_bound}, we have
\begin{align*}
\mathbb{E}_{\{x_i, y_i\}_{i=1}^n} \left[ \mathbb{E}_{y} \left[ {\rm TV} \left( \hat{P}_{t_0}(\cdot | y), P_0(\cdot | y) \right) \right] \right]
    = \calO \left(n^{-\frac{\beta_0}{2(\tilde{\nu}_{4}+1)(d_{0}+d_{y}+\beta_0)}}\right),    
\end{align*}
where
$\tilde{\nu}_4 = \frac{144\tilded\beta_0(\tildeL+2)(d_0 + 2\tilded + 1)}{d_0 + d_y} + 96\tilded C_{\sigma}(\tildeL+2)(d_0 + 2\tilded + 1) - 4\beta_0$.
\end{corollary}

\begin{proof}
    The proof closely follows \cref{cor:low_dim_distribution_est}, with differences highlighted in \hyperref[item:K1]{(K1)} and \hyperref[item:K2]{(K2)}. 
    By replacing the input dimension $d$, $L$ to $\tilded$ and $\tildeL$, and the input dimension $d_x$ with $d_0$ in \cref{cor:low_dim_distribution_est}, and under the the $\beta_0$-\holder smoothness assumption on latent data detailed in \hyperref[item:K2]{(K2)}, the proof is complete. 
    Please see \cref{app:proof_distribution_TV_bound} for a detailed proof.
\end{proof}

\clearpage
\section{Latent Conditional DiT with Lipschitz Assumption}
\label{sec:appendix_latent_dit_lipschitz}

In this section, we apply our techniques to the setting of \cite{hu2024statistical} on DiT approximation and estimation theory.
Specifically, we extend their work by using the one-layer self-attention transformer universal approximation framework introduced in \cref{sec:trans_univeral_approx}.

Compared to \cite{hu2024statistical}, we consider classifier-free conditional DiTs, providing a holistic view of the theoretical guarantees under various assumptions.
In particular,  our sample complexity bounds avoid the gigantic double exponential term $2^{(1/\epsilon)^{2L}}$ reported in  \cite{hu2024statistical}.
We adopt the following three assumptions considered by \citet{hu2024statistical}:
\begin{itemize}[leftmargin=2.4em]
    \item [\textbf{(A1)}]\label{item:lip_a1} \textbf{Low-Dimensional Linear Latent Space Data Assumption.}
    \begin{assumption}[Low-Dimensional Linear Latent Space (\cref{assumption:low_dim_linear_latent_space} Restated)]\label{assumption:low_dim_linear_latent_space_copy}
    Data point $x = U h$, where $U \in \mathbb{R}^{d_x \times d_0}$ is an unknown matrix with orthonormal columns. 
    The latent variable $h \in \mathbb{R}^{d_0}$ follows a distribution $P_h$ with a density function $p_h$.
    \end{assumption}
    Under this data assumption, \citet{chen2023plot} show that the latent score function endows a neat decomposition into on-support $s_{+}$ and orthogonal $s_{-}$ terms (see \cref{lemma:subspace_score}).
    \begin{lemma}[Score Decomposition, Lemma 1 of \cite{chen2023score} (\cref{lemma:subspace_score} Restated)]\label{lemma:subspace_score_restate}
    Let data $x = Uh$ follow \cref{assumption:low_dim_linear_latent_space}. 
    The decomposition of score function $\nabla \log p_t(x)$ is
    \begin{align}\label{eqn:score_decom_copy}
    ~\nabla \log p_t(x) = \underbrace{U\nabla \log p_t^{h}(\Bar{h} | y)}_{s_+(\Bar{h}, y, t)}   \underbrace{-  \left(I_D - UU^\top \right) x/\sigma_{t}^2}_{s_-(x, t)}, ~\Bar{h}=U^\top x,
    \end{align}
    where
    $p_t^{h}\left(\Bar{h} | y \right) \coloneqq  \int  \psi_t(\Bar{h}|h)p_h\left(h | y \right) \dd h$, 
    $\psi_t( \cdot | h)$ is the Gaussian density function of $N(\beta_{t}h, \sigma_{t}^2I_{d_0})$, $\beta_{t} = e^{-t/2}$ and $\sigma_{t}^2 = 1 - e^{-t}$.
    \end{lemma}

    \item [\textbf{(A2)}]\label{item:lip_a2} \textbf{Lipschitz Score Assumption.}
    We assume the on-support score function $s_+(\bar{h}, y, t)$ to be $L_{s_+}$-Lipschitz for any $\barh$ and $y$.
    \begin{assumption}[$L_{s_+}$-Lipschitz of $s_+(\bar{h}, y, t)$]\label{assumption:latent_on_support_lipschitz}
    The on-support score function $s_+(\bar{h}, y, t)$ is $L_{s_+}$-Lipschitz with respect to any $\bar{h} \in \R^{d_0}$ and $y \in \R^{d_y}$ for any $t \in [0, T]$.
    i.e., there exist a constant $L_{s_+}$, such that for any $\bar{h}$, $y$ and $\bar{h}^{\prime}$, $y^{\prime}$:
    \begin{align*}
    \|s_+(\bar{h}, y, t) - s_+(\bar{h}^\prime, y^{\prime}, t)\|_2 \leq L_{s_+} \|\bar{h} - \bar{h}^{\prime}\|_2 + L_{s_+} \|y - y^{\prime}\|_2.
    \end{align*}
    \end{assumption}

    \item [\textbf{(A3)}]\label{item:lip_a3}  \textbf{Light Tail Data Assumption.}
    \begin{assumption}[Tail Behavior of $P_h$]\label{assumption:latent_tail_behavior}
    The density function $p_h > 0$ is twice continuously differentiable. 
    Moreover, there exist positive constants $A_0, A_1, A_2$ such that when $\norm{h}_2 \geq A_0$, the density function $p_h\left(h | y \right) \leq (2\pi)^{-d_0/2} A_1 {\exp} (-A_2  \| h \|_2^2 / 2)$.
\end{assumption}

\end{itemize}

We note that, the assumptions \hyperref[item:lip_a1]{(A1)} and \hyperref[item:lip_a3]{(A3)} are on data, and \hyperref[item:lip_a2]{(A2)} are on the score function.
Notably, 
\hyperref[item:lip_a2]{(A2)} on the smoothness of score function is stronger than \holder data smoothness assumptions considered in \cref{sec:con_dit,sec:latent_con_dit}.

\paragraph{Organization.}
We study latent conditional DiTs under low-dimensional data \cref{assumption:low_dim_linear_latent_space_copy}, Lipschitz smoothness \cref{assumption:latent_on_support_lipschitz}, and tail behavior of $P_h$ \cref{assumption:latent_tail_behavior} and states the results in \cref{sec:appendix_latent_score_approx_lipschitz,sec:appendix_latent_score_est_lipschitz,sec:appendix_latent_dist_est_lipschitz}, respectively. 
\cref{sec:appendix_latent_score_approx_lipschitz} discusses score approximation. 
\cref{sec:appendix_latent_score_est_lipschitz} discusses score estimation.
\cref{sec:appendix_latent_dist_est_lipschitz} discusses distribution estimation. 
The proof in this section provided in \cref{sec:appendix_latent_score_approx_lipschitz_proof,sec:appendix_latent_score_est_lipschitz_proof,sec:appendix_latent_dist_est_lipschitz_proof}.
The proof strategy in this section follows \cite{hu2024statistical}.

Here we summarize the key settings of this section:

\begin{itemize}[leftmargin=2.0em]
    \item [\textbf{S1.}]\label{item:S1} \textbf{Lipschitz Smooth Assumption and Tail Behavior.} 
    Following \cite{hu2024statistical}, we introduce two assumptions on Lipschitz smoothness for on-support score function $s_{+}$ and tail behavior of $P_h$ in \cref{assumption:latent_tail_behavior,assumption:latent_on_support_lipschitz}, respectively. 
    The on-support score function is defined as $s_{+}(U^\top x, y , t) = U \nabla \log p_t^{h}\left(U^\top x | y \right)$ (see \cref{lemma:subspace_score} for score decomposition). 
    
    \item [\textbf{S2.}]\label{item:S2} \textbf{Low-Dimensional Space.} 
    We consider the setting of latent representation that is the data lives in a ``Low-Dimensional Subspace'' under \cref{assumption:low_dim_linear_latent_space}, following \cite{hu2024statistical,chen2023score}.
    The raw data $x \in \R^{d_x}$ is supported by latent $h \in \R^{d_0}$ where $d_0 \le d_x$.

    \item [\textbf{S3.}]\label{item:S3} \textbf{Transformer Network.} 
    We follow the standard setting of ``in-context'' conditional DiTs by \citet{peebles2023scalable} on latent representation.
    The network settings refer to \cref{sec:latent_con_dit}.
    Here we apply transformer-block $g_{\calT} \in \R^{d_0}$ for the approximation of on-support score function $s_{+}$.
    For each input $x \in \R^{d_x}$ and corresponding label $y \in \R^{d_y}$, we use an adapted transformer network to obtain a score estimator $s_W \in \R^{d_0}$.
    The adapted transformer network as the score estimator has the following components. 
    We utilize reshape layer to convert vector input $h\in\R^{d_0}$ to matrix (sequence) input $H\in\R^{\tilde{d}\times \tilde{L}}$.
    Specifically,    
    the reshape layer in the network \cref{fig:Latent_DiT} is defined as $\tilde{R}(\cdot): \mathbb{R}^{d_0} \to \mathbb{R}^{\tilde{d} \times \tilde{L}}$ and its reverse $\tilde{R}^{-1} (\cdot): \mathbb{R}^{\tilde{d} \times \tilde{L}} \to \mathbb{R}^{d_0}$, where $d_0 \le d_x$, $\tilde{d}\le d$, and $\tilde{L} \le L$.
    For raw data $x\in\R^{d_x}$, we utilize linear encoder $W_U^\top\in\R^{d_0\times d_x}$ and decoder $W_U\in \R^{d_x\times d_0}$ to convert the raw $x\in\R^{d_x}$ to latent $h\in\R^{d_0}$ data representations.
    Importantly, $x=Uh$ with $U\in \R^{d_x\times d_0}$ by \cref{assumption:low_dim_linear_latent_space}.

\end{itemize}

Under the \cref{assumption:low_dim_linear_latent_space_copy,assumption:latent_on_support_lipschitz,assumption:latent_tail_behavior} with the network setting following \hyperref[item:S3]{(S3)}, the theoretical results in \cref{sec:appendix_latent_score_approx_lipschitz,sec:appendix_latent_score_est_lipschitz,sec:appendix_latent_dist_est_lipschitz} achieve tighter approximation rates and efficient recovery accuracy of latent data detailed in \hyperref[item:R1]{(R1)}, \hyperref[item:R2]{(R2)}, and \hyperref[item:R3]{(R3)}.

We summarize the theoretical comparisons from  \cref{sec:appendix_latent_dit_holder} and \cref{sec:appendix_latent_dit_lipschitz} as follows:
\begin{itemize}[leftmargin=2.0em]
    \item [\textbf{R1.}]\label{item:R1} For score approximation (see \cref{theorem:latent_main1_formal,theorem:latent_main2_formal,theorem:latent_lipschitz_formal}):
    
    \begin{itemize}
        \item Under \holder data assumption the approximation rates gives $\tilde{\calO}\qty(\epsilon^{1/(d_0+d_y)})$, where $\tilde{\calO}$ ignores $B_0$, $\log \epsilon$, and $\log n$.
        \item Under Lipschitz score assumption the approximation rate gives $\tilde{\calO}\qty(\epsilon \cdot \sqrt{d_0+d_y})$, where $\tilde{\calO}$ ignores $B_0$, $\log \epsilon$, and $\log n$.
        \item For any precision $0 < \epsilon < 1$, the Lipschitz score assumption provides a tighter approximate rate for high dimension data $d_0 \gg 1$ compared with under \holder data assumption.
    \end{itemize}
    
    \item [\textbf{R2.}]\label{item:R2} For score estimation (see \cref{thm:latent_holder_est,thm:latent_score_est}):
    
    \begin{itemize}
        \item Under \holder data assumption the score estimation error gives $\tilde{\calO} \left( n^{-\frac{1}{\tilde{\nu}_{3}}\cdot\frac{\beta_0}{d_{0}+d_{y}+2\beta_0}} \right)$, where $\tilde{\calO}$ ignores $B_0$, $\log \epsilon$, and $\log n$.
        \item Under Lipschitz score assumption the score estimation error gives \blue{$\tilde\calO \left( n^{\frac{-3}{2{\qty(1+3/\tilded+4\tildeL)}}} \right)$}, where $\tilde{\calO}$ ignores $B_0$, $\log \epsilon$, and $\log n$.
        \item Under minimax optimal condition (see \cref{subsec:mini_op_con_dit}) by setting $\tilde{\nu}_3=1/2$, \holder data assumption gives $\tilde{\calO} \left( n^{-\frac{\beta_0}{2(d_{0}+d_{y}+2\beta_0)}} \right)$.
        On the other hand, Lipschitz assumption gives \blue{$\tilde\calO \left( n^{-\frac{\tilded}{\qty(3/4)d_0+\qty(2/3)\tilded+2}} \right)$}.
        Therefore, the Lipschitz assumption gives a better sample complexity guarantee for high dimensional data $d_0=\tilded \tildeL \gg 1$.
    \end{itemize}
    
    \item [\textbf{R3.}]\label{item:R3} For distribution estimation (see \cref{thm:lantent_distribution_TV_bound,thm:latent_dist_est_formal}):
    
    \begin{itemize}
        \item Under \holder data assumption: $\tilde{\calO} \left( n^{-\frac{1}{\tilde{\nu}_{3}}\frac{\beta_0}{2(d_{0}+d_{y}+2\beta_0)}} \right)$.
        \item Under Lipschitz score assumption: \blue{$\tilde\calO \qty(n^{\frac{-3}{2{\qty(1+3/\tilded+4\tildeL)}}})$.}
        \item Follow the arguments in \hyperref[item:R2]{(R2)}, Lipschitz assumption gives a better distribution estimation guarantee for high dimensional data.
    \end{itemize}
    
\end{itemize}

Note that $d_0$, $d_y$ is the latent data dimension and conditioning label dimension and \blue{$\tilde{\nu}_{3}={\frac{4(12\beta_0 d_{0}+31\beta_0 \tilded+ 6\beta_0)}{\tilded(d_0+d_y)} + \frac{12(12C_{\alpha}d_0 + 25C_{\alpha}\cdot \tilded + 6C_{\alpha})}{\tilded}}
+72C_{\sigma}$}.

From \hyperref[item:R1]{(R1)}, \hyperref[item:R2]{(R2)}, and \hyperref[item:R3]{(R3)}, we conclude that stronger approximations yield sharper rates.

\subsection{Score Approximation}\label{sec:appendix_latent_score_approx_lipschitz}

For completeness, we follow the proofs from \citep{hu2024statistical} for score approximation of the conditional latent diffusion model.

Here we use stricter assumptions on the latent density function, instead of assuming H\"{o}lder smoothness of the initial conditional data distribution as in \cref{sec:latent_con_dit}.
To be specific, we directly approximate the on-support latent score function, instead of approximating the denominator and nominator separately.
From the score decomposition in \eqref{eqn:score_docom_rearange}, we define the on-support score function $s_{+}$ as following:
\begin{align}
s_{+}(U^\top  x, y , t) 
= & ~   
U \int\frac{\nabla_{\Bar{h}}\psi_t(\Bar{h} | h) p_h\left(h | y \right)}{\int \psi_t(\Bar{h} | h') p_{h'}\left(h' | y \right) \dd  h'} \dd  h \nonumber \\
= & ~ 
U \nabla \log p_t^{h}\left(U^\top x | y \right).
\label{eqn:on_support_score}
\end{align}
Here we require two assumptions following the proof of \cite{hu2024statistical} on tail behavior of density function and Lipschitz continuous for on-support score function.
\cref{assumption:latent_tail_behavior} is the analogy of \cref{assumption:conditional_density_function_assumption_1} for assuming the tail behavior of the density function.
On the other hand, \cref{assumption:latent_on_support_lipschitz} further assume the on-support score function $s_{+}$ to be $L_{s_+}$-Lipshitz.
Note that this assumption is stricter than \cref{assumption:conditional_density_function_assumption_1} since we make the Lipschitz assumption directly on the score function instead of on the latent density function.

\begin{theorem}[Latent Score Approximation of Conditional DiT, modified from Theorem 3.1 in \citet{hu2024statistical}]
\label{theorem:latent_lipschitz_formal}
For any approximation error $\epsilon>0$ and any data distribution $P_0$ under \cref{assumption:low_dim_linear_latent_space,assumption:latent_tail_behavior,assumption:latent_on_support_lipschitz}, there exists a DiT score network $\calT_{\text{score}}(\barh,y,t)\in\calT_{\tilde{R}}^{\textcolor{blue}{h,s,r}}$ where 
${W}=\{{W}_U, \calT_{\text{score}}\}$, such that for any $t\in [t_0,T]$, we have:
\begin{align*}
\norm{\calT_{\text{score}}(\cdot,t)-\nabla\log p_t(\cdot)}_{L^2(P_t)}\leq \epsilon\cdot \sqrt{d_0+d_y}/\sigma_t^2,
\end{align*}
where $\sigma_t^2=1-e^{-t}$ and the parameter bounds in the transformer network class satisfy

\blue{
\begin{align*}
& 
\norm{W_{Q}}_{2}=\norm{W_{K}}_{2}
=
\calO\qty(\tilded\cdot\epsilon^{-(\frac{1}{\tilded}+2\tildeL)}(\log{\tildeL})^{\frac{1}{2}});\\
&\norm{W_{Q}}_{2,\infty}=\norm{W_{K}}_{2,\infty}
=
\calO\qty(\tilded^{\frac{3}{2}}\cdot\epsilon^{-(\frac{1}{\tilded}+2\tildeL)}(\log{\tildeL})^{\frac{1}{2}}); \\
& \norm{W_O}_{2}=\calO\left({\tilded}^{\frac{1}{2}}\epsilon^{\frac{1}{\tilded}}\right); 
\norm{W_O}_{2,\infty}=\calO\left(\epsilon^{\frac{1}{\tilded}}\right);\\
&\norm{W_{V}}_{2}=\calO({\tilded}^{\frac{1}{2}});
\norm{W_{V}}_{2,\infty}=\calO(\tilded); \\ 
& \norm{W_{1}}_{2}=\calO\left(\tilded\epsilon^{-\frac{1}{\tilded}}\right), \norm{W_{1}}_{2,\infty}=\calO\left({\tilded}^{\frac{1}{2}}\epsilon^{-\frac{1}{\tilded}}\right);\\
& \norm{W_{2}}_{2}=\calO\left(\tilded\epsilon^{-\frac{1}{\tilded}}\right);
\norm{W_{2}}_{2,\infty}=\calO\left({\tilded}^{\frac{1}{2}}\epsilon^{-\frac{1}{\tilded}}\right);\\
&\norm{E^{\top}}_{2,\infty}=\calO\left(\tilded^{\frac{1}{2}}\tildeL^{\frac{3}{2}}\right).
\end{align*}
}
\end{theorem}
\begin{proof}
Please see \cref{sec:appendix_latent_score_approx_lipschitz_proof} for a detailed proof.
\end{proof}

\begin{remark}[Comparing with H\"{o}lder Assumption Results in Low-Dimensional Data]
Under \cref{assumption:conditional_density_function_assumption_1,assumption:conditional_density_function_assumption_2}, the score approximation give us $\tilde{\calO}\left(\epsilon^{\frac{1}{d_{x}+d_{y}}}/\sigma_t^4\right)$ and $\tilde{\calO}\left(\epsilon^{\frac{1}{d_{x}+d_{y}}}/\sigma_t^2\right)$ in \cref{theorem:latent_main1_formal,theorem:latent_main2_formal}, respectively.
On the other hand, the direct approximation of the Lipschitz smooth on-support score function gives us the approximation error of $\calO \left( \epsilon\cdot \sqrt{d_0+d_y}/\sigma_t^2 \right)$.
For $(d_0+d_y) \gg 1$, \cref{theorem:latent_lipschitz_formal} delivers superior approximation error compare with \cref{theorem:latent_main1_formal,theorem:latent_main2_formal}.
    
\end{remark}

\subsection{Score Estimation}\label{sec:appendix_latent_score_est_lipschitz}

\begin{theorem}
[Score Estimation of Latent DiT]
\label{thm:latent_score_est}
    Under the \cref{assumption:low_dim_linear_latent_space_copy,assumption:latent_tail_behavior,assumption:latent_on_support_lipschitz}, we choose the score network $\calT_{\text{score}}(x,y,t)\in\calT_{\tilde{R}}^{\textcolor{blue}{h,s,r}}$ from  \cref{theorem:latent_lipschitz_formal} using $\epsilon \in (0,1)$ and $\tildeL> 1$.
    With probability $1-1/\poly(n)$, we have
    \begin{align*}
       &~\frac{1}{T-t_0}\int_{t_0}^{T}\norm{ \calT_{\text{score}}(\cdot,t)-\nabla\log p_t(\cdot)}_{L^2(P_t)} \dd t 
       =\tilde{\calO} \left( \frac{1}{t_0^2}  n^{\frac{-3}{2{\qty(1+3/\tilded+4\tildeL)}}} \log^{3}\tilde{L} \log^{3}n \right),
    \end{align*}
    where $\tilde{\calO}$ hides the factor about $d_x, d_y, d_0, \tilde{d}, L_{s_+}$ and $\delta (n)$ is negligible for sufficiently large $n$. 
\end{theorem}
\begin{proof}
    Please see \cref{sec:appendix_latent_score_est_lipschitz_proof} for a detailed proof.
\end{proof}

\begin{remark}[Comparing Score Estimation in \cref{thm:latent_holder_est,thm:latent_score_est}]
Under H\"older data assumption, the sample complexity of $L_2$ estimator for achieving $\epsilon$-error are bound by $\tilde{\calO}\left(\epsilon^{-{\tilde{\nu}_{1}(d_{0}+d_{y}+\beta_0)}/{\beta_0}}\right)$ and $\tilde{\calO}\left( \epsilon^{-{\tilde{\nu}_3(d_{0}+d_{y}+2\beta_0)}/{\beta_0}} \right)$.
In contrast, \cref{thm:latent_score_est} has the sample complexity bound of $\tilde{\calO}\qty( \epsilon^{-{2(1+3/\tilde{d}+4\tilde{L})}/3 })$.
Therefore, a direct approximation of the \lip smooth score function offers a better sample complexity bound than \holder data assumption.
\end{remark}

\subsection{Distribution Estimation}
\label{sec:appendix_latent_dist_est_lipschitz}

In practice, DiTs generate data using the discretized version with step size $\mu$.
Let $\hat{P}_{t_0}$ be the distribution generated by $\calT_{\text{score}}(x,y,t)$ in \cref{thm:latent_score_est}.
Let $P_{t_0}^h$ and $p_{t_0}^h$ be the distribution and density function of on-support latent variable $\Bar{h}$ at $t_0$.
We have the following results for distribution estimation.

\begin{theorem}
[Distribution Estimation of DiT, Modified From Theorem 3 of \cite{chen2023score}]
\label{thm:latent_dist_est_formal}
    Let $T =\calO(\log n), t_0 =\calO(\min\{c_0,1/L_{s_+}\})$, where $c_0$ is the minimum eigenvalue of $\EE_{P_h}[h h^\top]$.
    With the estimated DiT score network $\calT_{\text{score}}(x,y,t)$ in \cref{thm:latent_score_est}, we have the following with probability $1-1/\poly(n)$.
    \begin{itemize}[leftmargin=2em]
        \item [(i)]
        The accuracy to recover the subspace $U$ is
        \begin{align}\label{eq:tv_item1}
          \norm{W_U W_U^\top - UU^\top}_F^2=\tilde\calO\left(\frac{1}{c_0} n^{\frac{-3}{2{\qty(1+3/\tilded+4\tildeL)}}} \cdot \log^{3} n\right).
        \end{align}
        \item [(ii)]
        $(W_BU)^\top_{\sharp} \hat{P}_{t_0}$ denotes the pushforward distribution.
        With the conditions 
        ${\rm KL}(P_h || N (0, I_{d_0})) < \infty$, and step size $\mu \leq \xi(n,t_0,L)\cdot t_0^2/ ( d_0\sqrt{\log d_0} )$.
        There exists an orthogonal matrix $U \in \RR^{d \times d}$ such that we have the following upper bound for the total variation distance 
        \begin{align}
        \label{eq:tv_item2}
                {\rm TV} (P_{t_0}^{h}, (W_BU)^\top_{\sharp} \hat{P}_{t_0})= \tilde\calO \left(
                \frac{1}{t_0\sqrt{c_0}}n^{\frac{-3}{4{\qty(1+3/\tilded+4\tildeL)}}} \cdot \log^{4} n
                \right),
        \end{align}
        where $\tilde{\calO}$ hides the factor about $d_x, d_0, d$, and $L_{s_+}$.

        \item [(iii)] 
        For the generated data distribution $\hat{P}_{t_0}$, the orthogonal pushforward $(I-W_BW_B^\top)_{\sharp} \hat{P}_{t_0}$ is ${N}(0, \Sigma)$, where $\Sigma \preceq at_0 I$ for a constant $a > 0$.
    \end{itemize}
\end{theorem}
\begin{proof}
    Please see \cref{sec:appendix_latent_dist_est_lipschitz_proof} for a detailed proof. 
\end{proof}

\begin{remark}[Compare with Existing Work]
In \cite[Theorem 3]{chen2023score}, the upper bound for total variation distance with $\rm{ReLU}$ network is $\tilde{\calO}\left(\sqrt{{1}/{(c_0t_0)}} n^{-{1 }/{(d+5)}}\log^{2} n\right)$.
Therefore, for $n \gg 1$, \cref{thm:latent_dist_est_formal} gives tighter accuracy if $3d + 11 > 12/\tilded + 16\tildeL$ where $\tilded \le d$ and $\tildeL \le L$.
On the other hand, under similar conditions for $d$ and $L$, \cref{thm:latent_dist_est_formal} suggest to achieve similar total variation distance we only require $\sqrt{t_0}$ early stopping time which is beneficial for empirical setting.
\end{remark}

\subsection{Proof of Score Approximation (\texorpdfstring{\cref{theorem:latent_lipschitz_formal}}{})}\label{sec:appendix_latent_score_approx_lipschitz_proof}

To begin the proof of the approximate theorem, we first restate some auxiliary lemmas and their proofs here from \cite{chen2023score} for later convenience.
Note that some of the proofs extend to the latent density function.

\begin{lemma}[Modified from Lemma 16 in \cite{chen2023score}]
\label{lemma:Tool_lemma_1_t1}
    Consider a probability density function $p_h\left(h | y \right)=\exp(-C\norm{h}_2^2 /2)$ for $h\in \R^{d_0}$ and constant $C>0$. Let $r_h>0$ be a fixed radius. Then it holds
    \begin{align*}
        & \int_{\norm{h}_2>r_h} p_h\left(h | y \right) \dd  h \leq \frac{2d_0 \pi^{d_0/2}}{C\Gamma(d_0/2+1)}r_h^{d_0-2}\exp(-Cr_h^2/2),\\
        & \int_{\norm{h}_2>r_h}\norm{h}_2^2 p_h\left(h | y \right) \dd  h \leq \frac{2d_0 \pi^{d_0/2}}{C\Gamma(d_0/2+1)}r_h^{d_0}\exp(-Cr_h^2/2).
    \end{align*}
\end{lemma}

\begin{lemma}[Modified from Lemma 2 in \cite{chen2023score}]
\label{lemma:g_out_of_range}
    Suppose Assumption \cref{assumption:latent_tail_behavior} holds and $q$ is defined as:
    \begin{align*}
        q\left(\Bar{h}, y, t \right)=\int \frac{h \psi_t\left(\Bar{h} | h \right) p_h\left(h | y \right)}{\int \psi_t\left(\Bar{h} | h \right) p_h\left(h | y \right) \dd  h} \dd h, \quad
        \Bar{h}=B^\top x.
    \end{align*}
    Given $\epsilon > 0$, with $r_h = c \left(\sqrt{d_0\log (d_0/t_0) + \log (1/\epsilon)} \right)$ for an absolute constant $c$, it holds
    \begin{align*}
    \norm{q\left(\Bar{h}, y, t \right)\one\{\norm{\Bar{h}}_2 \geq r_h\}}_{L^2(P_t)} \leq \epsilon, ~ \text{for} ~ t \in [t_0, T].
    \end{align*}
\end{lemma}

\begin{lemma}[Modified from Theorem 1 in \cite{chen2023score}]
\label{lemma:coarse_upper_bound}
    We denote 
    \begin{align*}
\tau(r_h) = 
\sup_{t \in [t_0, T]} 
\sup_{\Bar{h} \in [0, r_h]^{d_0}}
\sup_{y \in [0, 1]^{d_{y}}}
\norm{\frac{\partial}{\partial t} q(\Bar{h}, y, t)}_2.
\end{align*}
With $q(\Bar{h},y,t)=\int h\psi_t(\Bar{h} | h)p_h(h | y)/(\int \psi_t(\Bar{h} | h)p_h(h | y)\dd  h) \dd  h$ and $p_h$ satisfies \cref{assumption:latent_tail_behavior}, we have a coarse upper bound for $\tau (r_h)$
\begin{align*}
    \tau(r_h) = \calO\left(\frac{1+\beta^2_t}{\beta_t} \left(L_{s_+} + \frac{1}{\sigma_t^2}\right) \sqrt{d_0} r_h\right) = \calO\left(e^{T/2}L_{s_+} r_h \sqrt{d_0}\right).
\end{align*}
\end{lemma}

\begin{proof}[Proof of \cref{lemma:coarse_upper_bound}]
\begin{align*}
\frac{\partial}{\partial t} q(\Bar{h}, y, t) 
= & ~ U \int \frac{h \frac{\partial}{\partial t}\psi_t(\Bar{h} | h) p_h(h | y)}{\int \psi_t(\Bar{h} | h) p_h(h | y) \dd  h} \dd  h 
- U \int \frac{h \psi_t(\Bar{h} | h) p_h(h | y) \int \frac{\partial}{\partial t} \psi_t(\Bar{h} | h) p_h(h | y) \dd  h}{\left(\int \psi_t(\Bar{h} | h) p_h(h | y) \dd  h\right)^2} \dd  h \\
= & ~ U \int \frac{h \frac{\beta_t}{\sigma^2_t} \left(\norm{h}_2^2 - (1 + \beta_t^2) h^\top \Bar{h} + \beta_t \norm{\Bar{h}}_2^2\right) \psi_t(\Bar{h} | h) p_h(h | y)}{\int \psi_t(\Bar{h} | h) p_h(h | y) \dd  h} \dd  h \\
& ~ -  U \int \frac{h \psi_t(\Bar{h} | h) p_h(h | y) \int \frac{\beta_t}{\sigma_t^2} \left(\norm{h}_2^2 - (1 + \beta^2_t) h^\top \Bar{h} + \beta_t \norm{\Bar{h}}_2^2\right) \psi_t(\Bar{h} | h) p_h(h | y) \dd  h}{\left(\int \psi_t(\Bar{h} | h) p_h(h | y) \dd  h\right)^2} \dd  h \\
\overset{(i)}{=} & ~ \frac{\beta_t}{\sigma^2_t} U \left[\E_{P_h}\left[h \norm{h}_2^2\right] - (1+\beta^2_t) \cov \left[h | \Bar{h} \right] \Bar{h}\right],
\end{align*}
where we plug in 
$\partial \psi_t(\Bar{h} | h)/\partial t = \beta_t \left(\norm{h}_2^2 - (1 + \beta^2_t) h^\top \Bar{h} + \beta_t \norm{\Bar{h}}_2^2\right) \psi_t(\Bar{h} | h)/\sigma^2_t$ and collect terms in $(i)$. 
Since $P_h$ has a Gaussian tail, its third moment is bounded. 

Then we bound $\norm{\cov[h | \Bar{h}]}_{\rm op}$ by taking derivative of $s_{+}(\Bar{h}, y,t)$ with respect to $\Bar{h}$, here
$$
s_{+}(\Bar{h}, y, t)= U \frac{\beta_t}{\sigma_t^2}\int \frac{h \cdot \psi_t(\Bar{h} | h) p_h(h | y)}{\int \psi_t(\Bar{h} | h) p_h(h | y) \dd  h} \dd  h  - U\frac{\Bar{h}}{\sigma_t^2}.
$$ 
Then we have
\begin{align*}
    \frac{\partial}{\partial \Bar{h}} s_{+}(\Bar{h}, y, t) 
    & = ~ \left(\frac{\beta_t}{\sigma_t^2}\right)^2 U \left[ 
    \int {hh^\top \varphi(\bar{h},y)} \dd h
    - \int {h \varphi(\bar{h},y)} \dd h
    \int {h^\top \varphi(\bar{h},y)} \dd h
    \right] - \frac{1}{\sigma_t^2}U\\
    & = ~ \left(\frac{\beta_t}{\sigma_t^2}\right)^2 U \left[{\rm \cov}(h | \Bar{h}) - \frac{1}{\sigma_t^2}I_{d_0}\right],
\end{align*}
where
\begin{align*}
\varphi(\bar{h},y) = \frac{\psi_t(\Bar{h} | h) p_h(h | y)}{\int \psi_t(\Bar{h} | h) p_h(h | y) \dd h}.
\end{align*}
Along with the $L_{s_+}$-Lipschitz property of $s_{+}$, we obtain
\begin{align*}
\norm{\cov(h | \Bar{h})}_{\rm op} \leq \frac{\sigma^4_t}{\beta^2_t} \left(L_{s_+} + \frac{1}{\sigma_t^2}\right).
\end{align*}
Therefore, we deduce
\begin{align*}
\tau(r_h) = \calO\left(\frac{1+\beta^2_t}{\beta_t} \left(L_{s_+} + \frac{1}{\sigma_t^2}\right) \sqrt{d_0} r_h\right) = \calO\left(e^{T/2}L_{s_+} r_h \sqrt{d_0}\right),
\end{align*}
as $P_h$ having sub-Gaussian tail implies $\E_{P_h}\left[h \norm{h}_2^2\right]$ is bounded.
\end{proof}

\begin{lemma}[Modified from Lemma 10 in \cite{chen2023score}]
\label{lemma:approximation_property_for_lipschitz}
    For any given $\epsilon>0$, and $L$-Lipschitz function $g$ defined on $[0,1]^{d_0} \times [0,1]^{d_y}$, there exists a continuous function $\Bar{f}$ constructed by trapezoid function that
    \begin{align*}
        \norm{g-\Bar{f}}_\infty \leq \epsilon.
    \end{align*}
    Moreover, the Lipschitz continuity of $\bar{f}$ is bounded by
\begin{align*}
\left\lvert \bar{f}(x,y) - \bar{f}(x^{\prime}, y^{\prime}) \right\rvert \leq 
10d_0L \norm{x - x^{\prime}}_2 + 
10d_yL \norm{y - y^{\prime}}_2,
\end{align*}
for any $x, x^{\prime} \in [0, 1]^{d_0}$ and $y, y^{\prime} \in [0, 1]^{d_y}$.

\end{lemma}
\begin{proof}[Proof of \cref{lemma:approximation_property_for_lipschitz}]
\blue{
This proof closely follows Lemma 10 in \cite{chen2023score}.
}
\blue{
We divide the proof into two parts: First, we use a collection of Trapezoid function $\bar{f}$ to approximate the function $g$ defined on $[0,1]^{d_0} \times [0,1]^{d_y}$.
Then we establish the Lipschitz continuity of the function $\bar{f}$ to facilitate the approximation with a transformer.}

\begin{enumerate}
    \item {\bf Approximation by Trapezoid Function. }
    Given an integer $N>0$, we choose $(N+1)^{d_0}$ points in the hypercube $[0,1]^{d_0}$ and $(N+1)^{d_y}$ points in the hypercube $[0,1]^{d_y}$. 
    \blue{We denote the index of the hypercubes as} $m=\qty[m_1,m_2,\cdots,m_{d_0}]^\top \in \qty{0,\cdots,N}$ and $n=\qty[n_1,n_2,\cdots,n_{d_y}]^\top \in \qty{0,\cdots,N}$.
    \blue{Next}, we define a univariate trapezoid function (see \cref{fig:trapezoid}) \blue{as follow}
    \begin{align}
    \label{func:trapezoid_func}
    \phi(a) = \begin{cases}
    1, & \vert a\vert < 1 \\
    2 - \vert a\vert, & \vert a\vert \in [1, 2] \\
    0, & \vert a\vert > 2 \\
    \end{cases}.
    \end{align}
    
    \begin{figure}[h]
    \centering
    \begin{tikzpicture}
        \begin{axis}[
            scale=1.0, 
            axis x line=middle,
            axis y line=none,
            xlabel={$x_k$},  
            ylabel={},
            ymin=-0.2, ymax=1.0, 
            xmin=-3, xmax=3,
            samples=100,
            domain=-3:3,
            smooth,
            thick,
            xtick={-2, -1, 0, 1, 2},
            xticklabels=\empty,
            ytick=\empty,
            tick style={thick}
        ]
        \addplot [
            domain=-2:-1,
            samples=2,
            thick,
            firebrick 
        ] {1 - 0.5*abs(x)}; 
        
        \addplot [
            domain=-1:1,
            samples=100,
            thick,
            firebrick
        ] {0.5}; 
        
        \addplot [
            domain=1:2,
            samples=2,
            thick,
            firebrick
        ] {1 - 0.5*abs(x)};
        
        \draw[dashed, thick] (axis cs:-1,0) -- (axis cs:-1,0.5);
        \draw[dashed, thick] (axis cs:1,0) -- (axis cs:1,0.5);
        
        \node at (axis cs:0,0.60) {$\phi\left(3N\left(x_k - \frac{m_k}{N}\right)\right)$};  
        \node at (axis cs:0,+0.10) {$m_k/N$};
        \end{axis}
    \end{tikzpicture}
    \vspace{-1.8em}
    \caption{\textbf{Trapezoid function.}}\label{fig:trapezoid}
\end{figure}
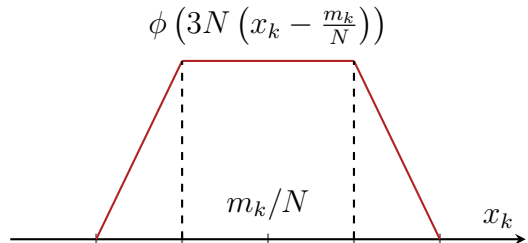

For any $x \in [0, 1]^{d_0}$ and $y \in [0, 1]^{d_y}$, we define a partition of unity based on a product of trapezoid functions indexed by $m$ and $n$,
\begin{align}\label{eqn:unity_func}
\xi_{m,n}(x, y) = 
\mathds{1} \left\{ y \in \biggl( \frac{n-1}{N}, \frac{n}{N} \biggr] \right\} 
\prod_{k=1}^{d_0} \phi\left(3N\left(x_k - \frac{m}{N}\right)\right).
\end{align}
For example, the product of trapezoid function $\xi_{m,n}(x, y) \neq 0$ only if $y \in \bigl( \frac{n-1}{N}, \frac{n}{N} \bigr]$ and $x \in \bigl[ \frac{m-2\cdot\mathds{1}\cdot 3}{N}, \frac{m+2\cdot\mathds{1}\cdot 3}{N} \bigr]$.
For any target $L$-Lipschitz function $g$ with respect to $x$ and $y$, it is more convenient to write its Lipschitz continuity with respect to the $\ell_\infty$ norm, i.e.,
\begin{align}\label{eqn:lip_linfty}
\vert g(x,y) - g(x^\prime, y^\prime)\vert 
\leq &~ L \norm{x - x^\prime}_2 + L \norm{y - y^\prime}_2 \nonumber \\
\leq &~ L \sqrt{d_0} \norm{x - x^\prime}_\infty + L \sqrt{d_y} \norm{y - y^\prime}_\infty.
\end{align}
We now define a collection of piecewise-constant functions as
\begin{align*}
P_{m,n}(x,y) = g(m,n) \quad \text{for}\quad m \in \{0, \dots, N\}^{d_0} \;\text{and}\; n \in \{0, \dots, N\}^{d_y}.
\end{align*}
We claim that $\bar{f}(x,y) = \sum_{m,n} \xi_{m,n}(x,y) P_{m,n}(x,y)$ is an approximation of $g$, with an approximation error evaluated as
\begin{align*}
& ~ \sup_{x \in [0, 1]^{d_0}} 
\sup_{y \in [0, 1]^{d_y}} 
\left\vert \bar{f}(x,y) - g(x,y)\right\vert \\
= &~ \sup_{x \in [0, 1]^{d_0}} 
\sup_{y \in [0, 1]^{d_y}} 
\left\vert \sum_{m,n} \xi_{m,n}(x,y) \left(P_{m,n}(x,y) - g(x,y)\right)\right\vert \\
\leq &~ \sup_{x \in [0, 1]^{d_0}} 
\sup_{y \in [0, 1]^{d_y}} 
\sum_{\substack{
m: \vert x_k - m_k/N\vert \leq \frac{2}{3N} \\ 
n: \vert y_j - n_j/N \vert \in ( -\frac{1}{2N}, \frac{1}{2N} ]}
} 
\left\vert P_{m,n}(x,y) - g(x,y) \right\vert \\
= &~ \sup_{x \in [0, 1]^{d_0}} 
\sup_{y \in [0, 1]^{d_y}} 
\sum_{\substack{
m: \vert x_k - m_k/N\vert \leq \frac{2}{3N} \\ 
n: \vert y_j - n_j/N \vert \in ( -\frac{1}{2N}, \frac{1}{2N} ]}
} 
\left\vert g(m,n) - g(x,y) \right\vert \\
\leq &~ 
L \sqrt{d_0} 2^{d_0+1} \frac{1}{3N} + 
L \sqrt{d_y} 1^{d_y} \frac{1}{2N} \annot{By Lipschitz continuity in \eqref{eqn:lip_linfty}}\\
= &~ \frac{L}{N} \left( \frac{\sqrt{d_0} 2^{d_0+1}}{3} + \frac{\sqrt{d_y}}{2} \right),
\end{align*}
\blue{
where the last inequality follows the Lipschitz continuity in \eqref{eqn:lip_linfty} and 
using the fact that there are at most $2^{d_0}$ terms in the summation of $m$ and at most $1^{d_y}$ terms in the summation of $n$. 
By choosing $N=\lceil L\left( \sqrt{d_0} 2^{d_0+1} / 3 + \sqrt{d_y} / 2 \right) / \epsilon \rceil$, we have $\norm{g-\bar{f}}_\infty\leq\epsilon$.
}

\item {\bf Lipschitz Continuity.} \blue{Next we compute the Lipschitz of the function $\Bar{f}$ with respect to $x$ and $y$}.
Suppose the approximation error $\epsilon>0$ is small enough, then we have
\begin{align*}
    & ~ \left\vert\Bar{f}(x,y)-\Bar{f}(x^\prime, y^\prime)\right\vert \\
    \leq &~ 
    \left\vert\Bar{f}(x, y)-g(x, y)\right\vert
    +\left\vert g(x, y)-g(x^\prime, y^\prime)\right\vert
    +\left\vert g(x^\prime,y^\prime)-\Bar{f}(x^\prime,y^\prime)\right\vert\\
    \leq &~ 2\epsilon+L\sqrt{d_0}\norm{x-x^\prime}_\infty+L\sqrt{d_y}\norm{y-y^\prime}_\infty \\
    \leq &~ 10 L\sqrt{d_0}\norm{x-x^\prime}_\infty+10 L\sqrt{d_y}\norm{y-y^\prime}_\infty\\
    \leq &~ 10Ld_0\norm{x-x^\prime}_2 + 10Ld_y\norm{y-y^\prime}_2.
\end{align*}
\end{enumerate}
This completes the proof.
\end{proof}

\paragraph{Main Proof of \cref{theorem:latent_lipschitz_formal}.}
Now we are ready to state the main proof.

\begin{proof}[Proof of \cref{theorem:latent_lipschitz_formal}]
\blue{
From low-dimensional data assumption, the score function $\log p_t(x | y)$ decomposes as the on-support and orthogonal component (see \cref{lemma:subspace_score}).
}
\blue{Recall the on-support score function is given by} $\nabla \log p_t^{h} \left( \Bar{h} | y \right)=U^\top s_{+} (\Bar{h}, y, t) $ from \eqref{eqn:lip_linfty}. 
\blue{ 
We use a latent score network to approximate the score function (see \hyperref[item:K3]{(K3)}).
Specifically, the latent score network includes a latent encoder and a latent decoder.
The encoder approximates $U^{\top} \in \R^{d_0\times d_x}$ ,and decoder approximates $U\in \R^{d_x\times d_0}$.
At its core, we use the transformer $g_{\calT}(W_U^\top x, y, t) \in \calT^{\textcolor{blue}{h,s,r}}$ to approximate $q\qty( \Bar{h}, y, t)$ as defined in \eqref{eqn:score_docom_rearange_appendix}.
The expression for $q \qty( \Bar{h}, y, t)$ is given by:
}
\begin{align}
    q(\Bar{h}, y, t)= \sigma_{t}^2\nabla \log p_t^{h} ( U^\top x | y)
    +U^{\top}x = \sigma_t^{2}U^\top (s_{+} (\Bar{h},y,t) + x/\sigma_t^{2}).
\end{align}

We proceed as follows:
\begin{itemize}
    \item {\bf Step 1.}
    Approximate $q(\Bar{h}, y, t)$ with a compact-supported continuous function $\bar{f}(\Bar{h}, y, t)$.
    
    \item {\bf Step 2.}
    Approximate $\bar{f}(\Bar{h}, y, t)$ with a one-layer single-head transformer network.
\end{itemize}

\textbf{Step 1. Approximate $q(\Bar{h}, y, t)$ with a Compact-Supported Continuous Function $\bar{f}(\Bar{h}, y, t)$. }
\blue{First, we}
partition $\R^{d_0}$ into a compact subset $H_1:=\{\Bar{h}\mid \norm{\Bar{h}}_2 \leq r_h \}$ and its complement $H_2$, \blue{where the choice of $r_h$ comes from \cref{lemma:g_out_of_range}.}
\blue{Next,}
we approximate $q(\Bar{h},y,t)$ on the two subsets \blue{by using the compact-supported continuous function $\bar{f}(\Bar{h}, y, t)$}. 
\blue{Finally, calculating the continuity of $\Bar{f}$ gives an estimation error of $\sqrt{d_0+d_y}\epsilon$ between $q(\Bar{h},y,t)$ and $\Bar{f}(\Bar{h},y,t)$. 
}
We present the main proof as follows. 

\begin{itemize}
\item 
\textbf{Approximation on $H_2 \times [0,1] \times [t_0,T]$. }For any $\epsilon >0$, \blue{by taking $r_h=c(\sqrt{d_0\log (d_0/t_0)-\log \epsilon})$, we obtain from \cref{lemma:g_out_of_range} that}
\begin{align*}
   \norm{q(\Bar{h}, y, t)\one\{\norm{\Bar{h}}_2 \geq r_h\}}_{L^2(P_t)} \leq \epsilon \quad \text{for}\quad t \in [t_0, T] \quad \text{and} \quad y \in [0, 1].
\end{align*}
So we set $\Bar{f}(\Bar{h},y,t)=0$ on $H_2 \times [0,1] \times [t_0,T]$.

\item 
\textbf{Approximation on  $H_1 \times [0,1] \times [t_0,T]$. }
On $H_1 \times [0,1] \times [t_0,T]$, we approximate 
\begin{align*}
    q(\Bar{h},y,t)=[q_1(\Bar{h},y,t),q_2(\Bar{h},y,t),\cdots ,q_{d_0}(\Bar{h},y,t)],
\end{align*}
by approximating each coordinate $q_k(\Bar{h},y,t)$ separately. 

We firstly rescale the input by $h^\prime =(\Bar{h} + r_h \one)/2r_h$ and $t^\prime = t/T$, so that the transformed input space is $[0, 1]^{d_0} \times [0, 1]^{d_y} \times [t_0/T, 1]$. 
Here we do not need to rescale $y$, since it is already in $[0,1]$ by definition.
We implement such transformation by a single feed-forward layer. 

By \cref{assumption:latent_on_support_lipschitz}, the on-support score $s_{+}(\Bar{h},y, t)$ is $L_{s_{+}}$-Lipschitz with respect to any $\Bar{h} \in \R^{d_0}$ and $y \in \R^{d_y}$. 
This implies $q(\Bar{h}, y, t)$ is $(1+L_{s_{+}})$-Lipschitz in $\Bar{h}$ and $y$. When taking the transformed inputs, $g(h^\prime, y, t^\prime) = q(2r_h h^\prime-r_h \one, T t^\prime)$ becomes $2r_h(1+L_{s_{+}})$-Lipschitz in $h^{\prime}$; each coordinate $g_k(h^{\prime},y,t)$ \blue{is also $2r_h(1+L_{s_{+}})$-Lipschitz in $h^{\prime}$}. 
Here we denote $L_{*} = 1 + L_{s_{+}}$.

Besides, $g(h^{\prime}, y, t^{\prime})$ is $T\tau(r_h)$-Lipsichitz with respect to $t$, where 
\begin{align*}
\tau(r_h) = \sup_{t \in [t_0, T]} \sup_{\Bar{h} \in [0, r_h]^d}\sup_{y \in [0, 1]^{d_{y}}} \norm{\frac{\partial}{\partial t} q(\Bar{h}, y, t)}_2.
\end{align*}
We have a coarse upper bound for $\tau(r_h)$ in \cref{lemma:coarse_upper_bound}. 
We \blue{restate it as follows:}
\begin{align*}
    \tau(r_h) = \calO\left(\frac{1+\beta^2_t}{\beta_t} \left(L_{s_{+}} + \frac{1}{\sigma_t^2}\right) \sqrt{d_0} r_h\right) = \calO\left(e^{T/2}L_{s_{+}} r_h \sqrt{d_0}\right).
\end{align*}

Since each $g_k(h^{\prime},y,t)$ is Lipsichitz continuous,  we apply \cref{lemma:approximation_property_for_lipschitz} to \blue{construct a collection of coordinate-wise functions, denoted as} $\Bar{f}_k(h^{\prime},y,t)$.
We concatenate $\bar{f}_k$'s together and construct $\bar{f} = [\bar{f}_1, \dots, \bar{f}_{d_0}]^\top$.
According to the construction of trapezoid function in \cref{lemma:approximation_property_for_lipschitz}, for any given $\epsilon$, \blue{we have the following relations:}
\begin{align*}
\sup_{h^{\prime}, y, t^{\prime} \in [0, 1]^d_0 \times [0, 1]^{d_y} \times [t_0/T, 1]} \norm{\bar{f}(h^{\prime},y, t^{\prime}) - g(h^{\prime},y, t^{\prime})}_\infty \leq \epsilon.
\end{align*}

Considering the input rescaling (i.e., $\Bar{h} \to h^{\prime}$, $y \to y$ and $t \to t'$), we obtain:
\begin{itemize}
    \item The constructed function is Lipschitz continuous in $\Bar{h}$ and $y$, i.e., for any $\Bar{h}_1, \Bar{h}_2 \in H_1$, $y_1, y_2 \in [0, 1]$ and $t \in [t_0, T]$, it holds
    \begin{align}
    \label{eqn:lip_h}
    \norm{\bar{f}(\Bar{h}_1, y_1, t) - \bar{f}(\Bar{h}_2, y_2, t)}_{\infty} \leq 10d_0 L_{*} \norm{\Bar{h}_1 - \Bar{h}_2}_2 + 10d_y L_{*}\norm{y_1 - y_2}_2.
    \end{align}
    
    \item The function is also Lipschitz in $t$, i.e., for any $t_1, t_2 \in [t_0, T]$ and $\norm{\Bar{h}}_2 \leq r_h$, it holds
    \begin{align*}
    \norm{\bar{f}(\Bar{h}, y, t_1) - \bar{f}(\Bar{h}, y, t_2)}_{\infty} \leq 10 \tau(r_h) \norm{t_1 - t_2}_2.
    \end{align*}
\end{itemize}
\blue{
To conclude, the construction of $\Bar{f}\qty( \Bar{h}, y, t )$ uses a collection of trapezoid functions, as described in \cref{lemma:approximation_property_for_lipschitz}. This ensures that $\bar f (\Bar{h}, y, t) = 0$ for $\norm{\Bar{h}}_2 > r_h$, for all $t \in [t_0, T]$ and $y \in [0, 1]$. 
Consequently, the Lipschitz continuity of $\Bar{f}\qty( \Bar{h}, y, t )$ with respect to $\Bar{h}$ extends over the entire space $\R^{d_0}$.
}

\item 
\textbf{Approximation Error Analysis under $L^2$ Norm. }
We first decompose the 
 $L^2$ approximation error of $\bar{f}$ into two terms \blue{($\norm{\Bar{h}}_2 < r_h$ and $\norm{\Bar{h}}_2 < r_h$)}:
\begin{align*}
& ~ \norm{q\qty(\Bar{h}, y, t) - \Bar{f}\qty(\Bar{h}, y, t)}_{L^2\qty(P_t^{h})} 
\\
= & ~ 
\norm{\qty(q\qty(\Bar{h}, y, t) - \bar{f}\qty(\Bar{h}, y, t)) \mathds{1}\qty{\norm{\Bar{h}}_2 < r_h}}_{L^2\qty(P_t^{h})} + \norm{q\qty(\Bar{h}, y, t) \mathds{1}\qty{\norm{\Bar{h}}_2 > r_h}}_{L^2\qty(P_t^{h})}.
\end{align*}
\blue{By selecting $r_h = \calO \qty(\sqrt{d_0\log (d_0/t_0) + \log (1/\epsilon)} ) $ (see \cref{lemma:g_out_of_range}), we bound the second term on the RHS of above expression as:}
\begin{align*}
    \norm{g(\Bar{h}, y, t) \mathds{1}\{\norm{\Bar{h}}_2 > r_h\}}_{L^2(P_t^{h})} \leq \epsilon.
\end{align*}
For the first term, we bound 
\begin{align*}
     & ~ \norm{\qty(q\qty(\Bar{h}, y, t) - \bar{f}\qty(\Bar{h}, y, t)) \mathds{1}\qty{\norm{\Bar{h}}_2 
    <  r_h}}_{L^2\qty(P_t^{h})} \\
    \leq & ~
    \sqrt{d_0+d_y}\sup_{h^\prime, y, t^\prime \in [0, 1]^{d_0} \times [0,1]^{d_y} \times [t_0/T, 1]} \norm{\bar{f}\qty(h^\prime, y, t^\prime) - g\qty(h^\prime, y, t^\prime)}_\infty \\
    \leq  & ~ \sqrt{d_0+d_y}\epsilon.
\end{align*}
So we obtain
\begin{align*}
    \norm{q\qty(\Bar{h}, y, t) - \bar{f}\qty(\Bar{h}, y, t)}_{L^2\qty(P_t^{h})}\leq \qty(\sqrt{d_0+d_y}+1)\epsilon.
\end{align*}
Substituting $\epsilon$ with $\epsilon/2$ gives an approximation error for $\bar f(\Bar{h}, y, t)$ of $\sqrt{d_0 + d_y} \epsilon$.

\end{itemize}

\textbf{Step 2. Approximate  $\Bar{f}(\Bar{h}, y,t)$ with One-Layer Self-Attention.}
This step is based on the universal approximation of single-layer single-head transformers for compact-supported continuous function in \cref{thm:Transformer_as_universal_approximators}.

Recall  the reshape layer $\tildeR(\cdot)$ from \cref{def:reshape_layer}.
We use $f(\cdot): ={\tildeR ^{-1}\circ \hat{g}_{\calT} \circ \tildeR}(\cdot)$ to approximate \blue{$\Bar{f}_t(\cdot): =\Bar{f}(\cdot,t)$}, where $\hat{g}_{\calT}(\cdot)\in\calT^{\textcolor{blue}{h,s,r}} = 
\{f^{(\text{FF})}_{2}
\circ f^{(\text{SA})}
\circ f^{(\text{FF})}_{1}:
\R^{\tilded \times \tildeL}\rightarrow\R^{\tilded\times \tildeL}\}$.

We first use $\hat{f}_t(\cdot):={\tildeR^{-1}\circ \hat{g}_{\calT} \circ \tildeR}(\cdot)$ to approximate the function $\Bar{f}_t(\cdot)$ constructed at Step 1 and denote $H=R(\bar{h})$.
Using \cref{thm:Transformer_as_universal_approximators}, we have:
\begin{align}
\label{eqn:app_error}
    \norm{\Bar{f}_t(\Bar{h}, y)-\hat{f}(\Bar{h}, y)}_{L^2(P_t^{h})}
    = &~ \left(\int_{P_t^{h}} \norm{\Bar{f}_t(\Bar{h}, y)-\hat{f}(\Bar{h}, y)}_2^2 \dd  h\right)^{1/2} 
    \\ \nonumber
    = &~ \left(\int_{P_t^{h}} \norm{\tildeR \circ \Bar{f}_t \circ \tildeR^{-1}(H) - \tildeR \circ \hat{g}_{\calT}\circ \tildeR^{-1}(H)}_F^2 \dd  h\right)^{1/2} \nonumber
    \\
    = &~ \left(\int_{P_t^{h}} \norm{\tildeR\circ \Bar{f}_t \circ \tildeR^{-1}(H) -\hat{g}_{\calT}(H)}_F^2 \dd  h\right)^{1/2} \nonumber
    \\
    \leq &~ \epsilon.
\end{align}
Along with Step 1, we obtain
\begin{align*}
     \norm{q(\Bar{h}, y,t)-\hat{f}(\Bar{h}, y)}_{L^2(P_t^{h})}
     \leq &~ \norm{q(\Bar{h},y,t)-\Bar{f}(\Bar{h},y,t)}_{L^2(P_t^{h})}+\norm{\Bar{f}(\Bar{h},y,t) - \hat{g}_{\calT}(\Bar{h}, y)}_{L^2(P_t^{h})} \\
     \leq &~ \qty(1+\sqrt{d_0+d_y})\epsilon.
\end{align*}
\blue{The approximator $s_{\hat{W}}$ for the score function $\nabla \log p_t(\bar{h}|y)$ is define in \eqref{eqn:score_network_S_formal} where $s_{\hat{W}}=(W_U\hat{f}(U^\top x, y,t)-x)/\sigma_t^2$.}
The approximation error for such an approximator is
\begin{align*}
    \norm{\nabla \log p_t(\cdot)-s_{\hat{W}}(\cdot,t)}_{L^2(P_t)}\leq \frac{1+\sqrt{d_0+d_y}}{\sigma_t^2}\epsilon,\quad \text{for all } t\in[t_0,T].
\end{align*}

Finally, 
the parameter bounds in the transformer network class satisfy

\blue{
\begin{align*}
& 
\norm{W_{Q}}_{2}=\norm{W_{K}}_{2}
=
\calO\qty(\tilded\cdot\epsilon^{-(\frac{1}{\tilded}+2\tildeL)}(\log{\tildeL})^{\frac{1}{2}});\\
&\norm{W_{Q}}_{2,\infty}=\norm{W_{K}}_{2,\infty}
=
\calO\qty(\tilded^{\frac{3}{2}}\cdot\epsilon^{-(\frac{1}{\tilded}+2\tildeL)}(\log{\tildeL})^{\frac{1}{2}}); \\
& \norm{W_O}_{2}=\calO\left({\tilded}^{\frac{1}{2}}\epsilon^{\frac{1}{\tilded}}\right); 
\norm{W_O}_{2,\infty}=\calO\left(\epsilon^{\frac{1}{\tilded}}\right);\\
&\norm{W_{V}}_{2}=\calO({\tilded}^{\frac{1}{2}});
\norm{W_{V}}_{2,\infty}=\calO(\tilded); \\ 
& \norm{W_{1}}_{2}=\calO\left(\tilded\epsilon^{-\frac{1}{\tilded}}\right), \norm{W_{1}}_{2,\infty}=\calO\left({\tilded}^{\frac{1}{2}}\epsilon^{-\frac{1}{\tilded}}\right);\\
& \norm{W_{2}}_{2}=\calO\left(\tilded\epsilon^{-\frac{1}{\tilded}}\right);
\norm{W_{2}}_{2,\infty}=\calO\left({\tilded}^{\frac{1}{2}}\epsilon^{-\frac{1}{\tilded}}\right);\\
&\norm{E^{\top}}_{2,\infty}=\calO\left(\tilded^{\frac{1}{2}}\tildeL^{\frac{3}{2}}\right).
\end{align*}
}
We refer to \cref{sec:appendix_transformer_parameter} for the calculation of the hyperparameters configuration of this network.

This completes the proof.
\end{proof}

\subsection{Proof of Score Estimation (\texorpdfstring{\cref{thm:latent_score_est}}{})}\label{sec:appendix_latent_score_est_lipschitz_proof}

\begin{lemma}[Lemma 15 of \cite{chen2023score}]
\label{lemma:chen_lemma15}
    Let $\mathcal{G}$ be a bounded function class, i.e., there exists a constant $b$ such that any function $g\in \mathcal{G}:\RR^{d_0}\mapsto[0,b]$. 
    Let $z_1,z_2,\cdots,z_n\in \RR^{d_0}$ be i.i.d. random variables. 
    For any $\delta\in(0,1),a\leq 1$, and $c >0$, we have
    \begin{align*}
         &P\left(\sup_{g\in\mathcal{G}}\frac{1}{n}\sum_{i=1}^n g(z_i)-(1+a)\EE\left[g(z)\right]>\frac{(1+3/a)B}{3n}\log\frac{\mathcal{N}(c,\mathcal{G},\norm{\cdot}_\infty)}{\delta}+(2+a)c\right)\leq \delta, \\
         &P\left(\sup_{g\in\mathcal{G}}\EE\left[g(z)\right]-\frac{1+a}{n}\sum_{i=1}^n g(z_i)>\frac{(1+6/a)B}{3n}\log\frac{\mathcal{N}(c,\mathcal{G},\norm{\cdot}_\infty)}{\delta}+(2+a)c\right)\leq \delta .
    \end{align*}
\end{lemma}

\paragraph{Main Proof of \cref{thm:latent_score_est}.}
Now we are ready to state the main proof.

\begin{proof}[Proof of \cref{thm:latent_score_est}]
Our proof is built on \cite[Appendix B.2]{chen2023score}.

\blue{
Recall that the empirical score-matching loss is
\begin{align}
    \mathcal{L}(s_{\hat{W}}) = \frac{1}{n} \sum_{i=1}^{n} \ell(x_i, y_i; s_{\hat{W}}),
\end{align}
with the loss function $\ell$ for a data sample $(x,y)$ is defined as
\begin{align*}
\ell(x,y,s_{\hat{W}}) = \int_{t_{0}}^{T}\frac{1}{T-t_{0}}\E_{(x_{t}|x_{0}=x,\tau)}\left[\norm{s(x_{t},\tau y,t)-\nabla\log{\phi_{t}(x_{t}|x_{0})}}_{2}^{2}\right] \dd t.
\end{align*}
We organize the proof into the following three steps:}
\blue{
\begin{itemize}
    \item \textbf{Step 1. Decomposing $\calL\left( s_{\hat{W}}\right)$:} 
    We first decompose $\calL$ into three terms $(A)$, $(B)$, and $(C)$.
    \item \textbf{Step 2. Bounding Each Term:} We then bound three terms separately using some helper from \cref{lemma:Tool_lemma_1_t1} and \cref{lemma:chen_lemma15}.
    \item \textbf{Step 3. Putting All Together:} Finally, we combine the above bounds and substitute the covering number of $\calS \left( C_x \right)$ from \cref{lemma:covering_number_truncated_loss}.
\end{itemize} 
}

\begin{itemize}
\item \textbf{Step 1. Decomposing $\calL\left( s_{\hat{W}}\right)$:}

Following \cite[Appendix B.2]{chen2023score}, for any $a\in (0,1)$, we have:
\begin{align*}
& ~ \calL(s_{\hat{W}}) \\
\leq & ~
\underbrace{\calL^{\rm trunc}(s_{\hat{W}}) - (1+a)\hat{\calL}^{\rm trunc}(s_{\hat{W}})}_{(A)} + \underbrace{\calL(s_{\hat{W}}) - \calL^{\rm trunc}(s_{\hat{W}})}_{(B)} + (1+a) \underbrace{\inf_{s_{W} \in {\calT^{\textcolor{blue}{h,s,r}}_{\tildeR}}} \hat{\calL}(s_{W})}_{(C)}.
\end{align*}
where
\begin{align*}
\calL^{\rm trunc}(s_{\hat{W}}) \coloneqq 
 \EE_{ x  \sim P_{0}}  \qty[\ell( x,\tau y,s_{\hat{W}})\mathds{1}\{\norm{ x }_2 \leq r_x\} ], \quad r_x > B,
\end{align*}
We denote 
\begin{align*}
    \eta & \coloneqq ~ 4C_{\calT}(C_{\calT}+r_x) \qty({r_x}/{d_x} )^{d_x-2} \cdot {\exp(- r_x^2/\sigma_t^2)}/{t_0(T - t_0)},\\
    r_x & \coloneqq ~ \calO\qty( \sqrt{d_0\log d_0 + \log C_{\calT} + \log (n/\Bar{\delta})} ).
\end{align*}

\item \textbf{Step 2. Bounding Each Term:}
We bound ${(A)}$, ${(B)}$, and ${(C)}$ term separately using some helper from \cref{lemma:Tool_lemma_1_t1} and \cref{lemma:chen_lemma15}.

\textbf{Bounding term $(A)$.} For any $\Bar{\delta} >0$, following \cite[Appendix B.2]{chen2023score} and applying \cref{lemma:chen_lemma15}, we have the following for term $(A)$ with probability $1-\Bar{\delta}$,
\begin{align*}
(A) = \calO\left(\frac{(1+3/a)(C_{\calT}^2 + r_x^2)}{n t_0(T-t_0)} \log \frac{\calN\left(\frac{(T-t_0)(\epsilon_{c}-\eta)}{(C_{\calT} + r_x)\log (T/t_0)}, {\calT^{\textcolor{blue}{h,s,r}}}, \norm{\cdot}_2 \right)}{\Bar{\delta}} + (2+a)c\right),
\end{align*}
where $c \leq 0$ is a constant, and $\epsilon_{c} >0$ is another constant to be determined later.

By setting  $ \epsilon_{c} = \log( {2}/{\qty(n t_0\qty(T-t_0))})$,
then we have
\begin{align}\label{eqn:score_decomp_A}
(A) = \calO\qty(\frac{\qty(1+3/a) \qty(C_{\calT}^2 + r_x^2)}{n t_0 \qty(T-t_0)} \log \frac{\calN\qty(\qty(n \qty(C_{\calT} + r_x)t_0\log \qty(T/t_0))^{-1}, {\calT^{\textcolor{blue}{h,s,r}}}, \norm{\cdot}_2 )}{\Bar{\delta}} + \frac{1}{n}),
\end{align}
with probability $1-\Bar{\delta}$. 

\textbf{Bounding term $(B)$.}
Following \cite[Appendix B.2]{chen2023score} and applying \cref{lemma:Tool_lemma_1_t1}, we has the following bound for term $(B)$:
\begin{align}\label{eqn:score_decomp_B}
   (B) = \calO\left(\frac{1}{t_0(T - t_0)} C_{\calT}^2 r_x^{d_0} \frac{2^{-2/d_{0}+2}d_0}{\Gamma(d_{0}/2+1)} \exp\left(-C_2 r_{x}^2 / 2\right)\right).
\end{align}

\textbf{Bounding term $(C)$.} In \cref{theorem:latent_lipschitz_formal}, we approximate the score function with the network $\hat{s}_{W}$ for any $\epsilon > 0$.
We decompose the term $(C)$ into statistical error $(C_1)$ and approximation error $(C_2)$:
\begin{align*}
(C) \leq \underbrace{\hat{\calL}(\hat{s}_{W}) - (1+a)\calL^{\rm trunc}(\hat{s}_{W})}_{(C_1)} + (1+a) \underbrace{\calL^{\rm trunc}(\hat{s}_{W})}_{(C_2)}.
\end{align*}
Following \cite[Appendix B.2]{chen2023score} and applying \cref{lemma:Tool_lemma_1_t1} and \cref{lemma:chen_lemma15}, we have the following bound for term $(C_1)$:
\begin{align*}
(C_1) = 
\hat{\calL}^{\rm trunc}(\hat{s}_{W}) - (1+a)\calL^{\rm trunc}(\hat{s}_{W}) = \calO \left(\frac{(1+6/a)(C_{\calT}^2 + r_x^2)}{n t_0(T-t_0)} \log \frac{1}{\bar{\delta}}\right),
\end{align*}
with probability $1 - \delta$.

Finally, for the term $(C_2)$ we use \cref{theorem:latent_lipschitz_formal} for score function approximation of $\calL(\hat{s}_{W})$:
\begin{align*}
(C_2) = \calO\left(\frac{d_0+d_y}{t_0(T-t_0)} \epsilon^2\right) + ({\rm const.}).
\end{align*}
This give us the bound for term $(C) \le (C_1) + (1+a)(C_2)$ as
\begin{align}\label{eqn:score_decomp_C}
    (C)\leq  \calO 
    \left(
    \frac{(1+6/a)(C_{\calT}^2 + r_x^2)}{n t_0(T-t_0)} \log \frac{1}{\bar{\delta}}
    + \frac{d_0+d_y}{t_0(T-t_0)} \epsilon^2
    \right)
    + ({\rm const.}).
\end{align}

\item \blue{\textbf{Step 3. Putting All Together:}
In the final steps, we combine three terms and substitute the covering number to get the score estimation bound for latent DiT.}

\textbf{Combining $(A)$, $(B)$ and $(C)$.} Following \cite[Appendix B.2]{chen2023score}, we set $a=\epsilon^2$ and get the overall bound:
{
\begin{align}\label{eqn:bound_with_covering}
&~ \frac{1}{T-t_0} \int_{t_0}^T \norm{s_{\hat{W}}(\cdot, t) - \nabla \log p_t(\cdot)}_{L^2(P_t)}^2 \dd  t \nonumber 
\\
= &~ \calO\left(\frac{\left(C_{\calT}^2 + r_x^2\right)}{\epsilon^2 n t_0(T - t_0)}\log \frac{\calN\left((n(C_{\calT} + r_x)t_0\log (T/t_0))^{-1}, \calS_{\calT^{\textcolor{blue}{h,s,r}}}, \norm{\cdot}_2 \right)}{\Bar{\delta}} + \frac{1}{n} + \frac{d_0+d_y}{t_0(T-t_0)} \epsilon^2 \right),
\end{align}
}
with probability $1 - 3\Bar{\delta}$.

\blue{
Before we move on to the covering number of $\calT^{\textcolor{blue}{h,s,r}}_{\tildeR}$, we first compute the Lipschitz upper bound $L_{\calT}$ and model output bound $C_{\calT}$.}

\textbf{Lipschitz Upper Bound $L_{\calT}$ and Model Output Bound $C_{\calT}$.}
We then compute the Lipschitz upper bound $L_{\calT}$ for the transformer.
We denote $\Bar{f}_{t,R}(\cdot) = \tildeR \circ \hat{g}_t \circ \tildeR^{-1}(\cdot)$ and $H=\left( \tildeR(\Bar{h}), y\right)$.
We get the Lipschitz upper bound for $\hat{f}_{\calT} \in \calT^{\textcolor{blue}{h,s,r}}_{\tildeR}$:
\begin{align*}
\left\|\hat{f}_{\calT}\left(H_{1}\right)-\hat{f}_{\calT}\left(H_{2}\right)\right\|_{F}
\leq &~ \left\|\hat{f}_{\calT}\left(H_1\right)-\Bar{f}_{t,\tildeR}\left(H_1\right)\right\|_{F} + \left\|\Bar{f}_{t,\tildeR}\left(H_1\right)-\Bar{f}_{t,\tildeR}\left(H_2\right)\right\|_{F} 
\\ 
 &~ +
\left\|\Bar{f}_{t,\tildeR}\left(H_2\right)-\hat{f}_{\calT}\left(H_2\right)\right\|_{F} 
\\
\leq &~ 2\epsilon+\left\|\Bar{f}_{t,\tildeR}\left(H_1\right)-\Bar{f}_{t,\tildeR}\left(H_2\right)\right\|_{F} \annot{By \eqref{eqn:app_error}} 
\\
\leq &~ 2\epsilon+10 (d_{0} + d_{y}) L_{s_{+}}\norm{H_1 - H_2}_{F}. \annot{By \eqref{eqn:lip_h}}
\end{align*}
Then we get the upper bound of Lipschitzness of $\calT^{\textcolor{blue}{h,s,r}}_{\tildeR}$: 
\begin{align}
\label{eq:L_tau_est}
    L_{\calT}=\calO\left( \left(d_{0}+d_{y}\right)L_{s_{+}}\right).
\end{align}

Next, we compute the model output bound for $\calT^{\textcolor{blue}{h,s,r}}_{\tildeR}$.
For the output of the constructed transformer $\hat{f}_{\calT} \in \calT^{\textcolor{blue}{h,s,r}}$, according to \eqref{eqn:bump_function_def}, the output of the network is lower bounded by $\calO(1)$. 
Thus with the Lipschitz upper bound $L_\calT= \calO((d_0+d_y) L_{s_{+}})$, we have $\|\hat{f}_{\calT}(H)\|_F = \calO((d_0+d_y) L_{s_{+}}r_h)$, where $\norm{H}_F \leq r_h$.
With $r_h=c(\sqrt{d_0\log (d_0/t_0) + \log (1/ \epsilon ) })$, we obtain
\begin{align}
\label{eq:C_tau_est}
    C_{\calT}=\calO\left((d_{0}+d_{y}) L_{s_{+}} \cdot \sqrt{d_0\log (d_0/t_0)+\log (1/\epsilon)}\right).
\end{align}

\textbf{Covering Number of ${\calT^{\textcolor{blue}{h,s,r}}_{\tildeR}}$.} 
The next step is to calculate the covering number of ${\calT^{\textcolor{blue}{h,s,r}}_{\tildeR}}$. 
In particular, ${\calT^{\textcolor{blue}{h,s,r}}_{\tildeR}}$ consists of two components: 
(i) Matrix $W_U$ with orthonormal columns; 
(ii) Network function $g_{\calT}$. 
Suppose we have $W_{U1}, W_{U2}$ and $g_{1}, g_{2}$ such that $\norm{W_{U1} - W_{U2}}_{F} \leq \delta_1$ and $\sup_{\norm{ x }_2 \leq 3r_x+ \sqrt{d_x\log d_x}, y\in [0,1], t \in [t_0, T]} \norm{g_{1}( x, y , t) - g_{2}( x, y , t)}_2 \leq \delta_2$, where $g_{1}=\tildeR^{-1} \circ g_{\calT 1} \circ \tildeR$ and $g_{2}=\tildeR^{-1} \circ g_{\calT 2} \circ \tildeR$.
Then we evaluate
\begin{align}
\label{eqn:lip_score_cover}
& \quad \sup_{\norm{ x }_2 \leq 3r_x + \sqrt{d_x\log d_x}, y\in [0,1], t \in [t_0, T]} \norm{s_{W_{U1}, g_{\calT 1}}( x , y , t) - s_{W_{U2}, g_{\calT 2}}( x , y , t)}_2 \nonumber
\\
= &~ \frac{1}{\sigma_t^2} \sup_{\norm{ x }_2 \leq 3r_x + \sqrt{d_x\log d_x}, y \in [0,1], t \in [t_0, T]} \norm{W_{U1} g_{1}(W_{U1}^\top  x , y , t) - W_{U2}g_{2}(W_{U2}^\top  x , y , t)}_2 \nonumber
\\
\leq &~ \frac{1}{\sigma_t^2} \sup_{\norm{ x }_2 \leq 3r_x + \sqrt{d_x\log d_x}, y \in [0,1], t \in [t_0, T]} \Bigg(
\underbrace{\norm{W_{U1} g_{1}(W_{U1}^\top  x , y , t) - W_{U1} g_{1}(W_{U2}^\top  x , y , t)}_2}_{\rm 1^{st}~term} \nonumber
\\
&~ + \underbrace{\norm{W_{U1} g_{1}(W_{U2}^\top  x , y , t) - W_{U1} g_{2}(W_{U2}^\top  x , y , t)}_2}_{\rm 2^{nd}~term} 
+ \underbrace{\norm{W_{U1} g_{2}(W_{U2}^\top  x , y , t) - W_{U2} g_{2}(W_{U2}^\top  x , y , t)}_2}_{\rm 3^{rd}~term} \Bigg) \nonumber
\\
\leq &~ \frac{1}{\sigma_t^2} 
\left( \underbrace{L_{\calT} \delta_1 
\sqrt{d_0}(3r_x + \sqrt{d_x\log d_x})}_{\rm 1^{st}~term} + \underbrace{\delta_2}_{\rm 2^{nd} term} + \underbrace{\delta_1}_{\rm 3^{rd}~term} \right),
\end{align}
where $L_{\calT}$ upper bounds the Lipschitz constant of $g_{\calT}$ (see \eqref{eq:L_tau_est}). 

For the set $\{W_B \in \RR^{d_x \times d_0}: \norm{W_B}_{\rm 2} \leq 1\}$, its $\delta_1$-covering number is $\left(1 + 2 \sqrt{d_0}/\delta_1\right)^{{d_x}{d_0}}$ \citep[Lemma 8]{chen2023score}. 
The $\delta_2$-covering number of $f$ needs further discussion as there is a reshaping process in our network. 
For the input reshaped from $ \Bar{h} \in \RR^{d_0}$ to $H \in \RR^{\tilded\times \tildeL}$, we have 
\begin{align*}
    \norm{ \Bar{h} }_2 \leq r_x \Longleftrightarrow \norm{H}_{F} \leq r_x,
\end{align*}
Thus we have
\begin{align*}
    &~\sup_{\norm{ \Bar{h} }_2 \leq 3r_x+ \sqrt{D\log D}, y \in [0, 1], t \in [t_0, T]} \norm{g_{1}( \Bar{h}, y , t) - g_{2}( \Bar{h}, y , t)}_2 \leq \delta_2,
\end{align*}
\begin{align*}
    \Longleftrightarrow &~
    \sup_{\norm{H}_F \leq 3r_x + \sqrt{D\log D}, y \in [0, 1], t \in [t_0, T]} \norm{g_{\calT 1}(H) -g_{\calT 2}(H)}_2 \leq \delta_2.
\end{align*}
Next we follow the covering number property for sequence-to-sequence transformer $\calT^{\textcolor{blue}{h,s,r}}_{\tildeR}$, i.e., \cref{lemma:covering_number} and get the following $\delta_2$-covering number
{\footnotesize
\begin{align}\label{eqn:delta2_covering_number}
& ~ \log\mathcal{N}\left(\epsilon_c,\calT^{\textcolor{blue}{h,s,r}}_{\tildeR},\norm{\cdot}_2\right)\\
\leq & ~ 
\frac{\log (nL)}{\epsilon_c^2}\cdot\alpha^{2}\left(
\underbrace{\left((C_F)^2 C_{OV}^{2,\infty}\right)^{\frac{2}{3}}}_{\rm 1^{st}~term}
+\underbrace{(d+d_{y})^{\frac{2}{3}}\left(C_F^{2,\infty}\right)^{\frac{4}{3}}}_{\rm 2^{nd}~term}
+\underbrace{(d+d_{y})^{\frac{2}{3}}\left(2(C_F)^2 C_{OV} C_{KQ}^{2,\infty}\right)^{\frac{2}{3}}}_{\rm 3^{rd}~term}
\right)^3,
\end{align}
}
where 
\begin{align*}
    \alpha \coloneqq \prod_{j<i} (C_F)^2 C_{OV} (1+4C_{KQ})(C_X+C_E).
\end{align*}

Recall that from the network configuration in \cref{theorem:latent_lipschitz_formal}, we have the following bound:

\blue{
\begin{align*}
& 
\norm{W_{Q}}_{2}=\norm{W_{K}}_{2}
=
\calO\qty(\tilded\cdot\epsilon^{-(\frac{1}{\tilded}+2\tildeL)}(\log{\tildeL})^{\frac{1}{2}});\\
&\norm{W_{Q}}_{2,\infty}=\norm{W_{K}}_{2,\infty}
=
\calO\qty(\tilded^{\frac{3}{2}}\cdot\epsilon^{-(\frac{1}{\tilded}+2\tildeL)}(\log{\tildeL})^{\frac{1}{2}}); \\
& \norm{W_O}_{2}=\calO\left({\tilded}^{\frac{1}{2}}\epsilon^{\frac{1}{\tilded}}\right); 
\norm{W_O}_{2,\infty}=\calO\left(\epsilon^{\frac{1}{\tilded}}\right);\\
&\norm{W_{V}}_{2}=\calO({\tilded}^{\frac{1}{2}});
\norm{W_{V}}_{2,\infty}=\calO(\tilded); \\ 
& \norm{W_{1}}_{2}=\calO\left(\tilded\epsilon^{-\frac{1}{\tilded}}\right), \norm{W_{1}}_{2,\infty}=\calO\left({\tilded}^{\frac{1}{2}}\epsilon^{-\frac{1}{\tilded}}\right);\\
& \norm{W_{2}}_{2}=\calO\left(\tilded\epsilon^{-\frac{1}{\tilded}}\right);
\norm{W_{2}}_{2,\infty}=\calO\left({\tilded}^{\frac{1}{2}}\epsilon^{-\frac{1}{\tilded}}\right);\\
&\norm{E^{\top}}_{2,\infty}=\calO\left(\tilded^{\frac{1}{2}}\tildeL^{\frac{3}{2}}\right).
\end{align*}
}

Note that $W_{K,Q}=W_{Q}W_{K}^{\top}$ and $W_{O,V}=W_OW_{V}^{\top}$.
\blue{
Combining every component and substitute into \eqref{eqn:delta2_covering_number}, we have three respective terms bounded as
}

\blue{
\begin{align*}
{\rm 1^{st}~term} &= \mathcal{O}\qty(\tilded^{2} \epsilon^{-{2}/\qty({3\tilded})}), \\
{\rm 2^{nd}~term} &= \mathcal{O}\qty( \qty(d_0 + d_y)^{2/3} \tilded^{{2}/{3}} \epsilon^{-4 / \qty(3\tilded)} ), \\
{\rm 3^{rd}~term} &= \mathcal{O}\qty( \qty(d_0 + d_{y})^{{2}/{3}} \cdot \qty(\log{\tildeL})^{{2}/{3}} \cdot \tilded^{4} \cdot \epsilon^{\qty(-2/3)\qty(3/\tilded +4\tildeL)} ).
\end{align*}
}
Apparently the ${\rm 3^{rd}~term}$ dominates the other two.
For the $\alpha^2$ term,
we write

\blue{
\begin{align*}
\alpha^{2}=\calO\qty( \tilded^{10} \epsilon^{-2\qty(3/\tilded+4\tildeL)} \qty(\log{\tildeL}) C_{x}^{\prime} ),
\end{align*}
}
where $C_{x}^{\prime}=\left(C_{x}+(d_0+d_{y})^{3/2}\right)^{2}$.

Combining the above bound we get the log-covering number of $\calT_{2}$ as

\blue{
\begin{align}
\label{eqn:covering_number_lipschitz}
\log\mathcal{N}\left(\epsilon_c,\calT^{\textcolor{blue}{h,s,r}}_{\tildeR},\norm{\cdot}_2\right)\lesssim\calO\left(
\frac{\log{(n\tildeL)\log^{3}{(\tildeL)}}}{\epsilon_{c}^{2}}
\tilded^{22} (d_0+d_y)^2 \epsilon^{-4\qty(3/\tilded+4\tildeL)} C_{x}^{2}
\right).
\end{align}
}

\blue{
Substituting the log-covering number of ${\calT^{\textcolor{blue}{h,s,r}}_{\tildeR}}$ into \eqref{eqn:bound_with_covering}, we have
\begin{align*}
& ~ \frac{1}{T-t_0} \int_{t_0}^T \norm{s_{\hat{W}}(\cdot, t) - \nabla \log p_t(\cdot)}_{L^2(P_t)}^2 \dd  t 
\\
= & ~ \calO \left( \frac{\left(C_{\calT}^2 + \log (n/\Bar{\delta})\right)}{\epsilon^2 n t_0(T - t_0)}
\left( \frac{\log{(n\tildeL)\log^{3}{(\tildeL)}}}{(T-t_0) n^2}
\tilded^{22} (d_0+d_y)^2 \epsilon^{-4\qty(3/\tilded+4\tildeL)} C_{x}^{2} \right) 
+ \frac{1}{n} + \frac{d_0+d_y}{t_0(T-t_0)} \epsilon^2 \right) 
\annot{By \eqref{eqn:bound_with_covering}}
\\
= & ~ \calO \Biggl( \frac{ ( (\tilded+d_0)^2 L_{s_{+}}^2 (d_0\log (d_0/t_0)+\log (1/\epsilon)) + \log (n/\Bar{\delta}) )}{\epsilon^2 n t_0(T - t_0)} \left( \frac{\log{(n\tildeL)\log^{3}{(\tildeL)}}}{(T-t_0) n^2}
\tilded^{22}(\tilded+d_y)^2 \epsilon^{-4\qty(3/\tilded+4\tildeL)} C_{x}^{2} \right) 
\\
& \quad \quad + \frac{d_0+d_y}{t_0(T-t_0)} \epsilon^2
 \Biggl).
\annot{By \eqref{eq:L_tau_est} and \eqref{eq:C_tau_est}}
\end{align*}
}

\textbf{Balancing Error Terms.}
To balance the error term, we set \blue{$\epsilon=n^{{-3}/4{\qty(1+3/\tilded+4\tildeL)}}$.}
Also setting $\bar{\delta}={1}/{3n}$ then we have
\blue{
\begin{align}
\frac{1}{T-t_0} \int_{t_0}^T \norm{s_{\hat{W}}(\cdot, t) - \nabla \log p_t(\cdot)}_{L^2(P_t)}^2 \dd  t 
= \calO \left( \frac{\tilded^{22} (\tilded+d_0)^2 (\tilded+d_y)^2 }{t_0^2} 
n^{\frac{-3}{2{\qty(1+3/\tilded+4\tildeL)}}}
\log^{3}\tildeL \log^{3}n
\right)
\label{eqn:above_eqn}
\end{align}
}
with probability of $1-\frac{1}{n}$.

\end{itemize}
This completes the proof.
\end{proof}

\subsection{Proof of Distribution Estimation (\texorpdfstring{\cref{thm:latent_dist_est_formal}}{})}\label{sec:appendix_latent_dist_est_lipschitz_proof}
Our proof is built on \citet[Appendix C]{chen2023score}. 
The main difference between our work and \citet{chen2023score} is our score estimation error from \cref{thm:latent_score_est}. 
\blue{
This is based on our universal approximation of transformers in \cref{cor:transformer_class}.
}
Consequently, only the subspace error and the total variation distance differ from \citet[Theorem 3]{chen2023score}.

\paragraph{Proof Sketch of (i).}
We show that if the orthogonal score increases significantly, the mismatch between the column span of $U$ and $W_U$ will be greatly amplified. 
Therefore, an accurate score network estimator forces $U$ and $W_U$ to align with each other.

\paragraph{Proof Sketch of (ii).} 
We conduct the proof via 2 steps:
\begin{itemize}
\item \textbf{Step 1: Total Variation Distance Bound.} We obtain the discrete result from the continuous-time generated distribution $\hat{P}_{t_0}$ by adding discretization error \cite[Lemma 4]{chen2023score}. It suffices to bound the divergence between the following two stochastic processes:
\begin{itemize}
    \item 
    For the ground-truth backward process, consider $h_t^{\leftarrow} = B^\top y_t$ and the following SDE:
    \begin{align*}
    \dd  h_t^{\leftarrow} = \left[\frac{1}{2} h_t^{\leftarrow} + \nabla \log p^{h}{T-t}(h_t^{\leftarrow})\right] \dd  t + \dd  \Bar{U}_t^h.
    \end{align*}
    Denote the marginal distribution of the ground-truth process as $P_{t_0}^{h}$.
    
    \item 
    For the learned process, consider ${\tilde h}^{\leftarrow, r}_{t}$ and the following SDE:
    \begin{align*}
    \dd  {\tilde h}^{\leftarrow, r}_{t} = \left[\frac{1}{2} {\tilde h}^{\leftarrow, r}_{t} + \tilde{s}^h_{f, M}({\tilde h}^{\leftarrow, r}_{t}, T-t)\right] \dd t + \dd  \Bar{U}^h_t,
    \end{align*}
    where $\tilde{s}_{f, M}^h(z, t) \coloneqq [M^\top f(M z, t) - z]/\sigma_t^2$ and $M$ is an orthogonal matrix.
    Following the notation in \cite{chen2023score}, we use $(W_U M)_{\sharp}^\top \hat{P}_{t_0}$ to denote the marginal distribution of $\hat{P}_{t_0}$.
    We first calculate the latent score matching error, i.e., the error between $\nabla \log p^h_{t}(h,y)$ and $\tilde{s}_{M, f}^h(h, y, t)$. 
    Then, we adopt Girsanov's Theorem \citep{chen2022sampling} and bound the difference in the KL divergence of the above two processes to derive the score-matching error bound.
    
\end{itemize}

\end{itemize}

\paragraph{Proof Sketch of (iii).}
We derive item (iii) by solving the orthogonal backward process of the diffusion model.

\begin{definition}
For later convenience, let us define \blue{ $\xi(n,t_0,\tilded,\tilde{L}):=\frac{1}{t_0^2}  n^{\frac{-3}{2{\qty(1+3/\tilded+4\tildeL)}}} \log^{3}n$.}
\end{definition}

Here we include a few auxiliary lemmas from \citet{chen2023score} without proofs.
Recall the definition of Lipschitz norm: for a given function $f$, $\norm{f(\cdot)}_{Lip} = \sup_{x\neq y}(\norm{f(x)-f(y)}_2/\norm{x-y}_2)$.
\begin{lemma}[Lemma 3 of \citet{chen2023score}]
\label{lemma_3_used}
    Assume that the following holds
    \begin{align*}
        \EE_{h\sim P_h}\norm{\nabla \log p_h(h|y)}_2^2\leq C_{sh},\quad
        \lambda_{\rm min}\EE_{h\sim P_h}[hh^\top]\geq c_0,\quad
        \EE_{h\sim P_h}\norm{h}_2^2\leq C_{h},
    \end{align*}
    where $\lambda_{\rm min}$ denotes the smallest eigenvalue.
    We denote
    \begin{align*}
        \Bar{\EE}[\phi(\cdot,t)] = \int_{t_0}^{T}\frac{1}{\sigma^4_t}\EE_{x \sim P_t}[\phi(\cdot, t)]dt.
    \end{align*}
    
    We set $t_0\leq \min \{2 \log(d_0/C_{sh}),1, 2 \log(c_0),c_0 \}$ and $T\geq \max\{ 2 \log(C_{h}/d_0),1\}$. 
    Suppose we have
    \begin{align*}
        \Bar{\EE}\norm{W_B f(W_B ^\top x, y ,t)-U q(B^\top x, y ,t)}_2^2\leq \epsilon.
    \end{align*}
    Then we have 
    \begin{align*}
        \norm{W_UW_U^\top-UU^\top}_{\rm F}^2=\calO(\epsilon t_0 /c_0),
    \end{align*}
    and there exists an orthogonal matrix $M\in \RR^{d_0\times d_0}$, such that:
    \begin{align*}
        &\quad\Bar{\EE}\norm{M^\top f(Mh,y,t)-q(h,y,t)}_2^2\\
        &=\epsilon\cdot\calO
        \qty(
        1+\frac{t_0}{c_0}\left[(T-\log t_0)d_0\cdot \max_{t}\norm{f(\cdot,t)}_{\rm Lip}^2+C_sh\right]+\frac{\max_t\norm{f(\cdot,t)}_{\rm Lip}^2\cdot C_{h} }{c_0}
        ).
    \end{align*}
\end{lemma}

\begin{lemma}[Lemma 4 of \citet{chen2023score}]
\label{lemma_4_used}
Assume that $P_h$ is sub-Gaussian, $f(h, y, t)$ and $\nabla \log p_t^{h}(h|y)$ are Lipschitz in both $h$, $y$ and $t$. 
Assume we have the latent score matching error-bound
\begin{align*}
\int_{t_0}^T \mathbb{E}_{h \sim P_t^{h}}\left\|\widetilde{s}_{M, f}^{h}\left(h_t, y, t\right)-\nabla \log p_t^{h}\left(h_t | y\right)\right\|_2^2 \mathrm{~d} t \leq \epsilon_{\text {latent }}(T-t_0).
\end{align*}
Then we have the following latent distribution estimation error for the undiscretized backward SDE
$$
\operatorname{TV}\left(P_{t_0}^{h}, \widehat{P}_{t_0}^{h}\right) \lesssim \sqrt{\epsilon_{\text {latent }}(T-t_0)}+\sqrt{\mathrm{KL}\left(P_h \| N\left(0, I_{d_0}\right)\right)} \cdot \exp (-T).
$$

Furthermore, we have the following latent distribution estimation error for the discretized backward SDE
$$
\operatorname{TV}\left(P_{t_0}^{h}, \widehat{P}_{t_0}^{h, \mathrm{dis}}\right) \lesssim \sqrt{\epsilon_{\text {latent}}(T-t_0)}+\sqrt{\mathrm{KL}\left(P_h \| N\left(0, I_{d_0}\right)\right)} \cdot \exp (-T)+\sqrt{\epsilon_{\text {dis}}(T-t_0)},
$$
where
\begin{align*}
    \epsilon_{\rm d i s}=
    &\left(\frac{\max _{h}\left\|f(h, y, \cdot)\right\|_{\text {Lip }}}{\sigma\left(t_0\right)}+\frac{\max _{h, t}\left\|f(h, y, t)\right\|_2}{t_0^2}\right)^2 \eta^2 \\
    &~ +\left(\frac{\max _t\left\|f(\cdot, y, t)\right\|_{\text {Lip }}}{\sigma\left(t_0\right)}\right)^2 \eta^2 \max \left\{\mathbb{E}\left\|h_0\right\|^2, d_0\right\}+\eta d_0 ,
\end{align*}
and $\eta$ is the step size in the backward process.
\end{lemma}

\begin{lemma}[Lemma 6 of \citet{chen2023score}]
\label{lemma_6_used}
Consider the following discretized SDE with step size $\mu$ satisfying $T-t_0=K_T \mu$
$$
\mathrm{d} y_t=\left[\frac{1}{2}-\frac{1}{\sigma(T-k \mu)}\right] {y}_{k \mu} \mathrm{d} t+\mathrm{d} {U}_t, \text { for } t \in[k \mu,(k+1) \mu),
$$
where ${Y}_0 \sim N(0, I)$.
Then when $T>1$ and $t_0+\mu \leq 1$, we have ${Y}_{T-t_0} \sim N\left(0, \sigma^2 I\right)$ with $\sigma^2 \leq e\left(t_0+\mu\right)$.
\end{lemma}

\begin{lemma}[Lemma 10 in \citet{chen2023score}]
\label{lemma_10_used}
    Assume that $\nabla \log p_h(h|y)$ is $L_{h}$-Lipschitz. 
    Then we have $\mathbb{E}_{h \sim P_h}\left\|\nabla \log p_h(h|y)\right\|_2^2 \leq d_0 L_{h}$.
\end{lemma}

\paragraph{Main Proof of \cref{thm:latent_dist_est_formal}.}
Now we are ready to state the main proof.

\begin{proof}[Proof of \cref{thm:latent_dist_est_formal}]
    \blue{
    Recall that in \eqref{eqn:above_eqn}, we have
    }
    \blue{
    \begin{align*}
    \xi(n,t_0,\tilded,\tildeL):=\frac{1}{t_0^2}  n^{\frac{-3}{2{\qty(1+3/\tilded+4\tildeL)}}} \log^{3}L \log^{3}n.
    \end{align*}
    }
    
    \begin{itemize}
    \item \textbf{Proof of  (i).} 
    With \cref{lemma_3_used}, we replace $\epsilon$ to be $\epsilon(T-t_0)^2$ and we set $C_{sh}=L_{h} d_0$ by 
    \cref{lemma_10_used}, we have
    \begin{align*}
        \norm{W_U W_U^\top - UU^\top}_F^2=\calO\Bigg(\frac{t_0^2\xi(n,t_0,\tilded,\tildeL)}{c_0}\Bigg).
    \end{align*}
    We substitute the score estimation error in \cref{thm:latent_score_est} and $T=\calO(\log n)$ into the bound above, we deduce
    \begin{align*}
        \norm{W_U W_U^\top - UU^\top}_F^2=\tilde\calO\left(\frac{1}{c_0} n^{\frac{-3}{2{\qty(1+3/\tilded+4\tildeL)}}} \cdot \log^{3} n\right).
    \end{align*}
    
    We note that $\log n$ is great enough to make $T$ satisfies $T\geq\max\{\log(C_{h}/d_0+1),1\}$ where $C_{h}\geq \EE_{h\sim P_h}\norm{h}_2^2$.

    \item \textbf{Proof of (ii).}
    \cref{lemma_3_used} and \cref{lemma_10_used} imply that 
    \begin{align*}
        \Bar{\EE}\norm{M^\top f(Mh, y,t)-q(h,y,t)}_2^2=\calO(\epsilon_{\text{latent}}(T-t_0)),
    \end{align*}
    where
    \begin{align*}
        \epsilon_{\text{latent}}=\epsilon\cdot \calO\left(\frac{t_0}{c_0}\left[(T-\log t_0)d_0\cdot L_{s_{+}}^2+d_0L_{h}\right]+\frac{L_{s_{+}}^2\cdot C_{h}}{c_0}\right). 
    \end{align*}
    Through the algebra calculation, we get
    \begin{align*}
        \bar\EE\norm{M^\top f(Mh,y,t)-q(h,y,t)}_2^2 
        = & \int_{t_0}^T \EE_{h\sim P_t^{h}}\norm{\frac{U^\top f(Uh,y,t)-h}{\sigma_t^2}-\nabla\log p_t^{h}(h|y)}_2^2 \dd  t \\
        \leq &~ \epsilon_{\text{latent}}(T-t_0). 
    \end{align*}
    With $\epsilon_{\text{latent}}$ and \cref{lemma_4_used}, we obtain
    \blue{
    \begin{align*}
        & ~ {\rm TV} (P_{t_0}^{h}, (W_UM)^\top_{\sharp} \hat{P}_{t_0}^{\rm dis})\\
        \lesssim & ~  \sqrt{\epsilon_{\text {latent }}(T-t_0)}+\sqrt{\mathrm{KL}\left(P_h \| N\left(0, I_{d_0}\right)\right)} \exp (-T)+\sqrt{\epsilon_{\text {dis }}(T-t_0)}\\
        = & ~  \tilde\calO\left(\frac{1}{t_0\sqrt{c_0}}n^{\frac{-3}{2{\qty(1+3/\tilded+4\tildeL)}}} \cdot \log^{3} n+\frac{1}{n}+\mu\frac{\sqrt{d_0^2\log d_0}}{t_0^2}+\sqrt{\mu}\sqrt{d_0}\right).
    \end{align*}
    }
    As we choose time step \blue{$\mu=\calO \qty( t_0^2/d_0\sqrt{\log d_0} n^{\frac{-3}{4{\qty(1+3/\tilded+4\tildeL)}}})$}, we obtain
    \blue{
    \begin{align*}
        {\rm TV} (P_{t_0}^{h}, (W_U M)^\top_{\sharp} \hat{P}_{t_0}^{\rm dis})= \tilde\calO\left(\frac{1}{t_0\sqrt{c_0}}n^{\frac{-3}{2{\qty(1+3/\tilded+4\tildeL)}}} \cdot \log^{3} n\right).
    \end{align*}
    }
    By definition, $\hat{P}_{t_0}^{h,{\rm dis}}=(UW_B)_{\sharp}^\top \hat{P}_{t_0}^{\rm dis}$. 
    This completes the proof of 
    the total variation distance.

    \item \textbf{Proof of (iii).}
    We apply \cref{lemma_6_used} due to our score decomposition. 
    With the marginal distribution at time $T-t_0$ and observing $\mu\ll t_0$, we obtain the last property.
    \end{itemize}
    This completes the proof.
 \end{proof}

\clearpage
\section{Supplementary Theoretical Background}\label{sec:appendix_supp_theo_background}

In this section, we provide an overview of the conditional diffusion model \blue{and classifier guidance} in \cref{app:back_con_diff} and classifier-free
guidance in \cref{app:class_free_guide}.

\subsection{Conditional Diffusion Process}
\label{app:back_con_diff}

Conditional diffusion models use the conditional information (guidance) $y$ to generate samples from conditional data distribution $P(\cdot | y=\text{guidance})$.
Depending on the model's objective, the guidance is either a label for generating categorical images, a text prompt for generating images from input sentences, or an image region for tasks like image editing and restoration.
Throughout this paper, we coin diffusion models with label guidance $y$ as conditional diffusion models (CDMs).
Practically, implement a conditional diffusion model characterized as classifier and classifier-free guidance.
The classifier guidance diffusion model combines the unconditional score function with the gradient of an external classifier trained on corrupted data.
On the other hand, classifier-free guidance integrates the conditional and unconditional score function by randomly ignoring $y$ with mask signal (see \eqref{eqn:unified_score_def}).
In this paper, we focus on the latter approach.

Specifically,
we consider data $x\in\R^{d_{x}}$ and label $y\in\R^{d_{y}}$ with initial conditional distribution $P(x | y).$
The diffusion process (forward Ornstein–Uhlenbeck process) is characterized by:
\begin{align}
\label{eqn:forward_process}
\dd X_{t} = -\frac{1}{2}X_{t} \dd t + \dd W_{t}\quad \text{with} \quad X_{0}\sim P(x | y),
\end{align}
where $W_{t}$ is a Wiener process.
The distribution at any finite time $t$ is denoted by $P_{t}(x | y),$
and $X_{\infty}$ follows standard Gaussian distribution.
Up to a sufficiently large terminating time $T,$
we generate samples by the reverse process:
\begin{align}
\label{eqn:reverse_process}
\dd X_{t}^{\leftarrow} = \left[\frac{1}{2}X_{t}^{\leftarrow} + \nabla\log{p_{T-t}}(X_{t}^{\leftarrow} | y)\right] \dd t + \dd \bar{W}_{t}\quad \text{with} \quad X_{0}^{\leftarrow}\sim P_{T}(x | y),
\end{align}
where the term $\nabla\log{p_{T-t}}(X_{t}^{\leftarrow} | y)$ represents the conditional score function.
We have $X_{t}|X_{0}\sim N(\alpha_{t}X_{0},\sigma_{t}^{2}I)$ with $\alpha_{t}=e^{-t/2}$ and $\sigma_{t}^{2}=1-e^{-t}$.

We use a score network $\hat{s}$ to estimate the conditional score function $\nabla \log{p_t (x|y)}$, and the quadratic loss of the conditional diffusion model is given by
\begin{align} \label{eqn:Cond_Score_Objective}
\hat{s}\coloneqq\underset{s\in\calT^{\textcolor{blue}{h,s,r}}_{R}}{\operatorname{argmin}} \, \E_{t}\left[\E_{(x_{0},y)}\left[\E_{(x^{\prime}\sim x^{\prime}|x_{0})}\left[\norm{s(x^{\prime},y,t)-\nabla_{x^{\prime}}\log{p_{t}(x^{\prime}|x_{0})}}_{2}^{2}\right]\right]\right],
\end{align}
where $t\sim\text{Unif}(t_{0},T)$.

With the estimate score network $\hat{s}$ in \eqref{eqn:Cond_Score_Objective}, we generates the conditional sample in the backward process as follows:
\begin{align}\label{eqn:est_score_generation}
    \dd \tilde{X}_{t}^{\leftarrow} = \left[\frac{1}{2}\tilde{X}_{t}^{\leftarrow} 
    + \hat{s}\left( \tilde{X}_{t}^{\leftarrow}, y, T-t \right)
    \right]  \dd t 
    + \dd \bar{W}_{t}\quad 
    \text{with} \quad \tilde{X}_{0}^{\leftarrow} \sim N(0, I_d).
\end{align}

Classifier guidance \citep{song2021score,dhariwal2021diffusion} and classifier-free guidance \citep{ho2022classifier} are piratical implementations for conditional score estimation.
For classifier guidance \citep{song2021score,dhariwal2021diffusion}, it use the gradient of the classifier to improve the conditional sample quality of the diffusion model. 
According to Bayes rule, the conditional score function has the relation: 
\begin{align}
    \nabla_{x}\log{p_{t}(x_t | y)} = 
    \underbrace{\nabla \log{p_{t}(x_t)}}_{\text{Approximate by } \hat{s}} +
    \underbrace{\nabla_{x}\log{p_{t}(y | x_t)}}_{\text{Guidance from classifier}}.
\end{align}
It uses the neural network to approximate the unconditional score function $\nabla \log{\hat{p}_{t}(x_t)}$ along with external classifier to approximate $\hat{p}_{t}(y | x_t)$ and compute the gradient of the classifier logits as the guidance $\nabla \log{\hat{p}_{t}(y | x_t)}$.

\subsection{Classifier-free Guidance}
\label{app:class_free_guide}
Classifier-free guidance \citep{ho2022classifier} provides a widely used approach for training condition diffusion models.
It not only simplifies the training pipeline but also improves performance and removes the need for an external classifier.
Classifier-free guidance diffusion model approximates both conditional and unconditional score functions by neural networks $s_W$, where $W$ is the network parameters.

Our primary goal is to establish the theoretical guarantee for selecting
conditional score estimator $\hat{s}(x,y,t)$ chosen from the transformer architecture class and bound the error for such estimation.
Based on previous work by \citet{dhariwal2021diffusion,fu2024unveil,sohl2015deep,ho2022classifier},
we adopt the unified setting for the conditional diffusion model.
First we define the mask signal as $\tau\coloneqq\{\emptyset,\text{id}\},$
where $\emptyset$ denotes the the absence of guidance $y$
and $\text{id}$ denotes otherwise.
Unites the learning of conditional and unconditional scores by randomly ignoring the guidance $y$.
Therefore we write the function class of the score estimator as
\begin{align}
\label{eqn:unified_score_def}
s(x,y,t) = 
\begin{dcases}
    s_{1}(x,y,t), & \quad\text{if}\quad y\in\R^{d_{y}} \\
    s_{2}(x,t), &\quad\text{if}\quad y=\emptyset.
\end{dcases}
\end{align}
Both $s_{1}(x,y,t)$ and $s_{2}(x,t)$ belong to the transformer function class with slight adaption.
Following \citet{fu2024unveil},
we consider $P(\tau=\text{id})=P(\tau=\emptyset)=\frac{1}{2}$ without loss of generality,
and we have the following objective function for score matching:
\begin{align*}
\hat{s}\coloneqq\underset{{s}_W\in\calT^{\textcolor{blue}{h,s,r}}_{R}}{\operatorname{argmin}} \, \E_{t}\left[\E_{(x_{0},y)}\left[\E_{(\tau,x^{\prime}\sim x^{\prime}|x_{0})}\left[\norm{{s}_W(x^{\prime},\tau y,t)-\nabla_{x^{\prime}}\log{p_{t}(x^{\prime}|x_{0})}}_{2}^{2}\right]\right]\right].
\end{align*}

In practice, \blue{the loss function is given by}
\begin{align}
    \label{eqn:diff_loss_app}
    \ell(x_0, y ; {s}_{W})=\int_{T_0}^T \frac{1}{T-T_0} \EE_{\tau, x_t|x_0 \sim N(\alpha_t x_0, \sigma_t^2 I_{d_x})} \left[\left\|{s}_W (x_t, \tau y, t)-\nabla_{x_t} \log p_t\left(x_t | x_0 \right)\right\|_2^2\right]\dd t,
\end{align}
where $T_0$ is a small value for stabilize training \cite{vahdat2021score}. 
To train ${s}_W$, we select $n$ i.i.d. training samples $\{x_{0,i}, y_i \}_{i=1}^{n}$, where $x_{0,i} \sim P_0(\cdot | y_i)$.
We utilize the following empirical loss: 
\begin{align}
\label{eqn:empirical_loss_app}
    \hat{\calL}({s}_W) & = \frac{1}{n} \sum_{i=1}^n \ell(x_{0,i}, y_i ; {s}_{W}).
\end{align}

With the estimate score function ${s}_{W}(x,y,t)$ from minimizing the empirical loss in \eqref{eqn:empirical_loss_app},
\blue{we use ${s}_{W}(x,y,t)$ to generate new samples.}
In the classifier-free guidance \blue{setting}, we generate a new conditional sample by replacing the \blue{approximation} ${s}_{W}$ in \eqref{eqn:est_score_generation} with $\tilde{s}_{W}$, \blue{defined as}:
\begin{align}
    \tilde{s}_{W}(x,y,t) = (1 + \eta) \cdot {s}_{W}(x, y, t) - \eta \cdot {s}_{W}(x, \emptyset, t),
\end{align}
where the strength of guidance $\eta > 0$. 
\blue{The proper choice of $\eta$ is crucial for balancing trade-offs between conditional guidance and unconditional ones. 
The choice directly impacts the performance of the generation process.}
\citet{wu2024theoretical} theoretically study the effect of guidance $\eta$ on Gaussian mixture model.
\blue{They demonstrate that strong guidance improves classification confidence but reduces sample diversity.}
For more detailed related work, refer to \cref{sec:related}.

\clearpage
\section{Universal Approximation of Transformers}
In this section, we discuss the universal approximation theory of transformers.

In \cref{sec:trans_univeral_approx}, we present the universal approximation results of transformers for score approximation in \cref{sec:con_dit}. 
We emphasize that most of the material in \cref{sec:trans_univeral_approx} is not original and is drawn from \cite{hu2024fundamental,kajitsuka2023transformers,yun2019transformers}.

In \cref{sec:appendix_transformer_parameter}, we compute the parameter norm bounds of the transformers used for score approximation. These bounds are crucial for calculating the covering number of the transformers and are essential for score and distribution estimation in \cref{sec:score_est_dist_est}.

\subsection{Transformers as Universal Approximators}
\label{sec:trans_univeral_approx}

The key idea for demonstrating the transformers' ability to capture the entire sequence lies in the concept of contextual mapping \cite{hu2024fundamental,kajitsuka2023transformers,yun2019transformers}.
We first restate the background of a $(\gamma, \delta)$-contextual mapping in \cref{def:contextual_mapping_new}, using the definition of vocabulary (\cref{def:vocabulary}) and token separation (\cref{def:token_seperate_new}) in the input sequences.

\paragraph{Background: Contextual Mapping.}
Let $Z, Y \in \mathbb{R}^{d\times L}$ represent input embeddings and output label sequences, respectively, where $Z_{:,k} \in \mathbb{R}^{d}$ denotes the $k$-th token (column) of each $Z$ sequence.
The vocabulary corresponding to the $i$-th sequence at the $k$-th index is defined in \cref{def:vocabulary}.

\begin{definition}[Vocabulary]
\label{def:vocabulary}
We define the $i$-th vocabulary set for $i \in [N]$ by $\mathcal{V}^{(i)}=\bigcup_{k \in[L]} Z_{:, k}^{(i)} \subset \mathbb{R}^{d}$, and the whole vocabulary set $\mathcal{V}$ is defined by $\mathcal{V}=\bigcup_{i \in[N]} \mathcal{V}^{(i)} \subset \mathbb{R}^{d}$.
\end{definition}
In line with prior works \cite{hu2024fundamental,kajitsuka2023transformers,kim2022provable,yun2019transformers}, we assume the embeddings separateness to be $\left(\gamma_{\min }, \gamma_{\max }, \delta\right)$-separated, as defined in \cref{def:token_seperate_new}.
\begin{definition}[Tokenwise Separateness] \label{def:token_seperate_new}
    Let $Z^{(1)}, \ldots, Z^{(N)} \in \mathbb{R}^{d\times L}$ be embeddings. 
    Then, $Z^{(1)}, \ldots, Z^{(N)}$ are called tokenwise $\left(\gamma_{\min }, \gamma_{\max }, \delta\right)$-separated if the following three conditions hold.
\begin{itemize}[leftmargin=2em]
    \item [(i)] For any $i \in[N]$ and $k \in[n],\|Z_{:, k}^{(i)}\|> \gamma_{\min }$ holds.
    \item [(ii)]
    For any $i \in[N]$ and $k \in[n], \|Z_{:, k}^{(i)}\|<\gamma_{\max }$ holds.
    \item [(iii)]
    For any $i, j \in [N]$ and $k, l \in [n]$ if $Z_{:, k}^{(i)} \neq Z_{:, l}^{(j)}, $ then $ \|Z_{:, k}^{(i)}-Z_{:, l}^{(j)}\| > \delta$ holds.
\end{itemize}
Note that when only conditions (ii) and (iii) hold, we denote this as $(\gamma, \delta)$-separateness. 
Moreover, if only condition (iii) holds, we denote it as $(\delta)$-separateness.
\end{definition}
Next, we define a $(\gamma, \delta)$-contextual mapping, building from conditions (ii) and (iii) in the definition of \cref{def:token_seperate_new}.
The contextual mapping extends the concept of token separateness to captures the relationships between tokens across different input sequences effectively. 
This allows transformers' to utilize self-attention for full context representation.
\begin{definition} [Contextual Mapping] \label{def:contextual_mapping_new}
    Let $Z^{(1)}, \ldots, Z^{(N)} \in \mathbb{R}^{d\times L}$ be embeddings. 
    Then, a map $q: \mathbb{R}^{d\times L} \rightarrow \mathbb{R}^{d\times L}$ is called an $(\gamma, \delta)$-contextual mapping if the following two conditions hold:
    \begin{enumerate}
        \item For any $i \in[N]$ and $k \in[L],  \|q (Z^{(i)})_{:, k}\| < \gamma$ holds.

        \item For any $i, j \in[N]$ and $k, l \in[L]$ such that $\mathcal{V}^{(i)} \neq \mathcal{V}^{(j)}$ or $Z_{:, k}^{(i)} \neq Z_{:, l}^{(j)}$, $\|q(Z^{(i)})_{:, k}-q(Z^{(j)})_{:, l}\| > \delta$ holds.
    \end{enumerate}
Note that $q\left(Z^{(i)}\right)$ for $i \in[N]$ is called a \textit{context ID} of $Z^{(i)}$.
\end{definition}

\blue{
\paragraph{Helper Lemmas.}
For completeness, we restate some helper lemmas before presenting the proof that a one-layer single-head attention mechanism is a contextual mapping in \cref{lemma:contextual_map_self_attn_new}, following \citet{hu2024fundamental}.
}

\begin{lemma}[${\rm Boltz}$ Preserves Distance, Lemma~1 of \cite{kajitsuka2023transformers}] 
\label{lemma:boltz_new}    
    Given  $(\gamma, \delta)$-tokenwise separated vectors $z^{(1)}, \ldots, z^{(N)} \in \mathbb{R}^{n}$ with no duplicate entries in each vector, that is
    \begin{align*}
        z_s^{(i)} \neq z_t^{(i)}, 
    \end{align*}
    where $i \in [N]$ and $ s,t \in [L], s \neq t$. 
    Also, let 
    \begin{align*}
        \delta \geq 4 \ln  n.
    \end{align*}

    Then, the outputs of the Boltzmann operator has the following property:
\begin{align}
\label{eqn:boltz_gamma_bound}
 \abs{{\rm Boltz}\left(z^{(i)}\right)} & \leq \gamma,  
\\
\label{eqn:boltz_delta_sep}
 \abs{{\rm Boltz}\left(z^{(i)}\right)-{\rm Boltz}\left(z^{(j)}\right)} 
 & >
 \delta^{\prime}
 = \ln ^2 (n) \cdot e^{-2 \gamma} 
\end{align}
for all $ i,j \in [N], i \neq j$. 
\end{lemma}

\begin{lemma}[Lemma 13 of \cite{park2021provable}]   \label{lemma:13_park2021_new}
    For any finite subset $\mathcal{X} \subset \mathbb{R}^{d}$, there exists at least one unit vector $u \in \mathbb{R}^{d}$ such that
\begin{align*} 
\frac{1}{ \abs{\mathcal{X} }^{2}} \sqrt{\frac{8}{\pi d}} \norm{x-x^{\prime}} 
\leq
\abs{u^{\top}\left(x-x^{\prime}\right)} 
\leq
\norm{x-x^{\prime}} 
\end{align*}
for any $x, x^{\prime} \in \mathcal{X}$.
\end{lemma}

{\color{blue}
\cref{lemma:13_park2021_new} provides the existence of a unit vector $u \in \R^d$ that bounds the inner product of the difference between points in a finite subset $\mathcal{X} \subset \R^d$.

We are now ready to restate the construction of rank-$\rho$ weight matrices in a self-attention layer following \cite{hu2024fundamental} in \cref{lemma:tokio_vuu_new}.
}
\begin{lemma}[Construction of Weight Matrices, Lemma D.2 of \cite{hu2024fundamental}] 
\label{lemma:tokio_vuu_new}
    Given a dataset with a $\left(\gamma_{\min }, \gamma_{\max }, \epsilon\right)$-separated finite vocabulary $\mathcal{V} \subset \mathbb{R}^{d}$.
    There exists rank-$\rho$ weight matrices $W_K, W_Q \in \mathbb{R}^{s \times d}$ such that 
\begin{align*}
\abs{\left(W_K v_{a}\right)^{\top}\left(W_Q v_{c}\right)-\left(W_K v_{b}\right)^{\top}\left(W_Q v_{c}\right)} 
>\delta, 
\end{align*}
for any $\delta >0$, any $\min \left( d,s \right) \geq \rho \geq 1$ and any $v_{a}, v_{b}, v_{c} \in \mathcal{V}$ with $v_{a} \neq v_{b}$.
In addition, the matrices are constructed as
\begin{align*}
    & W_K
    = 
    \sum_{i=1}^{\rho} p_i q_i^{\top} \in \mathbb{R}^{s \times d},
    \quad
    W_Q= \sum_{j=1}^{\rho} p^\prime_j q_j^{\prime \top} \in \mathbb{R}^{s \times d},
\end{align*}
where for at least one $i$, $q_i, q^\prime_i \in \mathbb{R}^d$ are unit vectors that satisfy \cref{lemma:13_park2021_new}, and $p_i, p^\prime_i \in \mathbb{R}^s$ satisfies
\begin{align*}
    \abs{p_i^{\top} p_i^{\prime}} 
= 
5 \left( |\mathcal{V}| + 1 \right)^{4} d \frac{\delta}{\epsilon \gamma_{\min }}.
\end{align*}
\end{lemma}

\begin{proof}[Proof of \cref{lemma:tokio_vuu_new}]

{\color{blue}
For completeness, we restate the key point from the proof in \cite{hu2024fundamental}.

First, applying \cref{lemma:13_park2021_new} to $\mathcal{V} \cup \{0\}$,
there exists at least one unit vector $q \in \mathbb{R}^{d}$ such that for any $v_{a}, v_{b} \in \mathcal{V} \cup\{0\}$ and $v_{a} \neq v_{b}$ the following holds:
\begin{align*}
\frac{1}{ \left( |\mathcal{V}| + 1 \right)^{2} d^{0.5} } 
\norm{v_{a}-v_{b}}  
\leq
\abs{q^{\top} \left(v_{a}-v_{b} \right)} 
\leq
\norm{v_{a}-v_{b}}.
\end{align*}
Following \cite{hu2024fundamental}, let $v_b=0$ and all unit vector $q \in \{ q \in \mathbb{R}^n : \|q\| = 1 \}$, and select some arbitrary vector pairs $p_i, p_i^{\prime} \in \mathbb{R}^{s}$ that satisfy the constraint:
\begin{align} \label{eqn:pp'_constraint}
\abs{p_i^{\top} p_i^{\prime}} 
=
 \left( |\mathcal{V}| + 1 \right)^{4} d \frac{\delta}{\epsilon \gamma_{\min }}.
\end{align}
By constructing the weight matrices as follows:
\begin{align*}
    W_K
    = 
    \sum_{i=1}^{\rho} p_i q_i^{\top} \in \mathbb{R}^{s \times d},
    \quad
    W_Q= \sum_{j=1}^{\rho} p^\prime_j q_j^{\prime \top} \in \mathbb{R}^{s \times d},
\end{align*}
where for at least one $i$, $p_i, p^\prime_i$ satisfies \eqref{eqn:pp'_constraint} and $q_i, q^\prime_j \in \mathbb{Q}$,
we are able to demonstrate that: 
\begin{align*}
     \abs{ \left(W_K v_{a}\right)^{\top}\left(W_Q v_{c}\right)-\left(W_K v_{b}\right)^{\top}\left(W_Q v_{c}\right)} > \delta.
\end{align*}
This completes the proof.
}
\end{proof}

\blue{
\paragraph{Any-Rank Attention is Contextual Mapping.}
Next, we present the generalized result where self-attention mechanisms of any rank serve as contextual mappings, extending the low-rank analysis in \cite{kajitsuka2023transformers} to \textit{any-rank} attention weights, as shown in \cite{hu2024fundamental}.
}

\begin{theorem}[Any-Rank Attention is $(\gamma, \delta)$-Contextual Mapping, \blue{Lemma 2.2 of \cite{hu2024fundamental}}]  
\label{lemma:contextual_map_self_attn_new}
    Given embeddings $Z^{(1)}, \ldots, Z^{(N)} \in \mathbb{R}^{d \times L}$ which are  $\left(\gamma_{\min }, \gamma_{\max }, \epsilon\right)$-tokenwise separated and
    vocabulary set $\mathcal{V}=\bigcup_{i \in[N]} \mathcal{V}^{(i)} \subset \mathbb{R}^{d}$.
    Also, let $Z^{(1)}, \ldots, Z^{(N)} \in \mathbb{R}^{d \times L}$ be embedding sequences with no duplicate word token in each sequence, that is, $Z_{:, k}^{(i)} \neq Z_{:, l}^{(i)}$, 
    for any $i \in[N]$ and $k, l \in[L]$.
    Then, there exists a 1-layer single head attention with weight matrices $W_O \in \mathbb{R}^{d \times s}$ and $W_V, W_K, W_Q \in$ $\mathbb{R}^{s \times d}$, that is a $(\gamma, \delta)$-contextual mapping for the embeddings $Z^{(1)}, \ldots, Z^{(N)}$ with
    \begin{align*}
    \gamma =  \gamma_{\max} +\epsilon/4, 
    \quad
    \delta =  \exp( - 5 \epsilon^{-1} |{\cal V}|^4 d \kappa \gamma_{\max}  \log L  ),
    \end{align*}
    where $\kappa = \gamma_{\max}/\gamma_{\min}$.
\end{theorem}

\blue{
\cref{lemma:contextual_map_self_attn_new} shows that any-rank self-attention is able to distinguish two identical tokens in distinct contexts.
Specifically, this holds for embeddings $Z_{:, k}^{(i)}=Z_{:, l}^{(j)}$ when the vocabulary sets $\mathcal{V}^{(i)} \neq \mathcal{V}^{(j)}$.
}

\blue{
We remark that 
the proof of \cref{lemma:contextual_map_self_attn_new} is crucial for the subsequent analysis.
Therefore, we restate it below for later convenience.
}

\begin{proof}[Proof Sketch]
\blue{    
\citet{hu2024fundamental} generalize the results of \cite[Theorem 2]{kajitsuka2023transformers} where all weight matrices have to be rank-$1$.
This is achieved by constructing the weight matrices as an outer product sum $\sum_i^\rho u_i v_i^\top$, where $u_i \in \mathbb{R}^s, v_i \in \mathbb{R}^d$.
The proof in \cite{hu2024fundamental} is diveded into two parts: 
\begin{itemize}
    \item \textbf{Construction of Softmax Self-Attention}: Different input tokens are mapped to unique contextual embeddings by configuring the weight matrices according to \cref{lemma:tokio_vuu_new}.
    \item \textbf{Handling Identical Tokens in Different Contexts}: Use the tokenwise separateness guaranteed by \cref{lemma:tokio_vuu_new} to handle identical tokens appearing in different contexts.
    Additionally, \cref{lemma:boltz_new} shows ${\rm Boltz}$ preserves separateness properties.
\end{itemize}
With these, we prove that self-attention distinguish input embeddings $Z_{:, k}^{(i)}=Z_{:, l}^{(j)}$ when the vocabulary sets $\mathcal{V}^{(i)} \neq \mathcal{V}^{(j)}$.
}
\end{proof}

\begin{proof}[Proof of \cref{lemma:contextual_map_self_attn_new}]

For completeness, we restate the key point from the proof in \cite{hu2024fundamental}.

The proof consists of two parts: First, we show that the attention layer maps different tokens to unique IDs.
Second, we show that self-attention distinguishes duplicate input tokens when they appear in different contexts.

For the first part, by utilizing \cref{lemma:tokio_vuu_new} and set the weight matrices as follows:
\begin{itemize}

\item \textbf{Weight Matrices $W_K$ and $W_Q$:}
\begin{align*}
& W_K
= 
\sum_{i=1}^{\rho} p_i q_i^{\top} \in \mathbb{R}^{s \times d},
\\
& W_Q= \sum_{j=1}^{\rho} p^\prime_j q_j^{\prime \top} \in \mathbb{R}^{s \times d},
\end{align*}    
where $p_i, p_j^{\prime} \in \mathbb{R}^{s}$ and $q_i, q_j^\prime \in \mathbb{R}^{d}$.
In addition, let $\delta=4 \ln  n$ and $p_1, p_1^{\prime} \in \mathbb{R}^{s}$ be an arbitrary vector pair that satisfies 
\begin{align} \label{eqn:tokio_lemma3_uTu_new}
\abs{p_1^{\top} p_1^{\prime}}
=
\left( |\mathcal{V}| + 1 \right)^{4} d 
\frac{\delta}{\epsilon \gamma_{\min }}.
\end{align} 

\item \textbf{Weight Matrices $W_V$ and $W_O$:}
In addition, for the other two weight matrices $W_O \in \mathbb{R}^{d \times s}$ and $W_V \in \mathbb{R}^{s \times d}$, we  set 
\begin{align} \label{eqn:Wv_construct}
    W_V = \sum_{i=1}^{\rho} p^{ \prime \prime}_i q_i^{\prime \prime \top} \in \mathbb{R}^{s \times d},
\end{align}
where $q^{\prime \prime} \in \mathbb{R}^d $, $q^{\prime \prime}_1 = q_1$ and $p_i^{\prime \prime} \in \mathbb{R}^{s}$ is some nonzero vector that satisfies
\begin{align} \label{eqn:construct_WO_new}
    \norm{W_O p_i^{\prime \prime}} = \frac{\epsilon}{4 \rho \gamma_{\max}}.
\end{align}
This can be accomplished, e.g., $W_O = \sum_{i=1}^{\rho} p_i^{\prime \prime \prime} {p_i^{\prime \prime}}^{\top}$ for any vector $p_i^{\prime \prime \prime}$ which satisfies $\norm{p_i^{\prime \prime \prime}} = \epsilon / (4 \rho^2 \gamma_{\max} \norm{p_i^{\prime \prime}}^2)$ for any $i \in [\rho]$.

\item \textbf{Mapping Condition:}
With above weights construction, for $i \in[N]$ and $k \in[L]$,
we have 
\begin{align}
\label{eqn:dif_token_to_diff_value_condition_new}
 \norm{W_O
 \left(W_V Z^{(i)}\right)
 \Softmax
 \left[ \left(W_K Z^{(i)}\right)^{\top}\left(W_Q Z_{:, k}^{(i)}\right)\right]}
<
\frac{\epsilon}{4}.
\end{align}
\end{itemize}

For the second part, we prove that with the weight matrices $W_O, W_V, W_K, W_Q$ configured above, the attention layer distinguishes duplicate input tokens with different context, $Z_{:, k}^{(i)}=Z_{:, l}^{(j)}$ with different vocabulary sets $\mathcal{V}^{(i)} \neq \mathcal{V}^{(j)}$.

We define $a^{(i)}, a^{(j)}$ as
\begin{align*}
 a^{(i)} = \left( W_K Z^{(i)}\right)^{\top}\left(W_Q Z_{:, k}^{(i)}\right) \in \mathbb{R}^{n},  
\quad
 a^{(j)}=\left(W_K Z^{(j)}\right)^{\top}\left(W_Q Z_{:, l}^{(j)}\right) \in \mathbb{R}^{n},
\end{align*}
where $a^{(i)}$ and $a^{(j)}$ are tokenwise $(\gamma, \delta)$-separated. 
Specifically, the following inequality holds 
\begin{align*}
|{a_{k^{\prime}}^{(i)}}| \leq  \left( |\mathcal{V}| + 1 \right) ^ {4}  d \frac{\delta}{\epsilon \gamma_{\min }} \gamma_{\max }^{2}.     
\end{align*}

Since $\mathcal{V}^{(i)} \neq \mathcal{V}^{(j)}$ and there is no duplicate token in $Z^{(i)}$ and $Z^{(j)}$ respectively, we use \cref{lemma:boltz_new} and obtain 
\begin{align}  
\label{eqn:a_t_soft_a_1_new}
& ~ \abs{ {\rm Boltz}\left(a^{(i)} \right) -{\rm Boltz}\left(a^{(j)}\right) } \nonumber \\ 
= &~
\abs{\left(a^{(i)}\right)^{\top} \Softmax\left[a^{(i)}\right]-\left(a^{(j)}\right)^{\top} \Softmax\left[a^{(j)}\right]}
\\ \nonumber
> &~
\delta^{\prime}
\\ \nonumber
= &~
(\ln  n)^{2} e^{-2 \gamma}.
\end{align}
Additionally, using \cref{lemma:tokio_vuu_new} and \eqref{eqn:tokio_lemma3_uTu_new}, and assuming $Z_{:, k}^{(i)}=Z_{:, l}^{(j)}$, we have 
\begin{align} \label{eqn:a_t_soft_a_2_new}
&~ \abs{ \left( a^{(i)}\right)^{\top} \Softmax \left[a^{(i)}\right]-\left(a^{(j)}\right)^{\top} \Softmax \left[a^{(j)}\right]}  
\\ \nonumber
\leq &~ 
\sum_{i=1}^{\rho}
\gamma_{\max } 
\cdot 
(|\mathcal{V}|+1)^{4} \frac{\pi d}{8} \frac{\delta}{\epsilon \gamma_{\min }} 
\cdot
\abs{\left( q_i^{\top} Z^{(i)}\right) \Softmax\left[a^{(i)}\right]
-
\left(q_i^{\top} Z^{(j)}\right) \Softmax\left[a^{(j)}\right]}.
\end{align}

By combining \eqref{eqn:a_t_soft_a_1_new} and \eqref{eqn:a_t_soft_a_2_new}, we have 
\begin{align} \label{eqn:v_T_X_soft_a_new}
    \sum_{i=1}^{\rho}
    \abs{\left(q_i^{\top} Z^{(i)}\right) \Softmax\left[a^{(i)}\right]
    -
    \left(q_i^{\top} Z^{(j)}\right) \Softmax\left[a^{(j)}\right]}
    >
    \frac{\delta^{\prime}}{ \left( |\mathcal{V}| + 1 \right)^{4}} \frac{ \epsilon \gamma_{\min }}{d \delta \gamma_{\max }}.
\end{align}

Finally, using \eqref{eqn:construct_WO_new} and \eqref{eqn:v_T_X_soft_a_new}. we derive the lower bound of the difference between the self-attention outputs of $Z^{(i)}, Z^{(j)}$ as follows:
\begin{align} \label{eqn:SA_minus_SA}
\norm{f_{S}^{(\text{SA})}\left(Z^{(i)}\right)_{:, k}
-
f_{S}^{(\text{SA})}\left(Z^{(j)}\right)_{:, l}}  
> 
\frac{\epsilon}{4 \gamma_{\max }} 
\frac{\delta^{\prime}}{\left( |\mathcal{V}| + 1 \right)^{4}} 
\frac{\epsilon \gamma_{\min }}{ d \delta \gamma_{\max }},
\end{align}
where 
$\delta  = 4\ln  L$ and $\delta^{\prime}   = \ln ^2 (L)  e^{-2 \gamma}$ with $\gamma = \left( |\mathcal{V}| + 1 \right)^{4}
 d \delta \gamma_{\max }^{2} / (  \epsilon \gamma_{\min } )$.

This completes the proof.
\end{proof}

\blue{With \cref{lemma:contextual_map_self_attn_new} establishing that any-rank attention is a contextual mapping, we restate the universal approximation result for transformers with a single self-attention layer from \cite{hu2024fundamental,kajitsuka2023transformers}.}

\begin{theorem}[Transformers with $1$-Layer Self-Attention are Universal Approximators, Modified from Proposition 1 of \cite{kajitsuka2023transformers}]
\label{thm:Transformer_as_universal_approximators}
Let $0\leq p<\infty$ 
and $ f^{(\text{FF})}, f^{(\text{SA})}$ be feed-forward neural network layers and a single-head self-attention layer with softmax function respectively.
Then,
for any permutation equivariant, continuous function $f$ with compact support and $\epsilon>0,$
there exists $f^{\prime}\in\calT_{R}^{\textcolor{blue}{h,s,r}}$ such that $d_{p}(f,f^{\prime})<\epsilon$ holds
\end{theorem}
\begin{proof}[Proof of \cref{thm:Transformer_as_universal_approximators}]

We restate the proof from \cite{kajitsuka2023transformers} for completeness.

The proof consists of the following steps: 
\begin{enumerate}
    \item Approximate by Step Function: Given a permutation equivariant continuous function $f$ on a compact set, there exists a Transformer $f^{\prime} \in \calT_R^{\textcolor{blue}{h,s,r}}$ with one self-attention layer to approximate $f$ by step function with arbitrary precision in terms of $p$-norm.
    \item Quantization via $f^{\text{FF}}_{1}$: The first feed-forward network $f^{\text{FF}}_{1}$ quantize the input domain, reducing the problem to memorization of finite samples.
    \item Contextual Mapping $ f^{(\text{SA})}$ and Memorization $f^{\text{FF}}_2$: According to \cref{lemma:contextual_map_self_attn_new}, we construct any-rank attention $ f^{(\text{SA})}$ to be contextual mapping. Then use the second feed-forward $f^{\text{FF}}_2$ to memorize the \textit{context ID} with its corresponding label.
\end{enumerate}

The details for the three steps are below.
\begin{enumerate}
    \item Since $f$ is a continuous function on a compact set, $f$ has maximum and minimum values on the domain.
    By scaling with $f^{\text{FF}}_1$ and $f^{\text{FF}}_2$, $f$ is assumed to be normalized without loss of generality: 
    That is for any $ Z \in \mathbb{R}^{d \times L} \setminus [0,1]^{d \times L} $, we have $ f(Z) = 0 $. 
    For any $ X \in [-1,1]^{d \times L} $, the function $ f(X) $ satisfies $ -1 \leq f(X) \leq 1 $.
    
    Let $D \in \mathbb{N}$ be the granularity of a grid
    \begin{align*}
        \mathbb{G}_D
        = \{1/D, 2/D, \dots, 1\}^{d \times L} \subset \R^{d \times L}
    \end{align*}
    such that a piece-wise constant approximation
    \begin{align*}
        \bar{f}(X)
        = \sum_{L \in \mathbb{G}_D} f\left(L\right) 1_{Z \in L + [-1/D,0)^{d \times L}}
    \end{align*}
    satisfies 
    \begin{align}
        d_p(f, \bar{f}) < \epsilon/3.
        \label{eqn:est_of_step_function}
    \end{align}
    Such a $D$ always exists because of uniform continuity of $f$.
    
    \item We use $f^{\text{FF}}_1$ to quantize the input domain into $\mathbb{G}_D$.

    We first define the following two terms for first feed-forward neural network to approximate.

    \begin{itemize}
        \item  The quantize term ($\mathrm{quant}_D^{d \times L}: \R^{d \times L} \to \R^{d \times L}$): 
        Quantize $[0,1]$ into $\{1/D,\dots,1\}$, while it projects $\R \setminus [0,1]$ to $0$ by shifting and stacking step function.
        \begin{align}
        &~ \sum_{t = 0}^{D-1} \frac{{\rm ReLU}\left[x/\delta-t/\delta D\right]
        -
        {\rm ReLU}\left[x/\delta - 1-t/\delta D\right]
        }{D}  \notag \\
        \approx &~
        \mathrm{quant}_D(x)
        =
        \begin{cases}
            0 & x < 0 \\
            1/D & 0\leq x < 1/D \\
            \vdots & \vdots \\
            1 & 1-1/D \leq x
        \end{cases} \label{eqn:quantize_term_f1}.
        \end{align}
        \item The penalty term ($\mathrm{penalty}$): Identify whether an input sequence is in $[0,1]^{d \times L}$.
        This is defined by
        \begin{align}
        &
        {\rm ReLU}\left[(x-1)/\delta\right]
        -
        {\rm ReLU}\left[(x-1)/\delta - 1\right]
        -
        {\rm ReLU}\left[-x/\delta\right]
        -
        {\rm ReLU}\left[-x/\delta - 1\right] \notag \\
        \approx &~ \mathrm{penalty}(x)
        =
        \begin{cases}
            -1 & x \leq 0 \\
            0 & 0 < x \leq 1 \\
            -1 & 1 < x
        \end{cases} \label{eqn:penalty_term_f1}.
    \end{align}
    \end{itemize}

    Combining these components together, the first feed-forward neural network layer $f^{\text{FF}}_1$ approximates the following function:
    \begin{align} \label{eqn:f1_approx_comb}
        \bar{f}^{({\text{FF}})}_1(X) 
        = \mathrm{quant}_D^{d \times L}(X)
        + \sum_{t=1}^d \sum_{k=1}^L \mathrm{penalty}(X_{t,k})
    \end{align} 
    Note that this function quantizes inputs in $[0,1]^{d \times L}$ with granularity $D$, while every element of the output is non-positive for inputs outside $[0,1]^{d \times L}$.
    In particular, the norm of the output is upper-bounded by
    \begin{align}
        \max_{X \in \R^{d \times L}}\left\|f^{\text{FF}}_1(X)_{:,k}\right\|
        = \underbrace{dL}_{\text{Total number of elements in X}} \times \underbrace{\sqrt{d}}_{\text{Maximum Euclidean norm in } d\text{-dimensional space}} \label{eqn:maximum_norm_of_quantization}
    \end{align}
    for any $k \in [L]$.
    
    \item Let $\tilde{\mathbb{G}}_D \subset \mathbb{G}_D$ be a sub-grid
    \begin{align*}
        \tilde{\mathbb{G}}_D
        = \left\{
        G \in \mathbb{G}_D
        \mathrel{}\middle|\mathrel{}
        \forall k,l \in [L],\ G_{:,k} \neq G_{:,l}
        \right\},
    \end{align*}
    and consider memorization of $\tilde{\mathbb{G}}_D$ with its labels given by $f(G)$ for each $G \in \tilde{\mathbb{G}}_D$.
    Using our modified any-rank attention is contextual mapping in \cref{lemma:contextual_map_self_attn_new} allows us to construct a self-attention $ f^{(\text{SA})}$ to be a contextual mapping for such input sequences, because $ \tilde{\mathbb{G}}_D$ can be regarded as tokenwise $(1/D, \sqrt{d}, 1/D)$-separated input sequences.
    By taking sufficiently large granularity $D$ of $\mathbb{G}_D$, the number of cells with duplicate tokens, that is, $|\mathbb{G}_D \setminus \tilde{\mathbb{G}}_D|$ is negligible.
    
    From the way the self-attention $ f^{(\text{SA})}$ is constructed, we have
    \begin{align*}
        \left\| f^{(\text{SA})}(X)_{:,k} - X_{:,k}\right\|
        < \frac{1}{4\sqrt{d}D}\max_{k' \in [L]}\left\|X_{:,k'}\right\|
    \end{align*}
    for any $k \in [L]$ and $X \in \R^{d \times L}$.
    
    If we take large enough $D$, every element of the output for $X \in \R^{d \times L} \setminus [0,1]^{d \times L}$ is upper-bounded by
    \begin{align*}
         f^{(\text{SA})} \circ f^{\text{FF}}_1\left(X\right)_{t,k} < \frac{1}{4D}
        \quad (\forall t \in [d],\ k \in [L]),
    \end{align*}
    while the output for $X \in [0,1]^{d \times L}$ is lower-bounded by
    \begin{align*}
         f^{(\text{SA})} \circ f^{\text{FF}}_1\left(X\right)_{t,k}
        > \frac{3}{4D}
        \quad (\forall t \in [d],\ k \in [L]).
    \end{align*}
    Finally, we construct $\mathrm{bump}$ function of scale $R > 0$ to     map each input sequence $L \in \tilde{\mathbb{G}}_D$ to its labels $f(L)$ and for input sequence outside the range $X \in (-\infty, 1/4D)^{d \times L}$ to 0 using the second feed-forward $f^{\text{FF}}_2$.
    Precisely, $\mathrm{bump}$ function of scale $R > 0$ is given by
    \begin{align} \label{eqn:bump_function_def}
        \mathrm{bump}_R(x)
        &=\frac{f(L)}{dL}\sum_{t=1}^d\sum_{k=1}^L ({\rm ReLU}\left[R(X_{t,k}-G_{t,k})-1\right]
        - {\rm ReLU}\left[R(Z_{t, k}-G_{t,k})\right] \notag \\
        &\quad\quad\quad\quad\quad\quad\quad
        + {\rm ReLU}\left[R(Z_{t,k}-G_{t,k})+1\right]) + {\rm ReLU}[R(G_{t,k}-Z_{t,k})]
    \end{align}
    for each input sequence $G \in \tilde{\mathbb{G}}_D$ and add up these functions to implement $f^{\text{FF}}_2$.

    In addition, the value of $f^{(\text{FF})}_2$ is always bounded: $0 \leq f^{(\text{FF})}_2 \leq 1$. 
    Thus, by taking sufficiently small $\delta > 0$ to quantize the step function, we have
    \begin{align}
        d_p\left(f^{(\text{FF})}_2 \circ  f^{(\text{SA})} \circ f^{(\text{FF})}_1, f^{(\text{FF})}_2 \circ  f^{(\text{SA})} \circ \overline{f}^{(\text{FF})}_1\right) < \frac{\epsilon}{3}.
        \label{eqn:estimate_of_quantization}
    \end{align}
     Taking large enough $D$ to make duplicate tokens negligible, we have
    \begin{align}
        d_p\left(f^{(\text{FF})}_2 \circ  f^{(\text{SA})} \circ \overline{f}^{(\text{FF})}_1,
        \overline{f}\right) < \frac{\epsilon}{3}. \label{eqn:estimate_of_duplicate_area}
    \end{align}
    
    Combining estimation of step function \eqref{eqn:est_of_step_function}, estimation of quantization \eqref{eqn:estimate_of_quantization} and estimatation of duplicate tokens \eqref{eqn:estimate_of_duplicate_area} together, we get the approximation error of the any-rank Transformer as
    \begin{align}
        d_p\left(f^{(\text{FF})}_2 \circ  f^{(\text{SA})} \circ \overline{f}^{(\text{FF})}_1,
        f\right) < \epsilon. 
    \end{align}
\end{enumerate}
    This completes the proof. 
\end{proof}

{\color{blue}
Lastly, we provide the next corollary stating that the required transformer configuration $(h,r,s)$ for universal approximation.

\begin{corollary}[Universal Approximation of Transformers]\label{cor:transformer_class}
    From \cref{thm:Transformer_as_universal_approximators}, for any permutation equivariant, continuous function $f$ with compact support and $\epsilon>0$, a transformer network 
        $f^{\prime} \in \calT^{1,1,4}_{R}$  with MLP dimension (width) $r=4$  and $=\calO((1/\epsilon)^{dL})$ FFN layers
    is sufficient to approximate $f$ such that $d_p\qty( f,f^{\prime} ) < \epsilon$.
\end{corollary}
\begin{remark}
    We remark that $\calT^{1,1,4}_{R}$ belongs to the considered transformer network function class \cref{def:transformer_class}.
\end{remark}
We establish in \cref{cor:transformer_class} the minimal transformer configuration required to achieve universal approximation for compactly supported functions. 
We remark that this configuration is minimally sufficient but not necessary.
More complex configurations can also achieve transformer universality, as reported in \cite{hu2024statistical,kajitsuka2023transformers,yun2019transformers}.

Throughout this paper, unless otherwise specified, we use the transformer class $\calT^{1,1,4}_{R}$ to construct score function approximations.
}

\subsection{Parameter Norm Bounds for Transformer Approximation}\label{sec:appendix_transformer_parameter}

In the analysis of the approximation ability of transformers in \cite{kajitsuka2023transformers}, universal approximation is ensured by using a sufficiently large granularity $D$, a sufficiently small $\delta$ in $f_{1}^{(\text{FF})}$, and an appropriate scaling factor $R$ in $f_{2}^{(\text{FF})}$. Here, we provide a detailed discussion on parameter bounds for matrices in $\calT_{R}^{h, r, s}$, focusing on the choice of granularity and scaling factor.

\begin{lemma}[Order of Granularity and  Scaling Factor]
\label{lemma:setting_D_R}
Consider the universal approximation theorem for transformers in \cref{thm:Transformer_as_universal_approximators}.
The order for the granularity and the scaling factor follows \blue{$D=\calO(\epsilon^{-1/d})$} and $R=\calO(D),$
and the parameter $\delta$ for the first feed-forward layer in \eqref{eqn:quantize_term_f1} follows $\delta=o(D^{-1})$.
\end{lemma}

\begin{proof}
We investigate the more precise choice of $D$, $R$, and $\delta$ respectively.

\begin{itemize}
    \item 
    \textbf{Bound on Scaling Factor in $f_{2}^{(\text{FF})}$.}

First,
we need to ensure that $R>0$ is large enough such that it maps input $Z\in(-\infty,\frac{1}{4D})^{d\times L}$ to zero.

Because we have $Z_{t,k}-L_{t,k}\leq-\frac{3}{4D}$, 
we obtain the desired result from \eqref{eqn:bump_function_def} by taking $R=\calO(D)$ such that three ${\rm ReLU}(\cdot)$ output zero.

Second, we need to ensure that $R>0$ is large enough such that it  maps $L\in\tilde{\mathbb{G}}\subset(\frac{3}{4D},\infty)^{d\times L}$ to the corresponding label $f(L)$.

From \eqref{eqn:bump_function_def},
we achieve this by selecting proper $R$ such that
\begin{align*}
\sum_{t=1}^{d}\sum_{k=1}^{L}{\rm ReLU}\left[RS-1\right] - {\rm ReLU}\left[RS\right] + {\rm ReLU}\left[RS+1\right] {\rm ReLU}[-RS] = dL,
\end{align*}
where $S\coloneqq Z_{t,k}-L_{t,k}=\calO(D^{-1})$.

For any $S\in\R$,
we take $R=\calO(D)$ such that $\abs{RS}\leq1$.

\item 
\textbf{Bound on Granularity $D$.} 

In \cite{kajitsuka2023transformers}, there are $\calO(D^{-d}\abs{\mathbb{G}_{D}})$ omitted duplicated input.
Clearly,
by taking sufficiently large granularity $\abs{\mathbb{G}_{D}\setminus\tilde{\mathbb{G}}_{D}}$ becomes negligible,
but here we aim to evaluate the corresponding order of $D$.

First,
by the extreme value theorem,
the continuous function $f$ on $[0,1]^{d\times L}$ here is bounded by some constant, denoted by $B$.

Second, the total omitted points are $\calO(D^{d(L-1)})$.

Third, the probability for each point in $\mathbb{G}_{D}$ is $1/D^{dL}$.

Therefore, the corresponding error is bounded by $\calO(D^{-d/p})$.
Since we require error to be bounded $\epsilon/3,$
setting \blue{$D=\calO(\epsilon^{-p/d})$ for some constant $p>0$} guarantees the result. 
We provide the detailed derivations as follows.

{\color{blue}
We follow \cite{kajitsuka2023transformers} considering Lipschitz (under $p$-norm) function class of continuous sequence-to-sequence. 
This consideration is practical as realistic input of transformer blocks are vector embedding in Euclidean space.
Let $f(\cdot):[0,1]^{d\times L}\to[0,1]^{d\times L}$ be the target function and $\bar{f}(\cdot)$ be the piece-wise constant approximation of regularity $D$.
Recall the $p$-norm difference between two function $f(\cdot)$ and $\bar{f}(\cdot)$.
\eqref{eqn:est_of_step_function} gives
\begin{align*}
d_p(f, \bar{f}) 
= & ~  (\int \|f(x)-\bar{f}(x)\|^p \dd x)^{1/p}\\
= & ~\calO(D^{dL-d}) \cdot  (B^p (1/D)^{dL})^{1/p} \\
= & ~ \calO(D^{(dL-d)/p}) \cdot \calO(D^{-dL/p})
\\
= & ~ \calO(D^{-d/p}).
\end{align*}
Here, $\calO(D^{-d/p}) = \epsilon$ implies $D = \calO(\epsilon^{-p/d})$ for some constant $p > 0$. For simplicity, we use $D = \calO(\epsilon^{-1/d})$ in our analysis without loss of generality.
}

\item
\textbf{Bound on Parameter $\delta$ in $f_{1}^{(\text{FF})}$.}

In the quantization operation realized by the network,
we need to ensure the error within region $(i/D,i/D+\delta)$ does not affect the desired interval $(i/D,(i+1)/D)$ for $i\in[D]$.

Thus,  we need $\delta=o(1/D)$.

\end{itemize}

This completes the proof.
\end{proof}

Building upon \cref{lemma:setting_D_R}, 
we extend the results to derive explicit parameter bounds for matrices regarding the transformer-based universal approximation framework. 
That is,
we ensure a more precise quantification of parameter constraints across the architecture.

\begin{lemma}[Transformer Matrices Bounds]
\label{lemma:trans_para_bound}
Consider an input sequence $Z\in[0,1]^{d\times L}.$
Let $f(Z):[0,1]^{d\times L}\to \R^{d\times L}$ be any permutation equivariant and continuous sequence-to-sequence function on compact support $[0,1]^{d\times L}$.
For the transformer network $f'\in\calT_R^{r,h,s}$ defined in \cref{def:Tran_reshape_class} to approximate $f$ within $\epsilon$ precision, i.e., $d_{p}(f,f^{\prime})<\epsilon$, the following parameter bounds must hold for $d\geq1$ and $L\geq2$:
\begin{align*}
& 
\norm{W_{Q}}_{2}=\norm{W_{K}}_{2}
=
\calO(d\cdot\epsilon^{-(\frac{2dL+1}{d})})(\log{L})^{\frac{1}{2}});\\
&\norm{W_{Q}}_{2,\infty}=\norm{W_{K}}_{2,\infty}
=
\calO(d^{\frac{3}{2}}\cdot\epsilon^{-(\frac{2dL+1}{d})}(\log{L})^{\frac{1}{2}}); \\
& \norm{W_O}_{2}=\calO\left(\sqrt{d}\epsilon^{\frac{1}{d}}\right); 
\norm{W_O}_{2,\infty}=\calO\left(\epsilon^{\frac{1}{d}}\right);\\
&\norm{W_{V}}_{2}=\calO(\sqrt{d});
\norm{W_{V}}_{2,\infty}=\calO(d); \\ 
& \norm{W_{1}}_{2}=\calO\left(d\epsilon^{-\frac{1}{d}}\right), \norm{W_{1}}_{2,\infty}=\calO\left(\sqrt{d}\epsilon^{-\frac{1}{d}}\right);\\
& \norm{W_{2}}_{2}=\calO\left(d\epsilon^{-\frac{1}{d}}\right);
\norm{W_{2}}_{2,\infty}=\calO\left(\sqrt{d}\epsilon^{-\frac{1}{d}}\right);\\
&\norm{E^{\top}}_{2,\infty}=\calO\left(d^{\frac{1}{2}}L^{\frac{3}{2}}\right).
\end{align*}
For the case $L=1$,
the parameter bounds remain valid with the substitution of $\log{L}$ with $1$.
\end{lemma}
\begin{proof}
For the self-attention layer,
we denote the separatedness of the input tokens by $(\gamma_{\min},\gamma_{\max},\epsilon_{s})$ and the separatedness of the output tokens by $(\gamma,\delta_{s})$.
Moreover,
in \eqref{eqn:quantize_term_f1} we denote the parameter taken in  $f_{1}^{\text{FF}}$ corresponding to the granularity by $\delta_{f_{1}}$.

\begin{itemize}
    \item 
    \textbf{Bounds for $W_{Q}$ and $W_{K}$ in $f^{\text{(SA)}}$.}

From the universal approximation theorem of transformer \cref{thm:Transformer_as_universal_approximators},
with $p_{i}, p_{i}^{\prime}\in\R^{s}$ and $q_{i}, q_{i}^{\prime}$,
being any unit vectors in $\R^{d}$,
we construct rank $\rho$ matrix $W_{Q}$ and $W_{K}$ as
\begin{align*}
    & W_{K}
    = \sum_{i=1}^{\rho}p_{i}q_{i}^{\top}\in\RR^{s\times d},
    \\
    & W_{Q}
    = \sum_{i=1}^{\rho} p^\prime_{i}q_{i}^{\prime \top} \in\R^{s\times d},
\end{align*}    
with the identity $p_{i}^{\top}p_{i}^{\prime}=(\abs{\calV}+1)^{4}d{\delta_{s}}/{(\epsilon_{s}\gamma_{\min})}$.
With this, we have the bound for $p_i, p_i^{\prime}$:
\begin{align} \label{eqn:p_q_bound}
    & \norm{p_i}= 
    \calO \left( 
    \abs{\calV}^{2}\sqrt{d\frac{\delta_{s}}{\epsilon_{s}\gamma_{\min}}}
    \right),
    \;
    & \norm{p_i^{\prime}}= 
    \calO \left( 
     \abs{\calV}^{2}\sqrt{d \frac{\delta_{s}}{\epsilon_{s}\gamma_{\min}}}
    \right).
    \end{align}
Summing over the set of $p_{i}^{\top}p_{i}^{\prime}$ for $i=1,\ldots,\rho$,
we obtain the bound for rank $\rho$ matrix $W_{Q}$ and $W_{K}$
\begin{align*}
        & \norm{W_Q}_2 = \sup_{\norm{x}_2=1}\norm{W_Q x}_2 \le C_{Q} = 
        \calO \left(
        \sqrt{\rho}  \abs{\calV}^{2} \sqrt{d\frac{\delta_{c}}{\epsilon_{c} \gamma_{\min}}}
        \right),
        \\
        & \norm{W_Q}_{2,\infty} 
        = \max_{1\le i \le d}\| (W_Q)_{(i,:)} \|_{2}
        \le C_{Q}^{2,\infty}
        = \calO \left(
        \rho  \abs{\calV}^{2} \sqrt{d \frac{\delta_{s}}{\epsilon_{s}\gamma_{\min }} }
        \right),
        \\
        & \| W_K \|_2 
        = \sup_{\norm{x}_2=1}\norm{W_K x}_2
        \le  C_{K} =
        \calO \left(
        \sqrt{\rho}  \abs{\calV}^{2}\sqrt{d\frac{\delta_{s}}{\epsilon_{s} \gamma_{\min}}}
        \right),
        \\
        & \| W_K \|_{2,\infty} 
        =\max_{1\le i \le d}\| (W_K)_{(i,:)} \|_{2}
        \le  C_{K}^{2,\infty} =
        \calO \left(
        \rho  \abs{\calV}^{2}\sqrt{d \frac{\delta_{s}}{\epsilon_{s}\gamma_{\min }} }
        \right),
    \end{align*}
where $\rho\leq s$ and the head size $s\leq d$.

After the first step quantization,
we obtain vocabulary bounds $\abs{\calV}=\calO(D^{dL})$ and output sequences with $(1/D,\sqrt{d},1/D)$ tokenwise separatedness.
Also,
in \cref{thm:Transformer_as_universal_approximators} we take $\delta_{s}=4\log{L}$ so that $f^{\text{(SA)}}$ is a contextual mapping.

Next, 
by \cref{lemma:setting_D_R}, 
we need $D=\calO(\epsilon^{1/(dL)})$ for \cref{thm:Transformer_as_universal_approximators} to hold.

Combining all the components,
we have the bounds for $W_Q$ and $W_K$
\begin{align*}
& \norm{W_{Q}}_{2},\norm{W_{K}}_{2}=\calO\left(dD^{2dL+1}(\log{L})^{\frac{1}{2}}\right)
=
\calO(d\epsilon^{\frac{2dL+1}{dL}}(\log{L})^{\frac{1}{2}}),\quad\\
&\norm{W_{Q}}_{2,\infty},\norm{W_{K}}_{2,\infty}=\calO\left(d^\frac{3}{2}D^{2dL+1}(\log{L})^{\frac{1}{2}}\right)
=
\calO(d^{\frac{3}{2}}\epsilon^{\frac{2dL+1}{dL}}(\log{L})^{\frac{1}{2}})
\end{align*}

\item 
\textbf{Bounds for $W_O$ and $W_{V}$ in $f^{\text{(SA)}}$.}

Following the construction of $W_{Q}$ and $W_{K}$
in \cref{thm:Transformer_as_universal_approximators}, we have the relation for $W_V$ and $W_O$ as 
\begin{align*}
& W_{V}
= \sum_{i=1}^{\rho} p^{\prime\prime}_{i}q_i^{\prime \prime \top}\in\R^{s\times d},
\\
& W_O 
= \sum_{i=1}^{\rho} p_{i}^{\prime\prime\prime} {p_{i}^{\prime\prime}}^{\top}\in\R^{d\times s},
\end{align*}
with the identity $\norm{p_i^{\prime\prime\prime}}\lesssim {\epsilon_{s}}/{(4 \rho \gamma_{\max}\norm{p^{\prime\prime}_{i}})}$ from \eqref{eqn:construct_WO_new},
and $p^{\prime\prime}_{i} \in \R^s$ is any nonzero vector.

Along with the $(\gamma_{\min}=1/D,\gamma_{\max}=\sqrt{d},\epsilon_s=1/D)$ separateness and taking $D=\calO(\epsilon^{1/(dL)})$,
we have the following bounds for $W_V$ and $W_O$:
\begin{align*}
        & \| W_V \|_2 
        = \sup_{\norm{x}_2=1}\norm{W_V x}_2
        \le C_{V} =
        \calO \left( \sqrt{\rho}\right),
        \\
        & \| W_V \|_{2,\infty} 
        =\max_{1\le i \le d}\| (W_V)_{(i,:)} \|_{2}
        \le C_{V}^{2,\infty} =
        \calO \left(\rho\right),
        \\
        & \| W_{O} \|_2 
        = \sup_{\norm{x}_2=1}\norm{W_O x}_2
        \le C_{O} = 
        \calO \left(\sqrt{\rho}\cdot\rho^{-1}\cdot \gamma_{\max}^{-1}\cdot\epsilon_{s}\right)
        =
        \calO \left(d^{-1}\epsilon^{-\frac{1}{dL}}\right)
        \\
        & \| W_{O} \|_{2,\infty} 
        =\max_{1\le i \le s}\| (W_O)_{(i,:)} \|_{2}
        \le C_{O}^{2,\infty} = 
        \calO \left(\rho\cdot\rho^{-1}\cdot \gamma_{\max}^{-1}\cdot\epsilon_{s}\right)
        =
        \calO \left(d^{-\frac{1}{2}}\epsilon^{-\frac{1}{dL}}\right).
    \end{align*}
    Note that we use the fact $\max \rho = d$ in the last two lines.
    
\item 
\textbf{Bounds for $W_{1}$ in $f_{1}^{\text{FF}}$.}

In order to approximate the quantization in \cref{thm:Transformer_as_universal_approximators},
we set up $f_{1}^{\text{FF}}$ as in \eqref{eqn:quantize_term_f1}
where every entry of $W_{1}$ in the layer is bounded by $\calO(1/\delta)$.
Therefore we have
\begin{align}
\label{eqn:W_1_bound}
        & \norm{W_{1}}_{2, \infty} \le C_{F_{1}}^{2,\infty} = \calO \left( \dfrac{\sqrt{d}}{\delta} \right), \\
        & \norm{W_{1}}_{2} \le \norm{W_{1}}_{F}\leq C_{F_{1}} = \calO \left( \dfrac{d}{\delta} \right),
\end{align}
where the bound for $\delta$ is given from \cref{lemma:setting_D_R}.
We set $\delta=\nu D^{-1}$ for some $\nu\in(0,1)$ such that we have the bounds $\calO(\sqrt{d}\epsilon^{1/(dL)})$ and $\calO(d\epsilon^{1/(dL)})$ respectively.

\item 
\textbf{Bounds on $W_{2}$ in $f^{\text{FF}}$.}

The bounds for $\norm{W_{2}}_{2}, \norm{W_{2}}_{2,\infty}$ in \eqref{eqn:bump_function_def} follow the same argument as for $W_{1}$,
with the replacement of the largest element with the scaling factor $R$.
So we have
\begin{align}
        & \norm{W_{2}}_{2, \infty} \le C_{F_{2}}^{2,\infty} = \calO \left( {\sqrt{d}R} \right), \\
        & \norm{W_{2}}_{2} \le C_{F_{2}} = \calO \left( dR\right).
\end{align}
Again,
by \cref{lemma:setting_D_R},
we take $R=\calO(D)=\calO(\epsilon^{1/(dL)})$ such that we have the bounds $\calO(\sqrt{d}\epsilon^{1/(dL)})$ and $\calO(d\epsilon^{1/(dL)})$ respectively.

\item 
\textbf{Bounds on Positional Encoding Matrix $E$.}

For $\norm{E^{\top}}_{2}, \norm{E^{\top}}_{2,\infty},$
following \cite{kajitsuka2023transformers},
it suffices to set the positional encoding:
\begin{align*}
E=\begin{pmatrix}
  2\gamma_{\text{max}} & 4\gamma_{\text{max}} & \cdots & 2L\gamma_{\text{max}} \\
  \vdots & \vdots & \ddots & \vdots \\
  2\gamma_{\text{max}} & 4\gamma_{\text{max}} & \cdots & 2L\gamma_{\text{max}}
\end{pmatrix}.
\end{align*}
Since the $\ell_{2}$ norm over every row is identical,
it suffices to derive 
\begin{align*}
\norm{{E^{\top}}}_{2,\infty}
=
\left({\sum_{i=1}^{L}(2i\gamma_{\max})^{2}}\right)^{\frac{1}{2}}
=
\left(
4\gamma_{\max}^{2}\frac{L(L+1)(2L+1)}{6}
\right)^{2}
=\calO\left(\gamma_{\max}L^{\frac{3}{2}}\right).
\end{align*}
Recall that we have the relation $\gamma_{\max}=\sqrt{d}$ in the self-attention layer.
Therefore,
we have the following bound for encoding matrix $E$:
\begin{align}
\norm{E^{\top}}_{2,\infty}\leq C_{E}=\calO(d^{1/2}L^{3/2}).
\end{align}
\end{itemize}
This completes the proof.
\end{proof}

\clearpage
\section{Proof of \texorpdfstring{\cref{thm:Main_1}}{}}\label{sec:appendix_proof_main1}
Our proof builds on the local smoothness properties of functions within H\"{o}lder spaces and the universal approximation of transformers.
While the universal approximation theory of transformers in \cref{sec:appendix_supp_theo_background} ensures arbitrarily small errors,
it does not account for the smoothness of functions in the result.
To incorporate the smoothness assumptions of interest, we propose the following three steps to integrate function smoothness into approximation theory of transformer architectures.

\begin{itemize}
    \item \textbf{Step 1.}
    Consider the integral form of $p_{t}(x_{t}|y)$ in \eqref{eqn:decomposed_cond}.
    We clip the input domain $\R^{d_{x}}$ into closed and bounded region $B_{x,N}$ in \eqref{eqn:discretize_x}.
    This facilitates the error analysis for the Taylor expansion approximation in the next step.
    The clipping error arises from the integral over the region outside $B_{x,N}$.
    We specify the clipping error in \cref{clipping_integral}.

    \item \textbf{Step 2.}
    We employ \blue{$k_1$-order and $k_2$-order} Taylor expansion for $p(x_{0}|y)$ and  $\exp(\cdot)$ in \eqref{eqn:decomposed_cond}.
    We construct \textit{the diffused local polynomial} in \cref{lem:f_1_expression} based on the Taylor expansion.
    We approximate $p_{t}$ and $\nabla p_{t}$ with \textit{the diffused local polynomial} $f_{1}(x,y,t)\in\R$ and $f_{2}(x,y,t)\in\R^{d_{x}}$ in \cref{lemma:diffused_local_polynomials} and \cref{lemma:diffused_local_polynomials_grad}.

    \item \textbf{Step 3.}
    We approximate $f_{1}(x,y,t)$, $f_{2}(x,y,t)$ with transformers in \cref{lemma:Trans_Approx_Poly,lemma:Trans_Approx_Poly_Gradient}.
    To construct the final score approximator with the transformer,
    we approximate necessary algebraic operators 
    in \cref{lemma:Clipping_function,lemma:approx_prod_with_trans,lemma:inverse_trans,lemma:trans_approx_mean,lemma:trans_approx_variance}.
    We provide the output bound of our transformer model in \cref{bounds_on_score}.
    We combine all components into
    \cref{lemma:Score_Approx_Trans}, and complete the proof of \cref{thm:Main_1}.
\end{itemize}

\paragraph{Organization.} \cref{sec:axuiliary_lem_thm1} includes details regarding the three steps with auxiliary lemmas for supporting our proof.
\cref{proof:thm:Main_1} includes the main proof of \cref{thm:Main_1}. 

\subsection{Auxiliary Lemmas}
\label{sec:axuiliary_lem_thm1}

\paragraph{Step 1: Clip
 $\R^{d_{x}}\times[0,1]^{d_{y}}$ for $p_{t}(x|y)$.}
 
We introduce a helper lemma on the clipping integral.
\begin{lemma}[Approximating Clipped Multi-Index Gaussian Integral, Lemma A.8 of \cite{fu2024unveil}]
\label{clipping_integral}
Assume \cref{assumption:conditional_density_function_assumption_1}.
Consider any integer vector $\kappa\in\mathbb{Z}_{+}^{d_{x}}$ with $\norm{\kappa}_{1}\leq n$.
There exists a constant $C(n,d_x)\geq1$, 
such that for any $x\in\R^{d_{x}}$ and $0<\epsilon\leq 1/e$,
it holds
\begin{align}\label{eqn:clipped_integral}
\int_{\R^{d_{x}}\setminus B_{x}}\abs{\left(\frac{\alpha_{t}x_{0}-x}{\sigma_{t}}\right)^{\kappa}} \cdot 
p(x_{0}|y) \cdot \frac{1}{\sigma_{t}^{d}(2\pi)^{d/2}}\exp(-\frac{\norm{\alpha_{t}x_{0}-x}^{2}}{2\sigma_{t}^{2}}) \dd x_{0}
\leq\epsilon,
\end{align}
where $\left(\frac{\alpha_{t}x_{0}-x}{\sigma_{t}}\right)^{\kappa}\coloneqq((\frac{\alpha_{t}x_0[1]_1-x[1]}{\sigma_{t}})^{\kappa[1]},(\frac{\alpha_{t}x_{0}[2]-x[2]}{\sigma_{t}})^{\kappa[2]},\ldots,(\frac{\alpha_{t}x_0[d_x]-x[d_x]}{\sigma_{t}})^{\kappa[d_x]})$ is a \textit{multi-indexed} vector and
\begin{align*} 
B_{x} 
\coloneqq 
\Big[ \frac{x-\sigma_{t}C(n,d_{x})\sqrt{\log{(1/\epsilon)}}}{\alpha_{t}},  \frac{x+\sigma_{t}C(n,d_{x})\sqrt{\log{(1/\epsilon)}}}{\alpha_{t}} \Big]
\\
\bigcap
\Big[-C(n,d_{x})\sqrt{\log{(1/\epsilon)}},C(n,d_{x})\sqrt{\log{(1/\epsilon)}}\Big]^{d_{x}}.
\end{align*}
\end{lemma}
\begin{remark}\label{remark:N_resolution}
    $B_x$ is a bounded domain. \cref{clipping_integral} provides the difference between integrals of the form \eqref{eqn:clipped_integral} on $\R^{d_x}$ and on $B_x$. 
    The difference becomes arbitrarily small with precision $\epsilon = 1/N$.
\end{remark}

Based on \cref{clipping_integral},
we have the following considerations:
\begin{itemize}

    \item For each $x\in\R^{d_x}$, consider a bounded domain
    \begin{align}
    & ~ B_{x,N} \label{eqn:discretize_x}\\
    \coloneqq & ~ 
    \underbrace{\left[ \frac{x-\sigma_{t}C(0,d_x)\sqrt{\beta\log{N}}}{\alpha_{t}},  \frac{x+\sigma_{t}C(0,d_x)\sqrt{\beta\log{N}}}{\alpha_{t}} \right]}_{\rm (I)}
    \bigcap
    \underbrace{\left[-C(0,d_x)\sqrt{\beta\log{N}},C(0,d_x)\sqrt{\beta\log{N}}\right]^{d_{x}}}_{\rm(II)},    
    \nonumber
    \end{align}
    where $C(0,d_x)$ is some positive constant depending on $d_x$ and $N$.
    Here, we pick $n=0$ for $C(n,d_x)$ to reduce \eqref{eqn:clipped_integral} to 
    \begin{align*}
    p_t(x|y)=\int_{\R^{d_{x}}\setminus B_{x,N}}  p(x_{0}|y) \cdot \frac{1}{\sigma_{t}^{d}(2\pi)^{d/2}}\exp(-\frac{\norm{\alpha_{t}x_{0}-x}^{2}}{2\sigma_{t}^{2}}) \dd x_{0}
    \leq\epsilon = 1/N.  
    \end{align*}
    This motivates a polynomial expansion of \eqref{eqn:decomposed_cond} on $B_{x,N}$ with precision $1/N$.
    
    \item Uniformly discretize each dimension of $B_{x,N}$ into $N$ segments.
    Note that while not necessary, it is possible to pick a $C(0,d_x)$ such that grids in $B_{x,N}$ are non-overlapping.

    \item Uniformly discretize each dimension of $[0,1]^{d_y}$ into $N$ segments of length $1/N$.
\end{itemize}
This discretization of domains leads to $N^{d_x+d_y}$ hypercubes on bounded domain $B_{x,N}\times [0,1]^{d_y}$.

\begin{remark}
    For any $x\in\R^{d_x}$, we shorthand \eqref{eqn:discretize_x} with
    \begin{align}\label{eqn:C_x}
        B_{x,N} = \[-C_x\sqrt{ \log N},C_x\sqrt{ \log N}\]^{d_x},
    \end{align}
    where $C_x$ summarize all factors except $\sqrt{\log N}$ in all dimensions of $x\in\R^{d_x}$.
    Moreover, when content is clear, we suppress the notation dependence on $d_x$ for \eqref{eqn:C_x}.
    Namely, we use the notation $B_{x,N} = \[-C_x\sqrt{ \log N},C_x\sqrt{ \log N}\]$ and $B_{x,N} = \[-C_x\sqrt{ \log N},C_x\sqrt{ \log N}\]^{d_x}$   interchangeably.
  
\end{remark}

\begin{remark}
    \cref{clipping_integral} ensures that we can approximate the Gaussian integral of any polynomial function of the form \eqref{eqn:clipped_integral} on $\R^{d_x}$ with the same integral on $B_x$ to an arbitrary precision $0 < \epsilon < 1/e$.
    This motivate us to approximate functions on $\R^{d_x}$ with polynomials evaluated at $x\in\R^{d_x}$ on $B_{x,N}$.
    A natural choice is through Taylor expansion around $x\in\R^{d_x}$,
    as the H\"{o}lder class assumption guarantees local smoothing behavior for our error analysis.
\end{remark}

\paragraph{Step 2: Approximate $p_{t}(x|y)$ and $\nabla p_{t}(x|y)$ with Taylor Expansion.}
We begin with the definition.
\begin{definition}
[Normalization of $B_{x,N}$]
\label{def:normal_B_x_N}
Consider the clipping in \cref{clipping_integral} and the initial conditional distribution $p(x_0|y)$ with closed and bounded support $B_{x,N}\times[0,1]^{d_y}$.
We define
$R_{B}\coloneqq (2C(0,d)\sqrt{\beta\log{N}})$ and 
$x_{0}^{\prime}\coloneqq x_0/R_{B} + 1/2$.
Moreover,
we define $M(x_{0}^{\prime},y)\coloneqq p(R_{B}(x_0^{\prime}-1/2)|y)$.
\end{definition}

\begin{remark}
\label{rmk:M_x0prime1}
The purpose of \cref{def:normal_B_x_N} is to simplify the process of discretizing $B_{x,N}\times[0,1]^{d_y}$ into $N^{d_x+d_y}$ hypercubes.
In particular, $M(x_0', y)$ has compact support on $[0,1]^{d_x + d_y}$, where $R_B$ denotes the length of each coordinate of $B_{x, N}$, and $x_0' \in [0,1]^{d_x}$ represents $x_0$ normalized on $B_{x, N}$.
\end{remark}

\begin{remark}
\label{rmk:M_x0prime2}
The only difference between $M(x_0^{\prime},y)$ and $p(x_0|y)$ lies in their respective domains,
leading to the difference in the size of the \holder ball radius.
Recall that under \cref{assumption:conditional_density_function_assumption_1},
we have $p(x_0|y)\in\calH^{\beta}(\R^{d_x}\times[0,1]^{d_y},B)$.
Here we have
$M(x_0^{\prime},y)\in\calH([0,1]^{d_{x}+d_{y}},BR_{B}^{\textcolor{blue}{k_1}})$.
This follows from the fact that $p(\cdot|y)$ is \blue{$k_1$}-time differentiable so that the radius scale by a factor of $R_B^{\textcolor{blue}{k_1}}$.
\end{remark}

\begin{lemma}[Diffused Local Polynomial, Modified from \cite{fu2024diffusion}]
\label{lem:f_1_expression}
Assume \cref{assumption:conditional_density_function_assumption_1}.
We write $p_t(x|y)$ into the product of $p(x_0|y)$ and $\exp(\cdot)$:
\begin{align*}
p_{t}(x|y) = 
\int_{\R^{d_{x}}}p(x_{0}|y)p_{t}(x|x_{0})\dd x_{0} =\int_{\R^{d_{x}}}\frac{1}{\sigma_{t}^{d_x}(2\pi)^{d_x/2}} {p(x_{0}|y)}
{\exp(-\frac{\norm{\alpha_{t}x_{0}-x}^{2}}{2\sigma_{t}^{2}})}
\dd x_{0}.
\end{align*}
Then we approximate $p(x_{0}|y)$ and $\exp(-\frac{\norm{\alpha_{t}x_{0}-x}^{2}}{2\sigma_{t}^{2}})$ with $\textcolor{blue}{k_1}$-order Taylor polynomial and $\textcolor{blue}{k_2}$-order Taylor polynomial within $B_{x,N}$ respectively.
Altogether, 
we approximate $p_t(x|y)$ with the following \textit{diffused local polynomial} with the bounded domain $B_{x,N}$ around $x$ in \eqref{eqn:C_x}: 
\begin{align}\label{eqn:diff_local_poly}
f_{1}(x,y,t) = \sum_{v\in[N]^{d},w\in[N]^{d_{y}}}
\sum_{\norm{n_{x}}_{1}+\norm{n_{y}}_{1}\leq \textcolor{blue}{k_1}} 
\frac{R_{B}^{\norm{n_{x}}}}{n_{x}!n_{y}!}\frac{\partial^{n_{x}+n_{y}}{p}}{\partial{x}^{n_{x}}\partial{{y}}^{n_{y}}}\Bigg|_{x=R_{B}(\frac{v}{N}-\frac{1}{2}),y=\frac{w}{N}}\Phi_{n_{x},n_{y},v,w}(x,y,t),
\end{align}
where
\begin{itemize}
    \item $\phi(\cdot)$ is the trapezoid function.

    \item $g(x,n_{x},v,\textcolor{blue}{k_2}) \coloneqq \frac{1}{\sigma_{t}\sqrt{2\pi}}\int\left(\frac{x_{0}}{R}+\frac{1}{2}-\frac{v}{N}\right)^{n_{x}}\frac{1}{\textcolor{blue}{k_2}!}\left(-\frac{\abs{x-\sigma_{t}x_{0}^{2}}}{2\sigma_{t}^{2}}\right)^{\textcolor{blue}{k_2}}\dd x_{0}$.
    
    \item $\Phi_{n_{x},n_{y},v,w}(x,y,t) \coloneqq \left(y-\frac{w}{N}\right)^{n_{y}}\prod_{j=1}^{d_{y}}\phi\left(3N(y[j]-\frac{w}{N})\right)\prod_{i=1}^{d_{x}}\sum_{\textcolor{blue}{k_2}<p}g(x[i],n_{x}[i],v[i],\textcolor{blue}{k_2})$.
\end{itemize} 
\end{lemma}

\begin{remark}%
    The form of the diffused local polynomial arises from the Taylor expansion approximation applied on each grid point within $[0,1]^{d_x+d_y}$,
    with $v\in[N]^{d_{x}}$ and $w\in[N]^{d_{y}}$ denoting the specific grid point undergoing approximation.
    
\end{remark}

\begin{remark}%
    The H\"{o}lder space assumption in \cref{assumption:conditional_density_function_assumption_1} establishes an upper bound on the error arising from the remainder term in the Taylor expansion. 
    This ensures the approximation accuracy is well-controlled.
\end{remark}

\begin{proof}[Proof Sketch]
\label{proofsketch:local_diffused}
We provide the proof overview of \cref{lem:f_1_expression}. 
with the following three steps.

\textbf{Step A: Clip $\R^{d_x}\times[0,1]^{d_y}$.}

We clip the domain $\R^{d_x}\times[0,1]^{d_y}$ into closed and bounded region $B_{x,N}$.%

\textbf{Step B: Replace $p(x_{0}|y)$ with $\textcolor{blue}{k_1}$-order Taylor Polynomials.}

We discretize $[0,1]^{d_{x}+d_{y}}$ into $N^{d_x+d_y}$ hypercubes.
We apply Taylor expansion to each grid point.
For areas not located on any grid point,
we construct a trapezoid function and an indicator function to control the approximation error.

\textbf{Step C: Replace $\exp(\cdot)$ with $\textcolor{blue}{k_2}$-order Taylor Polynomials.}

We apply Taylor expansion to approximate regions within $B_{x,N}$ for $\exp(\cdot)$.
Note that we leverage the explicit form of the exponential function to achieve accurate approximation without additional discretization as in previous step.

\textbf{Step D: Altogether, the Diffused Local Polynomials.}

We combine these 4 steps and construct \textit{the diffused local polynomial} \eqref{eqn:diff_local_poly}.
\end{proof}

\begin{proof}[Proof of \cref{lem:f_1_expression}]

We demonstrate details regarding the three steps.
\begin{itemize}
    \item 
\textbf{Step A: Clip $\R^{d_x}\times[0,1]^{d_{y}}$.}

We take $\kappa[i]=0$ for $i=[d_x]$ and set $\epsilon=N^{-\beta}$ in \cref{clipping_integral}.
This gives closed and bounded domain $B_{x,N}$ specified in \eqref{eqn:C_x} and clipping-induced error:
\begin{align}
\label{eqn:diff_proof_step_A}
\abs{
p_{t}(x|y)
-
\int_{B_{x,N}}p(x_{0}|y) \cdot \frac{1}{\sigma_{t}^{d}(2\pi)^{d/2}}\exp(-\frac{\norm{\alpha_{t}x_{0}-x}^{2}}{2\sigma_{t}^{2}}) \dd x_{0}}
\leq N^{-\beta}.
\end{align}

   \item 
\textbf{Step B: Replace $p_{0}(x_{0}|y)$ with $\textcolor{blue}{k_1}$-order Taylor Expansion.}

We construct a approximator $Q(x_0^{\prime},y)$ for $M(x_0^{\prime},y)$ with domain $[0,1]^{d_x+d_y}$.\footnote{Recall
$R_{B}\coloneqq (2C(0,d)\sqrt{\beta\log{N}})$,  
$x_{0}^{\prime}\coloneqq x_0/R_{B} + 1/2$,
and  $M(x_{0}^{\prime},y)\coloneqq p(R_{B}(x_0^{\prime}-1/2)|y)$ from \cref{def:normal_B_x_N}.}
At the end of this step,
we reset $x_0^{\prime}=x_{0}/R_B+1/2$ in $Q(x_0^{\prime},y)$ as the final approximator of $p(x_0|y)$.

\begin{itemize}
    \item \textbf{Step B.1: Discretize $[0,1]^{d_x+d_y}$.}
    
    We uniformly discretize $[0,1]^{d_x+d_y}$ into grid points $[0,1/N,2/N,\ldots,(N-1)/N,1]^{d_x+d_y}$.

    \item \textbf{Step B.2: Implement Taylor Expansion.}

We construct the $\textcolor{blue}{k_1}$-order Taylor polynomial $P_{v,w}(x,y)$ at point $(v/N,w/N)$ for $M(x_0^{\prime},y)$:\footnote{Please see \cref{rmk:M_x0prime1,rmk:M_x0prime2} for  details.}
\begin{align}
\label{eqn:s_order_P}
P_{v,w}(x_0^{\prime},y)
\coloneqq
\sum_{\norm{n_{x}}_{1}+\norm{n_{y}}_{1}\leq \textcolor{blue}{k_1}}\frac{1}{n_{x}!n_{y}!}\frac{\partial^{n_{x}+n_{y}}M}{\partial x^{n_{x}}\partial y^{n_{y}}}\Bigg|_{x_0^{\prime}=\frac{v}{N},y=\frac{w}{N}}\left(x_0^{\prime}-\frac{v}{N}\right)^{n_{x}}\left(y-\frac{w}{N}\right)^{n_{y}}.
\end{align}

For $x_0^{\prime}$ and $y$ not located on any grid point,
we construct an indicator function that ensures $\norm{x_0^{\prime}-v/N}_{\infty}<1/N$ and $\norm{y-w/N}_{\infty}<1/N$ in the next step.
For now, 
we assume these conditions hold.

To analyze the error,
we expand the target function $M(x_0^{\prime},y)$.
By Taylor's theorem,
there exist $\theta_{x}\in[0,1]^{d_{x}}$ and $\theta_{y}\in[0,1]^{d_{y}}$ such that
\begin{align*}
M(x_0^{\prime},y) = 
&~ \sum_{\norm{n_{x}}_{1}+\norm{n_{y}}_{1}<\textcolor{blue}{k_1}}\frac{1}{n_{x}!n_{y}!}\cdot\frac{\partial^{n_{x}+n_{y}}M}{\partial x_0^{\prime n_{x}}\partial{y}^{n_{y}}}\Bigg|_{x_0^{\prime}=\frac{v}{N},y=\frac{w}{N}}\left(x_0^{\prime}-\frac{v}{N}\right)^{n_{x}}\left(y-\frac{w}{N}\right)^{n_{y}}\\
&~ +
\sum_{\norm{n_{x}}_{1}+\norm{n_{y}}_{1}=\textcolor{blue}{k_1}}\frac{1}{n_{x}!n_{y}!}\cdot\frac{\partial^{n_{x}+n_{y}}M}{\partial x_0^{\prime n_{x}}\partial{y}^{n_{y}}}\Bigg|_{x_0^{\prime}=x_1,y=y_1}\left(x_0^{\prime}-\frac{v}{N}\right)^{n_{x}}\left(y-\frac{w}{N}\right)^{n_{y}},
\end{align*}
where $x_1=(1-\theta_{x})v/N + \theta_{x}x_0^{\prime}$ and $y_1=(1-\theta_y)w/N+\theta_y y$.
This ensures $x_1$ lies between $x_0^{\prime}$ and $v/N$,
and $y_1$ lies between $y$ and $w/N$.

Note that the difference between $P_{v,w}(x_0^{\prime},y)$ and $M(x_0^{\prime},y)$ stems from the different value taken in $\partial^{n_{x}+n_{y}}M/(\partial x_0^{\prime n_{x}}\partial y^{n_{y}})$ for all terms in the series with $\norm{n_{x}}_{1}+\norm{n_{y}}_{1}=\textcolor{blue}{k_1}$.

To study the error,
let $z=(x_0^{\prime},y)$ and
recall from the definition of H\"{o}lder norm (\cref{def:holder_norm_space}):
\begin{align}
\label{eqn:holder_regularity}
\max\limits_{\alpha:\norm{\alpha}_{1}=\textcolor{blue}{k_1}}\sup\limits_{z\neq z^{\prime}}\frac{\abs{\partial^{\textcolor{blue}{k_1}}M(z)-\partial^{\textcolor{blue}{k_1}}M(z^{\prime})}}{\norm{z-z^{\prime}}_{\infty}^{\gamma}}
<
\norm{M(x_0^{\prime},y)}_{\calH^{\beta}
([0,1]^{d_{x}+d_{y}})}
<
R_B^{\textcolor{blue}{k_1}}B.
\end{align}

We rewrite the error as
\begin{align*}
&~\abs{P_{v,w}(x_0^{\prime},y)-M(x_0^{\prime},y)}\\
\leq&~
\sum_{\norm{n_{x}}_{1}+\norm{n_{y}}_{1}=\textcolor{blue}{k_1}}\frac{1}{n_{x}!n_{y}!}\left(x_0^{\prime}-\frac{v}{N}\right)^{n_{x}}\left(y-\frac{w}{N}\right)^{n_{y}}
\abs{\underbrace{
\left(\frac{\partial^{n_{x}+n_{y}}M}{\partial x_0^{\prime n_{x}}\partial{y}^{n_{y}}}\Bigg|_{x_0^{\prime}=x_1,y=y_1}-\frac{\partial^{n_{x}+n_{y}}M}{\partial x_0^{\prime n_{x}}\partial{y}^{n_{y}}}\Bigg|_{x_0^{\prime}=\frac{v}{N},y=\frac{w}{N}}\right)}_{\text{Apply H\"{o}lder Regularity}}
}\\
\leq&~
\sum_{\norm{n_{x}}_{1}+\norm{n_{y}}_{1}=\textcolor{blue}{k_1}}\frac{1}{n_{x}!n_{y}!}\left(x_0^{\prime}-\frac{v}{N}\right)^{n_{x}}\left(y-\frac{w}{N}\right)^{n_{y}}
\underbrace{\norm{M(x_0^{\prime},y)}_{\calH^{\beta}([0,1]^{d_{x}+d_{y}})}}_{ \eqref{eqn:holder_regularity}}
\underbrace{\norm{[\theta_x x_0^{\prime},\theta_{y}y]-\frac{1}{N}[\theta_{x} v,\theta_{y}w]}_{\infty}^{\gamma}}_{\text{Controlled by indicator function \eqref{eqn:indicator}}}
\\
\leq&~
\sum_{\norm{n_{x}}_{1}+\norm{n_{y}}_{1}=\textcolor{blue}{k_1}}\frac{BR_B^{\textcolor{blue}{k_1}}}{n_{x}!n_{y}!N^{\norm{n_{x}}_{1}+\norm{n_{y}}_{1}+\gamma}}
=\frac{BR_B^{\textcolor{blue}{k_1}}(d_{x}+d_{y})^{\textcolor{blue}{k_1}}}{N^{\beta}\textcolor{blue}{k_1}!}.
\end{align*}

\item \textbf{B.3: Control Error  for the Off-Grid Regions.}

For regions not located on any grid point $(v/N, w/N)$, we construct an indicator function $\psi(x_0^{\prime}, y)$ to ensure that our Taylor approximation at $(v/N, w/N)$ does not deviate from $(x_0^{\prime}, y)$ by more than $1/N$ in $\ell_{\infty}$ distance.

Specifically, we define
\begin{align}
\label{eqn:indicator}
\psi_{v,w}(x_0^{\prime},y)
\coloneqq
\mathbbm{1}\left\{x_0^{\prime}\in\left(\frac{v-1}{N},\frac{v}{N}\right]\right\}\prod_{j=1}^{d_{y}}\phi\left(3N\left(y[j]-\frac{w}{N}\right)\right),
\end{align}
where $\phi(\cdot)$ is the trapezoid function:
\begin{align*}
\phi(\tau) = 
\begin{cases}
    1, & \vert \tau\vert < 1 \\
    2 - \vert \tau\vert, & \vert \tau\vert \in [1, 2] \\
    0, & \vert \tau\vert > 2.
\end{cases}
\end{align*}
Note that, $\psi_{v,w}$ is nonzero if and only if $x_0^{\prime}\in[(v-1)/N,v/N]$ and $y[j]\in[(w[j]-2/3)/N,(w[j]-2/3)/N)]$ for $j\in[d_y]$.
This guarantees $\norm{x_0^{\prime}-v/N}_{\infty}\leq1/N$ and $\norm{y-w/N}_{\infty}\leq1/N$.

\item 

\textbf{Step B.4: Construct the Final Approximator for $p(x_0|y)$.}

Combining \eqref{eqn:s_order_P} and \eqref{eqn:indicator},
we obtain an approximator of the form:
\begin{align*}
Q(x_0^{\prime},y)=\sum_{v,w}\psi_{v,w}(x,y)P_{v,w}(x_0^{\prime},y).
\end{align*}
Since for all $x\in(0,1]^{d_{x}}$ and $y\in[0,1]^{d_{y}}$ the indicator function $\psi_{v,w}(x_0^{\prime},y)$ sums to $1$,
it holds:
\begin{align}
\label{eqn:diffuse_local_step1.2}
\abs{M(x_0^{\prime},y)-Q(x_0^{\prime},y)}\leq\frac{BR^{\textcolor{blue}{k_1}}(d_{x}+d_{y})^{\textcolor{blue}{k_1}}}{\textcolor{blue}{k_1}!N^{\beta}}.
\end{align}
We conclude this step with the approximator $Q(x_0^{\prime},y)=Q(x_{0}/R_{B}+1/2,y)$ for $p(x_{0}|y)$.
\end{itemize}
\item 
\textbf{Step C: Replace $\exp(\cdot)$ with $k_2$-order Taylor Expansion.}

Recall that we set $B_{x,N}$ as
\begin{align*}
B_{x,N} 
= & ~ 
\left[ \frac{x-\sigma_{t}C(0,d_x)\sqrt{\beta\log{N}}}{\alpha_{t}},  \frac{x+\sigma_{t}C(0,d_x)\sqrt{\beta\log{N}}}{\alpha_{t}} \right]\\
 & ~ 
\bigcap
\left[-C(0,d_x)\sqrt{\beta\log{N}},C(0,d_x)\sqrt{\beta\log{N}}\right]^{d_{x}}.
\end{align*}

This gives $\abs{(x[i]-\alpha_{t}x_{0}[i])/\sigma_{t}}\leq C(0,d_x)\sqrt{\beta\log{N}}$ for any $i\in[d_{x}]$ and $x_{0}\in B_{x,N}$.

Furthermore,
we have
\begin{align}
\label{eqn:B_x_N_exp_bound}
\norm{(x-\alpha_t x_0)/\sigma_t}^2
=
\sum_{i=1}^{d_x}\abs{(x[i]-\alpha_{t}x_{0}[i])/\sigma_{t}}^2
\leq 
d_x\cdot\left(C(0,d_x)\sqrt{\beta\log{N}}\right)^2.
\end{align}

From this fact,
we implement the $\textcolor{blue}{k_2}$-order Taylor expansion to $\exp(-\norm{(x-\alpha_t x_0)/\sigma_t}^2/2)$:
\begin{align*}
& ~\abs{\exp(-\frac{\norm{x-\alpha_t x_0}^2}{2\sigma_t^2}) - \sum_{\textcolor{blue}{k_2}<u}\frac{1}{\textcolor{blue}{k_2}!}\left(-\frac{\norm{x-\alpha_t x_0}^2}{2\sigma_t^2}\right)^{\textcolor{blue}{k_2}}}
\annot{By Taylor theorem}\\
\leq & ~ 
\frac{1}{u!2^{u}}\left(\norm{\frac{x-\alpha_t x_0}{\sigma_t}}^2\right)^u \\
=  & ~ 
\frac{1}{u!2^{u}}\left(\sum_{i=1}^{d_x}\abs{(x[i]-\alpha_{t}x_{0}[i])/\sigma_{t}}^2\right)^u \\
\leq  & ~ 
\frac{1}{u!2^{u}}\left(d_x\cdot \left(C(0,d)\sqrt{\beta\log{N}}\right)^{2}\right)^{u}.
\end{align*}
for all $x_{0}\in B_{x,N}$,
and $u$ is a positive real number.

Following the choice of $u$ from \cite{fu2024unveil},
by utilizing the inequality $u!\geq(u/3)^u$ for $u\geq3$ and setting
\begin{align*}
u\coloneqq
\max\left(\frac{2}{3}C^2(0,d)\beta^2e\log{N},\beta\log{N}+\log{d_x}\right),
\end{align*}
we further write the bound as:
\begin{align}
\label{eqn:diffuse_local_step1.3_second}
\abs{\exp(-\frac{\norm{x-\alpha_t x_0}^2}{2\sigma_t^2}) - \sum_{\textcolor{blue}{k_2}<u}\frac{1}{\textcolor{blue}{k_2}!}\left(-\frac{\norm{x-\alpha_t x_0}^2}{2\sigma_t^2}\right)^{\textcolor{blue}{k_2}}}
\lesssim N^{-\beta}.
\end{align}

\item 

\textbf{Step D: The Diffused Local Polynomial.}

Substituting $p(x_0|y)$ and $\exp(\cdot)$ with their respective approximator in \eqref{eqn:diffuse_local_step1.2} and \eqref{eqn:diffuse_local_step1.3_second},
we obtain the following expression:
\begin{align}
\label{eqn:f_1_form_appendix}
f_{1}(x,y,t)=\frac{1}{\sigma_{t}^{d_x}(2\pi)^{\frac{d_x}{2}}}\int_{B_{x,N}}Q\left(\frac{x_{0}}{R_B}+\frac{1}{2}, y\right)\sum_{\textcolor{blue}{k_2}<u}\frac{1}{\textcolor{blue}{k_2}!}\left(-\frac{\norm{x-\alpha_{t}x_{0}}^{2}}{2\sigma_{t}^{2}}\right)^{\textcolor{blue}{k_2}}\dd x_{0}.
\end{align}

We term $f_1$ as \textit{diffused local polynomial}, following \cite{fu2024diffusion}.

\footnote{Further details regarding the derivation are in \cite[Appendix A.4]{fu2024unveil}.}Rearranging  \eqref{eqn:f_1_form_appendix},
we obtain the form
\begin{align}
f_{1}(x,y,t) = \sum_{v\in[N]^{d},w\in[N]^{d_{y}}}
\sum_{\norm{n_{x}}_{1}+\norm{n_{y}}_{1}\leq \textcolor{blue}{k_1}} 
\frac{R_B^{\norm{n_{x}}}}{n_{x}!n_{y}!}\frac{\partial^{n_{x}+n_{y}}{f}}{\partial{x}^{n_{x}}\partial{{y}}^{n_{y}}}\Bigg|_{x=\frac{v}{N},y=\frac{w}{N}}\Phi_{n_{x},n_{y},v,w}(x,y,t),
\end{align}
where 
\begin{itemize}
 \item $g(x,n_{x},v,\textcolor{blue}{k_2}) \coloneqq \frac{1}{\sigma_{t}\sqrt{2\pi}}\int\left(\frac{x_{0}}{R}+\frac{1}{2}-\frac{v}{N}\right)^{n_{x}}\frac{1}{\textcolor{blue}{k_2}!}\left(-\frac{\norm{x-\sigma_{t}x_{0}^{2}}}{2\sigma_{t}^{2}}\right)^{\textcolor{blue}{k_2}}\dd x_{0}$.

 \item $\Phi_{n_{x},n_{y},v,w}(x,y,t) \coloneqq \left(y-\frac{w}{N}\right)^{n_{y}}\prod_{j=1}^{d_{y}}\phi\left(3N(y[j]-\frac{w}{N})\right)\prod_{i=1}^{d_{x}}\sum_{\textcolor{blue}{k_2}<p}g(x[i],n_{x}[i],v[i],\textcolor{blue}{k_2})$.
\end{itemize}
\end{itemize}
This completes the proof.
\end{proof}

We specifies the error from the approximation of $p_{t}$ and $\nabla p_{t}$ with $f_1$ and $f_2$ in \cref{lemma:diffused_local_polynomials,lemma:diffused_local_polynomials_grad}.

\begin{lemma}[Approximation of $p_{t}(x|y)$ by Polynomials, Lemma A.4 of \cite{fu2024unveil}]
\label{lemma:diffused_local_polynomials}
Assume \cref{assumption:conditional_density_function_assumption_1}.
For any $x\in\R^{d_{x}}, y\in[0,1]^{d_{y}}$, $t>0$, and a sufficiently larger $N>0$, there exists a diffused local polynomial $f_{1}(x,y,t)$ with at most $N^{d_{x}+d_{y}}(d_{x}+d_{y})^{\textcolor{blue}{k_1}}$ monomials such that
\begin{align*}
\abs{f_{1}(x,y,t) - p_{t}(x|y)}\lesssim BN^{-\beta}\log^{\frac{d_{x}+\textcolor{blue}{k_1}}{2}}{N}.
\end{align*}
\end{lemma}

\begin{lemma}[Approximation of $\nabla \log p_{t}(x|y)$ by Polynomials, Lemma A.6 of \cite{fu2024unveil}]
\label{lemma:diffused_local_polynomials_grad}
Assume \cref{assumption:conditional_density_function_assumption_1}.
For any $x\in\R^{d_{x}}, y\in[0,1]^{d_{y}}$, $t>0$, and a sufficiently larger $N>0$, there exists $f_{2}\coloneqq(f_{2}[1],\ldots,f_{2}[d_{x}])^{\top}\in\R^{d_{x}}$ with local diffused polynomial $f_2[i]$ such that
\begin{align*}
\abs{f_{2}(x,y,t)[i] - \sigma_{t}\nabla{p_{t}(x|y)}[i]}\lesssim BN^{-\beta}\log^{\frac{d_{x}+\textcolor{blue}{k_1}+1}{2}}{N},
\end{align*}
where each $f_{2}[i]$ contains at most $N^{d_{x}+d_{y}}(d_{x}+d_{y})^{\textcolor{blue}{k_1}}$ monomials.
\end{lemma}

We have finished the approximation of $p_{t}$ and $\nabla p_{t}$ with diffused local polynomial $f_{1}$ and $f_{2}$.

\paragraph{Step 3. Approximate Diffused Local Polynomials and Algebraic Operators with Transformers.}

First,
we utilize universal approximation capabilities of transformers to deal with $f_{1}, f_{2}$ established in previous step. 
Second,
we employ similar scheme to approximate several algebraic operators necessary in final score approximation.
Lastly,
we present the incorporation of these components in \cref{lemma:Score_Approx_Trans} with a unified transformer architecture and corresponding parameter configuration.

\begin{itemize}
    \item 
    \textbf{Step 3.1: Approximate the Diffused Local Polynomials $f_{1}$ and $f_{2}$.}
    
    We invoke the universal approximation theorem of transformer (\cref{thm:Transformer_as_universal_approximators}).
    We utilize network consisting of one transformer block and one feed-forward layer (see \cref{fig:condition_DiT} and \cref{def:transformer_class}).
    
    \begin{lemma}
    [Approximate Scalar Polynomials with Transformers]
    \label{lemma:Trans_Approx_Poly}
    Assume \cref{assumption:conditional_density_function_assumption_1}.
    Consider the diffused local polynomial $f_{1}$ in \cref{lemma:diffused_local_polynomials}.
    For any $\epsilon>0$, there exists a transformer $\calT_{f_{1}}\in\calT_R^{\textcolor{blue}{h,s,r}}$, such that for any $x\in[-C_{x}\sqrt{\log{N}},C_{x}\sqrt{\log{N}}]^{d_{x}}, y\in[0,1]^{d_{y}}$ and $t\in[N^{-C_{\sigma}},C_{\alpha}\log{N}]$
    it holds 
    \begin{align*}
    \abs{f_{1}(x,y,t)-\calT_{f_{1}}(x,y,t)[d_{x}]}\leq\epsilon.
    \end{align*}
    The parameter bounds in the Transformer network class satisfy
    \begin{align*}
    & ~ \norm{W_{Q}}_{2},\norm{W_{K}}_{2}=\calO\left(d\epsilon^{-\frac{2dL+4d+1}{d}}(\log{L})^{\frac{1}{2}}\right);
    \\
    &~ \norm{W_{Q}}_{2,\infty},\norm{W_{K}}_{2,\infty}=\calO\left(d^\frac{3}{2}\epsilon^{-\frac{2dL+4d+1}{d}}(\log{L})^{\frac{1}{2}}\right);
    \\
    & ~ \norm{W_{V}}_{2}=\calO(\sqrt{d});
    \norm{W_{V}}_{2,\infty}=\calO(d);
    \\
    &~
    \norm{W_{O}}_{2}=\calO\left(\sqrt{d}\epsilon^{\frac{1}{d}}\right); 
     \norm{W_{O}}_{2,\infty}=\calO\left(\epsilon^{\frac{1}{d}}\right);
    \\
    & ~ \norm{W_{1}}_{2}=\calO\left(d\epsilon^{-\frac{1}{d}}\cdot\log{N}\right);
    \norm{W_{1}}_{2,\infty}=\calO\left(\sqrt{d}\epsilon^{-\frac{1}{d}}\cdot\log{N}\right);
    \\
    & ~ \norm{W_{2}}_{2}=\calO\left(d\epsilon^{-\frac{1}{d}}\right);
    \norm{W_{2}}_{2,\infty}=\calO\left(\sqrt{d}\epsilon^{-\frac{1}{d}}\right); \norm{E^{\top}}_{2,\infty}=\calO\left(d^{\frac{1}{2}}L^{\frac{3}{2}}\right).
    \end{align*}
    \end{lemma}
    
    \begin{proof}[Proof of \cref{lemma:Trans_Approx_Poly}]
    We first skip the embedded dimension of $y$ and $t$ for the following proof without loss of generality.
    We put it back at the end of the derivation, by
    replacing $L$ with $L+2$.
    
    To implement a sequence-to-sequence model for approximating a function that outputs a scalar, we define a trivial function for converting the scalar target into a sequence represented by matrices.

    To begin with,
    for $x\in\R^{d_x}$ and  $f_1:\R^{d_x}\rightarrow\R$,
    we define a trivial function:
    \begin{align*}
        F_{1} (x)\coloneqq 
        (\underbrace{\alpha_1 f_1(x),\alpha_{2}f_1(x),\ldots,\alpha_{d_{x}-1}f_1(x)}_{\text{(padding $d_{x}-1$ elements)}},f_{1}(x))^{\top}
        \in\R^{d_{x}},
    \end{align*}
    for any set of non-repeated constants $\{\alpha_{i}\}_{i=1}^{d_{x}-1}\in \R \setminus \{1\}$. 
    
    By \emph{trivial}, we mean that $F_1$ transforms $f_1(x) \in \mathbb{R}$ into a vector $F_1(x) \in \mathbb{R}^{d_x}$ where only the last entry is meaningful.

    In order to apply the universal approximation of transformers in \cref{thm:Transformer_as_universal_approximators},
    we show the uniform continuity of $F_{1}$ as follows.

    \begin{itemize}
        \item 
        \textbf{Step A: Uniform Continuity.}

    For different input $x$, $x^{\prime}$,
    we start by writing
    \begin{align*}
    \norm{F_{1}(x)-F_{1}(x^{\prime})}_{p}
    & = \left\{\abs{f(x)-f(x^{\prime})}^{p} + \sum_{i=1}^{d_{x}-1}\abs{\alpha_{i}f(x)-\alpha_{i}f(x^{\prime})}^{p}\right\}^{1/p}\\
    & = \left\{\abs{f(x)-f(x^{\prime})}^{p}\left(1+\sum_{i=1}^{d_{x}-1}\abs{\alpha_{i}}^{p}\right)\right\}^{1/p}\\
    & = \eta\abs{f(x)-f(x^{\prime})},
    \end{align*}
    where $\eta=\left(1+\sum_{i=1}^{d_{x}-1}\abs{\alpha_{i}}^{p}\right)^{1/p}\in\R_{+}.$
    
    Next,
    we utilize the fact that the diffused local polynomials $f_{1}$ is continuous on compact support.
    That is,
    for all $\epsilon>0$, 
    there exists $\delta>0$
    such that for all $x$ and $x^{\prime}$,
    if $\norm{x-x^{\prime}}_{\infty}<\delta$,
    then $\abs{f_{1}(x)-f_{1}(x^{\prime})}<\epsilon$.
    
    From this fact,
    by taking $\epsilon = \epsilon^{\prime}/{\eta}$,
    we have that
    for all $\epsilon^{\prime}>0$, 
    there exists $\delta^{\prime}>0$
    such that for all $x$ and $x^{\prime}$,
    if $\norm{x-x^{\prime}}_{\infty}<\delta^{\prime}$,
    then $\abs{f_{1}(x)-f_{1}(x^{\prime})}<\epsilon^{\prime}=\epsilon\eta$.
    
    This gives $\norm{F_{1}(x)-F_{1}(x^{\prime})}_{p}\leq\epsilon^{\prime}$ and therefore we obtain the uniform continuity for $F_{1}$. 
    
    Also,
    the reshape layer $R(\cdot)$ that converts $x\in\R^{d_{x}}$ into sequential input $R(x)\in\R^{d\times L}$ does not harm this continuity due to its linearity.
    Therefore, 
    the map $R \circ F_{1}(x):\R^{d_{x}}\rightarrow\R^{d\times L}$ is also uniformly continuous.

    \item 

    \textbf{Step B: Universal Approximation.}
    
    We apply \cref{thm:Transformer_as_universal_approximators} that guarantees
    for any $\epsilon_{f_{1}}>0$, 
    there exists one transformer block and one feed-forward layer such that
    \begin{align*}
    \norm{R \circ F_{1}-f^{\textcolor{blue}{h,s,r}}\circ f^{\text{FF}}\circ R}_{p}\leq\epsilon_{f_{1}}.
    \end{align*}

    Adding a reverse reshape layer,
    we have $\calT_{f_{1}}=R^{-1}\circ f^{\textcolor{blue}{h,s,r}}\circ f^{\text{FF}}\circ R$ with
    $\norm{F_{1}-\calT_{f_{1}}}_{p}\leq\epsilon_{f_{1}}$.
    
    Next,
    observe that
    \begin{align}
    \label{eqn:trans_f_1}
    \abs{\calT_{f_{1}}[d_{x}]-f_{1}}\leq\left\{\sum_{i=1}^{d_{x}}\abs{\calT_{f_{1}}[i]-\alpha_{i}f_{1}}^{p}\right\}^{1/p}=\norm{\calT_{f_{1}}-F_{1}}_{p}\leq\epsilon_{f_{1}},
    \end{align}
    with $\alpha_{d_{x}}=1$.
    \eqref{eqn:trans_f_1} completes the proof of the approximation error.

    \item 

        \textbf{Step C: Parameter Bounds.}

    To establish the approximation \eqref{eqn:trans_f_1}, we need the parameter bounds in \cref{lemma:trans_para_bound} to hold.
    This requires transforming the input domain from $[-C_{x}\sqrt{\log{N}},C_{x}\sqrt{\log{N}}]$ to normalized compact support $[0,1]$ for all dimensions (i.e., $x[i]$ for all $i\in[d_x]$.)

Recall that \eqref{eqn:W_1_bound},
we have bound for $W_1$:    
    \begin{align}
        & 
        \norm{W_{1}}_{2, \infty} 
        = 
        \calO 
        \left(\sqrt{d}D \right)
        = 
        \calO 
        \left(\sqrt{d}\epsilon^{-dL} \right), \\
        &
        \norm{W_{1}}_{2} 
        =
        \calO 
        \left(dD \right)
        = 
        \calO 
        \left(d\epsilon^{-dL} \right),
\end{align}
that is, the bounds on each element in $W_{1}$ scales up as the granularity increases.
Because for a fixed precision level,
the granularity is proportional to the length of the interval in each dimension of the input domain,
    we conclude that $\norm{W_{1}}_{2}=\calO\left(d\epsilon^{-dL}\log{N}\right)$ and $\norm{W_{1}}_{2,\infty}=\calO\left(\sqrt{d}\epsilon^{-dL}\log{N}\right)$.
    
    The rest of bounds for each operation follows \cref{lemma:trans_para_bound}.
    Lastly,
    we incorporate the embedded dimensions of $y$ and $t$ by replacing $L$ with $L+2$ (see \cref{fig:condition_DiT}).
    \end{itemize}

    This completes the proof.
    \end{proof}
    
Similarly,
we have the corresponding $\calT_{f_{2}}\in\calT_{R}^{\textcolor{blue}{h,s,r}}$ for the approximation of $f_{2}(x,y,t)$.
    
\begin{lemma}[Approximate Vector-Valued Polynomials with Transformers]
\label{lemma:Trans_Approx_Poly_Gradient}
Assume \cref{assumption:conditional_density_function_assumption_1} and 
consider $f_{2}(x,y,t)\in\R^{d_{x}}$ with every entry $f_2[1],\ldots,f_2[d_x]$ is a local diffused polynomial defined in \cref{lem:f_1_expression}.
For any $\epsilon>0$, there exists a transformer $\calT_{f_{2}}\in\calT_R^{\textcolor{blue}{h,s,r}} $ such that 
\begin{align*}
\norm{f_{2}(x,y,t)-\calT_{f_{2}}}_{\infty}\leq\epsilon,    \end{align*}
for any $x\in[-C_{x}\sqrt{\log{N}},C_{x}\sqrt{\log{N}}]^{d_{x}}, y\in[0,1]^{d_{y}}$ and $t\in[N^{-C_{\sigma}},C_{\alpha}\log{N}]$. The parameter bounds in the transformer network class follows \cref{lemma:Trans_Approx_Poly}.
    \end{lemma}
    
\begin{proof}[Proof of \cref{lemma:Trans_Approx_Poly_Gradient}]
Since each entry of the diffused local polynomials in $f_{2}$ is continuous on compact support, $f_{2} \in \mathbb{R}^{d_{x}}$ is uniformly continuous by the same argument as in the proof of \cref{lemma:Trans_Approx_Poly}.

Similarly,
by \cref{thm:Transformer_as_universal_approximators},
 for any $\epsilon_{f_{2}} > 0$, there exists a transformer block and a feed-forward layer such that $\norm{R \circ f_{2} - f^{\textcolor{blue}{h,s,r}} \circ f^{\text{FF}} \circ R}_{p} \leq \epsilon{f_{2}}$.

    By adding the reversed reshape layer, we obtain $\calT_{f_{2}} \in \calT_{R}^{\textcolor{blue}{h,s,r}}$, satisfying $\norm{f_{2} - \calT_{f_{2}}}{p} \leq \epsilon{f_{2}}$.
    
    Then we have,
    \begin{align*}
    \abs{\calT_{f_{2}}[j]-f_{2}[j]}\leq\left\{\sum_{j=1}^{d_{x}}\abs{\calT_{f_{2}}[j]-f_{2}[j]}^{p}\right\}^{1/p}\leq\epsilon_{f_{2}}
    \end{align*}
    for all $j=1,\ldots,d_{x}$.
    Thus the result with $\ell_{\infty}$ bound also holds.
    
    The network configuration follows the argument as in the proof of \cref{lemma:Trans_Approx_Poly}.
    
    This completes the proof.
    \end{proof}
    
    So far, we have obtained approximation results for $f_{1}$ and $f_{2}$. 
    To complete the full approximation of the score decomposition $\nabla \log p = \frac{\nabla p}{p}$, we still need to approximate several key algebraic operators, 
    including the product (\cref{lemma:approx_prod_with_trans}), inverse (\cref{lemma:inverse_trans})...etc.
    
    We establish their approximations as follows.

    \item 
    \textbf{Step 3.2: Approximate Algebraic Operators with Transformers.}
    
    We give transformer approximation theory for
    the clipping operator,
    the inverse operator,
    the product operator,
    and functions that evolve with time $t$:

    \begin{itemize}
        \item Clipping operation (\cref{lemma:Clipping_function})
        
        \item Product operation
        (\cref{lemma:approx_prod_with_trans})
        
        \item Inverse operation
        (\cref{lemma:inverse_trans})

        \item Mean $\alpha_t=\exp(-t/2)$
        (\cref{lemma:trans_approx_mean})
        
        \item Standard deviation $\sigma_t=\sqrt{1-e^{-t}}$
        (\cref{lemma:trans_approx_variance})
    \end{itemize}
    
    The approximations for these operators are common with the network structure consisting of $\rm ReLU$ activation function and fully connected feed-forward layers, 
    such as the product approximation by
    \citet{schmidt2020nonparametric} and the inverse approximation by
    \citet{telgarsky2017neural}.

    In their works, 
    the general network structure is as follows.
    \begin{definition}
    \label{def:relu_ff_net}
    
    A family of fully-connected neural networks with length $L$, width $W$, sparsity constraint $S$, and norm constraint $B$ is defined as:
    \begin{align*}
    \Phi(L, W, S, B) \coloneqq A^{(L)} {\rm ReLU}(\cdot) + b^{(L)} \circ \cdots \circ A^{(1)} x + b^{(1)},    
    \end{align*}
    where $A^{(i)}$ and $b^{(i)}$ represent the matrix operator and bias in the $i$-th layer.
    Specifically:
    \begin{itemize}
        \item {Length}: $L \in \mathbb{R}$ denotes the number of hidden layers plus one.
        \item {Width}: $W \in \mathbb{N}^{L+1}$ is a vector representing the output dimension of each layer.
        \item {Sparsity Constraint}: $\sum_{i=1}^{L} \|A^{(i)}\|_{0,0} + \|b^{(i)}\|_{0} \leq S$ specifies the maximum number of non-zero terms.
        \item {Norm Constraint}: $\max\limits_{1 \leq i \leq L} \|A^{(i)}\|_{\infty, \infty} \vee \|b^{(i)}\|_{\infty} \leq B$ specifies the upper bound on the parameter norms.
    \end{itemize}
    Here $\vee$ denotes the maximum of two values.
    \end{definition}

    \begin{remark}[Generalization \relu Networks with Transformers]
    Transformers are more general network class that encompasses $\rm{ReLU}$-based networks defined in \cref{def:relu_ff_net}. 
    By setting all self-attention layers in the transformer to identity maps,
    we recover the \relu feed-forward network structure. 
    Therefore, our work on approximating with transformers extends previous works \citet{fu2024unveil,oko2023diffusion} by incorporating the flexibility of self-attention mechanisms.
    \end{remark}
    
    The following lemma provides a network that executes the clipping operation.
    
    \begin{lemma}[Clipping Operation, Lemma F.4 of \cite{oko2023diffusion}]
    \label{lemma:Clipping_function}
    For any $a,b\in\R^{d}$ with $a[i]\leq b[i]$ for all $i\in[d]$, there exist a neural network $\phi_{\text{clip}}(x;a,b)\in\Phi(L,W,S,B)$ such that for all $i\in[d]$,
    it holds 
    \begin{align*}
    \phi_{\text{clip}}(x;a,b)[i] = \min(b[i],\max(x[i],a[i])),
    \end{align*}
    with
    \begin{align}
    L = 2, \quad W = (d,2d,d)^{\top}, \quad 
    S = 7d, \quad B = \max\limits_{1\leq i \leq d}\text{max}(\abs{a[i]},b[i]).
    \end{align}
    Moreover, suppose $a[i] = c$ and $b[i] = C$ for all $i\in[d]$ with $c$ and $C$ being some constant, $\phi_{\text{clip}}(x;a,b)$ is denoted as $\phi_{\text{clip}}(x;c,C).$
    \end{lemma}
    \begin{proof}
    It suffices to show the result for $i$-th coordinate, 
    and implement the parallelization to complete the proof that holds for the entire vector $\phi_{\text{clip}}(x;a,b)$.\footnote{For a more detailed description regarding parallelization please see Appendix F of \cite{oko2023diffusion}.}
    The clipping operation yields the middle among $a[i], b[i]$ and the input $x[i].$
    Following \cite{oko2023diffusion}, we achieve the task by setting:
    \begin{align*}
    \min\left(b[i],\max (x[i],a[i])\right)
    = {\rm ReLU}(x[i] - a[i]) - {\rm ReLU}(x[i]-b[i]) + a[i].
    \end{align*}
    Note that the RHS is realized by the network with one hidden layer:
    \begin{align*}
    (1,-1)
    {\rm ReLU}\left(
    (1,1)
    x[i] + 
    \begin{pmatrix}
    -a[i]\\
    -b[i]
    \end{pmatrix}\right)
    +a[i],
    \end{align*}
    with $7$ non-zero parameters,
    and the scale of parameter is $\text{max}(\abs{a[i]},b[i]).$
    So there exists $\phi_{\text{clip}}(x[i];a[i],b[i])\in\Phi(2,(1,2,1)^{\top},7,\text{max}(\abs{a[i]},b[i]))$ executing the clipping operation.
    Then the proof is complete by the parallelization for all the components $i=1,\ldots,d.$
    
    This completes the proof.
    \end{proof}
    
    Next, we deal with the approximation of products with Transformer.

    \begin{lemma}[Approximation of the Product Operator with Transformer.]
    \label{lemma:approx_prod_with_trans}
    Let $m\geq2$ and $C\geq1$.
    For any $0<\epsilon_{\text{mult}}<1$,
    there exists $\calT_{\text{mult}}(\cdot) \in\calT_R^{\textcolor{blue}{h,s,r}}$ such that for all $x\in[-C,C]^{m},$ $x^{\prime}\in\R^{m}$
    with $\norm{x-x^{\prime}}_{\infty}\leq\epsilon_{\text{error}},$
    it holds 
    \begin{align*}
    \abs{\calT_{\text{mult}}(x^\prime)-\prod_{i=1}^{m}x_{i}}\leq\epsilon_{\text{mult}} + mC^{m-1}\epsilon_{\text{error}}.
    \end{align*}
    The parameter bounds in the transformer network class $\calT^{\textcolor{blue}{h,s,r}}_R$ satisfy
    \begin{align*}
    & 
    \norm{W_{Q}}_{2},
    \norm{W_{K}}_{2},
    \norm{W_{Q}}_{2,\infty},
    \norm{W_{K}}_{2,\infty}
    =
    \calO\left(\epsilon_{\text{mult}}^{-(2m+1)}(\log{m})^{\frac{1}{2}}\right)
    ;\\
    & \norm{W_{O}}_{2},
    \norm{W_{O}}_{2,\infty}
    =
    \calO\left(\epsilon_{\text{mult}}^{m}\right);\quad
    \norm{W_{V}}_{2},
    \norm{W_{V}}_{2,\infty}
    =
    \calO(1)
    ;\\ 
    & \norm{W_{1}}_{2},
    \norm{W_{1}}_{2,\infty}
    =
    \calO\left(C\epsilon_{\text{mult}}^{-m}\right);\quad
    \norm{W_{2}}_{2},
    \norm{W_{2}}_{2,\infty}=\calO\left(\epsilon_{\text{mult}}^{-m}\right).
    \end{align*}
    \end{lemma}
    
    \begin{proof}
    We build our proof on \cite[Lemma F.6]{oko2023diffusion}.
    
    Unlike approximation for input $x\in[-C_{x}\sqrt{\log{N}},C_{x}\sqrt{\log{N}}]^{d_{x}}$ in \cref{lemma:Trans_Approx_Poly},
    the input dimension for the product operator is sufficiently smaller so that we skip the reshape layer by setting $R$ and $R^{-1}$ as identity map.

    Next,
    let $f(x)=\prod_{i=1}^{m}x[i]$,
    and define a trivial function $F(\cdot): \R^m\to \R^{1\times m}$ as 
    \begin{align*}
    F(x)\coloneqq 
    (\underbrace{\alpha_1 f(x),\alpha_{2}f(x),\ldots,\alpha_{m-1}f(x)}_{\text{(padding $m-1$ elements)}},f(x))\in\R^{1\times m}.
    \end{align*}
    
    The idea of padding a scalar into a row vector again stems from the purpose of utilizing sequence-to-sequence model to approximate functions that output a scalar. 
    
    By the same argument as in the proof of \cref{lemma:Trans_Approx_Poly},
    the uniform continuity of $f$ guarantees the uniform continuity of $F$ with respect to the $L_{p}$ norm.
    
    By \cref{thm:Transformer_as_universal_approximators} ,
    for any $\epsilon>0$,
    there exist $\calT_{\text{mult}}\in\calT_{R}^{\textcolor{blue}{h,s,r}}$ with $R, R^{-1}$ being identity map such that
    \begin{align*}
    \norm{\calT_{\text{mult}} - F}_{p}\leq\epsilon.
    \end{align*}
    
    Clearly, $\abs{\calT_{\text{mult}}[m] - F[m]}\leq\norm{\calT_{\text{mult}}-F}_{p}\leq\epsilon$.
    
    To extend the input to $x^{\prime}\in\R^{m}$ with $\norm{x-x^{\prime}}\leq\epsilon_{\text{error}}$,
    we adopt \cref{lemma:Clipping_function} and write
    \begin{align*}
     & ~  \abs{C^{m}\calT_{\text{mult}}(\phi_{\text{clip}}(x^{\prime};-C,C)/C) - \prod_{i=1}^{m}x[i]} \\
    \leq & ~  \abs{C^{m}\calT_{\text{mult}}(\phi_{\text{clip}}(x^{\prime};-C,C)/C) - \prod_{i=1}^{m}\text{min}(C, \text{max}(x^{\prime}[i]],-C))} + \abs{\prod_{i=1}^{m}\text{min}(C, \text{max}(x^{\prime}[i],-C)) - \prod_{i=1}^{m}x[i]} \\
    \leq & ~   C^{m}C^{-m}\epsilon + C^{m-1}\sum_{i=1}^{m}\abs{x[i] - \text{min}(C, \text{max}(x^{\prime}[i]],-C))}\\
    = & ~  \epsilon + mC^{m-1}\epsilon_{\text{error}}.
    \end{align*}
    Further details regarding the product approximation are in Appendix F.2 of \cite{oko2023diffusion}. 
    
    For the parameter bounds,
    following the same argument in the proof of \cref{lemma:Trans_Approx_Poly},
    it suffices to take $\calO(C\epsilon^{-1})$ for $W_{1}$.
    The rest of bounds for each operation follows \cref{lemma:trans_para_bound} with $d=1$ and $L=m$.
    
    This completes the proof.
    \end{proof}

    Next,
    we introduce the next lemma to approximate the inverse operator.
    
    \begin{lemma}[Approximation of the Reciprocal Function with Transformer.]
    \label{lemma:inverse_trans}
    For any $0<\epsilon_{\text{rec}}<1$ there exists a $\calT_{\text{rec}}(\cdot)\in\calT_R^{\textcolor{blue}{h,s,r}}$ such that for all $x\in[\epsilon_{\text{rec}},\epsilon_{\text{rec}}^{-1}]$ and $x^{\prime}\in\R$.
    It holds that
    \begin{align*}
    \abs{\calT_{\text{rec}}(x^{\prime})-\frac{1}{x}}\leq\epsilon_{\text{rec}} + \frac{\abs{x-x^{\prime}}}{\epsilon_{\text{rec}}^{2}}.
    \end{align*}
    The parameter bounds in the Transformer network class satisfy
    \begin{align*}
    & \norm{W_{Q}}_{2},\norm{W_{Q}}_{2,\infty},\norm{W_{K}}_{2},\norm{W_{K}}_{2,\infty}=\calO\left(\epsilon_{\text{rec}}^{-3}\right);\\
    & \norm{W_{O}}_{2},\norm{W_{O}}_{2,\infty}=\calO\left(\epsilon_{\text{rec}}\right); 
    \norm{W_{V}}_{2},\norm{W_{V}}_{2,\infty}=\calO(1);\\ 
    & \norm{W_{1}}_{2},\norm{W_{1}}_{2,\infty}
    =
    \calO\left(\epsilon_{\text{rec}}^{-2}\right);
    \norm{W_{2}}_{2},
    \norm{W_{2}}_{2,\infty}
    =
    \calO\left(\epsilon_{\text{rec}}^{-1}\right).
    \end{align*}
    \end{lemma}
    
    \begin{proof}
    We build our proof on \cite[Lemma F.7]{oko2023diffusion}.
    For any $\epsilon_{\text{rec}}\in(0,1)$,
    since $1/x$ is continuous on $x\in[\epsilon_{\text{rec}},\epsilon_{\text{rec}}^{-1}]$,
    by \cref{thm:Transformer_as_universal_approximators}, there exist a transformer $\calT_{\text{rec}}\in\calT_{R}^{\textcolor{blue}{h,s,r}}$ such that
    \begin{align*}
    \abs{\calT_{\text{rec}}-\frac{1}{x}}\leq\epsilon_{\text{rec}}.
    \end{align*}
    
    Extending to network with input $x^{\prime}\in\R,$
    the sensitivity analysis follows:
    \begin{align*}
    \abs{\calT_{\text{rec}}(x^{\prime})-\frac{1}{x}}\leq\abs{\calT_{\text{rec}}(x^{\prime})-\frac{1}{\text{max}({x^{\prime},\epsilon})}}+\abs{\frac{1}{x}-\frac{1}{\text{max}({x^{\prime},\epsilon})}}.
    \end{align*}
    This yields the result.
    
    For the parameter bounds,
    by the same discussion in the proof of \cref{lemma:approx_prod_with_trans},
    we scale $W_{1}$ up by $\epsilon_{\text{rec}}$ such that the quantization in \eqref{eqn:quantize_term_f1} works on normalized $[0,1]$.
    The rest of the bounds follow \cref{lemma:trans_para_bound}.
    
    This completes the proof.
    \end{proof}
    
    Next, we state approximation results using Transformer for $\alpha_{t}$ and $\sigma_{t}.$
    From \eqref{eqn:reverse_process} we have $\alpha_{t} = \exp({-{t}/{2}})$ and $\sigma_{t}=\sqrt{1-\alpha_{t}^{2}}.$
    \begin{lemma}[Approximation of $\alpha_{t}=\exp({-{t}/{2}})$ with Transformer.]
    \label{lemma:trans_approx_mean}
    For any $\epsilon_{\alpha}\in(0,1),$
    there exists Transformer $\calT_{\alpha}(t)\in\calT_R^{\textcolor{blue}{h,s,r}}$ such that for all $t\geq0,$
    we have
    \begin{align*}
    \abs{\calT_{\alpha}(t)-\alpha_{t}}\leq\epsilon_{\alpha}.
    \end{align*}
    The parameter bounds in the Transformer network class satisfy
    \begin{align*}
    & \norm{W_{Q}}_{2},\norm{W_{Q}}_{2,\infty},\norm{W_{K}}_{2},\norm{W_{K}}_{2,\infty}=\calO\left(\epsilon_{\alpha}^{-3}\right);\quad\\
    & \norm{W_O}_{2},\norm{W_O}_{2,\infty}=\calO\left(\epsilon_{\alpha}^{-1}\right); 
    \norm{W_{V}}_{2},\norm{W_{V}}_{2,\infty}=\calO(1);\quad\\ 
    & \norm{W_{1}}_{2},
    \norm{W_{1}}_{2,\infty}
    =\calO\left((\log{\epsilon_{\alpha}^{-1}})\epsilon_{\alpha}^{-1}\right);
    \norm{W_{2}}_{2},
    \norm{W_{2}}_{2,\infty}=\calO\left(\epsilon_{\alpha}^{-1}\right).
    \end{align*}
    \end{lemma}
    \begin{proof}
    We build our proof on \cite[Lemma F.8]{fu2024unveil}.
    The proof consists of four steps.

    \begin{itemize}
    \item 
    
    \textbf{Step A: Approximate $\exp(\cdot)$ with Taylor polynomial for $t\in[0,T]$.}
        
    By Taylor theorem,
    there exist some $\theta\in[0,T]$ such that
    \begin{align*}
    \exp(-\frac{t}{2}) = \sum_{i=0}^{s-1}\frac{(-1)^{i}}{i!}\left(\frac{t}{2}\right)^{i} + \frac{(-1)^{s}}{s!}\left(\frac{\theta}{2}\right)^{s}\exp(-\frac{\theta
    }{2}).
    \end{align*}

    We further bound the error from the remainder by
    \begin{align}
    \label{eqn:mean_approx_first}
    \abs{\exp(-\frac{t}{2}) - \sum_{i=0}^{s-1}\frac{(-1)^{i}}{i!}\left(\frac{t}{2}\right)^{i}}\leq\frac{T^{s}}{2^{s}s!},
    \end{align}
    with $T$ and $s$ to be chosen later.

    \item 

    \textbf{Step B: Approximate Taylor polynomial with transformer for $t\in[0,T]$.}
        
    We take $t$ as a sequence with length $1$ and one-dimensional token.
    
    For $t\in[0,T]$,
    Taylor polynomial is a continuous function with compact support.
    
    Therefore,
    by \cref{thm:Transformer_as_universal_approximators}.
    for any $\epsilon$
    there exist a transformer $\calT^{\prime}_{\alpha}\in\calT_{R}^{\textcolor{blue}{h,s,r}}$ such that
    \begin{align}
    \label{eqn:mean_approx_second}
    \abs{\calT_{\alpha}^{\prime} - \sum_{i=1}^{s-1}\frac{(-1)^{i}}{i!}\left(\frac{t}{2}\right)^{i}}\leq\epsilon.
    \end{align}

    \item 

     \textbf{Step C: Extend the two approximation results from \textbf{Step 1.} and \textbf{Step 2.} to $t>T$.}
    
    We define $\calT_{\alpha}$ as
    
    (i) $\calT_{\alpha}(t)=\calT_{\alpha}^{\prime}(t)$ for $t\in[0,T]$.
    
    (ii)$\calT_{\alpha}(t)=\calT_{\alpha}^{\prime}(T)$ for $t\geq T$.
    
    Next,
    we bound the error for $t>T$ by
    \begin{align}
    \label{eqn:mean_approx_third}
    \abs{\exp(-\frac{t}{2}) - \calT_{\alpha}(t)}\leq
    \abs{\exp(-\frac{T}{2})-\exp(-\frac{t}{2})} + \abs{\calT_{\alpha}(t)-\exp(-\frac{T}{2})}.
    \end{align}

    \item 

    \textbf{Step D: Select $T$, $s$ and transformer approximation error such that the result holds for all $t\geq0$.}
        
    For any $\epsilon_{\alpha}>0$,
    we ensure $\abs{\calT_{\alpha}-\exp(-t/2)}\leq\epsilon_{\alpha}$ holds for all $t\geq0$. 
        
    To achieve this, 
    apply Stirling formula to \eqref{eqn:mean_approx_first} and set $s =eT$, $T=2\log{3\epsilon_{\alpha}^{-1}}$,
    we have
    \begin{align*}
    \abs{e^{-\frac{t}{2}} - \sum_{i=0}^{s-1}\frac{(-1)^{i}}{i!}\left(\frac{t}{2}\right)^{i}}\leq\left(\frac{1}{2}\right)^{eT}=\left(\frac{\epsilon_{\alpha}}{3}\right)^{\frac{2e}{\log_{2}{e}}}\leq\frac{\epsilon_{\alpha}}{3}.
    \end{align*}
    
    Next we set the transformer error $\epsilon=\epsilon_{\alpha}/3$.
    Combining \eqref{eqn:mean_approx_first} and \eqref{eqn:mean_approx_second},
    for $t\in[0,T]$ we obtain
    \begin{align*}
    \abs{\calT_{t}-\exp(-\frac{t}{2})}\leq\frac{2}{3}\epsilon_{\alpha}.
    \end{align*}
    
    Furthermore,
    since $\exp(-T/2)=\epsilon_{\alpha}/3$,
    \eqref{eqn:mean_approx_third} becomes
    \begin{align*}
    \abs{\exp(-\frac{t}{2}) - \calT_{\alpha}(t)}\leq\frac{\epsilon_{\alpha}}{3} + \frac{2\epsilon_{\alpha}}{3}=\epsilon_{\alpha}.
    \end{align*}
    
    For the parameter bounds,
    by the same argument as in the proof of \cref{lemma:Trans_Approx_Poly},
    we normalize the domain from $[0,T]$ to $[0,1]$ for the quantization,
    and then the rest of the step follows \cref{thm:Transformer_as_universal_approximators}.
    
    This results in
    parameter bound $\calO(\log{\epsilon_{\alpha}^{-1}}\epsilon_{\alpha}^{-\frac{1}{d}})$ for $\norm{W_{1}}_{2}$ and $\norm{W_{1}}_{2,\infty}$,
    and the rest of the bounds follow the result in \cref{lemma:trans_para_bound} with $d=1$ and $L=1$.
    \end{itemize}
This completes the proof.
    \end{proof}

    \begin{lemma}[Approximation of $\sigma_{t}=\sqrt{1-e^{-t}}$ with transformer]
    \label{lemma:trans_approx_variance}
    For any $\sigma_{\sigma}\in(0,1),$
    there exists a transformer $\calT_{\sigma}(t)\in\calT_R^{\textcolor{blue}{h,s,r}}$ such that for any $t\in[t_{0},T]$ with $t_{0}<1$
    we have
    \begin{align*}
    \abs{\calT_{\sigma}(t)-\sigma_{t}}\leq\epsilon_{\sigma}.
    \end{align*}
    The parameter bounds in the transformer network class satisfy
    \begin{align*}
    & \norm{W_{Q}}_{2},\norm{W_{Q}}_{2,\infty},\norm{W_{K}}_{2},\norm{W_{K}}_{2,\infty}=\calO\left(\epsilon_{\sigma}^{-3}\right);\\
    & \norm{W_{O}}_{2},\norm{W_{O}}_{2,\infty}=\calO\left(\epsilon_{\sigma}\right); 
    \norm{W_{V}}_{2},\norm{W_{V}}_{2,\infty}=\calO(1);\\ 
    & \norm{W_{1}}_{2}=\calO\left(T\epsilon_{\sigma}^{-1}\right);\quad
    \norm{W_{1}}_{2,\infty}=\calO\left(T\epsilon_{\sigma}^{-1}\right);\\ &\norm{W_{2}}_{2}=\calO\left(\epsilon_{\sigma}^{-1}\right);\quad
    \norm{W_{2}}_{2,\infty}=\calO\left(\epsilon_{\sigma}^{-1}\right).
    \end{align*}
    \end{lemma}
    \begin{proof}
    We follow the proof structure of \cite[Lemma F.10]{fu2024unveil}.
    
    Since $f(t)=\sqrt{1-e^{-t}}$ with $t\in[t_{0},T]$ is a continuous on compact domain.
    The first part of the proof is complete by applying \cref{thm:Transformer_as_universal_approximators}.
    
    For the parameter bounds,
    we take $\calO(T\epsilon_{\sigma}^{-1})$ for $\norm{W_{1}}_{2}$ and $\norm{W_{1}}_{2\infty}$ in the first feed-forward layer.
    This follows from the argument in the proof of \cref{lemma:Trans_Approx_Poly}.
    
    The rest of the bounds follow \cref{lemma:trans_para_bound} with $d=1$ and $L=1$
    
    This completes the proof.
    \end{proof}

    We have finished the approximation of every key component for the proof of \cref{thm:Main_1}.
    We now proceed to the detailed assembly and integration of these components to finalize the proof.

    \item
    \textbf{Step 3.3: Unified Transformer-Based Score Function Approximation.}

    First,
    we establish a theoretical upper bound for transformer model output
    by analyzing the upper bound of the score function in $\ell_{\infty}$ distance under \cref{assumption:conditional_density_function_assumption_1} as follows.
    \begin{itemize}
    
    \item 

    \textbf{Bound on $p_{t}(x|y)$:}
    
    Recall that the conditional distribution at time $t$ has the form:
    \begin{align*}
    p_{t}(x|y) = \frac{1}{\sigma_{t}^{d}(2\pi)^{\frac{d}{2}}}\int p(x_{0}|y)\exp(-\frac{\norm{x-\alpha_{t}x_{0}}^{2}}{2\sigma_{t}^{2}})\dd x_{0}.
    \end{align*}
dk    Applying the light tail property in \cref{assumption:conditional_density_function_assumption_1},
    the upper bound follows:
    \begin{align}
    \label{eqn:upper_bound_p_t}
    p_{t}(x|y)\leq\frac{C_{1}}{\sigma_{t}^{d}(2\pi)^{\frac{d}{2}}}\int\exp(-\frac{C_{2}\norm{x_{0}}^{2}}{2})\exp(-\frac{\norm{x-\alpha_{t}x_{0}}^{2}}{2\sigma_{t}^{2}})\dd x_{0}.
    \end{align}
    
    On the other hand,
    the lower bound follows:
    \begin{align}
    \label{eqn:lower_bound_p_t}
    p_{t}(x|y) \geq
    \frac{1}{\sigma_{t}^{d}(2\pi)^{\frac{d}{2}}}\int_{\norm{x_{0}}\leq1}p(x_{0}|y)\exp(-\frac{\norm{x-\alpha_{t}x_{0}}^{2}}{2\sigma_{t}^{2}})\dd x_{0}.
    \end{align}

\item 
\textbf{Bound on $\nabla p_{t}(x|y)$:}
The first element of the gradient has the form:
\begin{align}
\label{eqn:bound_grad_p_t}
\abs{\qty(\nabla p_{t})[1]}=\frac{1}{\sigma_{t}^{2}(2\pi)^{\frac{d}{2}}}\cdot \abs{\int\left(\frac{x[1]-\alpha_{t}x_{0}[1]}{\sigma_{t}^{2}}\right)p(x_{0}|y)\exp(-\frac{\norm{x-\alpha_{t}x_{0}}^{2}}{2\sigma_{t}^{2}})\dd x_{0}}.
\end{align}

The $\ell_{\infty}$ bound on $\nabla p_{t}$ follows by applying light tail property to each coordinate as in \eqref{eqn:upper_bound_p_t}.
\end{itemize}

Combining \eqref{eqn:upper_bound_p_t}, \eqref{eqn:lower_bound_p_t} and \eqref{eqn:bound_grad_p_t},
we provide the $\ell_{\infty}$ bounds on the score.

\begin{lemma}[Bounds on Score, Lemma A.10 of \cite{fu2024unveil}]
\label{bounds_on_score} 
Assume \cref{assumption:conditional_density_function_assumption_1}. There exists a constant $K$ such that
\begin{align*}
\norm{\nabla \log{p_{t}(x|y)}}_{\infty}\leq\frac{K}{\sigma_{t}^{2}}(\norm{x}+1).
\end{align*}
\end{lemma}

Further details regarding the derivation are in Appendix A.7 of \cite{fu2024unveil}.

Next lemma incorporates previous approximation results into an unified transformer architecture.
\begin{lemma}
[Approximation Score Function with Transformer on Supported Domain]
\label{lemma:Score_Approx_Trans}
Assume \cref{assumption:conditional_density_function_assumption_1}.
Consider $t\in[N^{-C_{\sigma}},C_{\alpha}\log{N}]$,
for constant $C_{\sigma}, C_{\alpha}$,
and $(x,y)\in-[C_{x}\sqrt{\log{N}},C_{x}\sqrt{\log{N}}]^{d_{x}}\times[0,1]^{d_{y}}$,
where $N\in\mathbb{N}$ and $C_{x}$ depends on $d,\beta,B,C_{1},C_{2}$.
There exist a transformer network $\calT_{\text{score}}(x,y,t)\in\calT_R^{\textcolor{blue}{h,s,r}} $ such that
\begin{align*}
p_{t}(x|y)\norm{\nabla \log{p_{t}(x|y)}-\calT_{\text{score}}(x,y,t)}_{\infty}\lesssim\frac{B}{\sigma_{t}^{2}}N^{-\beta}(\log{N})^{\frac{d_{x}+\textcolor{blue}{k_1}+1}{2}}.
\end{align*}
The parameter bounds in the Transformer network class satisfy
\begin{align*}
& \norm{W_{Q}}_{2}, \norm{W_{K}}_{2},
\norm{W_{Q}}_{2,\infty}, \norm{W_{K}}_{2,\infty}
=
\calO\left(N^{(7\beta+6C_{\sigma})}\right); \\
&
\norm{W_{O}}_{2},\norm{W_{O}}_{2,\infty}=\calO\left(N^{-(3\beta+6C_{\sigma})}(\log{N})^{3(d_x+\beta)}\right);\\
&
\norm{W_{V}}_{2}=\calO(\sqrt{d}); \norm{W_{V}}_{2,\infty}=\calO (d);
\norm{E^{\top}}_{2,\infty}=\calO\left(d^{\frac{1}{2}}L^{\frac{3}{2}}\right);\\ 
& \norm{W_{1}}_{2}, \norm{W_{1}}_{2,\infty}
=
\calO\left(N^{(2\beta+4C_{\sigma})}\right);
C_\calT=\calO\left(\sqrt{\log{N}}/\sigma_{t}^{2}\right);
\\
& \norm{W_{2}}_{2}, \norm{W_{2}}_{2,\infty}
=
\calO\left(N^{(3\beta+2C_{\sigma})}\right). 
\end{align*}
\end{lemma}

\begin{proof}[Proof of \cref{lemma:Score_Approx_Trans}]
Our poof follows the structure of \citet[Proposition A.3]{fu2024unveil}.

Recall that from \cref{bounds_on_score},
we have $\norm{\nabla\log{p_{t}(x|y)}}_{\infty}\leq K(C_{x}\sqrt{d_x\log{N}}+1)/\sigma_{t}^{2}$,
along with the diffused local polynomial $f_1$ and $f_2$,
we define first-step score approximator $f_{3}(x,y,t)$ as
\begin{align*}
f_{3}(x,y,t) = \min\left(\frac{f_{2}}{\sigma_{t}f_{1,\text{clip}}},\frac{K}{\sigma_{t}^{2}}(C_{x}\sqrt{d_x\log{N}}+1)\right),
\end{align*}
where we set $f_{1,\text{clip}} = \{f_{1},\epsilon_{\text{low}}\}$ to prevent score from blowing up and
we set $\epsilon_{\text{low}}$ later.

We proceed with the following three steps:

\begin{itemize}
    \item 
    \textbf{Step A. Approximate Score Function with $f_{3}$.}
    
Without loss of generality,
we first derive error bound on the difference between the first component in $f_{3}$ and the score.
\begin{align*}
\abs{(\nabla\log{p_{t}})[1] - f_{3}[1]}
& \leq\abs{(\nabla\log{p_{t}})[1]-\frac{f_{2}[1]}{\sigma_{t}f_{1,\text{clip}}}}\\
& \leq\abs{\frac{(\nabla p_{t})[1]}{p_{t}} - \frac{(\nabla p_{t})[1]]}{f_{1,\text{clip}}}} + \abs{\frac{(\nabla p_{t})[1]}{f_{1,\text{clip}}} - \frac{f_{2}[1]}{\sigma_{t}f_{1,\text{clip}}}}.
\end{align*}
From \cref{bounds_on_score},
the bound on the score implies $(\nabla p_{t})[1]\leq K(\sqrt{d_x\log{N}}+1)p_{t}/{\sigma_{t}^{2}}$.

Therefore,
\begin{align*}
& ~ \abs{(\nabla\log{p_{t}})[1] - f_{3}[1]} \\
\leq & ~
\frac{K}{\sigma_{t}^{2}}(\sqrt{d\log{N}}+1)p_{t}\abs{\frac{1}{p_{t}} - \frac{1}{f_{1,\text{clip}}}} + \frac{1}{f_{1,\text{clip}}}\abs{\frac{(\nabla\sigma_{t}p_{t})[1]-f_{2}[1]}{\sigma_{t}}}\\
\lesssim & ~
\frac{1}{f_{1,\text{clip}}}\left(\frac{1}{\sigma_{t}^{2}}\sqrt{\log{N}}\abs{p_{t}-f_{1,\text{clip}}} + \frac{(\nabla\sigma_{t}p_{t})[1]-f_{2}[1]}{\sigma_{t}}\right).
\annot{By dropping Constant Terms}
\end{align*}

From \cref{lemma:Trans_Approx_Poly}, 
we have
\begin{align*}
\abs{f_{1}-p_{t}}\leq BN^{-\beta}\log^{\frac{d_{x}+\textcolor{blue}{k_1}}{2}}{N}.
\end{align*}
We set $\epsilon_{\text{low}}=C_{3}N^{-\beta}\log^{(d_{x}+\textcolor{blue}{k_1})/2}{N}\leq p_{t}$ such that $f_{1}\geq p_{t}/2$ by the choice of constant $C_{3}$.

We further write
\begin{align*}
& ~ \abs{(\nabla\log{p_{t}})[1] - f_{3}[1]} \\
\lesssim & ~
\frac{1}{p_{t}}\left(\frac{1}{\sigma_{t}^{2}}\sqrt{\log{N}}\abs{p_{t}-f_{1,\text{clip}}} + \frac{(\nabla\sigma_{t}p_{t})[1]-f_{2}[1]}{\sigma_{t}}\right)
\annot{By the choice of $\epsilon_{\text{low}}$}\\
\lesssim & ~
\frac{B}{\sigma_{t}^{2}p_{t}}N^{-\beta}(\log{N})^{\frac{d_{x}+\textcolor{blue}{k_1}+1}{2}}.
\annot{By \cref{lemma:diffused_local_polynomials} and \cref{lemma:diffused_local_polynomials_grad}}
\end{align*}
By the symmetry of each coordinate, the infinity bound for the score holds as well:
\begin{align}
\label{eqn:ground_truth_f3}
\norm{\nabla\log{p_{t}} - f_{3}}_{\infty}\lesssim\frac{B}{\sigma_{t}^{2}p_{t}}N^{-\beta}(\log{N})^{\frac{d_{x}+\textcolor{blue}{k_1}+1}{2}}.
\end{align}

\item 

\textbf{Step B: Approximate $f_{3}$ with Transformer $\calT_{\text{score}}$.}

In this step, we utilize transformers to approximate $f_{3}$ to an accuracy of order $N^{-\beta}$ such that it aligns with the error order in \eqref{eqn:ground_truth_f3}.

Since $f_{3}$ is the minimum between two components,
we approximate each of them as follows.

\begin{itemize}
    \item \textbf{Step B.1: Approximate $\frac{1}{\sigma_{t}}\cdot \frac{f_{2}}{{f_{1,\text{clip}}}}$.}
    
First,
we utilize $\calT_{f_{1}}$, $\calT_{f_{2}}$ and $\calT_{\sigma,1}$ in
\cref{lemma:Trans_Approx_Poly}, \cref{lemma:Trans_Approx_Poly_Gradient}, and \cref{lemma:trans_approx_variance} for $f_{1}$, $f_{2}$, and $\sigma_{t}$ respectively.
This gives error $\epsilon_{f_{1}}$, $\epsilon_{f_{2}}$ and $\epsilon_{\sigma,1}$,
and we address the clipping of $f_{1}$ in later paragraph.

Next,
We utilize $\calT_{\text{rec},1}$ and $\calT_{\text{rec},2}$ in \cref{lemma:inverse_trans} for the approximation of the inverse of $f_{1}$ and $\sigma_{t}$.

This gives error 
\begin{align*}
\abs{\calT_{\text{rec},1}-\frac{1}{f_{1}}}\leq
\epsilon_{\text{rec},1}+\frac{\abs{\calT_{f_{1}}-f_{1}}}{\epsilon_{\text{rec},1}^{2}}
\leq
\epsilon_{\text{rec},1}+\frac{\epsilon_{f_{1}}}{\epsilon_{\text{rec},1}^{2}},
\end{align*}
and 
\begin{align*}
\abs{\calT_{\text{rec},2}-\frac{1}{\sigma_{t}}}
\leq
\epsilon_{\text{rec},2}+\frac{\abs{\calT_{\sigma,1}-\sigma_{t}}}{\epsilon_{\text{rec},2}^{2}}
\leq
\epsilon_{\text{rec},2}+\frac{\epsilon_{\sigma,1}}{\epsilon_{\text{rec},2}^{2}}.
\end{align*}

Note that all the approximation error propagates to the next approximation.
    
Next,
we utilize $\calT_{\text{mult},1}$ in \cref{lemma:approx_prod_with_trans} for the approximation of
the product of $f_{1}^{-1}$, $f_{2}$ and $\sigma_{t}^{-1}$.

This gives error of
\begin{align*}
\abs{\calT_{\text{mult},1}-\frac{f_{2}}{\sigma_{t}f_{1}}}
\leq & ~ 
\epsilon_{\text{mult},1} + 3K_{2}^{2}\underbrace{\max\left(\epsilon_{\text{rec},1} + \frac{\epsilon_{f_{1}}}{\epsilon_{\text{rec},1}^{2}},\epsilon_{f_{2}},\epsilon_{\text{rec},2} + \frac{\epsilon_{\sigma,1}}{\epsilon_{\text{rec},2}^{2}}\right)}_{\coloneqq\epsilon_{1}} \\
= & ~ 
\epsilon_{\text{mult},1} + 3K_{2}^{2}\epsilon_{1},
\end{align*}     
and $K_{2}$ is a positive constant.
From \cref{lemma:approx_prod_with_trans} we require that $[-K_{2},K_{2}]$ covers the domain for all of $f_{1}^{-1}$, $f_{2}$ and $f_{\sigma}^{-1}$.

To be more specific,
we reiterate three facts that determines the choice of $K_{2}$.
\begin{itemize}
    \item Recall that in the \textbf{Step A.},
we set $f_{1,\text{clip}}=\{f_{1},\epsilon_{\text{low}}\}$.

   \item \cref{bounds_on_score} states $K(C_{x}\sqrt{d_x\log{N}}+1)/\sigma_{t}^{2}$ is 
the $\ell_{\infty}$ bound on the score.

   \item The maximum value of $\sigma_{t}^{-1}$ happens at $t=t_{0}$.
\end{itemize}

As a result,
we set $K_{2}$ as
\begin{align*}
K_{2}=\max\left(\frac{1}{\epsilon_{\text{low}}},\frac{K}{\sigma_{t_{0}}}(C_{x}\sqrt{d_x\log{N}}+1),\frac{1}{\sigma_{t_{0}}}\right).
\end{align*}

By the earlier choice of  $\epsilon_{\text{low}}$,
we have $\epsilon_{\text{low}}^{-1}=\calO(N^{\beta}\log{N}^{-(d_{x}+\textcolor{blue}{k_1})/2})$,
and next we expand $\sigma_{t_{0}}$.
\begin{align*}
\sigma_{t_{0}}
=\sqrt{1-\exp(N^{-C_{\sigma}})}
=1-\left(1-\calO(N^{-C_{\sigma}})\right).
\end{align*}
Therefore we have $\sigma_{t_{0}}^{-1}=\calO(N^{C_{\sigma}})$.
Putting all together, we have 
\begin{align}
\label{eqn:main_1_K2}
K_{2}=\calO\left(N^{\beta+C_{\sigma}}\log^{-\frac{d_{x}+\beta}{2}}{N}\right),
\end{align}
where we use $\textcolor{blue}{k_1}\leq\beta$.

\item \textbf{Step B.2 : Approximate $K(C_{x}\sqrt{d_x\log{N}}+1)/{\sigma_{t}^{2}}$.}

We invoke $\calT_{\sigma,2}$ in \cref{lemma:trans_approx_variance} for the approximation of $\sigma_{t}$, and this gives error $\epsilon_{\sigma,2}$.

Next,
we utilize $\calT_{\text{rec},3}$ in \cref{lemma:approx_prod_with_trans} for the approximation of the inverse of $\sigma_{t}$.

This gives error
\begin{align*}
\abs{\calT_{\text{rec},3}-\frac{1}{\sigma_{t}}}
\leq
\epsilon_{\text{rec},3}+\frac{\abs{\calT_{\sigma,3}-\sigma_{t}}}{\epsilon_{\text{rec},3}^{2}}
\leq
\epsilon_{\text{rec},3}+\frac{\epsilon_{\sigma,2}}{\epsilon_{\text{rec},3}^{2}}.
\end{align*}

Next,
we utilize $\calT_{\text{mult},2}$ for the approximation of the square of $\sigma_{t}^{-1}$.
     
This gives error of
\begin{align*}
\abs{\calT_{\text{mult},2}-\left(\frac{1}{\sigma_{t}}\right)^{2}}
\leq
\epsilon_{\text{mult},2} + 2K_{1}\left(\epsilon_{\text{rec},3}+\frac{\epsilon_{\sigma,2}}{\epsilon_{\text{rec},3}^{2}}\right),
\end{align*}
and $K_{1}$ is constant to be chosen such that $\sigma_{t}\in[-K_{1},K_{1}]$.

With the same argument for $K_2$,
it suffices to take $\calO(\sigma_{t}^{-1})$:
\begin{align}
\label{eqn:main_1_K1}
K_{1}=\calO\left(N^{C_{\sigma}}\right).
\end{align}

\item \textbf{Step B.3: Error Bound on Every Approximation Combined.}

Combining \textbf{Step B.1} and \textbf{Step B.2},
the total error is bounded by
\begin{align*}
\epsilon_{\text{score}}
\leq
\max\left(\epsilon_{\text{mult},2}+ 2K_{1}\left(\epsilon_{\text{rec},3}+ 
\frac{\epsilon_{\sigma,2}}{\epsilon_{\text{rec},3}^{2}}\right), \epsilon_{\text{mult},1}+3K_{2}^{2}\epsilon_{1}\right).
\end{align*}
The goal is to guarantee the final error $\epsilon_{\text{score}}\leq N^{-\beta}$ such that it matches the order of the approximation error in \textbf{Step A.}
We list all the error choice to achieve the goal.\footnote{Further details regarding the choice of each one of $\epsilon$ are in Appendix F.4 of \cite{fu2024unveil}.}

\begin{itemize}
\item \textbf{For the Error of the First Two Inverse Operators:}
\begin{align*}
\epsilon_{\text{rec},1}, \epsilon_{\text{rec},2}
=\calO\left(N^{-(3\beta+2C_{\sigma})}(\log{N})^{(d_{x}+\beta)}\right).
\end{align*}

\item \textbf{For the Error of the Last Inverse Operator:}
\begin{align*}
\epsilon_{\text{rec},3}
=\calO\left(N^{-(\beta+2C_{\sigma})}\right).
\end{align*}

\item \textbf{For the Error of $f_{1}$:}
\begin{align*}
\epsilon_{f_{1}}=\calO\left(N^{-(9\beta+6C_{\sigma})}(\log{N})^{3(d_{x}+\beta)}\right).
\end{align*}

\item  \textbf{For the Error of $f_{2}$:}
\begin{align*}
\epsilon_{f_{2}}=\calO\left(N^{-(3\beta+2C_{\sigma})}(\log{N})^{(d_{x}+\beta)}\right).
\end{align*}

\item \textbf{For the Error of the First Variance:}
\begin{align*}
\epsilon_{\sigma,1}=\calO\left(N^{-(9\beta+6C_{\sigma})}(\log{N})^{3(d_{x}+\beta)}\right).
\end{align*}

\item \textbf{For the Error of the Second Variance:}
\begin{align*}
\epsilon_{\sigma,2}=\calO\left(N^{-(7\beta+5C_{\sigma})}(\log{N})^{2(d_{x}+\beta)}\right).
\end{align*}

\item \textbf{For the Error of the Two Product Operators:}
\begin{align*}
\epsilon_{\text{mult},1}, \epsilon_{\text{mult},2} = \calO(N^{-\beta}).
\end{align*}
\end{itemize}

The above error choice renders $\epsilon_{\text{score}}\leq N^{-\beta}$.

Therefore we conclude that there exist a transformer $\calT_{\text{score}}\in\calT_R^{\textcolor{blue}{h,s,r}}$ such that
\begin{align}\label{eqn:fscore_f3}
\norm{\calT_{\text{score}}(x,y,t) - f_{3}(x,y,t)}_{\infty}\leq N^{-\beta}.
\end{align}

Combining \eqref{eqn:ground_truth_f3} and \eqref{eqn:fscore_f3}
we obtain
\begin{align*}
\norm{\nabla \log{p_{t}}-\calT_{\text{score}}(x,y,t)}_{\infty}\lesssim\frac{1}{p_{t}}\frac{B}{\sigma_{t}^{2}}N^{-\beta}(\log{N})^{\frac{d_{x}+\textcolor{blue}{k_1}+1}{2}}.
\end{align*}
\end{itemize}

We have completed the first part of the proof.
We next give the norm bounds for the transformer parameters.
Specifically,
we select the parameter bounds that are consistent across all operations.
including \cref{lemma:Trans_Approx_Poly}, \cref{lemma:Trans_Approx_Poly_Gradient}, \cref{lemma:approx_prod_with_trans}, \cref{lemma:inverse_trans} and \cref{lemma:trans_approx_variance}.

\item 
\textbf{Step C: Transformer Parameter Bound.}

Our result highlights the influence of $N$ under varying $d_x$.
Therefore,
for the transformer parameter bounds,
we keep terms with $d_x, d, L$ appearing in the exponent of $N$ and $\log{N}$.

Note that the following parameter selection is based on high-dimensional case where $\log{N}$ term dominates $N$
term. 
\begin{itemize}
    \item \textbf{Parameter Bound on $W_{Q}$ and $W_{K}$.}

Given error $\epsilon$,
the bound on each operation follows:
\begin{itemize}
\item 
    \textbf{For $\epsilon_{f_1}$:}
    By \cref{lemma:Trans_Approx_Poly},
    we have
    \begin{align*}
    & ~ \norm{W_{Q}}_{2},\norm{W_{K}}_{2}, \norm{W_{Q}}_{2,\infty},\norm{W_{K}}_{2,\infty}
    =
    \calO\left(N^{(9\beta+6C_{\sigma})\cdot\frac{2dL+4d+1}{d}}\cdot(\log{N})^{-3(d_x+\beta)\cdot\frac{2dL+4d+1}{d}}\right).
    \end{align*}

\item 
    \textbf{For $\epsilon_{f_2}$:}
    By \cref{lemma:Trans_Approx_Poly_Gradient},
    we have
    \begin{align*}
    & ~ \norm{W_{Q}}_{2},\norm{W_{K}}_{2}, \norm{W_{Q}}_{2,\infty},\norm{W_{K}}_{2,\infty}
    =
    \calO\left(N^{(3\beta+2C_{\sigma})\cdot\frac{2dL+4d+1}{d}}\cdot(\log{N})^{-(d_x+\beta)\cdot\frac{2dL+4d+1}{d}}\right).
    \end{align*}

\item 
    \textbf{For $\epsilon_{\text{mult},1}$:}
    By \cref{lemma:approx_prod_with_trans}
    with $m=3$,
    we have
\begin{align*}
    & ~ \norm{W_{Q}}_{2},\norm{W_{K}}_{2},
    \norm{W_{Q}}_{2,\infty},\norm{W_{K}}_{2,\infty}
    =\calO\left(N^{7\beta}\right).
\end{align*}

\item 
    \textbf{For $\epsilon_{\text{mult},2}$:}
    By \cref{lemma:approx_prod_with_trans}
    with $m=2$,
    we have
\begin{align*}
    & ~ \norm{W_{Q}}_{2},\norm{W_{K}}_{2},
    \norm{W_{Q}}_{2,\infty},\norm{W_{K}}_{2,\infty}
    =\calO\left(N^{5\beta}\right).
\end{align*}

\item 
    \textbf{For $\epsilon_{\text{rec},1}$, $\epsilon_{\text{rec},2}$:}
    By \cref{lemma:inverse_trans},
    we have
    \begin{align*}
    & ~ \norm{W_{Q}}_{2},\norm{W_{K}}_{2}
    ,\norm{W_{Q}}_{2,\infty},
    \norm{W_{K}}_{2,\infty}
    =
    \calO\left(N^{(9\beta+6C_{\sigma})}(\log{N})^{-3(d_x+\beta)}\right).
    \end{align*}

\item 
    \textbf{For $\epsilon_{\text{rec},3}$:}
    By \cref{lemma:inverse_trans},
    we have
    \begin{align*}
    & ~ \norm{W_{Q}}_{2},\norm{W_{K}}_{2},
    \norm{W_{Q}}_{2,\infty},\norm{W_{K}}_{2,\infty}
    =\calO\left(N^{(3\beta+6C_{\sigma})}\right).
    \end{align*}

\item 
    \textbf{For $\epsilon_{\sigma_1}$:}
    By \cref{lemma:trans_approx_variance},
    we have
    \begin{align*}
    \norm{W_{Q}}_{2},\norm{W_{K}}_{2},\norm{W_{Q}}_{2,\infty},\norm{W_{Q}}_{2,\infty}
    =
    \calO\left(N^{(27\beta+18C_{\sigma})}(\log{N})^{-9(d_x+\beta)}\right).
    \end{align*}

\item 
    \textbf{For $\epsilon_{\sigma_2}$:}
    By \cref{lemma:trans_approx_variance},
    we have
    \begin{align*}
    \norm{W_{Q}}_{2},\norm{W_{K}}_{2},\norm{W_{Q}}_{2,\infty},\norm{W_{Q}}_{2,\infty}
    =
    \calO\left(N^{(21\beta+15C_{\sigma})}(\log{N})^{-6(d_x+\beta)}\right).
    \end{align*}
\end{itemize}

We select the largest parameter bound from $\epsilon_{\text{mult},1}$  and $\epsilon_{\text{rec},3}$ that remains valid across all other approximations.
That is,
we take $N^{(7\beta+6C_{\sigma}})$ as the upper-bound.

\item \textbf{Parameter Bound on $W_{O}$ and $W_{V}$.}

Given error $\epsilon$,
the bound on each operation follows:
\begin{itemize}
\item 
    \textbf{For $\epsilon_{f_1}$:}
    By \cref{lemma:Trans_Approx_Poly},
    we have
\begin{align*}
    & ~ 
    \norm{W_O}_{2},    \norm{W_O}_{2,\infty}
    =\calO\left(N^{-\frac{(9\beta+6C_{\sigma})}{d}}(\log{N})^{\frac{3(d_x+\beta)}{d}}\right).
\end{align*}

\item 
    \textbf{For $\epsilon_{f_2}$:}
    By \cref{lemma:Trans_Approx_Poly_Gradient},
    we have
\begin{align*}
    & ~ 
    \norm{W_O}_{2},    \norm{W_O}_{2,\infty}
    =\calO\left(N^{-\frac{(3\beta+2C_{\sigma})}{d}}(\log{N})^{\frac{(d_x+\beta)}{d}}\right).
\end{align*}

\item 
    \textbf{For $\epsilon_{\text{mult},1}$:}
    By \cref{lemma:approx_prod_with_trans}
    with $m=3$,
    we have
\begin{align*}
    & ~ \norm{W_O}_{2},\norm{W_O}_{2,\infty}
    =\calO\left(N^{-3\beta}\right).
\end{align*}

\item 
    \textbf{For $\epsilon_{\text{mult},2}$:}
    By \cref{lemma:approx_prod_with_trans}
    with $m=2$,
    we have
\begin{align*}
    & ~ \norm{W_O}_{2},\norm{W_O}_{2,\infty}
    =\calO\left(N^{-2\beta}\right).
\end{align*}

\item 
    \textbf{For $\epsilon_{\text{rec},1}$, $\epsilon_{\text{rec},2}$:}
    By \cref{lemma:inverse_trans},
    we have
    \begin{align*}
    & ~ 
    \norm{W_O}_{2}, \norm{W_O}_{2,\infty}
    =\calO\left(N^{-(3\beta+C_{\sigma})}(\log{N})^{d_x+\beta}\right).
    \end{align*}

\item 
    \textbf{For $\epsilon_{\text{rec},3}$:}
    By \cref{lemma:inverse_trans},
    we have
    \begin{align*}
    & ~ 
    \norm{W_O}_{2}, \norm{W_O}_{2,\infty}
    =\calO\left(N^{-(\beta+2C_{\sigma})}\right).
    \end{align*}

\item 
    \textbf{For $\epsilon_{\sigma_1}$:}
    By \cref{lemma:trans_approx_variance},
    we have
    \begin{align*}
    & ~ 
    \norm{W_O}_{2}, \norm{W_O}_{2,\infty}
    =\calO\left(N^{-(9\beta+6C_{\sigma})}(\log{N})^{3(d_x+\beta)}\right).
    \end{align*}

\item 
    \textbf{For $\epsilon_{\sigma_2}$:}
    By \cref{lemma:trans_approx_variance},
    we have
    \begin{align*}
    & ~ 
    \norm{W_O}_{2}, \norm{W_O}_{2,\infty}
    =\calO\left(N^{-(7\beta+5C_{\sigma})}(\log{N})^{2(d_x+\beta)}\right).
    \end{align*}

\end{itemize}

Note that only $\epsilon_{f_{1}}$ and $\epsilon_{f_{2}}$ involve the reshape operation.
From \cref{lemma:trans_para_bound},
we take $\calO(\sqrt{d})$ and $\calO(d)$ $\norm{W_{V}}_{2}$ and $\norm{W_{V}}_{2,\infty}$.
Moreover,
We select the largest parameter bound from $\epsilon_{\text{rec},1}$ and $\epsilon_{\sigma_1}$ that remains valid across all other approximations.
That is,
we take $N^{-(3\beta+6C_{\sigma})}(\log{N})^{3(d_x+\beta)}$ as the upper-bound.

\item \textbf{Parameter Bound on $W_{1}$.}

Given error $\epsilon$,
the bound on each operation follows:
\begin{itemize}
\item 
    \textbf{For $\epsilon_{f_1}$:}
    By \cref{lemma:Trans_Approx_Poly},
    we have
    \begin{align*}
    & ~ \norm{W_1}_{2}, \norm{W_{1}}_{2,\infty}
    =\calO\left(N^{\frac{(9\beta+6C_{\sigma})}{d}}(\log{N})^{-\frac{3(d_x+\beta)}{d}}\cdot(\log{N})\right).
    \end{align*}

\item 
    \textbf{For $\epsilon_{f_2}$:}
    By \cref{lemma:Trans_Approx_Poly_Gradient},
    we have
    \begin{align*}
    & ~ \norm{W_1}_{2}, \norm{W_{1}}_{2,\infty}
    =\calO\left(N^{\frac{(3\beta+2C_{\sigma})}{d}}(\log{N})^{-\frac{(d_x+\beta)}{d}}\cdot(\log{N})\right).
    \end{align*}
\item 
    \textbf{For $\epsilon_{\text{mult},1}$:}
    By \cref{lemma:approx_prod_with_trans} with $m=3$ and $C=K_2$ in \eqref{eqn:main_1_K2},
    we have
    \begin{align*}
    & ~ \norm{W_1}_{2},\norm{W_{1}}_{2,\infty}
    =
    \calO\left(K_2\cdot N^{3\beta}\right)
    =
    \calO\left(N^{(4\beta+C_{\sigma})}(\log{N})^{-\frac{1}{2}(d_x+\beta)}\right).
    \end{align*}

\item 
    \textbf{For $\epsilon_{\text{mult},2}$:}
    By \cref{lemma:approx_prod_with_trans} with $m=2$ and $C=K_1$ in \eqref{eqn:main_1_K1},
    we have
    \begin{align*}
    & ~ \norm{W_1}_{2},\norm{W_{1}}_{2,\infty}
    =
    \calO\left(K_1\cdot N^{2\beta}\right)
    =
    \calO\left(N^{(2\beta+C_{\sigma})}\right).
    \end{align*}

\item 
    \textbf{For $\epsilon_{\text{rec},1}$ , $\epsilon_{\text{rec},2}$:}
    By \cref{lemma:inverse_trans},
    we have
    \begin{align*}
    & ~ \norm{W_1}_{2}, \norm{W_{1}}_{2,\infty}
    =
    \calO\left(N^{(6\beta+4C_{\sigma})}(\log{N})^{-2(d_x+\beta)}\right).
    \end{align*}

\item 
    \textbf{For $\epsilon_{\text{rec},3}$:}
    By \cref{lemma:inverse_trans},
    we have
    \begin{align*}
    & ~ \norm{W_1}_{2}, \norm{W_{1}}_{2,\infty}
    =
    \calO\left(N^{(2\beta+4C_{\sigma})}\right).
    \end{align*}

\item 
    \textbf{For $\epsilon_{\sigma_1}$:}
    By \cref{lemma:trans_approx_variance},
    we have
   \begin{align*}
    & ~ \norm{W_1}_{2}, \norm{W_{1}}_{2,\infty}
    =
    \calO\left(N^{(9\beta+6C_{\sigma})}(\log{N})^{-3(d_x+\beta)}\cdot\log{N}\right).
    \end{align*}

\item 
    \textbf{For $\epsilon_{\sigma_2}$:}
    By \cref{lemma:trans_approx_variance},
    we have
   \begin{align*}
    & ~ \norm{W_1}_{2}, \norm{W_{1}}_{2,\infty}
    =
    \calO\left(N^{(7\beta+5C_{\sigma})}(\log{N})^{-2(d_x+\beta)}\cdot\log{N}\right).
    \end{align*}
\end{itemize}

We select the largest parameter bound from $\epsilon_{\text{rec},3}$ that remains valid across all other approximations.
That is,
we take $N^{(2\beta+4C_{\sigma})}$ as the upper-bound.

\item \textbf{Parameter Bound for $W_{2}$.}

Given error $\epsilon$,
the bound on each operation follows:
\begin{itemize}
\item 
    \textbf{For $\epsilon_{f_1}$:}
    By \cref{lemma:Trans_Approx_Poly},
    we have
    \begin{align*}
    & ~ \norm{W_2}_{2}, \norm{W_{2}}_{2,\infty}
    =
    \calO\left(N^{\frac{(9\beta+6C_{\sigma})}{d}}(\log{N})^{-3\frac{(d_x+\beta)}{d}}\right).
    \end{align*}

\item 
    \textbf{For $\epsilon_{f_2}$:}
    By \cref{lemma:Trans_Approx_Poly_Gradient},
    we have
   \textbf{For $\epsilon_{f_1}$:}
    By \cref{lemma:Trans_Approx_Poly},
    we have
    \begin{align*}
    & ~ \norm{W_2}_{2}, \norm{W_{2}}_{2,\infty}
    =
    \calO\left(N^{\frac{(3\beta+2C_{\sigma})}{d}}(\log{N})^{-\frac{(d_x+\beta)}{d}}\right).
    \end{align*}

\item 
    \textbf{For $\epsilon_{\text{mult},1}$:}
    By \cref{lemma:approx_prod_with_trans} with $m=3$,
    we have
   \begin{align*}
    & ~ \norm{W_2}_{2},\norm{W_2}_{2,\infty}
    =
    \calO\left(N^{3\beta}\right).
    \end{align*}

\item 
    \textbf{For $\epsilon_{\text{mult},2}$:}
    By \cref{lemma:approx_prod_with_trans} with $m=2$,
    we have
   \begin{align*}
    & ~ \norm{W_2}_{2},\norm{W_2}_{2,\infty}
    =
    \calO\left(N^{2\beta}\right).
    \end{align*}

\item 
    \textbf{For $\epsilon_{\text{rec},1}$, $\epsilon_{\text{rec},2}$:}
    By \cref{lemma:inverse_trans},
    we have
    \begin{align*}
    & ~ \norm{W_2}_{2}, \norm{W_2}_{2,\infty}
    =
    \calO\left(N^{(3\beta+2C_{\sigma})}(\log{N})^{-(d_x+\beta)}\right).
    \end{align*}

\item 
    \textbf{For $\epsilon_{\text{rec},3}$:}
    By \cref{lemma:inverse_trans},
    we have
    \begin{align*}
    & ~ \norm{W_2}_{2}, \norm{W_2}_{2,\infty}
    =
    \calO\left(N^{(\beta+2C_{\sigma})}\right).
    \end{align*}

\item 
    \textbf{For $\epsilon_{\sigma_1}$:}
    By \cref{lemma:trans_approx_variance},
    we have
   \begin{align*}
    & ~ \norm{W_2}_{2}, \norm{W_2}_{2,\infty}
    =
    \calO\left(N^{(9\beta+6C_{\sigma})}(\log{N})^{-3(d_x+\beta)}\right).
    \end{align*}

\item 
    \textbf{For $\epsilon_{\sigma_2}$:}
    By \cref{lemma:trans_approx_variance},
    we have
   \begin{align*}
    & ~ \norm{W_2}_{2}, \norm{W_2}_{2,\infty}
    =
    \calO\left(N^{(7\beta+5C_{\sigma})}(\log{N})^{-2(d_x+\beta)}\right).
    \end{align*}
\end{itemize}

We select the largest parameter bound from $\epsilon_{\text{mult},1}$ and $\epsilon_{\text{rec},3}$ that remains valid across all other approximations.
That is,
we take $N^{(3\beta+2C_{\sigma})}$ as the upper-bound.

\item \textbf{Parameter Bound for $E$.}

Since only $\epsilon_{f_{1}}$ and $\epsilon_{f_{2}}$ involve the reshape operation.
From \cref{lemma:trans_para_bound},
we take $\calO(d^{\frac{1}{2}}L^{\frac{3}{2}})$ for $\norm{E^{\top}}_{2,\infty}$.
\end{itemize}

By integrating results above, 
we derive the following parameter bounds for the transformer network, ensuring valid approximation across all nine approximations.

\begin{align*}
& \norm{W_{Q}}_{2}, \norm{W_{K}}_{2},
\norm{W_{Q}}_{2,\infty}, \norm{W_{K}}_{2,\infty}
=
\calO\left(N^{(7\beta+6C_{\sigma})}\right); \\
&
\norm{W_{O}}_{2},\norm{W_{O}}_{2,\infty}=\calO\left(N^{-(3\beta+6C_{\sigma})}(\log{N})^{3(d_x+\beta)}\right);\\
&
\norm{W_{V}}_{2}=\calO(\sqrt{d}); \norm{W_{V}}_{2,\infty}=\calO (d);
\norm{E^{\top}}_{2,\infty}=\calO\left(d^{\frac{1}{2}}L^{\frac{3}{2}}\right);\\ 
& \norm{W_{1}}_{2}, \norm{W_{1}}_{2,\infty}
=
\calO\left(N^{(2\beta+4C_{\sigma})}\right);
C_\calT=\calO\left(\sqrt{\log{N}}/\sigma_{t}^{2}\right);
\\
& \norm{W_{2}}_{2}, \norm{W_{2}}_{2,\infty}
=
\calO\left(N^{(3\beta+2C_{\sigma})}\right). 
\end{align*}

The last  network output bound $C_{\calT}=\calO(\sqrt{d_x\log{N}}/\sigma_{t}^{2})$ follows 
the entry-wise minimum bounds $K(C_{x}\sqrt{d\log{N}}+1)/\sigma_{t}^{2}$ in $\ell_{\infty}$ distance by \cref{bounds_on_score}.
\end{itemize}

This completes the proof.
\end{proof}
\end{itemize} 

\subsection{Main Proof of \texorpdfstring{\cref{thm:Main_1}}{}}
\label{proof:thm:Main_1}

In \cref{lemma:Score_Approx_Trans},
we establish the score approximation with transformer that incorporates every essential components and encodes the H\"{o}lder smoothness in the final result.
However,
it is only valid within the input domain $[C_{x}\sqrt{\log{N}},C_{x}\sqrt{\log{N}}]^{d_{x}}\times[0,1]^{d_{y}}$,
and we also excludes region $p_{t}<\epsilon_{\text{low}}$ where the problem of score explosion remains unaddressed.

To combat this,
we introduce two additional lemmas.
The first lemma gives us the error caused by the truncation of $\R^{d_{x}}$ within a radius $R_{1}$ in $\ell_{2}$ distance. 
\begin{lemma}[Truncate $x$ for Score Function, Lemma A.1 of \cite{fu2024unveil}]
\label{lemma:truncate_x}
Assume \cref{assumption:conditional_density_function_assumption_1}.
For any $R_{1}>1,$ $y,t>0$ we have
\begin{align*}
& \int_{\norm{x}_{\infty}\geq R_{1}}p_{t}(x|y)dx\leq R_{1}\exp(-C_{2}^{\prime}R_{1}^{2}),\\
& \int_{\norm{x}_{\infty}\geq R_{1}}\norm{\nabla \log{p_{t}(x|y)}}_{2}^{2}p_{t}(x|y)dx\leq\frac{R_{1}^{3}}{\sigma_{t}^{4}}\exp(-C_{2}^{\prime}R_{1}^{2}),
\end{align*}
where $C_{2}^{\prime}=C_{2}/(2\max(C_{2},1))$.
\end{lemma}

\begin{remark}
Because we only impose assumption on the light tail property of the conditional distribution in \cref{assumption:conditional_density_function_assumption_1},
the unboundedness of $x$ necessitates a truncation for integrals regarding $x$, or else the result would diverge.    
\end{remark}

Furthermore,
we address the explosion of score function with the second lemma.
\begin{lemma}[Lemma A.2 of \cite{fu2024unveil}]
\label{lemma:truncate_p_x_cond_y}
Assume
\cref{assumption:conditional_density_function_assumption_1}.
For any $R_{2},y,\epsilon_{\text{low}}>0$ we have
\begin{align*}
& \int_{\norm{x}_{\infty}\leq R_{2}}\mathbbm{1}\{\abs{p_{t}(x|y)}<\epsilon_{\text{low}}\}
\cdot
p_{t}(x|y) \dd x \leq R_{2}^{d_{x}}\epsilon_{\text{low}},\\
& \int_{\norm{x}_{\infty}\leq R_{2}}\mathbbm{1}\{\abs{p_{t}(x|y)}<\epsilon_{\text{low}}\}\cdot
\norm{\nabla \log{p_{t}(x|y)}}_2^2 p_{t}(x|y) \dd x \leq \frac{1}{\sigma_{t}^{4}}R_{2}^{d_{x}+2}\epsilon_{\text{low}}.
\end{align*}
\end{lemma}

\begin{remark}
Recall that the score function has the form $\nabla\log{p_{t}}(x|y)=\nabla p_{t}(x|y)/p_{t}(x|y)$.
It is essential to set a threshold for $p_{t}(x|y)$ prevents the explosion of the score function. 
\end{remark}

We begin the proof of \cref{thm:Main_1}.

\begin{proof}[Proof Sketch of \cref{thm:Main_1}]
In the following proof, 
we give error bound for the three terms:

\begin{itemize}
    \item \textbf{(A.1): The approximation for $\norm{x}_{\infty}>R_{1}$.} 
    
    This step controls the error from truncation of $\R^{d_x}$ with radius $R_1$ in $\ell_2$ distance.  
    We approximate the error with \cref{lemma:truncate_x}  
    
    \item 
    \textbf
    {(A.2): The approximation for $\mathbf{1}\{p_{t}(x|y)<\epsilon_{\text{low}}\}$ and 
$\{\norm{x}_{\infty}\leq{R_{1}}\}$.} 

    This step controls the error from setting a threshold to prevent score explosion within the bounded domain $\norm{x}_\infty\leq R_1$.
    We approximate the error with \cref{lemma:truncate_p_x_cond_y}.
    
    \item \textbf{(A.3) The approximation for  $\mathbf{1}\{p_{t}(x|y)\geq\epsilon_{\text{low}}\}$  and $\{\norm{x}_{\infty}\leq{R_{1}}\}$.}

   With previous two steps ensuring the bounded domain and preventing the divergence of score function,
   we approximate with \cref{lemma:Score_Approx_Trans}.
\end{itemize}
\end{proof}

\begin{proof}[Proof of \cref{thm:Main_1}]
We apply $N=N^{1/(d_x+d_y)}$ in \cref{lemma:Score_Approx_Trans}.
Throughout the proof, 
we use $N$ as a notational simplification, 
with the understanding that 
$N$ represents $N^{1/(d_x+d_y)}$ in full form.
At the end of of the proof we replace $N$ by $N^{1/(d_x+d_y)}$.

To begin with,
we set $R_{1}=R_{2}=\sqrt{2\beta\log{N}/C_{2}^{\prime}}$ in \cref{lemma:truncate_x} and \cref{lemma:truncate_p_x_cond_y},
and we expand the target into three parts $(A_1)$, $(A_2)$, and $(A_3)$: 
\begin{align*}
& ~ \int_{\R^{d_{x}}}\norm{s(x,y,t)-\nabla\log{p_{t}(x|y)}}_{2}^{2}\cdot p_{t}(x|y) \dd x\\
= & ~ \underbrace{\int_{\norm{x}_{\infty}>\sqrt{\frac{2\beta}{C_{2}^{\prime}}\log{N}}}\norm{s(x,y,t)-\nabla\log{p_{t}(x|y)}}_{2}^{2}\cdot p_{t}(x|y) \dd x}_{(A_1)},\\
& + ~   \underbrace{\int_{\norm{x}_{\infty}\leq\sqrt{\frac{2\beta}{C_{2}^{\prime}}\log{N}}}\mathbbm{1}\{\abs{p_{t}(x|y)}<\epsilon_{\text{low}}\}\norm{s(x,y,t)-\nabla\log{p_{t}(x|y)}}_{2}^{2}\cdot p_{t}(x|y) \dd x}_{(A_2)}\\
& + ~  \underbrace{\int_{\norm{x}_{\infty}\leq\sqrt{\frac{2\beta}{C_{2}^{\prime}}\log{N}}}\mathbbm{1}\{\abs{p_{t}(x|y)}\geq\epsilon_{\text{low}}\}\norm{s(x,y,t)-\nabla \log{p_{t}(x|y)}}_{2}^{2}\cdot p_{t}(x|y) \dd x}_{(A_3)}.
\end{align*}

We derive the bound for $(A_1), (A_2), (A_3)$ and combine these results.

\begin{itemize}
\item 
\textbf{Bounding ($A_1$).}
We apply \cref{lemma:truncate_x}.
Note that we have $\norm{s(x,y,t)}_{\infty}\lesssim\sqrt{\log{N}}/\sigma_{t}^{2}$
from the construction of the score estimator in \cref{lemma:Score_Approx_Trans}.
\begin{align*}
& ~ \int_{\norm{x}_{\infty}>\sqrt{\frac{2\beta}{C_{2}^{\prime}}\log{N}}}\norm{s(x,y,t)-\nabla\log{p_{t}(x|y)}}_{2}^{2}\cdot p_{t}(x|y) \dd x
\annot{By expanding the $\ell_{2}$ norm}\\
\leq & ~ 
2\int_{\norm{x}_{\infty}>\sqrt{\frac{2\beta}{C_{2}^{\prime}}\log{N}}}\norm{s(x,y,t)}_{2}^{2}\cdot p_{t}(x|y)\dd x
+ 2\int_{\norm{x}_{\infty}>\sqrt{\frac{2\beta}{C_{2}^{\prime}}\log{N}}}\norm{\nabla\log{p_{t}(x|y)}}_{2}^{2}\cdot p_{t}(x|y) \dd x
\annot{By $\norm{\cdot}_{2}^{2}\leq d_{x}\norm{\cdot}_{\infty}^{2}$}\\
\leq & ~
2d_x\int_{\norm{x}_{\infty}>\sqrt{\frac{2\beta}{C_{2}^{\prime}}\log{N}}}
\norm{s(x,y,t)}_{\infty}^{2}\cdot p_{t}(x|y)\dd x
+ 2\int_{\norm{x}_{\infty}>\sqrt{\frac{2\beta}{C_{2}^{\prime}}\log{N}}}\norm{\nabla\log{p_{t}(x|y)}}_{2}^{2}\cdot p_{t}(x|y) \dd x
\annot{By the $\ell_{\infty}$ bound on the score function}\\
\lesssim & ~ 
2d_{x}\left(\frac{\sqrt{\log{N}}}{\sigma_{t}^{2}}\right)^{2}\int_{\norm{x}_{\infty}>\sqrt{\frac{2\beta}{C_{2}^{\prime}}\log{N}}}p_{t}(x|y)\dd x 
+ 2\int_{\norm{x}_{\infty}>\sqrt{\frac{2\beta}{C_{2}^{\prime}}\log{N}}}\norm{\nabla\log{p_{t}(x|y)}}_{2}^{2}\cdot p_{t}(x|y)\dd x
\annot{By \cref{lemma:truncate_x} and dropping constant}\\
\lesssim & ~ 
2d_{x}\left(\frac{\sqrt{\log{N}}}{\sigma_{t}^{2}}\right)^{2}\left(\sqrt{\frac{2\beta}{C_{2}^{\prime}}\log{N}}N^{-2\beta}\right) 
+ \frac{2}{\sigma_{t}^{4}}\left(\frac{2\beta}{C_{2}^{\prime}}\log{N}\right)^{\frac{3}{2}}N^{-2\beta}
\annot{By dropping constant and lower order term}\\
\lesssim & ~
\frac{1}{\sigma_{t}^{4}}N^{-2\beta}(\log{N})^{\frac{3}{2}}.
\end{align*}

\item 
\textbf{Bounding ($A_2$).}
We apply \cref{lemma:truncate_p_x_cond_y}.
Note that we set $\epsilon_{\text{low}} = C_{3}N^{-\beta}(\log{N})^{(d_{x}+\textcolor{blue}{k_1})/2}$ in \cref{lemma:Score_Approx_Trans}.
\begin{align*}
&~
\int_{\norm{x}_{\infty}\leq\sqrt{\frac{2\beta}{C_{2}^{\prime}}\log{N}}}\mathbbm{1}\{\abs{p_{t}(x|y)}<\epsilon_{\text{low}}\}\norm{s(x,y,t)-\nabla\log{p_{t}(x|y)}}_{2}^{2}\cdot p_{t}(x|y) \dd x
\annot{By expanding the $\ell_2$ norm}\\
\leq & ~
\int_{\norm{x}_{\infty}\leq\sqrt{\frac{2\beta}{C_{2}^{\prime}}\log{N}}}
2\mathbbm{1}\{\abs{p_{t}(x|y)}<\epsilon_{\text{low}}\}
\left(\norm{s(x,y,t)}_{2}^{2} +\norm{\nabla\log{p_{t}(x|y)}}_{2}^{2}\right)
\cdot p_{t}(x|y) \dd x
\annot{By $\norm{\cdot}_{2}^{2}\leq d_{x}\norm{\cdot}_{\infty}^{2}$}\\
\leq & ~
\int_{\norm{x}_{\infty}\leq\sqrt{\frac{2\beta}{C_{2}^{\prime}}\log{N}}}
\mathbbm{1}\{\abs{p_{t}(x|y)}<\epsilon_{\text{low}}\}
\left(d_x\norm{s(x,y,t)}_{\infty}^{2}+\norm{\nabla\log{p_{t}(x|y)}}_{2}^{2}\right)
\cdot p_{t}(x|y) \dd x
\annot{By the $\ell_{\infty}$ bound on the score function}\\
\lesssim & ~
\int_{\norm{x}_{\infty}\leq\sqrt{\frac{2\beta}{C_{2}^{\prime}}\log{N}}}
\mathbbm{1}\{\abs{p_{t}(x|y)}<\epsilon_{\text{low}}\}
\left(d_x\left(\frac{\sqrt{\log{N}}}{\sigma_{t}^{2}}\right)^{2}+\norm{\nabla\log{p_{t}(x|y)}}_{2}^{2}\right)
\cdot p_{t}(x|y) \dd x
\annot{By \cref{lemma:truncate_p_x_cond_y} and dropping constant}\\
\lesssim & ~
d_x\left(\frac{\sqrt{\log{N}}}{\sigma_{t}^{2}}\right)^{2}\left(\frac{2\beta}{C_{2}^{\prime}}\log{N}\right)^{\frac{d_x}{2}}\epsilon_{\text{low}}
+
\left(\frac{2\beta}{C_{2}^{\prime}}\log{N}\right)^{\frac{d_x+2}{2}}\frac{\epsilon_{\text{low}}}{\sigma_{t}^{4}}
\annot{By dropping constant and lower order term}\\
\lesssim & ~
\frac{1}{\sigma_{t}^{4}}\left(\log{N}\right)^{\frac{d_x+2}{2}}\epsilon_{\text{low}}.
\end{align*}

\item 
\textbf{Bounding ($A_3$).}
We apply \cref{lemma:Score_Approx_Trans}.
\begin{align*}
&~
\int_{\norm{x}_{\infty}\leq\sqrt{\frac{2\beta}{C_{2}^{\prime}}\log{N}}}\mathbbm{1}\{\abs{p_{t}(x|y)}\geq\epsilon_{\text{low}}\}\norm{s(x,y,t)-\nabla \log{p_{t}(x|y)}}_{2}^{2}\cdot p_{t}(x|y) \dd x
\annot{By $\norm{\cdot}_{2}^{2}\leq d_x\norm{\cdot}_{\infty}^{2}$}\\
\leq&~
\int_{\norm{x}_{\infty}\leq\sqrt{\frac{2\beta}{C_{2}^{\prime}}\log{N}}}{\mathbbm{1}\{\abs{p_{t}(x|y)}\geq\epsilon_{\text{low}}\}}d_{x}\norm{s(x,y,t)-\nabla \log{p_{t}(x|y)}}_{\infty}^{2}\cdot p_{t}(x|y) \dd x
\annot{Multiply with $p_t/p_t$}
\\
= & ~
\int_{\norm{x}_{\infty}\leq\sqrt{\frac{2\beta}{C_{2}^{\prime}}\log{N}}}\frac{\mathbbm{1}\{\abs{p_{t}(x|y)}\geq\epsilon_{\text{low}}\}}{p_{t}(x|y)}d_{x}\norm{s(x,y,t)-\nabla \log{p_{t}(x|y)}}_{\infty}^{2}\cdot p_{t}^{2}(x|y) \dd x
\annot{By \cref{lemma:Score_Approx_Trans}}\\
\lesssim & ~
\frac{B^{2}d_{x}}{\sigma_{t}^{2}}N^{-2\beta}(\log{N})^{d_{x}+\textcolor{blue}{k_1}+1} \int_{\norm{x}_{\infty}\leq\sqrt{\frac{2\beta}{C_{2}^{\prime}}\log{N}}}\mathbbm{1}\{\abs{p_{t}(x|y)}\geq\epsilon_{\text{low}}\}{p_{t}(x|y)}\dd x
\annot{Multiply with $\epsilon_{\text{low}}/\epsilon_{\text{low}}$}\\
= & ~
\frac{B^{2}d_{x}}{\sigma_{t}^{2}\epsilon_{\text{low}}}N^{-2\beta}(\log{N})^{d_{x}+\textcolor{blue}{k_1}+1} \int_{\norm{x}_{\infty}\leq\sqrt{\frac{2\beta}{C_{2}^{\prime}}\log{N}}}\mathbbm{1}\{\abs{p_{t}(x|y)}\geq\epsilon_{\text{low}}\}\frac{\epsilon_{\text{low}}}{p_{t}(x|y)}\dd x
\annot{By \cref{lemma:truncate_p_x_cond_y}}\\
\lesssim & ~
\frac{B^{2}d_{x}}{\sigma_{t}^{2}\epsilon_{\text{low}}}N^{-2\beta}(\log{N})^{d_{x}+\textcolor{blue}{k_1}+1}\cdot\left(\frac{2\beta}{C_{2}^{\prime}}\log{N}\right)^{\frac{d_{x}}{2}}
\annot{By the choice of $\epsilon_{\text{low}}$ and dropping lower order term}\\
\lesssim & ~
\frac{B^{2}d_{x}}{\sigma_{t}^{4}\epsilon_{\text{low}}}N^{-2\beta}(\log{N})^{\frac{3d_{x}}{2}+\textcolor{blue}{k_1}+1}.
\end{align*}

\item 
\textbf{Combining the Results.}

Combining ($A_1$), ($A_2$) and ($A_3$),
we have
\begin{align*}
&~\int_{\R^{d}}\norm{s(x,y,t)-\nabla\log{p_{t}(x|y)}}_{2}^{2}p_{t}(x|y)\dd x\\
\lesssim&~
\underbrace{
\frac{N^{-2\beta}(\log{N})^{\frac{3}{2}}}{\sigma_{t}^{4}}
}_{(A_1)}
+ \underbrace{
\frac{\epsilon_{\text{low}}(\log{N})^{\frac{d_{x}+2}{2}}}{\sigma_{t}^{4}}
}_{(A_2)}
+ \underbrace{
\frac{B^{2}d}{\sigma_{t}^{4}\epsilon_{\text{low}}}N^{-2\beta}(\log{N})^{\frac{3d_{x}}{2}+\textcolor{blue}{k_1}+1}
}_{(A_3)}.
\end{align*}
By replacing $\epsilon_{\text{low}}$ with $C_{3}N^{-\beta}(\log{N})^{{d_x+\textcolor{blue}{k_1}}/{2}}$ and using the relation $\textcolor{blue}{k_1}\leq\beta$,\footnote{Recall the definition of the \holder smoothness from \cref{def:holder_norm_space}.}
we obtain
\begin{align*}
\int_{\R^{d}}\norm{s(x,y,t)-\nabla\log{p_{t}(x|y)}}_{2}^{2}p_{t}(x|y)\dd x
= 
\calO\left(\frac{B^{2}}{\sigma_{t}^{4}}N^{-\beta}(\log{N})^{d_{x}+\frac{\beta}{2}+1}\right).
\end{align*}
\end{itemize}

Replacing $N$ with $N^{1/(d_x+d_y)}$ completes the first part of the proof.

The transformer parameter norm bounds follow \cref{lemma:Score_Approx_Trans},
with the replacement of $N$ with $N^{{1}/({d_{x}+d_{y}}})$ as well.
Note that this results in $t\in[N^{-C_\alpha/(d_x+d_y)},C_\sigma/((d_x+d_y))\log{N}]$.
For better interpretation of the cutoff and early stopping time parameter,
we reset $C_\alpha$ as $(d_x+d_y)C_\alpha$ and $C_\sigma$ as $(d_x+d_y)C_\sigma$ such that $t\in[N^{-C_\alpha},C_\sigma\log{N}]$.

This completes the proof.
\end{proof}

\clearpage
\section{Proof of \texorpdfstring{\cref{thm:Main_2_informal}}{}}\label{sec:appendix_proof_main2}

We provide the formal version of \cref{thm:Main_2_informal} at the end of \cref{sec:op_approx}.

\begin{itemize}
    \item \textbf{Step 0.}
    We decompose the density function and the score function under \cref{assumption:conditional_density_function_assumption_2}.
    In \cref{lemma:score_decomp},
    we provide details regarding the decomposed form of the score function presented in \eqref{eqn:stronger_holder_score_decom}.
    We specify the upper and lower bound on $h$ and $\nabla h$ in \cref{lemma:bounds_on_h}.

    \item \textbf{Step 1.}
    Similar to the domain discretization in the proof of previous main result,
    we discretize the input domain of the decomposed density function in \cref{lemma:clipping_h}.
    
    \item \textbf{Step 2.}
    We construct polynomial approximation based on Taylor expansion of $h$ and $\nabla h$ in \cref{lemma:diffused_local_polynomials_2,lemma:diffused_local_polynomials_grad_2}.
    The approximation result captures the local \holder smoothness,
    with improved precision relative to the analogous step in \cref{lemma:diffused_local_polynomials} and \cref{lemma:diffused_local_polynomials_grad}.

    \item \textbf{Step 3.}
    We approximate $h$  and $\nabla h$ with transformer in \cref{lemma:Trans_Approx_Poly_2,lemma:Trans_Approx_Poly_Gradient_2}.
    In order to construct the score approximator with transformer,
    we approximate several additional algebraic operators with transformer in \cref{lemma:approx_alpha_square}, \cref{lemma:approx_hat_alpha} and \cref{lemma:approx_hat_sigma}.
    We incorporate these results into a unified transformer architecture in \cref{lemma:score_approx_trans_2}.
\end{itemize}

\paragraph{Organization.}
\cref{sec:axuiliary_lem_thm32} includes the four steps and auxiliary lemmas supporting our proof.
\cref{sec:op_approx} includes the formal version and main proof
of \cref{thm:Main_2_informal}.

\subsection{Auxiliary Lemmas}
\label{sec:axuiliary_lem_thm32}

\paragraph{Step 0: Decompose the Score with Stronger H\"{o}lder Smoothness Assumption.}

We utilize the condition assumed in \cref{assumption:conditional_density_function_assumption_2} to achieve the decomposition.

\begin{lemma}[Lemma B.1 of \citet{fu2024unveil}]
\label{lemma:score_decomp}
Assume \cref{assumption:conditional_density_function_assumption_2}.
The conditional distribution at time $t$ has the following expression:
\begin{align*}
p_{t}(x|y) = \frac{1}{(\alpha_{t}^{2}+ C_{2}\sigma_{t}^{2})^{d_x/2}}\exp(-\frac{C_{2}\norm{x}_{2}^{2}}{2(\alpha_{t}^{2}+C_{2}\sigma_{t}^{2})})h(x,y,t).
\end{align*}
Moreover,
the score function has the following expression:
\begin{align*}
\nabla\log{p_{t}(x|y)} = \frac{-C_{2}x}{\alpha_{t}^{2}+C_{2}\sigma_{t}^{2}} + \frac{\nabla h(x,y,t)}{h(x,y,t)},
\end{align*}
where $h(x,y,t) = \int\frac{f(x_{0},y)}{\hat{\sigma}_{t}^{d}(2\pi)^{d/2}}\exp(-\frac{\norm{x_{0}-\hat{\alpha}_{t}x}^{2}}{2\hat{\sigma}_{t}^{2}})\dd x_{0}$, $\hat{\sigma}_{t} = \frac{\sigma_{t}}{(\alpha_{t}^{2}+C_{2}\sigma_{t}^{2})^{1/2}}$, and
$\hat{\alpha}_{t}=\frac{\alpha_{t}}{\alpha_{t}^{2}+C_{2}\sigma_{t}^{2}}$.
\end{lemma}

\begin{proof}
From \cref{assumption:conditional_density_function_assumption_2}, 
we have the initial conditional density with the form: $p(z|y)=\exp\left(-C_{2}\norm{z}_{2}^{2}/2\right)\cdot f(z,y)$.

This allows the decomposition:
\begin{align}
\label{eqn:assump2_first_decomp}
p_{t}(x|y) 
& = \int\frac{1}{\sigma_{t}^{d}(2\pi)^{d/2}}p(z|y)\exp\left(-\frac{\norm{x-\alpha_{t}z}^{2}}{2\sigma_{t}^{2}}\right) \dd z,\\
& = \frac{1}{\sigma_{t}^{d}(2\pi)^{d/2}}\int\exp\left(-\frac{C_{2}\norm{z}_{2}^{2}}{2}\right)f(z,y)\exp\left(-\frac{\norm{x-\alpha_{t}z}^{2}}{2\sigma_{t}^{2}}\right) \dd z.
\label{eqn:p_t_2exp}
\end{align}
We rearrange the two exponential terms in \eqref{eqn:p_t_2exp} into
\begin{align*}
\exp\left(-\frac{C_{2}\norm{z}_{2}^{2}}{2}\right)\exp\left(-\frac{\norm{x-\alpha_{t}z}^{2}}{2\sigma_{t}^{2}}\right) = \exp\left(-\frac{1}{2\sigma_{t}^{2}}\sum_{i=1}^{d}(x[i]^{2}-2\alpha_{t}x[i]z[i]+\alpha_{t}^{2}z[i]^{2}+C_{2}\sigma_{t}^{2}z[i]^{2})\right).
\end{align*}
Note that, 
we replace the summation in the exponents by first focusing on one coordinate and then do the product for all $d$ components.

Without loss of generality,
we derive the first coordinate of the fucntion:
\begin{align*}
&~ \exp\left(-\frac{1}{2\sigma_{t}^{2}}(x[1]^{2}-2\alpha_{t}x[1]z[1]+\alpha_{t}^{2}z[1]^{2}+C_{2}\sigma_{t}^{2}z[1]^{2})\right),\\
= &~ 
\exp\left(-\frac{1}{2\sigma_{t}^{2}}{(\alpha_{t}^{2}+C_{2}\sigma_{t}^{2})}\left(z[1]^{2}-\frac{2\alpha_{t}}{\alpha_{t}^{2}+C_{2}\sigma_{t}^{2}}x[1]z_[1] + \frac{x[1]^{2}}{\alpha_{t}^{2} + C_{2}\sigma_{t}^{2}}\right)\right),\\
=&~ 
\exp\left(-\frac{1}{2\sigma_{t}^{2}}{(\alpha_{t}^{2}+C_{2}\sigma_{t}^{2})}\left(z[1]-\frac{\alpha_{t}x[1]}{\alpha_{t}^{2}+C_{2}\sigma_{t}^{2}}\right)^{2} - \frac{1}{2\sigma_{t}^{2}}\left(\frac{-\alpha_{t}^{2}}{\alpha_{t}^{2}+C_{2}\sigma_{t}^{2}} + 1\right)x[1]^{2}\right),\\
=&~ 
\exp\left(-\frac{1}{2\sigma_{t}^{2}}{(\alpha_{t}^{2}+C_{2}\sigma_{t}^{2})}\left(z[1]-\frac{\alpha_{t}x[1]}{\alpha_{t}^{2}+C_{2}\sigma_{t}^{2}}\right)^{2}\right)\exp\left(-\frac{C_{2}x[1]^{2}}{2(\alpha_{t}^{2}+C_{2}\sigma_{t}^{2})}\right).
\end{align*}
The other $d_{x}-1$ coordinates abide by the same derivation.
Consider the product of them,
we have:
\begin{align}
\label{two exp result}
& ~ \exp\left(-\frac{C_{2}\norm{z}_{2}^{2}}{2}\right)\exp\left(-\frac{\norm{x-\alpha_{t}z}^{2}}{2\sigma_{t}^{2}}\right),\\
= & ~  \exp\left(-\frac{1}{2\sigma_{t}^{2}}{(\alpha_{t}^{2}+C_{2}\sigma_{t}^{2})}\norm{z-\frac{\alpha_{t}x}{\alpha_{t}^{2}+C_{2}\sigma_{t}^{2}}}^{2}\right)\exp\left(-\frac{C_{2}}{2(\alpha_{t}^{2}+C_{2}\sigma_{t}^{2})}\norm{x}_{2}^{2}\right).
\end{align}
Following \cite{fu2024unveil},
we plug \eqref{two exp result} into \eqref{eqn:assump2_first_decomp} and set $\hat{\alpha}_{t} =\frac{\alpha_{t}}{\alpha_{t}^{2}+C_{2}\sigma_{t}^{2}}$ and $\hat{\sigma}_{t}^{2} = \frac{\sigma_{t}^{2}}{\alpha_{t}^{2}+C_{2}\sigma_{t}^{2}}$ for simplicity.
Then we get:
\begin{align*}
&~ p_{t}(x|y)\\
= &~  \frac{1}{\sigma_{t}^{d}(2\pi)^{d/2}}\exp\left(-\frac{C_{2}\norm{x}_{2}^{2}}{2(\alpha_{t}^{2}+C_{2}\sigma_{t}^{2})}\right)\int f(z,y)\exp\left(-\frac{1}{2\sigma_{t}^{2}}{(\alpha_{t}^{2}+C_{2}\sigma_{t}^{2})}\norm{z-\frac{\alpha_{t}x}{\alpha_{t}^{2}+C_{2}\sigma_{t}^{2}}}^{2}\right) \dd z,\\
= &~  \frac{1}{\sigma_{t}^{d}(2\pi)^{d/2}}\exp\left(-\frac{C_{2}\norm{x}_{2}^{2}}{2(\alpha_{t}^{2}+C_{2}\sigma_{t}^{2})}\right)\int f(z,y)\exp\left(-\frac{\norm{z-\hat{\alpha}_{t}x}^{2}}{2\hat{\sigma}_{t}^{2}}\right) \dd z.
\end{align*}
Finally,
we define $h(x,y,t) = \int\frac{1}{\hat{\sigma}_{t}^{d}(2\pi)^{d/2}}f(z,y)\exp\left(-\frac{\norm{z-\hat{\alpha}_{t}x}^{2}}{2\hat{\sigma}_{t}^{2}}\right) \dd z$
and plug it back to the equation above.

The form of the score function is proved by simply implementing the logarithm and the gradient to the result of $p_{t}(x|y)$

This completes the proof.
\end{proof}

Next,
we provide lemma that provides bound on $h(x,y,t)$ and $\nabla h(x,y,t)$ in \cref{lemma:score_decomp} 
\begin{lemma}[Lemma B.8 of \cite{fu2024unveil}]
\label{lemma:bounds_on_h}
Under \cref{assumption:conditional_density_function_assumption_2}, we have the following bounds for $h(x,y,t)$ and $\frac{\hat{\sigma}_{t}}{\hat{\alpha}_{t}}\nabla h(x,y,t)$
\begin{align*}
C_{1}\leq h(x,y,t)\leq B, \quad \norm{{\frac{\hat{\sigma}_{t}}{\hat{\alpha}_{t}}}\nabla h(x,y,t)}_{\infty}\leq\sqrt{\frac{2}{\pi}}B,
\end{align*}
where $C_1$ and $B$ are the hyperparameters of $\calH^{\beta}(\R^{d_{x}}\times[0,1]^{d_{y}},B)$ in \cref{assumption:conditional_density_function_assumption_2}.
\end{lemma}
\begin{remark}[Bound on $h$ and $\nabla h$]
We reiterate that \cref{lemma:bounds_on_h} drives the key distinction between the analyses in \cref{thm:Main_1} and \cref{thm:Main_2_informal}.
Specifically,
in \cref{proof:thm:Main_1},
the decomposed term containing the threshold $\epsilon_{\text{low}}$ results in lower approximation rate,
while bounds on $h$ and $\nabla h$ eliminate the need of the threshold with $h$'s lower bound $C_1$,
rendering faster approximation rate.
\end{remark}

\paragraph{Step 1: Discretize $\R^{d_{x}}\times[0,1]^{d_{y}}$ for $h(x,y,t)$.}
This step parallels \cref{clipping_integral};
however,
the discretization differs due to the structure of $h$.
\begin{lemma}[Clipping Integral, Lemma B.10 of \citet{fu2024unveil}]
\label{lemma:clipping_h}
Assume \cref{assumption:conditional_density_function_assumption_2}.
Consider any integer vector $\kappa\in\mathbb{Z}_{+}^{d_{x}}$ with $\norm{\kappa}_{1}\leq n$.
There exists a constant $C(n,d_x)$, 
such that for any $x\in\R^{d_{x}}$ and $0<\epsilon\leq 0.99$,
it holds
\begin{align}\label{eqn:clipped_integral_h}
\int_{\R^{d_{x}}\setminus B_{x}}\abs{\left(\frac{\hat{\alpha}_{t}x_{0}-x}{\hat{\sigma}_{t}}\right)^{\kappa}} \cdot 
p(x_{0}|y) \cdot \frac{1}{\hat{\sigma}_{t}^{d}(2\pi)^{d/2}}\exp(-\frac{\norm{\hat{\alpha}_{t}x_{0}-x}^{2}}{2\hat{\sigma}_{t}^{2}}) \dd x_{0}
\leq\epsilon,
\end{align}
where $\left(\frac{\hat{\alpha}_{t}x_{0}-x}{\hat{\sigma}_{t}}\right)^{\kappa}\coloneqq((\frac{\hat{\alpha}_{t}x_0[1]_1-x[1]}{\hat{\sigma}_{t}})^{\kappa[1]},(\frac{\hat{\alpha}_{t}x_{0}[2]-x[2]}{\hat{\sigma}_{t}})^{\kappa[2]},\ldots,(\frac{\hat{\alpha}_{t}x_0[d_x]-x[d_x]}{\hat{\sigma}_{t}})^{\kappa[d_x]})$ 
and
\begin{align*} 
B_{x} 
\coloneqq 
\Big[\hat{\alpha}_{t}x-C(n,d)\hat{\sigma}_{t}\sqrt{\log{\epsilon^{-1}}}, \hat{\alpha}_{t}x+C(n,d)\hat{\sigma}_{t}\sqrt{\log{\epsilon^{-1}}}\Big]^{d_{x}}.
\end{align*}
\end{lemma}

\paragraph{Step 2: Approximate $h$ and $\nabla h$ with Polynomials.}

Similar to the construction of the diffused local polynomials in  \cref{lemma:Trans_Approx_Poly} and \cref{lemma:Trans_Approx_Poly_Gradient},
the following two lemmas render the first step approximation for $h(x,y,t)$ and $\nabla h(x,y,t)$ that captures the local smoothness.
\begin{lemma}[Approximation with Diffused Local Polynomials, Lemma B.4 of \cite{fu2024unveil}]
\label{lemma:diffused_local_polynomials_2}
Assume \cref{assumption:conditional_density_function_assumption_2}.
For sufficiently larger $N>0$ and constant $C_{2},$
there exists a diffused local polynomial $f_{1}(x,y,t)$ with at most $N^{d+d_{y}}(d+d_{y})^{\textcolor{blue}{k_1}}$ monomials such that
\begin{align*}
\abs{f_{1}(x,y,t) - h(x,y,t)}\lesssim BN^{-\beta}\log^{\frac{\textcolor{blue}{k_1}}{2}}{N},
\end{align*}
for any $x\in[-C_{x}\sqrt{\log{N}},C_{x}\sqrt{\log{N}}]^{d_{x}}, y\in[0,1]^{d_{y}}$ and $t>0.$
\end{lemma}

\begin{lemma}[Counterpart of \cref{lemma:diffused_local_polynomials_2}, Lemma B.6 of \cite{fu2024unveil}]
\label{lemma:diffused_local_polynomials_grad_2}
Assume \cref{assumption:conditional_density_function_assumption_2}.
For sufficiently larger $N>0$ and constant $C_{2},$
there exists a diffused local polynomial $f_{2}(x,y,t)\in\calT_{R}^{\textcolor{blue}{h,s,r}}$ with at most $N^{d_{x}+d_{y}}(d_{x}+d_{y})^{\textcolor{blue}{k_1}}$ monomials $f_{2}[i](x,y,t)$ such that
\begin{align*}
\abs{f_{2}[i](x,y,t) -  \left(\frac{\hat{\sigma}_{t}}{\hat{\alpha}_{t}}\nabla h(x,y,t)\right)[i]}\lesssim BN^{-\beta}\log^{\frac{\textcolor{blue}{k_1}+1}{2}}{N},
\end{align*}
for any $x\in\R^{d_{x}}, y\in[0,1]^{d_{y}}$ and $t>0.$
\end{lemma}

\paragraph{Step 3: Approximate Diffused Local Polynomials and Algebraic Operators with Transformers.}

First, we apply the universal approximation theory of transformers to $f_1$ and $f_2$.
Second, 
we adopt a comparable approach to approximate the algebraic operators essential for the final score computation. 
Last, 
we introduce \cref{lemma:score_approx_trans_2} that outlines how these components fit into a single transformer architecture with a specified parameter configuration.

\begin{itemize}
    \item 
    \textbf{Step 3.1: Approximate the Diffused Local Polynomials $f_{1}$ and $f_{2}$.}

We invoke the universal approximation theorem of transformer \cref{thm:Transformer_as_universal_approximators}.
We utilize network consisting of one transformer block and one feed-forward layer (see \cref{fig:condition_DiT} and \cref{def:transformer_class}).
\begin{lemma}
[Approximate Scalar Polynomials with Transformers]
\label{lemma:Trans_Approx_Poly_2}
Assume \cref{assumption:conditional_density_function_assumption_1}.
Consider the diffused local polynomial $f_{1}$ in \cref{lemma:diffused_local_polynomials_2}.
For any $\epsilon>0$, there exists a transformer $\calT_{f_{1}}\in\calT_R^{\textcolor{blue}{h,s,r}}$, such that for any $x\in[-C_{x}\sqrt{\log{N}},C_{x}\sqrt{\log{N}}]^{d_{x}}, y\in[0,1]^{d_{y}}$ and $t\in[N^{-C_{\sigma}},C_{\alpha}\log{N}]$,
it holds 
\begin{align*}
\abs{f_{1}(x,y,t)-\calT_{f_{1}}(x,y,t)[d_{x}]}\leq\epsilon,
\end{align*}
The parameter bounds in the transformer network class follows \cref{lemma:Trans_Approx_Poly}.
\end{lemma}

\begin{proof}[Proof of \cref{lemma:Trans_Approx_Poly_2}]
The proof closely follows \cref{lemma:Trans_Approx_Poly}
\end{proof}

\begin{lemma}[Approximate Vector-Valued Polynomials with Transformers]
\label{lemma:Trans_Approx_Poly_Gradient_2}
Assume
\cref{assumption:conditional_density_function_assumption_1} and 
consider $f_{2}(x,y,t)\in\R^{d_{x}}$ in \cref{lemma:diffused_local_polynomials_grad_2}.
For any $\epsilon>0$, there exists a transformer $\calT_{f_{2}}\in\calT_R^{\textcolor{blue}{h,s,r}} $ such that 
\begin{align*}
\norm{f_{2}(x,y,t)-\calT_{f_{2}}}_{\infty}\leq\epsilon,
\end{align*}
for any $x\in[-C_{x}\sqrt{\log{N}},C_{x}\sqrt{\log{N}}]^{d_{x}}, y\in[0,1]^{d_{y}}$ and $t\in[N^{-C_{\sigma}},C_{\alpha}\log{N}]$. The parameter bounds in the transformer network class follows \cref{lemma:Trans_Approx_Poly}.
\end{lemma}

\begin{proof}[Proof of \cref{lemma:Trans_Approx_Poly_Gradient_2}]
The proof closely follows \cref{lemma:Trans_Approx_Poly_Gradient}
\end{proof}

\item 
\textbf{Step 3.2: Approximate Algebraic Operators with Transformers.}

Next,
we introduce lemmas regarding the function of time.
These are also key components to the proof of \cref{thm:Main_2}.

\begin{lemma}
[Approximation of $\alpha^{2}$ with Transformer]
\label{lemma:approx_alpha_square}
For $t\in[t_{0},T]$ with $t_{0}<1$,
there exists Transformer $\calT_{\alpha^{2}}(t)\in\calT_{R}^{\textcolor{blue}{h,s,r}}$ such that 
\begin{align*}
\abs{\calT_{\alpha^{2}}-\alpha^{2}}\leq\epsilon_{\hat{\alpha}}.
\end{align*}
The parameter bounds in the Transformer network class follow \cref{lemma:trans_approx_variance}.
\end{lemma}
\begin{proof}
The proof closely follows \cref{lemma:trans_approx_variance}.
\end{proof}

Also,
we approximate $\hat{\alpha}$ and $\hat{\sigma}_{t}$ as well.
\begin{lemma}
[Approximation of $\hat{\alpha}$ with Transformer]
\label{lemma:approx_hat_alpha}
Consider $\hat{\alpha}_{t}=\frac{\alpha_{t}}{\alpha_{t}^{2}+C_{2}\sigma_{t}^{2}}$,
for $t\in[t_{0},T]$ with $t_{0}<1$,
there exists Transformer $\calT_{\hat{\alpha}}(t)\in\calT_{R}^{\textcolor{blue}{h,s,r}}$ such that 
\begin{align*}
\abs{\calT_{\hat{\alpha}}-\hat{\alpha}}\leq\epsilon_{\hat{\alpha}}.
\end{align*}
The parameter bounds in the transformer network class follow \cref{lemma:trans_approx_variance}.
\end{lemma}
\begin{proof}
The proof closely follows \cref{lemma:trans_approx_variance}.
\end{proof}

\begin{lemma}
[Approximation of $\hat{\sigma}$ with Transformer]
\label{lemma:approx_hat_sigma}
Consider $\hat{\sigma}_{t} = \frac{\sigma_{t}}{(\alpha_{t}^{2}+C_{2}\sigma_{t}^{2})^{1/2}}$,
for $t\in[t_{0},T]$ with $t_{0}<1$,
there exists Transformer $\calT_{\hat{\sigma}}(t)\in\calT_{R}^{\textcolor{blue}{h,s,r}}$ such that 
\begin{align*}
\abs{\calT_{\hat{\sigma}}-\hat{\sigma}}\leq\epsilon_{\hat{\sigma}}.
\end{align*}
The parameter bounds in the transformer network class follow \cref{lemma:trans_approx_variance}.
\end{lemma}
\begin{proof}
The proof closely follows \cref{lemma:trans_approx_variance}.
\end{proof}

We have finished establishing the approximation with transformer for every key component for the proof of \cref{thm:Main_2_informal}.

\item
\textbf{Step 3.3: Unified Transformer-Based Score Function Approximation.}

We introduce the counterpart of \cref{lemma:Score_Approx_Trans}.
It is the core of the proof for \cref{thm:Main_2}.

\begin{lemma}[Score Approximation with Transformer]
\label{lemma:score_approx_trans_2}
Assume \cref{assumption:conditional_density_function_assumption_2}.
For sufficiently large integer $N$,
there exists a mapping from transformer $\calT_{\text{score}}\in\calT_R^{\textcolor{blue}{h,s,r}}$ such that
\begin{align*}
\norm{\calT_{\text{score}}-\nabla\log{h(x,y,t)} + \frac{C_{2}x}{\alpha_{t}^{2} + C_{2}\sigma_{t}^{2}}}_{\infty}\leq\frac{B}{\sigma_{t}}N^{-\beta}(\log{N})^{\frac{\textcolor{blue}{k_1}+1}{2}},
\end{align*}
for any $x\in[-C_{x}\sqrt{\log{N}},C_{x}\sqrt{\log{N}}]^{d_x}$, $y\in[0,1]^{d_{y}}$ and $t\in[N^{-C_{\sigma}},C_{\alpha}\log{N}]$.

The parameter bounds in the transformer network class satisfy
\begin{align*}
& \norm{W_{Q}}_{2},\norm{W_{K}}_{2}, \norm{W_{Q}}_{2,\infty}, \norm{W_{K}}_{2,\infty}
=\calO\left(N^{(3\beta+9C_{\sigma})\frac{2dL+4d+1}{d}}\right);\\
&
\norm{W_{V}}_{2}=\calO(\sqrt{d}); \norm{W_{V}}_{2,\infty}=\calO(d); \norm{W_{O}}_{2},\norm{W_{O}}_{2,\infty}=\calO\left(N^{-\beta}\right);\\  
& \norm{W_{1}}_{2}, \norm{W_{1}}_{2,\infty}
=
\calO\left(N^{4\beta+9C_{\sigma}+\frac{3C_{\alpha}}{2}}\cdot\log{N}\right);
\norm{E^{\top}}_{2,\infty}=\calO\left(d^{\frac{1}{2}}L^{\frac{3}{2}}\right);\\
& 
\norm{W_2}_{2}, \norm{W_{2}}_{2,\infty}
=
\calO\left(N^{4\beta+9C_{\sigma}+\frac{3C_{\alpha}}{2}}\right);
C_{\calT}=\calO\left(\sqrt{\log{N}}/\sigma_{t}\right).
\end{align*}
\end{lemma}
\begin{proof}
Our proof follows the proof structure of  \cite[Proposition B.3]{fu2024unveil}.

Recall the decomposed score function presented in \textbf{Step 0},
we establish the the first-step approximator $f_{3}$ with the form:
\begin{align*}
f_{3}(x,y,t)\coloneqq\frac{\hat{\alpha}_{t}}{\hat{\sigma}_{t}}\cdot \frac{f_{2}(x,y,t)}{f_{1}(x,y,t)}-\frac{C_{2}x}{\alpha_{t}^{2} + C_{2}\sigma_{t}^{2}}.
\end{align*}
We derive the error bound on the approximation of the first term containing Taylor polynomials in $f_{3}$.
We incorporate second term containing the linear function in $x$ into the the transformer architecture.

We proceed as follows:
\begin{enumerate}
    \item \textbf{Step A:} Approximate $\nabla\log{p_{t}(x|y)}$ with $f_{3}$.

    \item \textbf{Step B:} Approximate $f_{3}$ with $\calT_{\text{score}}\in\calT_{R}^{\textcolor{blue}{h,s,r}}$.

    \item \textbf{Step C:} Derive the final Parameter Configuration
\end{enumerate}

\begin{itemize}
\item 
\textbf{Step A. Approximate Scroe Function with $f_{3}$.}

We first construct $f_{1}(x,y,t)$ and $f_{2}(x,y,t)$ from \cref{lemma:diffused_local_polynomials_2} and \cref{lemma:diffused_local_polynomials_grad_2} to approximate $h(x,y,t)$ and $\nabla h(x,y,t)$ respectively.

From \cref{lemma:bounds_on_h}, we have $C_{1} \leq h \leq B$ and $\left\| \frac{\hat{\sigma}_{t} \nabla h}{\hat{\alpha}_{t}} \right\|_{\infty} \leq \sqrt{\frac{2}{\pi}} B$.

Next, by \cref{lemma:diffused_local_polynomials_2} and \cref{lemma:diffused_local_polynomials_grad_2}, we select a sufficiently large $N$ such that $\frac{C_{1}}{2} \leq f_{1} \leq 2B$ and $f_{2} \leq B$.

Without loss of generality, we begin by bounding the first coordinate of $\nabla h$, denoted as $\nabla h[1]$:
\begin{align*}
\abs{\frac{\nabla h[1]}{h} - \frac{\hat{\alpha}_{t}}{\hat{\sigma}_{t}}\frac{f_{2}[1]}{f_{1}}}
\leq  
&~\abs{\frac{\nabla h[1]}{h} - \frac{\nabla h[1]]}{f_{1}}} + \abs{\frac{\nabla h[1]}{f_{1}} - \frac{\hat{\alpha}_{t}}{\hat{\sigma}_{t}}\frac{f_{2}[1]]}{f_{1}}},\\
\leq &~ 
\abs{\frac{\nabla h[1]]}{h\cdot f_{1}}}
\abs{f_1-h}
+
\frac{\hat{\alpha}_{t}}{\hat{\sigma}_{t}}\abs{\frac{1}{f_{1}}}
\abs{f_{2}-\frac{\hat{\sigma}_{t}}{\hat{\alpha}_{t}}\nabla h[1]]},\\
\lesssim &~
\frac{\hat{\alpha}_{t}}{\hat{\sigma}_{t}}\left(\abs{f_{1}-h} + \abs{f_{2}-\frac{\hat{\sigma}_{t}}{\hat{\alpha}_{t}}\nabla h[1]}\right),
\annot{By bounds on $h$, $\nabla h$, $f_{1}$, $f_{2}$}\\
\lesssim &~
\frac{\hat{\alpha}_{t}}{\hat{\sigma}_{t}}\left(BN^{-\beta}(\log{N}^{\frac{\textcolor{blue}{k_1}}{2}} + BN^{-\beta}(\log{N}^{\frac{\textcolor{blue}{k_1}+1}{2}})\right)
\annot{By \cref{lemma:diffused_local_polynomials_2} and \cref{lemma:diffused_local_polynomials_grad_2}},\\
\lesssim&~
\frac{1}{\sigma_{t}}\left(BN^{-\beta}(\log{N}^{\frac{\textcolor{blue}{k_1}+1}{2}})\right).
\end{align*}
Note that in the last line, we utilize
\begin{align*}
\frac{\hat{\alpha}_{t}}{\hat{\sigma}_{t}}
=
\frac{\alpha_{t}}{\sigma_
{t}}\frac{1}{\sqrt{\alpha_{t}^{2}+C_{2}\sigma_{t}^{2}}}
=
\frac{1}{\sigma_{t}}\frac{1}{\sqrt{1+C_{2}\left(\sigma_{t}/{\alpha_{t}}\right)^{2}}}
=
\frac{1}{\sigma_{t}}\frac{1}{\sqrt{1+C_{2}\frac{\sigma_{t}^{2}}{1-\sigma_{t}^{2}}}}
=
\calO(\sigma_{t}^{-1}).
\end{align*}

By the symmetry of each coordinate in $\nabla h$, we obtain the $\ell_{\infty}$ bounds:
\begin{align}
\label{eq:h_f2_f1}
\norm{\frac{\nabla h_(x,y,t)}{h(x,y,t)} - \frac{\hat{\alpha}_{t}}{\hat{\sigma}_{t}}\frac{f_{2}(x,y,t)}{f_{1}(x,y,t)}}_{\infty}\lesssim\frac{B}{\sigma_{t}}N^{-\beta}(\log{N})^{\frac{\textcolor{blue}{k_1}+1}{2}}.
\end{align}

\item     
\textbf{Step B. Approximate $f_{3}$ with Transformer $\calT_{\text{score}}$.}

Next,
we prove that there exist Transformer networks $\calT_{\text{score}}\in\calT_{R}^{\textcolor{blue}{h,s,r}}$ that approximates $f_{3}(x,y,t)$ with error of order $N^{-\beta}$.
We illustrate the overall approximation of $f_3$ in \cref{fig:error_main2}.

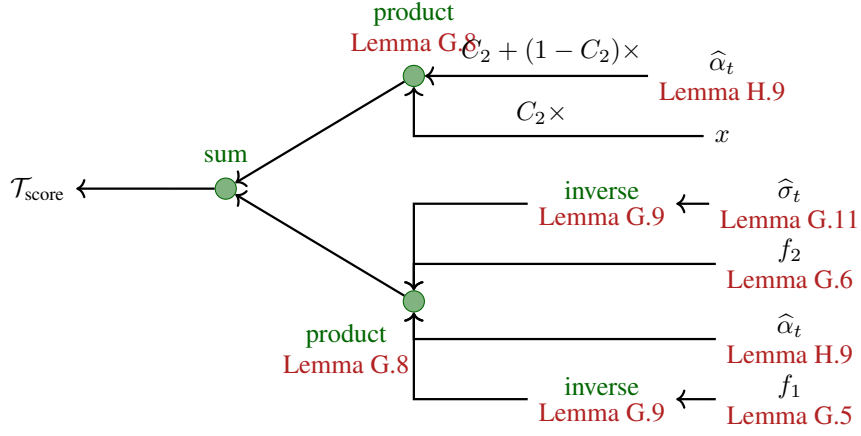
\begin{figure}
\centering
\definecolor{red}{rgb}{0.698,0.133,0.133}
\definecolor{green}{rgb}{0,0.40,0}
\definecolor{gray}{RGB}{169,169,169}

\begin{tikzpicture}[node distance=3.5cm and 2cm, auto, font=\footnotesize, scale=0.60]

\node (main) [below of=main1, yshift=0cm] {$\calT_{\text{score}}$};

\node (sum1) [right of=main, xshift=-1.0cm, yshift=0cm, draw=green, circle, fill=green!50, inner sep=0pt, minimum size=8pt] {};
\node at (sum1.north) [above, xshift=0.0cm, yshift=0.05cm, text=green] {sum};

\node (prod1) [right of=sum1, xshift=-1.0cm, yshift=1.5cm, draw=green, circle, fill=green!50, inner sep=0pt, minimum size=8pt] {};
\node at (prod1.north) [above, xshift=-0.0cm, yshift=0.03cm, text=green] {\shortstack{product \\\cref{lemma:approx_prod_with_trans}}};

\node (prod2) [right of=sum1, xshift=-1.0cm, yshift=-1.5cm, draw=green, circle, fill=green!50, inner sep=0pt, minimum size=8pt] {};
\node at (prod2.north) [below, xshift=-0.9cm, yshift=-0.3cm, text=green] {\shortstack{product \\\cref{lemma:approx_prod_with_trans}}};

\node (x) [right of=prod1,xshift=0.6cm,yshift=-0.8cm] {\shortstack{$x$}};

\node (alpha1) [right of=prod1,xshift=0.6cm,yshift=-0.0cm] {\shortstack{$\hat{\alpha}_t$\\ \cref{lemma:approx_hat_alpha}}};

\node (inv1) [right of=prod2, xshift=-1.0cm, yshift=1.3cm, color=green] { \shortstack{inverse \\ \cref{lemma:inverse_trans} }};

\node (inv2) [right of=prod2, xshift=-1.0cm, yshift=-1.3cm, color=green] { \shortstack{inverse \\ \cref{lemma:inverse_trans} }};

\node (sig1) [right of=inv1,xshift=-1.0cm,yshift=0.0cm] {\shortstack{$\hat{\sigma}_t$\\ \cref{lemma:trans_approx_variance}}};

\node (f1) [right of=inv2, xshift=-1.0cm, yshift=0.0cm] {\shortstack{$f_1$ \\ \cref{lemma:Trans_Approx_Poly}}};

\node (alpha2) [below of=sig1,xshift=+0.0cm,yshift=+1.7cm] {\shortstack{$\hat{\alpha}_t$\\ \cref{lemma:approx_hat_alpha}}};

\node (f2) [above of=f1, xshift=+0.0cm, yshift=-1.7cm] {\shortstack{$f_2$ \\ \cref{lemma:Trans_Approx_Poly_Gradient}}};

\draw[->, line width=0.3mm] (sum1) -- (main) node[midway, above, yshift=0.0cm,xshift=0.0cm] {};

\draw[->, line width=0.3mm] (prod1) -- (sum1) node[midway, above, yshift=0.0cm,xshift=0.0cm] {};

\draw[->, line width=0.3mm] (prod2) -- (sum1) node[midway, above, yshift=0.0cm,xshift=0.0cm] {};

\draw[->, line width=0.3mm] (x) -| (prod1) node[midway, above, yshift=0.0cm,xshift=1.7cm] {$C_2 \times$};

\draw[->, line width=0.3mm] (alpha1) -- (prod1) node[midway, above, yshift=0.0cm,xshift=0.2cm, sloped] {$C_2 + (1-C_2)\times$};

\draw[->, line width=0.3mm] (inv1) -| (prod2) node[midway, above, yshift=0.0cm,xshift=0.0cm] {};

\draw[->, line width=0.3mm] (inv2) -| (prod2) node[midway, above, yshift=0.0cm,xshift=0.0cm] {};

\draw[->, line width=0.3mm] (sig1) -- (inv1) node[midway, above, yshift=0.0cm,xshift=0.0cm] {};

\draw[->, line width=0.3mm] (f1) -- (inv2) node[midway, above, yshift=0.0cm,xshift=0.0cm] {};

\draw[->, line width=0.3mm] (f2) -| (prod2) node[midway, above, yshift=0.0cm,xshift=0.0cm] {};

\draw[->, line width=0.3mm] (alpha2) -| (prod2) node[midway, above, yshift=0.0cm,xshift=0.0cm] {};

\end{tikzpicture}
\caption{\small
\textbf{Approximate Score Function under \cref{assumption:conditional_density_function_assumption_2} with Transformer $\calT_{\text{score}}$.} 
The construction of the final score function consists of the approximation of diffused local polynomials $f_1$ and $f_2$ with transformer and transformer-approximate operators.
We highlight the overall pipeline and related lemmas to ensemble the Transformer network.
}
\label{fig:error_main2}
\end{figure}

In the following, we construct a transformer approximating the two terms in $f_{3}$,
and incorporate the result into a unified network architecture.

\begin{itemize}
    \item 
\textbf{Step B.1: Approximation for $\frac{\hat{\alpha}_{t}f_{2}}{\hat{\sigma}_{t}f_{1}}$.}

We utilize $\calT_{f_{1}}$, $\calT_{f_{2}}$,  $\calT_{\hat{\alpha}}$ and $\calT_{\hat{\sigma}}$
in \cref{lemma:Trans_Approx_Poly}, \cref{lemma:Trans_Approx_Poly_Gradient}, \cref{lemma:approx_hat_alpha} and \cref{lemma:approx_hat_sigma} to approximate each one of the component.
This gives error $\epsilon_{f_{1}}$, $\epsilon_{f_{2}}$, $\epsilon_{\hat{\alpha}}$ and $\epsilon_{\hat{\sigma}}$ respectively.

Next we utilize $\calT_{\text{rec},2}$ and $\calT_{\text{rec},3}$ in \cref{lemma:inverse_trans} for the approximation of the inverse of $f_{1}$ and $\hat{\sigma}_{t}$.
This gives error
\begin{align*}
\abs{\calT_{\text{rec},2}-\frac{1}{f_{1}}}\leq
\epsilon_{\text{rec},2}+\frac{\abs{\calT_{f_{1}}-f_{1}}}{\epsilon_{\text{rec},2}^{2}}
\leq
\epsilon_{\text{rec},2}+\frac{\epsilon_{f_{1}}}{\epsilon_{\text{rec},2}^{2}},
\end{align*}
and
\begin{align*}
\abs{\calT_{\text{rec},3}-\frac{1}{\hat{\sigma}_{t}}}
\leq
\epsilon_{\text{rec},3}+\frac{\abs{\calT_{\hat{\sigma}}-\hat{\sigma}_{t}}}{\epsilon_{\text{rec},2}^{2}}
\leq
\epsilon_{\text{rec},3}+\frac{\epsilon_{\hat{\sigma}}}{\epsilon_{\text{rec},3}^{2}}.
\end{align*}

Next we utilize $\calT_{\text{mult},1}$ in  \cref{lemma:approx_prod_with_trans} for the approximation of the product of $f_{1}^{-1}$, $f_{2}$, $\hat{\alpha}_{t}$ and $\hat{\sigma}_{t}^{-1}$.
This gives error 
\begin{align*}
& ~  \abs{\calT_{\text{mult},1}-\frac{\hat{\alpha}_{t}f_{2}}{\hat{\sigma}_{t}f_{1}}}\\
\leq & ~ 
\epsilon_{\text{mult},1} + 4K_{4}^{3}\underbrace{\max\left(
\epsilon_{\text{rec},2}+\frac{\epsilon_{f_{1}}}{\epsilon_{\text{rec},2}^{2}}, \epsilon_{f_{2}},  \epsilon_{\hat{\alpha}}, \epsilon_{\text{rec},3}+\frac{\epsilon_{\hat{\sigma}}}{\epsilon_{\text{rec},3}^{2}}
\right) }_{\coloneqq \epsilon_2}
\coloneqq  
\epsilon_{\text{mult},1} + 4K_{4}^{3}\epsilon_{2},
\end{align*}     
and $K_{3}$ is a positive constant.

From \cref{lemma:approx_prod_with_trans}, we require $[-K_{4}, K_{4}]$ to cover the domain of $f_{1}^{-1}$, $f_{2}$, $\hat{\alpha}$, and $\hat{\sigma}_{t}$. 
Recall that we give the upper and lower bounds for $f_{1}^{-1}$ and $f_{2}$ in the beginning of \textbf{Step 1.}
Thus, we set $K_{4} = \max\left( \hat{\sigma}_{t}^{-1}, \hat{\alpha}_{t} \right)$.

To derive the asymptotic behavior of $K_{4}$, we set the positive constant $C_{2} = 2$ without loss of generality and note that the maximum occurs at $t = t_{0}$. 
We then expand $\hat{\sigma}_{t_{0}}$ and $\hat{\alpha}_{t_{0}}^{-1}$:
\begin{align*}
\hat{\sigma}_{t_{0}}
=
\left(\frac{1-\exp(-t)}{2-\exp(-t_{0})}\right)^{\frac{1}{2}}
=
\left(1-\frac{1}{2-\exp(-t_{0})}\right)^{\frac{1}{2}}
=
\calO\left(N^{-C_{\sigma}}\right).
\end{align*}
and 
\begin{align*}
\hat{\alpha}_{t_{0}}^{-1}
=
\left(\frac{2-\exp(-t_{0})}{\exp(-\frac{t}{2})}\right)
=
2\exp(\frac{t_{0}}{2})-\exp(-\frac{t_{0}}{2})
=
\calO\left(N^{-C_{\sigma}}\right).
\end{align*}
So we take $K_{4}=\calO(N^{C_{\sigma}})$.

\item 
\textbf{Step B.2: Approximation for $-C_{2}x/(\alpha_{t}^{2}+C_{2}\sigma_{t}^{2})$.}

We use $\alpha_{t}^{2}+\sigma_{t}^{2}=1$ to
rewrite $(\alpha_{t}^{2}+C_{2}\sigma_{t}^{2})^{-1}$ as $(C_{2}+(1-C_{2})\alpha_{t}^{2})^{-1}$.

We first utilize $\calT_{\alpha^{2}}$ in \cref{lemma:approx_alpha_square} for the approximation of $\alpha_{t}^{2}$.
This gives error $\epsilon_{\alpha^{2}}$.

Next,
we utilize $\calT_{\text{rec},1}$ in \cref{lemma:approx_prod_with_trans} for the approximation of the inverse of $\alpha_{t}^{2}$.

This gives error
\begin{align*}
\abs{\calT_{\text{rec},1}-\frac{1}{\alpha_{t}^{2}}}
\leq
\epsilon_{\text{rec},1}+\frac{\abs{\calT_{\alpha_{t}^{2}}-\alpha_{t}^{2}}}{\epsilon_{\text{rec},1}^{2}}
\leq
\epsilon_{\text{rec},1}+\frac{\epsilon_{\alpha^{2}}}{\epsilon_{\text{rec},1}^{2}}.
\end{align*}

Next,
we utilize $\calT_{\text{mult},2}$ for the approximation of the product of $(C_{2}+(1-C_{2})\alpha_{t}^{2})^{-1}$ and $x$.
     
This gives error
\begin{align*}
\abs{\calT_{\text{mult},2}-\left(\frac{x}{C_{2}+(1-C_{2})\alpha_{t}^{2}}\right)}
\leq
\epsilon_{\text{mult},2} + 2K_{3}\left(\epsilon_{\text{rec},1}+\frac{\epsilon_{\alpha^{2}}}{\epsilon_{\text{rec},1}^{2}}\right),
\end{align*}
and from \cref{lemma:approx_prod_with_trans},
$K_{3}$ is positive constant 
such that $x\in[-K_{3},K_{3}]$ and $\alpha_{t}^{-1}\in[-K_{3},K_{3}]$.
Since $x\in[-C_{x}\sqrt{\log{N}},C_{x}\sqrt{\log{N}}]$ and $\alpha_{T}^{-1}=(\exp(-C_{\alpha}\log{N}/2))^{-1}=N^{C_{\alpha}/2}$,
we take $K_{3}=N^{C_{\alpha}/2}$.

\item
\textbf{Step B.3: Error Bound on Every Approximation Combined.}

Combining \textbf{Step B.1} and \textbf{Step B.2},
we obtain the total network with error bounded by
\begin{align*}
\epsilon_{\text{score}}\leq\epsilon_{\text{mult},2} + 2K_{3}\left(\epsilon_{\text{rec},1}+\frac{\epsilon_{\alpha^{2}}}{\epsilon_{\text{rec},1}^{2}}\right) + \epsilon_{\text{mult},1} + 4K_{4}^{3}\epsilon_{2}.
\end{align*}

Next,
we specify on the choice of $\epsilon$ in each approximation to attain a final approximation error of order $N^{-\beta}$.

\begin{itemize}
\item \textbf{For the Error of the First Inverse Operator:}
\begin{align*}
\epsilon_{\text{rec},1}
=\calO\left(N^{-(\beta+\frac{1}{2}C_{\alpha})}\right).
\end{align*}

\item \textbf{For the Error of the Second and Third Inverse Operator:}
\begin{align*}
\epsilon_{\text{rec},2}, \epsilon_{\text{rec},3}
=\calO\left(N^{-(\beta+3C_{\sigma})}\right).
\end{align*}

\item \textbf{For the Error of $f_{1}$:}
\begin{align*}
\epsilon_{f_{1}}=\calO\left(N^{-(3\beta+9C_{\sigma})}\right).
\end{align*}

\item \textbf{For the Error of $f_{2}$:}
\begin{align*}
\epsilon_{f_{2}}=\calO\left(N^{-(\beta+3C_{\sigma})}\right).
\end{align*}

\item \textbf{For the Error of $\hat{\sigma}$:}
\begin{align*}
\epsilon_{\hat{\sigma}}=\calO\left(N^{-(3\beta+9C_{\sigma})}\right).
\end{align*}

\item \textbf{For the Error of $\hat{\alpha}$:}
\begin{align*}
\epsilon_{\hat{\alpha}}=\calO\left(N^{-(\beta+3C_{\sigma)}}\right).
\end{align*}

\item \textbf{For the Error of $\alpha^{2}$:}
\begin{align*}
\epsilon_{\alpha^{2}}=\calO\left(N^{-(3\beta+\frac{3}{2}C_{\alpha})}\right).
\end{align*}

\item \textbf{For the Error of the Two Product Operators:}
\begin{align*}
\epsilon_{\text{mult},1}, \epsilon_{\text{mult},2} = \calO(N^{-\beta}).
\end{align*}
\end{itemize}
\end{itemize}

With above error choice,
we have
\begin{align}
\label{eq:fscore_f3_2}
\abs{\calT_{\text{score}}(x,y,t) - f_{3}(x,y,t)}\leq N^{-\beta}.
\end{align}
Combining  \eqref{eq:h_f2_f1}, \eqref{eq:fscore_f3_2} and dropping lower order term,
we obtain
\begin{align*}
\norm{\calT_{\text{score}} - \nabla\log{p_{t}(x|y)}}_{\infty}\lesssim\frac{B}{\sigma_{t}}N^{-\beta}(\log{N})^{\frac{\textcolor{blue}{k_1}+1}{2}}.
\end{align*}

We have completed the first part of the proof.
Next, we select the parameter bounds based on all the above approximations.
\end{itemize}

\item 
\textbf{Step C: Transformer Parameter Bound.}

Our result highlights the influence of $N$ under varying $d_x$.
Therefore,
for the transformer parameter bounds,
we keep terms with $d_x, d, L$ appearing in the exponent of $N$ and $\log{N}$.

\begin{itemize}
    \item \textbf{Parameter Bound on $W_{Q}$ and $W_{K}$.}

Given error $\epsilon$,
the bound on each operation follows:
\begin{itemize}
\item 
    \textbf{For $\epsilon_{f_1}$:}
    By \cref{lemma:Trans_Approx_Poly},
    we have
    \begin{align*}
    & ~ \norm{W_{Q}}_{2},\norm{W_{K}}_{2}, \norm{W_{Q}}_{2,\infty}, \norm{W_{K}}_{2,\infty}
    =\calO\left(N^{(3\beta+9C_{\sigma})\frac{2dL+4d+1}{d}}\right).
    \end{align*}

\item 
    \textbf{For $\epsilon_{f_2}$:}
    By \cref{lemma:Trans_Approx_Poly_Gradient},
    we have
    \begin{align*}
    & ~ \norm{W_{Q}}_{2},\norm{W_{K}}_{2}, \norm{W_{Q}}_{2,\infty}, \norm{W_{K}}_{2,\infty}
    =\calO\left(N^{(\beta+3C_{\sigma})\frac{2dL+4d+1}{d}}\right).
    \end{align*}

\item 
    \textbf{For $\epsilon_{\text{mult},1}$:}
    By \cref{lemma:approx_prod_with_trans}
    with $m=4$,
    we have
\begin{align*}
    & ~ \norm{W_{Q}}_{2},\norm{W_{K}}_{2},
    \norm{W_{Q}}_{2,\infty},\norm{W_{K}}_{2,\infty}
    =\calO\left(N^{9\beta}\right).
\end{align*}

\item 
    \textbf{For $\epsilon_{\text{mult},2}$:}
    By \cref{lemma:approx_prod_with_trans}
    with $m=2$,
    we have
\begin{align*}
    & ~ \norm{W_{Q}}_{2},\norm{W_{K}}_{2},
    \norm{W_{Q}}_{2,\infty},\norm{W_{K}}_{2,\infty}
    =\calO\left(N^{5\beta}\right).
\end{align*}

\item 
    \textbf{For $\epsilon_{\text{rec},1}$:}
    By \cref{lemma:inverse_trans},
    we have
    \begin{align*}
    & ~ \norm{W_{Q}}_{2},\norm{W_{K}}_{2}
    ,\norm{W_{Q}}_{2,\infty},
    \norm{W_{K}}_{2,\infty}
    =
    \calO\left(N^{3\beta+\frac{3C_{\alpha}}{2}}\right).
    \end{align*}

\item 
    \textbf{For $\epsilon_{\text{rec},2}$ and $\epsilon_{\text{rec},3}$:}
    By \cref{lemma:inverse_trans},
    we have
    \begin{align*}
    & ~ \norm{W_{Q}}_{2},\norm{W_{K}}_{2},
    \norm{W_{Q}}_{2,\infty},\norm{W_{K}}_{2,\infty}
    =\calO\left(N^{3\beta+9C_{\sigma}}\right).
    \end{align*}

\item 
    \textbf{For $\epsilon_{\hat{\alpha}}$:}
    By \cref{lemma:trans_approx_variance},
    we have
    \begin{align*}
    \norm{W_{Q}}_{2},\norm{W_{K}}_{2},\norm{W_{Q}}_{2,\infty},\norm{W_{Q}}_{2,\infty}
    =
    \calO\left(N^{3\beta+9C_{\sigma}}\right).
    \end{align*}
    
\item 
    \textbf{For $\epsilon_{\alpha^2}$:}
    By \cref{lemma:trans_approx_variance},
    we have
    \begin{align*}
    \norm{W_{Q}}_{2},\norm{W_{K}}_{2},\norm{W_{Q}}_{2,\infty},\norm{W_{Q}}_{2,\infty}
    =
    \calO\left(N^{9\beta+\frac{9C_{\alpha}}{2}}\right).
    \end{align*}

\item 
    \textbf{For $\epsilon_{\hat{\sigma}}$:}
    By \cref{lemma:trans_approx_variance},
    we have
    \begin{align*}
    \norm{W_{Q}}_{2},\norm{W_{K}}_{2},\norm{W_{Q}}_{2,\infty},\norm{W_{Q}}_{2,\infty}
    =
    \calO\left(N^{9\beta+27C_{\sigma}}\right).
    \end{align*}
\end{itemize}

We select the largest parameter bound from $\epsilon_{f_1}$ that remains valid across all other approximations.

\item \textbf{Parameter Bound on $W_{O}$ and $W_{V}$.}

Given error $\epsilon$,
the bound on each operation follows:
\begin{itemize}
\item 
    \textbf{For $\epsilon_{f_1}$:}
    By \cref{lemma:Trans_Approx_Poly},
    we have
\begin{align*}
    & ~ 
    \norm{W_O}_{2}, \norm{W_O}_{2,\infty}
    =\calO\left(N^{-\frac{(3\beta+9C_{\sigma})}{d}}\right).
\end{align*}

\item 
    \textbf{For $\epsilon_{f_2}$:}
    By \cref{lemma:Trans_Approx_Poly_Gradient},
    we have
    \begin{align*}
    & ~ 
    \norm{W_O}_{2},  \norm{W_O}_{2,\infty}
    =\calO\left(N^{-\frac{(\beta+3C_{\sigma})}{d}}
    \right).
    \end{align*}

\item 
    \textbf{For $\epsilon_{\text{mult},1}$:}
    By \cref{lemma:approx_prod_with_trans}
    with $m=4$,
    we have
\begin{align*}
    & ~ \norm{W_O}_{2},\norm{W_O}_{2,\infty}
    =\calO\left(N^{-4\beta}\right).
\end{align*}

\item 
    \textbf{For $\epsilon_{\text{mult},2}$:}
    By \cref{lemma:approx_prod_with_trans}
    with $m=2$,
    we have
\begin{align*}
    & ~ \norm{W_O}_{2},\norm{W_O}_{2,\infty}
    =\calO\left(N^{-2\beta}\right).
\end{align*}

\item 
    \textbf{For $\epsilon_{\text{rec},1}$:}
    By \cref{lemma:inverse_trans},
    we have
    \begin{align*}
    & ~ 
    \norm{W_O}_{2}, \norm{W_O}_{2,\infty}
    =\calO\left(N^{-(\beta+\frac{C_{\alpha}}{2})}\right).
    \end{align*}

\item 
    \textbf{For $\epsilon_{\text{rec},2}$ and $\epsilon_{\text{rec},3}$:}
    By \cref{lemma:inverse_trans},
    we have
    \begin{align*}
    & ~ 
    \norm{W_O}_{2}, \norm{W_O}_{2,\infty}
    =\calO\left(N^{-(\beta+3C_{\sigma})}\right).
    \end{align*}

\item 
    \textbf{For $\epsilon_{\hat{\alpha}}$:}
    By \cref{lemma:trans_approx_variance},
    we have
    \begin{align*}
    & ~ 
    \norm{W_O}_{2}, \norm{W_O}_{2,\infty}
    =\calO\left(N^{-(\beta+3C_{\sigma})}\right).
    \end{align*}

\item 
    \textbf{For $\epsilon_{\alpha^2}$:}
    By \cref{lemma:trans_approx_variance},
    we have
    \begin{align*}
    & ~ 
    \norm{W_O}_{2}, \norm{W_O}_{2,\infty}
    =\calO\left(N^{-(3\beta+\frac{3C_{\alpha}}{2})}\right).
    \end{align*}

\item 
    \textbf{For $\epsilon_{\hat{\sigma}}$:}
    By \cref{lemma:trans_approx_variance},
    we have
    \begin{align*}
    & ~ 
    \norm{W_O}_{2}, \norm{W_O}_{2,\infty}
    =\calO\left(N^{-(3\beta+9C_{\sigma})}\right).
    \end{align*}
\end{itemize}

Since we do not impose any relation on $C_{\sigma}$, $C_{\alpha}$ and $\beta$,
we simply take looser bound $\norm{W_{O}}_{2},\norm{W_{O}}_{2,\infty}=N^{-\beta}$.
Moreover,
since only $\epsilon_{f_{1}}$ and $\epsilon_{f_{2}}$ involve the reshape operation.
From \cref{lemma:trans_para_bound},
we take $\calO(\sqrt{d})$ and $\calO(d)$ $\norm{W_{V}}_{2}$ and $\norm{W_{V}}_{2,\infty}$.

\item \textbf{Parameter Bound for $W_{1}$.}

Given error $\epsilon$,
the bound on each operation follows:
\begin{itemize}
\item 
    \textbf{For $\epsilon_{f_1}$:}
    By \cref{lemma:Trans_Approx_Poly},
    we have
    \begin{align*}
    & ~ \norm{W_1}_{2}, \norm{W_{1}}_{2,\infty}
    =
    \calO\left(N^{\frac{(3\beta+9C_{\sigma})}{d}}\cdot\log{N}\right).
    \end{align*}

\item 
    \textbf{For $\epsilon_{f_2}$:}
    By \cref{lemma:Trans_Approx_Poly_Gradient},
    we have
    \begin{align*}
    & ~ \norm{W_1}_{2}, \norm{W_1}_{2,\infty}
    =
    \calO\left(N^{\frac{(\beta+3C_{\sigma})}{d}}\cdot\log{N}\right).
    \end{align*}
\item 
    \textbf{For $\epsilon_{\text{mult},1}$:}
    By \cref{lemma:approx_prod_with_trans} with $m=4$ and $C=K_4$ in \eqref{eqn:main_1_K2},
    we have
    \begin{align*}
    & ~ \norm{W_1}_{2},\norm{W_{1}}_{2,\infty}
    =
    \calO\left(K_4\cdot N^{4\beta}\right)
    =
    \calO\left(N^{(4\beta+C_{\sigma})}\right).
    \end{align*}

\item 
    \textbf{For $\epsilon_{\text{mult},2}$:}
    By \cref{lemma:approx_prod_with_trans} with $m=2$ and $C=K_3$ in \eqref{eqn:main_1_K1},
    we have
    \begin{align*}
    & ~ \norm{W_1}_{2},\norm{W_{1}}_{2,\infty}
    =
    \calO\left(K_3\cdot N^{2\beta}\right)
    =
    \calO\left(N^{(2\beta+\frac{C_{\alpha}}{2})}\right).
    \end{align*}

\item 
    \textbf{For $\epsilon_{\text{rec},1}$:}
    By \cref{lemma:inverse_trans},
    we have
    \begin{align*}
    & ~ \norm{W_1}_{2}, \norm{W_{1}}_{2,\infty}
    =
    \calO\left(N^{2\beta+C_{\alpha}}\right).
    \end{align*}

\item 
    \textbf{For $\epsilon_{\text{rec},2}$ and $\epsilon_{\text{rec},3}$:}
    By \cref{lemma:inverse_trans},
    we have
    \begin{align*}
    & ~ \norm{W_1}_{2}, \norm{W_{1}}_{2,\infty}
    =
    \calO\left(N^{(2\beta+6C_{\sigma})}\right).
    \end{align*}

\item 
    \textbf{For $\epsilon_{\hat{\alpha}}$:}
    By \cref{lemma:trans_approx_variance},
    we have
   \begin{align*}
    & ~ \norm{W_1}_{2}, \norm{W_{1}}_{2,\infty}
    =
    \calO\left(N^{(\beta+3C_{\sigma})}\cdot\log{N}\right).
    \end{align*}

\item 
    \textbf{For $\epsilon_{\alpha^2}$:}
    By \cref{lemma:trans_approx_variance},
    we have
   \begin{align*}
    & ~ \norm{W_1}_{2}, \norm{W_{1}}_{2,\infty}
    =
    \calO\left(N^{(3\beta+\frac{3C_{\alpha}}{2})}\cdot\log{N}\right).
    \end{align*}

\item 
    \textbf{For $\epsilon_{\hat{\sigma}}$:}
    By \cref{lemma:trans_approx_variance},
    we have
   \begin{align*}
    & ~ \norm{W_1}_{2}, \norm{W_{1}}_{2,\infty}
    =
    \calO\left(N^{(3\beta+9C_{\sigma})}\cdot\log{N}\right).
    \end{align*}
\end{itemize}

We select the largest parameter bound from $\epsilon_{f_1}$ that remains valid across all other approximations.

\item \textbf{Parameter Bound for $W_{2}$.}

Given error $\epsilon$,
the bound on each operation follows:
\begin{itemize}
\item 
    \textbf{For $\epsilon_{f_1}$:}
    By \cref{lemma:Trans_Approx_Poly},
    we have
    \begin{align*}
    & ~ \norm{W_2}_{2}, \norm{W_{2}}_{2,\infty}
    =
    \calO\left(N^{\frac{(3\beta+9C_{\sigma})}{d}}\right).
    \end{align*}

\item 
    \textbf{For $\epsilon_{f_2}$:}
    By \cref{lemma:Trans_Approx_Poly_Gradient},
    we have
    \begin{align*}
    & ~ \norm{W_2}_{2}, \norm{W_2}_{2,\infty}
    =
    \calO\left(N^{\frac{(\beta+3C_{\sigma})}{d}}\right).
    \end{align*}

\item 
    \textbf{For $\epsilon_{\text{mult},1}$:}
    By \cref{lemma:approx_prod_with_trans} with $m=4$,
    we have
   \begin{align*}
    & ~ \norm{W_2}_{2},\norm{W_2}_{2,\infty}
    =
    \calO\left(N^{4\beta}\right).
    \end{align*}

\item 
    \textbf{For $\epsilon_{\text{mult},2}$:}
    By \cref{lemma:approx_prod_with_trans} with $m=2$,
    we have
   \begin{align*}
    & ~ \norm{W_2}_{2},\norm{W_2}_{2,\infty}
    =
    \calO\left(N^{2\beta}\right).
    \end{align*}

\item 
    \textbf{For $\epsilon_{\text{rec},1}$:}
    By \cref{lemma:inverse_trans},
    we have
    \begin{align*}
    & ~ \norm{W_2}_{2}, \norm{W_2}_{2,\infty}
    =
    \calO\left(N^{(\beta+\frac{C_{\alpha}}{2})}\right).
    \end{align*}

\item 
    \textbf{For $\epsilon_{\text{rec},2}$ and $\epsilon_{\text{rec},3}$:}
    By \cref{lemma:inverse_trans},
    we have
    \begin{align*}
    & ~ \norm{W_2}_{2}, \norm{W_2}_{2,\infty}
    =
    \calO\left(N^{(\beta+3C_{\sigma})}\right).
    \end{align*}

\item 
    \textbf{For $\epsilon_{\hat{\alpha}}$:}
    By \cref{lemma:trans_approx_variance},
    we have
   \begin{align*}
    & ~ \norm{W_2}_{2}, \norm{W_2}_{2,\infty}
    =
    \calO\left(N^{(\beta+3C_{\sigma})}\right).
    \end{align*}

\item 
    \textbf{For $\epsilon_{\alpha^2}$:}
    By \cref{lemma:trans_approx_variance},
    we have
   \begin{align*}
    & ~ \norm{W_2}_{2}, \norm{W_2}_{2,\infty}
    =
    \calO\left(N^{(3\beta+\frac{3C_{\alpha}}{2})}\right).
    \end{align*}

\item 
    \textbf{For $\epsilon_{\hat{\sigma}}$:}
    By \cref{lemma:trans_approx_variance},
    we have
   \begin{align*}
    & ~ \norm{W_2}_{2}, \norm{W_2}_{2,\infty}
    =
    \calO\left(N^{(3\beta+9C_{\sigma})}\right).
    \end{align*}
\end{itemize}

We select the largest parameter bound from $\epsilon_{f_1}$ that remains valid across all other approximations.

\item \textbf{Parameter Bound for $E$.}

Since only $\epsilon_{f_{1}}$ and $\epsilon_{f_{2}}$ involve the reshape operation.
From \cref{lemma:trans_para_bound},
we take $\calO(d^{{1}/{2}}L^{{3}/{2}})$.
\end{itemize}

By integrating results above, 
we derive the following parameter bounds for the transformer network, ensuring valid approximation across all ten approximations.

\begin{align*}
& \norm{W_{Q}}_{2},\norm{W_{K}}_{2}, \norm{W_{Q}}_{2,\infty}, \norm{W_{K}}_{2,\infty}
=\calO\left(N^{(3\beta+9C_{\sigma})\frac{2dL+4d+1}{d}}\right);\\
&
\norm{W_{V}}_{2}=\calO(\sqrt{d}); \norm{W_{V}}_{2,\infty}=\calO(d); \norm{W_{O}}_{2},\norm{W_{O}}_{2,\infty}=\calO\left(N^{-\beta}\right);\\  
& \norm{W_{1}}_{2}, \norm{W_{1}}_{2,\infty}
=
\calO\left(N^{4\beta+9C_{\sigma}+\frac{3C_{\alpha}}{2}}\cdot\log{N}\right);
\norm{E^{\top}}_{2,\infty}=\calO\left(d^{\frac{1}{2}}L^{\frac{3}{2}}\right);\\
& 
\norm{W_2}_{2}, \norm{W_{2}}_{2,\infty}
=
\calO\left(N^{4\beta+9C_{\sigma}+\frac{3C_{\alpha}}{2}}\right);
C_{\calT}=\calO\left(\sqrt{\log{N}}/\sigma_{t}\right).
\end{align*}
This completes the proof.
\end{proof}
\end{itemize}

\clearpage
\subsection{Main Proof of \texorpdfstring{\cref{thm:Main_2_informal}}{}}
\label{sec:op_approx}

We state the formal version of \cref{thm:Main_2_informal}.

Next,
similar to the proof of \cref{thm:Main_1},
we need the truncation of $x$ due to the unboundedness as well.
\begin{lemma}[Truncate $x$, Lemma B.2 of \cite{fu2024unveil}.]
\label{lemma:truncate_x_2}
Assume \cref{assumption:conditional_density_function_assumption_2}.
For any $R_{3}>1,$ 
we have:
\begin{align*}
\int_{\norm{x}_{\infty}\geq R_{3}}p_{t}(x|y) \dd x\lesssim R_{3}\exp(-C_{2}^{\prime}R_{2}^{2}).
\end{align*}
\begin{align*}
\int_{\norm{x}_{\infty}\geq R_{3}}\norm{\nabla\log{p_{t}(x|y)}}_{2}^{2}p_{t}(x|y) \dd x\lesssim R_{3}\exp(-C_{2}^{\prime}R_{3}^{2})\lesssim\frac{1}{\sigma_{t}^{2}}R_{3}^{3}\exp(-C_{2}^{\prime}R_{3}^{2}),
\end{align*}
where $C_{2}^{\prime}=C_{2}/(2\max(1,C_{2}))$.
\end{lemma}
Again,
unlike result under \cref{assumption:conditional_density_function_assumption_1},
the explicit form of $p_{t}(x|y)$ in \eqref{eqn:assump2_first_decomp} and the upper and the lower bound of the joint distribution \cref{lemma:bounds_on_h} automatically allow us to skip the threshold $\epsilon_{\text{low}}$ as in \cref{lemma:truncate_p_x_cond_y}.

\begin{theorem}[Approximation Score Function with Transformer under Stronger H\"older Assumption (Formal Version of \cref{thm:Main_2_informal})]
\label{thm:Main_2}
Assume \cref{assumption:conditional_density_function_assumption_2} \blue{and $d_x=\Omega( \frac{\log N}{\log \log N})$}.
For any precision parameter $0 < \epsilon < 1$ and smoothness parameter $\beta > 0$, let $\epsilon \le \calO(N^{-\beta})$ for some $N \in \mathbb{N}$.
For some positive constants $C_{\alpha},C_{\sigma}>0$, for any $y \in [0,1]^{d_{y}}$ and $t \in [N^{-C_{\sigma}}, C_{\alpha} \log N]$, there exists a $\calT_{\text{score}}(x,y,t)\in\calT_R^{\textcolor{blue}{h,s,r}}$ such that the conditional score approximation satisfies
\begin{align*}
\int_{\R^{d_x}}\norm{\calT_{\text{score}}(x,y,t)-\nabla\log{p_{t}(x|y)}}_{2}^{2}\cdot p_{t}(x|y) \dd x = \calO\left(\frac{B^{2}}{\sigma_{t}^{2}}\cdot N^{-\frac{2\beta}{d_{x}+d_{y}}}\cdot(\log{N})^{{\beta}+1}\right).
\end{align*}
Notably, for $\epsilon=\calO(N^{-\beta})$, the approximation error has the upper bound 
$\tilde{\calO}( \epsilon^{2/(d_{x} + d_{y})}/\sigma_{t}^{2} )$.

The parameter bounds in the transformer network class satisfy
\begin{align*}
& \norm{W_{Q}}_{2},\norm{W_{K}}_{2}, \norm{W_{Q}}_{2,\infty}, \norm{W_{K}}_{2,\infty}
=\calO\left(N^{\frac{3\beta(2d_x+4d+1)}{d(d_x+d_y)}+\frac{9C_{\alpha}(2d_x+4d+1)}{d}}\right);\\
&
\norm{W_{V}}_{2}=\calO(\sqrt{d}); \norm{W_{V}}_{2,\infty}=\calO(d); \norm{W_{O}}_{2},\norm{W_{O}}_{2,\infty}=\calO\left(N^{-\frac{\beta}{d_x+d_y}}\right);\\  
& \norm{W_{1}}_{2}, \norm{W_{1}}_{2,\infty}
=
\calO\left(N^{\frac{4\beta}{d_x+d_y}+9C_{\sigma}+\frac{3C_{\alpha}}{2}}\cdot\log{N}\right);
\norm{E^{\top}}_{2,\infty}=\calO\left(d^{\frac{1}{2}}L^{\frac{3}{2}}\right);\\
& 
\norm{W_2}_{2}, \norm{W_{2}}_{2,\infty}
=
\calO\left(N^{\frac{4\beta}{d_x+d_y}+9C_{\sigma}+\frac{3C_{\alpha}}{2}}\right);
C_{\calT}=\calO\left(\sqrt{\log{N}}/\sigma_{t}\right).
\end{align*}
\end{theorem}

\begin{proof}[Proof Sketch of \cref{thm:Main_2}]
We decompose the integral into two terms based on \cref{lemma:truncate_x_2}.
\begin{itemize}
    \item \textbf{(A.1): The approximation for region outside of the truncation $\norm{x}>R_{3}$:} 
    
    We give the error bound via \cref{lemma:truncate_x_2}.
    
    \item \textbf{(A.2): The approximation for region within the truncation $\norm{x}_{\infty}\leq R_{3}$:} 
    
    We give the error bound via \cref{lemma:score_approx_trans_2}.
\end{itemize}
\end{proof}

\begin{proof}[Proof of \cref{thm:Main_2_informal}]
For simplicity, 
we change the variable $N$ to $N^{\frac{1}{d_{x}+d_{y}}}$ in the following subsection.
We put the original form back at the end of the proof.

We take $C_{x}=\sqrt{\frac{2\beta}{C_{2}^{\prime}}}$ in \cref{lemma:score_approx_trans_2} and $R_{3}=C_{x}\sqrt{\log{N}}$ in \cref{lemma:truncate_x_2}.

With the transformer parameter bounds in \cref{lemma:score_approx_trans_2},
we have $\norm{\calT_{\text{score}}}_{2}\leq\sqrt{\log{N}}/\sigma_{t}$ for any $x\in\R^{d_{x}}$, $y\in\R^{d_{y}}$ and $t>0$.
We start with the truncation on $x$
\begin{align*}
&~\int_{\R^{d_{x}}}\norm{\calT_{\text{score}}-\nabla\log{p_{t}}}_{2}^{2}p_{t}\dd x
\\ \annot{By expanding $\ell_{2}$ norm}
\leq&~
\underbrace{\int_{\norm{x}_{\infty}>\sqrt{\frac{2\beta}{C_{2}^{\prime}}\log{N}}}\left(2\norm{\calT_{\text{score}}}_{2}^{2} + 2\norm{\nabla\log{p_{t}}}_{2}^{2}\right)p_{t}\dd x}_{(A.1)}
+
\underbrace{\int_{\norm{x}_{\infty}\leq\sqrt{\frac{2\beta}{C_{2}^{\prime}}\log{N}}}\left(\norm{\calT_{\text{score}}-\nabla\log{p_{t}}}_{2}^{2}\right)p_{t}\dd x}_{A.2}
\\ \annot{By $\ell_{2}$ bound on $\calT_{\text{score}}$ and \cref{lemma:score_approx_trans_2}}
\lesssim&~
\int_{\norm{x}_{\infty}>\sqrt{\frac{2\beta}{C_{2}^{\prime}}\log{N}}}\left(2\left(\frac{\sqrt{\log{N}}}{\sigma_{t}}\right)^{2} + 2\norm{\nabla\log{p_{t}}}_{2}^{2}\right)p_{t}\dd x
+\frac{B^{2}}{\sigma_{t}^{2}}N^{-2\beta}(\log{N})^{\textcolor{blue}{k_1}+1}
\\ \annot{By \cref{lemma:truncate_x_2}}
\lesssim&~
2d_{x}\frac{\sqrt{\log{N}}}{\sigma_{t}^{2}}\left({\frac{2\beta}{C_{2}^{\prime}}\log{N}}\right)^{\frac{1}{2}}N^{-2\beta} 
+
\frac{2}{\sigma_{t}^{2}}\left(\frac{2\beta}{C_{2}^{\prime}}\log{N}\right)^{\frac{3}{2}}N^{-2\beta}
+
\frac{B^{2}}{\sigma_{t}^{2}}N^{-2\beta}(\log{N})^{\textcolor{blue}{k_1}+1}
\\ \annot{By dropping lower order term}
\lesssim&~
\frac{B^{2}}{\sigma_{t}^{2}}N^{-2\beta}(\log{N})^{\beta+1}.
\end{align*}

The transformer parameter norm bounds follow \cref{lemma:score_approx_trans_2},
with the replacement of $N$ with $N^{{1}/{d_{x}+d_{y}}}$.
This gives in $t\in[N^{-C_\alpha/(d_x+d_y)},C_\sigma(\log{N})^{1/(d_x+d_y)}]$.
For a better interpretation of the cutoff and early stopping time parameter,
we reset $C_\alpha=(d_x+d_y)C_\alpha$ and $C_\sigma=(d_x+d_y)C_\sigma$ such that $t\in[N^{-C_\alpha},C_\sigma\log{N}]$.

This completes the proof.
\end{proof}

\clearpage
\section{Proof of the Estimation Results for Conditional DiTs}

\textbf{Overview of Our Proof Strategy of \cref{thm:main_risk_bounds}.}

\begin{itemize}[leftmargin=3.4em]
    \item [\textbf{Step 0.}]
    \textbf{Preliminaries.}
    We introduce the mixed risk that accounts for risk with the distribution of the mask signal in \cref{def:mixed_risk}. 
    We restate the loss function and the score matching technique in \cref{def:score_matching_loss}.

    \item [\textbf{Step 1.}] \textbf{Truncate the Domain of the Risk.}
    We truncate the domain of the loss function in order to obtain finite covering number of transformer network class. 
    Precise definition of the truncated loss function class is in \cref{def:trunc_loss}.
    We bound the error from the truncation from the assumed light tail condition in
    \cref{lemma:error_trunc_risk}.
    
    \item [\textbf{Step 2.}] \textbf{Derive the Covering Number of Transformer Network.}
    We introduce the covering number of a given function class in \cref{def:covering_number}.
    We provide lemma detailing the calculation of the covering number for transformer architecture in \cref{lemma:covering_number}.
    We derive the covering numbers under the respective parameter configurations for our two previous main results in \cref{lemma:covering_number_truncated_loss}.

    \item [\textbf{Step 3.}] \textbf{Bound the True Risk on Truncated Domain.}
    With the previous steps,
    we present the upper-bound of the mixed risk in \cref{lemma:bound_on_trunc_ghost_vs_real}.
\end{itemize}

\textbf{Overview of Our Proof Strategy of \cref{thm:distribution_TV_bound}.}
We decompose the total variation into three components and we bound the separately.
\begin{itemize}[leftmargin=3.4em]

    \item [\textbf{Step 1.}]
    We bound the total variation distance between the true distributions evaluated at $t=0$ and early-stopping time $t=t_{0}$.
    
    \item [\textbf{Step 2.}] 
    We bound the total variation between the true distribution at $t_0$ and the reverse process distribution using the true score function.

    \item [\textbf{Step 3.}] 
    We bound the total variation between the reverse process distributions using the true and estimated score functions at $t_0$.
\end{itemize}

\paragraph{Organization.}
\cref{sec:aux_thm:tv_risk} includes auxiliary lemmas for supporting our proof of \cref{thm:main_risk_bounds}.
\cref{app:proof_main_risk_bounds} includes the main proof of \cref{thm:main_risk_bounds}. 
\cref{sec:aux_thm:tv_bound} includes auxiliary lemmas for supporting our proof of \cref{thm:distribution_TV_bound}.
\cref{app:proof_distribution_TV_bound} includes the main proof of \cref{thm:distribution_TV_bound}.

\subsection{Auxiliary Lemmas for \texorpdfstring{\cref{thm:main_risk_bounds}}{}}
\label{sec:aux_thm:tv_risk}

\paragraph{Step 0: Preliminary Framework.}

We evaluate the quality of the estimator $s_W$ through the risk:
\begin{align}
\label{eqn:conditional_risk}
\calR({s_W})\coloneqq\int_{t_{0}}^{T}\frac{1}{T-t_{0}}\E_{x_{t},y}\norm{s_W(x_{t},y,t)-\nabla\log{p_{t}(x_t|y)}}_{2}^{2}\dd t.
\end{align}

\begin{definition}[Mixed Risk]
\label{def:mixed_risk}
The risk \eqref{eqn:conditional_risk} considers guidance $y$ throughout whole the diffusion process.
We refer to it as the conditional score risk.
In contrast,
we have the mixed risk $\calR_{m}$ that accounts for the distribution of the mask signal $\tau=\{\emptyset,\text{id}\}$
with $P(\tau=\emptyset)=P(\tau=\mathrm{id})=0.5$:
\begin{align}
\label{mix_risk}
{\calR_m}(s_W)\coloneqq \int_{t_{0}}^{T}\frac{1}{T-t_{0}}\E_{(x_{t},y,\tau)}\left[\norm{s_W(x_{t},\tau y,t)-\nabla\log{p_{t}(x_{t}|\tau y)}}_{2}^{2}\right] \dd t,
\end{align}
\end{definition}

\begin{remark}
Given the score estimator $\hat{s}$ trained from the empirical loss \eqref{eqn:empirical_loss_app},
the conditional score risk is upper-bounded by twice of the mixed risk.
That is,
we have 
$\calR(\hat{s})\leq2{\calR_m}(\hat{s})$.
This follows from direct calculation:
\begin{align*}
{\calR_m}(\hat{s})
=
\frac{1}{2}\int_{t_{0}}^{T}\frac{1}{T-t_{0}}\E_{x_{t}}\left[\norm{\hat{s}(x_{t},\emptyset,t)-\nabla\log{p_{t}(x_{t})}}_{2}^{2}\right] \dd t
+
\frac{1}{2}\calR(\hat{s}).
\end{align*}
\end{remark}

\begin{definition}
[Loss Function and Score Matching]
\label{def:score_matching_loss}
Let $x=x_t|x_0$ denote the random variable following Gaussian distribution $N(\alpha_t x_0, \sigma_t^2 I_{d_x})$,
we define loss function and score matching loss:
\begin{align*}
\ell(x, y ; {s}_{W})
\coloneqq
\int_{T_0}^T \frac{1}{T-T_0} \EE_{\tau,x} \left[\left\|{s}_W (x_t, \tau y, t)-\nabla \log p_t\left(x_t | x_0 \right)\right\|_2^2\right]\dd t,
\end{align*}
\begin{align*}
\calL(s_W)\coloneqq \int_{t_{0}}^{T}\frac{1}{T-t_{0}}\E_{x_{0},y}\left[\E_{\tau,x}\left[\norm{s(x_{t},\tau y,t)-\nabla\log{p_t(x_{t}|x_{0})}}_{2}^{2}\right]\right] \dd t.
\end{align*}
\end{definition}

\begin{remark}
Given i.i.d samples $\{x_{0,i},y_i\}_{i=1}^{n}$,
we write $\ell(x_i,y_i;s_W)$ with the understanding that $x_i=x_t|x_{0,i}$.
When context is clear,
we use $\ell(x_i,y_i;s_W)$ and $\ell(x_{0,i},y_i;s_W)$; $\{x_{0,i},y_i\}_{i=1}^{n}$ and $\{x_{i},y_i\}_{i=1}^{n}$ interchangeably.
\end{remark}

\begin{remark}
By \cite{vincent2011connection},
$\calL(s_W)$ and $\calR_m(s_W)$ differ by a constant that is inconsequential to the minimization.
Therefore,
minimizing the mixed risk is equivalent to minimizing the score matching loss
\end{remark}

\begin{definition}
[Empirical Risk]
\label{def:empirical_risk}
Consider a score estimator $s_W\in\calT_R^{\textcolor{blue}{h,s,r}}$.
Recall the definition of empirical loss: $\hat{\calL}(s_W)=\sum_{i=1}^{n}\frac{1}{n}\ell(x_i,y_i;s_W)$.
Let $s^\circ\coloneqq\nabla\log{p_t(x|y)}$,
we define the empirical risk:
\begin{align*}
\hat{\calR}_m(s_W)\coloneqq\hat{\calL}(s_W)-\hat{\calL}(s^\circ)
=
\sum_{i=1}^{n}\frac{1}{n}\ell(x_i,y_i;s_W) - 
\sum_{i=1}^{n}\frac{1}{n}\ell(x_i,y_i;s^\circ).
\end{align*}
\end{definition}

\begin{remark}
The key distinction between $\calR_m$ and $\calL$ lies in their formulations. 
Specifically,
$\calR_m$ takes input $x_t$ and compares $s_W$ to the ground truth $\nabla \log p_t(x|y)$.
In contrast, 
the score matching loss $\calL$ provides an explicit calculation based on the sample. 
It averages the squared difference between $s_W$ and $\nabla \log p_t(x|x_0)$ over the sample and time interval.
\end{remark}

\begin{remark}
Observe that 
(I): $s^\circ = \nabla \log p_t(x|y)$ is the ground truth of score function with $\calR_m(s^\circ)=0$.
(II): By \cite{vincent2011connection},  $\calR_m$ and $\calL$ differ by a constant.
Based on these,
we define the empirical risk $\hat{\calR}_m$ using the score matching loss as an intermediary:
$
\calR_m(s_W) 
= 
\calR_m(s_W) - \calR_m(s^\circ)
=
\calL(s_W) - \calL(s^\circ)
$.
This establishes the empirical risk $\hat{\calR}_m$ as a practical approximation of the true risk difference $\calR_m(s_W) - \calR_m(s^\circ)$.
\end{remark}

\begin{remark}
For any score estimator $s_W\in\calT_R^{\textcolor{blue}{h,s,r}}$ obtained from the training with i.i.d samples $\{x_{i},y_i\}_{i=1}^{n}$,
it holds $\E_{\{x_{i},y_i\}_{i=1}^n}[\hat{\calR}_m(s_W)]=\calR_m(s_W)$.
This follows from direct calculation with \cref{def:empirical_risk} and the i.i.d assumption.
\end{remark}

\paragraph{Step 1:
Domain Truncation of the Risk.}
We define the loss function with truncated domain.
This is essential for obtaining finite covering number for transformer network class.

\begin{definition}[Truncated Loss]
\label{def:trunc_loss}
We define the truncated domain of the score function by $\calD\coloneqq[-R_{\calT},R_{\calT}]^{d_{x}}\times[0,1]^{d_{y}}\cup\emptyset$.
Given loss function $\ell(x,y;s_{W})$,
we define the truncated loss:
\begin{align}
\label{eqn:trunc_loss_function}
\ell^{\text{trunc}}(x,y;s_{W})\coloneqq\ell(x,y;s_{W})\mathbbm{1}\{\norm{x}_{\infty}\leq R_{\calT}\}.
\end{align}
Similarly,
we define $\calL^{\text{trunc}}(s_W)\coloneqq\calL(s_W)\mathbbm{1}\{\norm{x}_{\infty}\leq R_{\calT}\}$
,
$\calR_m^{\text{trunc}}(s_W)\coloneqq\calR_m(s_W)\mathbbm{1}\{\norm{x}_{\infty}\leq R_{\calT}\}$
and 
$\hat{\calR}_m^{\text{trunc}}(s_W)\coloneqq\hat{\calR}_m(s_W)\mathbbm{1}\{\norm{x}_{\infty}\leq R_{\calT}\}$.
We define the function class of the truncated loss by
\begin{align}
\label{eqn:domain_of_trunc_loss}
\calS(R_{\calT})\coloneqq\{\ell(\cdot,\cdot; s_{W}):\calD\rightarrow\R\mid s_{W}\in\calT_{R}^{\textcolor{blue}{h,s,r}}\}.
\end{align}
\end{definition}

Next,
we introduce the following lemma dealing with the error bound for the truncation of the loss.

\begin{lemma}
[Truncation Error, Lemma D.1 of \cite{fu2024unveil}]
\label{lemma:error_trunc_risk}
Consider the truncated loss $\ell^{\text{trunc}}(x,y;s_W)$ and $t\in[n^{-\calO(1)},\calO(\log{n})]$.
Under \cref{assumption:conditional_density_function_assumption_1},
we have $\abs{\ell(x,y;s_W)}\lesssim 1/t_0$.
Consider the parameter configuration in \cref{thm:Main_1},
it holds:
\begin{align*}
\E_{x,y}\left[\abs{\ell(x,y,t)-\ell^{\text{trunc}}(x,y,s)}\right]
\lesssim
\exp(-C_{2}R_{\calT}^{2})R_{\calT}\left(\frac{1}{t_0}\right).
\end{align*}
Moreover, 
under \cref{assumption:conditional_density_function_assumption_2},
we have $\abs{\ell(x,y;s_W)}\lesssim \log(1/t_0)$.
Consider the parameter configuration in \cref{thm:Main_2},
it holds:
\begin{align*}
\E_{x,y}\left[\abs{\ell(x,y,t)-\ell^{\text{trunc}}(x,y,s)}\right]
\lesssim
\exp(-C_{2}R_{\calT}^{2})R_{\calT}\log{\left(\frac{1}{t_0}\right)}.
\end{align*}
\end{lemma}

\paragraph{Step 2: Covering Number of Transformer Network Class.}

We begin with the definition.
\begin{definition}[Covering Number]
\label{def:covering_number}
    Given a function class $\calF$ and a data distribution $P$. 
    Sample n data points $\{X_i\}_{i=1}^{n}$ from $P$, then the covering number $\calN(\epsilon,\calF,\{X_i\}_{i=1}^{n},\norm{\cdot})$ is the smallest size of a collection (a cover) $\calC \in \calF$ such that for any $f \in \calF$, there exist $\hat{f} \in \calC$ satisfying 
    \begin{align*}
        \max_i \norm{f(X_i)-\hat{f}(X_i)} \leq \epsilon.
    \end{align*}
    Further, we define the covering number with respect to the data distribution as
    \begin{align*}
        \calN(\epsilon,\calF,\norm{\cdot}) = \sup_{\{X_i\}_{i=1}^{n} \sim P}\calN(\epsilon,\calF,\{X_i\}_{i=1}^{n},\norm{\cdot}).
    \end{align*}
\end{definition}

Next,
we introduce the following lemma that provides results for the calculation of the covering number for transformer networks. 

\begin{lemma}[Modified from Theorem A.17 of \citet{Benjamin2020transformer_covering_number}]
\label{lemma:covering_number}

\begin{align*}
\text{Let}\quad \calT_R^{\textcolor{blue}{h,s,r}}  & (C_{\calT}, C_{Q}^{2,\infty},           C_{Q}, 
        C_{K}^{2,\infty}, C_{K},
        C_{V}^{2,\infty}, C_{V},
        C_{O}^{2,\infty}, C_{O},
        C_E, 
        C_{f_1}^{2,\infty}, C_{f_1}, 
        C_{f_2}^{2,\infty}, C_{f_2}, 
        L_{\calT})~
\end{align*} 
represent the class of functions of one transformer block satisfying the norm bound for matrix and Lipsichitz property for feed-forward layers. Then for all data point $\norm{X}_{2,\infty}\leq R_{\calT}$ we have
\begin{align*}
    & \log\mathcal{N}(\epsilon_c,\calT_R^{\textcolor{blue}{h,s,r}},\norm{\cdot}_{2})\\
    \leq & \
    \frac{\log (nL_{\calT})}{\epsilon_c^2}\cdot 
    \left(
    \alpha^{\frac{2}{3}}\left(d^{\frac{2}{3}}\left(C_F^{2,\infty}\right)^{\frac{4}{3}} 
    +d^{\frac{2}{3}}\left(2(C_F)^2 C_{OV} C_{KQ}^{2,\infty}\right)^{\frac{2}{3}}
    +
    2\left((C_F)^2 C_{OV}^{2,\infty}\right)^{\frac{2}{3}}\right)
    \right)^3,
\end{align*}
where $\alpha \coloneqq (C_F)^2 C_{OV} (1+4C_{KQ})(R_{\calT}+C_E)$.
\end{lemma}

\begin{remark}
    We modify \cite[Theorem A.17]{Benjamin2020transformer_covering_number} in seven aspects:
    \begin{enumerate}
        \item We do not consider the last linear layer in the model: converting each column vector of the transformer output to a scalar. 
        Therefore, we ignore the item related to the last linear layer in \citet[Theorem A.17]{Benjamin2020transformer_covering_number}.
        \item We do not consider the normalization layer in our model.
        Because the normalization layer in the original proof only applies $\norm{\prod_{\rm norm}(X_1)-\prod_{\rm norm}(X_2)}_{2, \infty}\leq \norm{X_1 - X_2}_{2, \infty}$, ignoring this layer does not change the result.
        \item Our activation function is ${\rm ReLU}$, we replace the Lipschitz upper bound of the activate function by 1.
        \item We consider the positional encoding in our work, we need to replace the upper bound $R_{\calT}$ for the inputs with the upper bound $R_{\calT}+C_E$.
        Besides, for multi-layer transformer, the original conclusion in \citet[Theorem A.17]{Benjamin2020transformer_covering_number} considers the upper bound for the $2, \infty$-norm of inputs is 1, we add the upper bound for the inputs in \cref{lemma:covering_number}.
        \item We use the feed-forward layer, including two linear layers and a residual layer. 
        Thus, in \cref{lemma:covering_number}, we replace the original upper bound for the norm of the weight matrix with the upper bound for the norm of $I_d+W_2W_1$.
        In the following, we use $\calO$ to estimate the log-covering number, thus we ignore the item for $I_d$ here for convenience. 
        This is the same for the self-attention layer.
        \item We use multi-head attention, and we add the number of heads $\tau$ in our result, similar to \cite[Theorem A.12]{Benjamin2020transformer_covering_number}.
        \item 
        \blue{
        In our work, we use transformer $\mathcal{T}^{1,4,1}_R$, i.e., with $h=1$ head, $r=4$ MLP dimension, and $s=1$ hidden dimension, following the configuration for transformers' universality in \cref{thm:Transformer_as_universal_approximators} and \cref{cor:transformer_class}.
        We remark that this configuration is minimally sufficient to achieve DiTs' score approximation result \cref{thm:Main_1} but not necessary.
        More complex configurations can also achieve transformer universality, as reported in \cite{hu2024statistical,kajitsuka2023transformers,yun2019transformers}.
        }
    \end{enumerate}
\end{remark}
With \cref{lemma:covering_number},
we derive the covering number under transformer weights configuration in \cref{thm:Main_1}  and \cref{thm:Main_2}.

\begin{lemma}[Covering Number for $\calS(R_{\calT})$]
\label{lemma:covering_number_truncated_loss}
Given $\epsilon_{c}>0$ and consider $\norm{x}_{\infty}\leq R_{\calT}$.
With sample $\{x_{i},y_{i}\}_{i=1}^{n}$,
the $\epsilon_{c}$-covering number for $\calS(R_\calT)$ with respect to $\norm{\cdot}_{L_{\infty}}$ under the network configuration in \cref{thm:Main_1} satisfies
\begin{align*}
\log\mathcal{N}\left(\epsilon_c,\calS(R_{\calT}),\norm{\cdot}_\infty\right)
\lesssim
\frac{\log{n}}{\epsilon_{c}^{2}}N^{\nu_{1}}(\log{N})^{\nu_{2}}(R_{\calT})^{2},
\end{align*}
where $\nu_{1}=172\beta/(d_x+d_y)+104C_{\sigma}$
and
$\nu_{2}=12d_x+12\beta+2$.
Moreover,
under network configuration in \cref{thm:Main_2},
we have
\begin{align*}
\log\mathcal{N}\left(\epsilon_c,S(R_{\calT}),\norm{\cdot}_\infty\right)
\lesssim
\frac{\log{n}}{\epsilon_{c}^{2}}N^{\nu_{3}}(\log{N})^{10}(R_{\calT})^{2},
\end{align*}
where 
$\nu_{3}={48d\beta(L+2)(d_x+2d+1)}/({d_x+d_y})+144dC_{\sigma}(L+2)-8\beta$.
\end{lemma}
\begin{proof}

Applying \cref{lemma:covering_number},
we have
\begin{align}
\label{eqn:covering_eqn_first}
    &~ \log\mathcal{N}(\epsilon_c,\calT_R^{\textcolor{blue}{h,s,r}},\norm{\cdot}_{2}) \notag\\
    \leq  &~ 
    \frac{\log n}{\epsilon_c^2}\cdot 
    \alpha^{2}
    \Bigg(
    \underbrace{2\left((C_F)^2 C_{OV}^{2,\infty}\right)^{\frac{2}{3}}}_{\textbf{(I)}} 
    +
    \underbrace{(d^{\frac{2}{3}}\left(C_F^{2,\infty}\right)^{\frac{4}{3}}}_{\textbf{(II)}}
    +
    \underbrace{d^{\frac{2}{3}}\left(2(C_F)^2 C_{OV} C_{KQ}^{2,\infty}\right)^{\frac{2}{3}}}_{\textbf{(III)}}
    \Bigg)^3,
\end{align}
where $\alpha \coloneqq (C_F)^2 C_{OV} (1+4C_{KQ})(R_{\calT}+C_E)$.

Note that we drop $L_{\calT}$ because it is inconsequential under \cref{assumption:conditional_density_function_assumption_1,assumption:conditional_density_function_assumption_2}.

\begin{itemize}
    \item 
    \textbf{Step A: Covering Number for Transformer with Network Configuration in \cref{thm:Main_1} (under \cref{assumption:conditional_density_function_assumption_1}).}
    
Recall that from the network configuration in \cref{thm:Main_1}:
\begin{align*}
& \norm{W_{Q}}_{2}, \norm{W_{K}}_{2},
\norm{W_{Q}}_{2,\infty}, \norm{W_{K}}_{2,\infty}
=
\calO\left(N^{\frac{7\beta}{d_x+d_y}+6C_{\sigma}}\right); \\
&
\norm{W_{O}}_{2},\norm{W_{O}}_{2,\infty}=\calO\left(N^{-\frac{3\beta}{d_x+d_y}+6C_{\sigma}}(\log{N})^{3(d_x+\beta)}\right);\\
&
\norm{W_{V}}_{2}=\calO(\sqrt{d});
\quad \norm{W_{V}}_{2,\infty}=\calO (d);\\ 
& \norm{W_{1}}_{2}, \norm{W_{1}}_{2,\infty}
=
\calO\left(N^{\frac{2\beta}{d_x+d_y}+4C_{\sigma}}\right);
\norm{E^{\top}}_{2,\infty}=\calO\left(d^{\frac{1}{2}}L^{\frac{3}{2}}\right);
\\
& \norm{W_{2}}_{2}, \norm{W_{2}}_{2,\infty}
=
\calO\left(N^{\frac{3\beta}{d_x+d_y}+2C_{\sigma}}\right);
C_\calT=\calO\left(\sqrt{\log{N}}/\sigma_{t}^{2}\right).
\end{align*}
Note that $W_{K,Q}=W_{Q}W_{K}^{\top}$,
we take $\norm{W_{Q}}_{2,\infty}\cdot\norm{W_{K}}_{2,\infty}$ as the upper bound for $\norm{W_{KQ}}_{2,\infty}$.
Since $W_{Q}$, $W_{K}$ share identical upper-bound,
we calculate $(C_{K}^{2,\infty})^{4}$ for $(C_{K,Q}^{2,\infty})^{2}$.
Similarly we use $\norm{W_{O}}_{2,\infty}\cdot\norm{W_{V}}_{2,\infty}$ as the upper bound for $\norm{W_{OV}}_{2,\infty}$.
Moreover,
we take $C_{F}=\max\{C_{f_1},C_{f_2}\}$.
Since we do not impose any relation on $\beta$ and $C_{\sigma}$ here,
we take $N^{3\beta/(d_x+d_y)+4C_{\sigma}}$ such that the upper-bound holds for both $W_1$ and $W_2$.

Our result highlights the influence of $N$ under varying $d_x$.
Therefore,
for the transformer parameter bounds,
we keep terms with $d_x, d, L$ appearing in the exponent of $N$ and $\log{N}$.

Among three terms,
it is obvious that \textbf{(III)} dominates the other two.
so we begin with:
\begin{align*}
\textbf{(III)}
\lesssim
&~\left((C_{F})^{4}(C_{OV})^{2}(C_{KQ}^{2,\infty})^{2}\right)^{\frac{1}{3}}\\
\lesssim&~
\Bigg(
\underbrace{N^{\frac{12\beta}{d_{x}+d_{y}}+16C_{\sigma}}}_{(C_{F})^{4}}
\underbrace{N^{-\frac{6\beta}{d_{x}+d_{y}}+12C_{\sigma}}(\log{N})^{6(d_x+\beta)}}_{(C_{OV})^{2}}
\underbrace{N^{\frac{28\beta}{d_{x}+d_{y}}+24C_{\sigma}}}_{(C_{K}^{2,\infty})^{4}}\Bigg)^{\frac{1}{3}},\\
\lesssim&~
\left(N^{\frac{34\beta}{d_{x}+d_{y}}+52C_{\sigma}}(\log{N})^{6(d_x+\beta)}\right)^{\frac{1}{3}}.
\end{align*}

Recall
$\alpha \coloneqq (C_F)^2 C_{OV} (1+4C_{KQ})(R_{\calT}+C_E)$,
\begin{align*}
\alpha^{2}
\lesssim&~
(C_{F})^{4}(C_{OV})^{2}(C_{KQ})^{2}(R_{\calT}+C_{E})^{2},\\
\lesssim&~
\underbrace{N^{\frac{12\beta}{d_{x}+d_{y}}+16C_{\sigma}}}_{(C_{F})^{4}}
\underbrace{N^{-\frac{6\beta}{d_{x}+d_{y}}+12C_{\sigma}}(\log{N})^{6(d_x+\beta)}}_{(C_{OV})^{2}}
\underbrace{N^{\frac{28\beta}{d_{x}+d_{y}}+24C_{\sigma}}}_{{(C_{K}^{2,\infty})^{4}}}
\underbrace{R_{\calT}^{2}dL^{3}}_{(R_{\calT}^{2}C_{E}^{2})},
\\
\lesssim&~
\left(\underbrace{N^{\frac{34\beta}{d_{x}+d_{y}}+52C_{\sigma}}(\log{N})^{6(d_x+\beta)}}_{\textbf{(III)$^3$}}(R_{\calT})^{2}\right).
\end{align*}

Putting all together,
we obtain
\begin{align}
\label{eqn:covering_number_first_form_first_assump}
& ~
\log\mathcal{N}\left(\epsilon_c,\mathcal{T}_{R}^{h,s,r},\norm{\cdot}_2\right) 
\lesssim 
\frac{\log{n}}{\epsilon_{c}^{2}}N^{\frac{68\beta}{d_{x}+d_{y}}+104C_{\sigma}}(\log{N})^{12d_x+12\beta}(R_{\calT})^{2}.
\end{align}

\item 
\textbf{Step B: Covering Number for Transformer with Network Configuration in \cref{thm:Main_2} (under \cref{assumption:conditional_density_function_assumption_2})}.

Recall that from the network configuration in \cref{thm:Main_2}
\begin{align*}
& \norm{W_{Q}}_{2},\norm{W_{K}}_{2}, \norm{W_{Q}}_{2,\infty}, \norm{W_{K}}_{2,\infty}
=\calO\left(N^{\frac{3\beta(2d_x+4d+1)}{d(d_x+d_y)}+\frac{9C_{\alpha}(2d_x+4d+1)}{d}}\right);\\
&
\norm{W_{V}}_{2}=\calO(\sqrt{d}); \norm{W_{V}}_{2,\infty}=\calO(d); \norm{W_{O}}_{2},\norm{W_{O}}_{2,\infty}=\calO\left(N^{-\frac{\beta}{d_x+d_y}}\right);\\  
& \norm{W_{1}}_{2}, \norm{W_{1}}_{2,\infty}
=
\calO\left(N^{\frac{4\beta}{d_x+d_y}+9C_{\sigma}+\frac{3C_{\alpha}}{2}}\cdot\log{N}\right);
\norm{E^{\top}}_{2,\infty}=\calO\left(d^{\frac{1}{2}}L^{\frac{3}{2}}\right);\\
& 
\norm{W_2}_{2}, \norm{W_{2}}_{2,\infty}
=
\calO\left(N^{\frac{4\beta}{d_x+d_y}+9C_{\sigma}+\frac{3C_{\alpha}}{2}}\right);
C_{\calT}=\calO\left(\sqrt{\log{N}}/\sigma_{t}\right).
\end{align*}

We derive the covering number for result under second assumption by the same procedure.

Similar to previous step,
we bound \textbf{(III)} in \eqref{eqn:covering_eqn_first}.
First,
we calculate:
\begin{itemize}
    \item \textbf{Bound on $(C_{F})^{4}=(C_{f_1})^4$.}
    \begin{align*}
   (C_{f_1})^4\lesssim
    \calO\left(N^{\frac{16\beta}{d_x+d_y}+36C_{\sigma}+6C_{\alpha}}\cdot(\log{N})^4\right)
    \end{align*}
    
    \item \textbf{Bound on $(C_{K}^{2,\infty})^{4}$.}
    \begin{align*}
    (C_{K}^{2,\infty})^{4}\lesssim
    N^{{\frac{12\beta(2d_x+4d+1)}{d(d_x+d_y)}+\frac{36C_{\alpha}(2d_x+4d+1)}{d}}}
    \end{align*}
\end{itemize}

The upper-bound on $\textbf{(III)}$ follows:
\begin{align*}
\textbf{(III)}
\lesssim
&~\left(d^{2}(C_{f_{1}})^{4}(C_{OV})^{2}(C_{KQ}^{2,\infty})^{2}\right)^{\frac{1}{3}},\\
\lesssim
&~
\left(
\underbrace{N^{
{\frac{24\beta d_{x}+64\beta d+ 12\beta}{d(d_x+d_y)} + \frac{72C_{\alpha}d_x+150C_{\alpha}d + 36C_{\alpha}}{d}}
+36C_{\sigma}}(\log{N})^{4}}_{(C_{f_{1}})^{4}\cdot(C_{K}^{2,\infty})^{4}}
\underbrace{N^{-\frac{2\beta}{d_x+d_y}}}_{(C_{OV})^{2}}
\right)^{\frac{1}{3}}\\
&~
\left(
N^{
{\frac{24\beta d_{x}+62\beta d+ 12\beta}{d(d_x+d_y)} + \frac{72C_{\alpha}d_x+150C_{\alpha}d + 36C_{\alpha}}{d}}
+36C_{\sigma}}(\log{N})^{4}\right)
\end{align*}

Second we bound $\alpha$ in \eqref{eqn:covering_eqn_first}.

\begin{align*}
\alpha^{2}
\lesssim&~
(C_{f_{1}})^{4}(C_{OV})^{2}(C_{KQ})^{2}(R_{\calT}+C_{E})^{2}
\lesssim
\textbf{(III)}^3\cdot(R_{\calT})^{2}.
\end{align*}

Combining \textbf{(III)} and $\alpha^{2}$ for network configuration in \cref{thm:Main_2},
we obtain
\begin{align}
\label{eqn:covering_number_first_form_second_assump}
\log\mathcal{N}\left(\epsilon_c,\mathcal{T}_{R}^{h,s,r},\norm{\cdot}_2\right)
\lesssim
\frac{\log{n}}{\epsilon_{c}^{2}}N^{{\frac{4(12\beta d_{x}+31\beta d+ 6\beta)}{d(d_x+d_y)} + \frac{12(12C_{\alpha}d_x + 25C_{\alpha}\cdot d + 6C_{\alpha})}{d}}
+72C_{\sigma}}(\log{N})^{8}\cdot(R_{\calT})^{2}.
\end{align}

\item \textbf{Step C: Covering Number under Domain Truncation.}

We extend the result to the covering number for $\calS(R_{\calT})$ defined in \eqref{eqn:domain_of_trunc_loss}.

First note that we obtain the score estimator from $\calT_{2}$ by virtue of arranging $x,y,t$ into a row vector and treating them as a sequence for execution,
so we convert our $\ell_{2,\infty}$ case into $\ell_{\infty}$ as stated in \citet{fu2024unveil} without loss of generality.   

For two score estimator $s_{1}(x,y,t),s_{2}(x,y,t)\in\calT_R^{\textcolor{blue}{h,s,r}}$ such that $\norm{s_{1}-s_{2}}_{L{\infty},\calD}\leq\epsilon$,
Proof of lemma D.3 in \citet{fu2024unveil} shows the difference between the loss $\ell(\cdot,\cdot,s_{1})$ and $\ell(\cdot,\cdot,s_{2})$ in $L_{\infty}$ is bounded by
\begin{align}
\label{eqn:extend_loss_covering}
\abs{\ell(\cdot,\cdot,s_{1})-\ell(\cdot,\cdot,s_{2})}\lesssim\epsilon\log{N}.
\end{align}

Therefore,
by replacing $\epsilon_{c}$ with $\epsilon_{c}/\log{N}$ 
in \eqref{eqn:covering_number_first_form_first_assump} we obtain the log-covering number for transformer under \cref{assumption:conditional_density_function_assumption_1} 
\begin{align*}
\log\mathcal{N}\left(\epsilon_c,\calS(R_{\calT}),\norm{\cdot}_\infty\right)
\lesssim
&~
\frac{\log{n}}{\epsilon_{c}^{2}}N^{\frac{172\beta}{d_{x}+d_{y}}+104C_{\sigma}}(\log{N})^{12d_x+12\beta+2}(R_{\calT})^{2}\\
\coloneqq&~
\frac{\log{n}}{\epsilon_{c}^{2}}N^{\nu_{1}}(\log{N})^{\nu_{2}}(R_{\calT})^{2},
\end{align*}
where \blue{$\nu_{1}=68\beta/(d_x+d_y)+104C_{\sigma}$
and
$\nu_{2}=12d_x+12\beta+2$.}

Moreover,
by replacing $\epsilon_{c}$ with $\epsilon_{c}/\log{N}$ 
in \eqref{eqn:covering_number_first_form_second_assump}we obtain the log-covering number for transformer under \cref{assumption:conditional_density_function_assumption_2} 
\begin{align*}
\log\mathcal{N}\left(\epsilon_c,\calS(R_{\calT}),\norm{\cdot}_\infty\right)
=
\frac{\log{n}}{\epsilon_{c}^{2}}N^{\nu_{3}}(\log{N})^{10}(R_{\calT})^{2}.
\end{align*}
where  \blue{$\nu_{3}={\frac{4(12\beta d_{x}+31\beta d+ 6\beta)}{d(d_x+d_y)} + \frac{12(12C_{\alpha}d_x + 25C_{\alpha}\cdot d + 6C_{\alpha})}{d}}
+72C_{\sigma}$.}
\end{itemize}

This completes the proof.
\end{proof}

\paragraph{Step 3: Bound the True Risk on Truncated Domain.}

We begin with the definition.

\begin{definition}
\label{def:covering_number_omega}
Let $s^{\circ}\coloneqq \nabla\log{p_{t}(x|y)}$ denote the ground truth of score function for simplicity.
Given i.i.d samples $\{x_i,y_i\}_{i=1}^{n}$ and 
a score estimator $s_W\in\calT_R^{\textcolor{blue}{h,s,r}}$,
we define the difference function:
\begin{align*}
\Delta_n(s_W,s^\circ)
\coloneqq
\abs{
\E_{\{x_i,y_i\}_{i=1}^{n}}
\left[
\hat{\calR}_{m}^{\text{trunc}}(s_W)
-
\calR_m^{\text{trunc}}(s_W)
\right]
}.
\end{align*}
\end{definition}

\begin{remark}
Note that the difference function $\Delta_{n}(s_W,s^\circ)$ measures the expected difference between 
the truncated empirical risk  and the truncated mixed risk with respect to the training sample.
Since the true risk is unattainable,
we construct $\Delta_{n}(s_W,s^\circ)$ serving as an intermediate that allows us to derive the upper-bound on the mixed risk.
Surprisingly,
we are able to handle the upper-bound of the difference function,
presented in \cref{lemma:bound_on_trunc_ghost_vs_real}.
\end{remark}

\begin{definition}
\label{def:epsilon_c_covering}
Given the truncated loss function class $\calS(R_{\calT})$,
we define its $\epsilon_c$-covering with the minimum cardinality in the $L^\infty$ metric as $\calL_\calN\coloneqq\{\ell_1,\ell_2,\ldots,\ell_\calN\}$.
Moreover,
we define $\ell_J\in\calL_\calN$ with  random variable $J$.
By definition,
there exist $\ell_J\in\calL_\calN$
such that $\norm{\ell_{J}-\ell(x_i,y_i;s_W)}_\infty\leq\epsilon_c$.
\end{definition}

Note that \cref{lemma:covering_number_truncated_loss} provides the upper-bound on the $\epsilon_c$-covering number of $\calS(R_{\calT})$ for score estimator trained from transformer network class.
Next,
we bound the difference function.

\begin{lemma}[Bound on Difference Function]
\label{lemma:bound_on_trunc_ghost_vs_real}
Consider i.i.d training samples $\{x_{0,i},y_i\}_{i=1}^{n}$ and score estimator $\hat{s}$ from \eqref{eqn:empirical_loss}.
Under \cref{assumption:conditional_density_function_assumption_1} and parameter configuration in \cref{thm:Main_1},
it holds:
\begin{align*}
\Delta_n(\hat{s},s^\circ)
\lesssim
\E_{\{x_i,y_i\}_{i=1}^{n}}
\left[\hat{\calR}_m(\hat{s})\right] + \frac{1}{t_0}\left(R_{\calT}\exp(-C_{2}R_{\calT}^{2}) + \frac{1}{n}\log{\calN}\right) + 7\epsilon_c,
\end{align*}
where $\calN(\epsilon_c,\calT_R^{\textcolor{blue}{h,s,r}},\norm{\cdot}_2)$ is the covering number of transformer network class.
Moreover,
Under \cref{assumption:conditional_density_function_assumption_2} and parameter configuration in \cref{thm:Main_2},
it holds:
\begin{align*}
\Delta_n(\hat{s},s^\circ)
\lesssim
\E_{\{x_i,y_i\}_{i=1}^{n}}
\left[\hat{\calR}_m(\hat{s})\right] + \log{\frac{1}{t_0}}\left( R_{\calT}\exp(-C_{2}R_{\calT}^{2}) + \frac{1}{n}\log{\calN}\right) + 7\epsilon_c.
\end{align*}
\end{lemma}

\begin{proof}
In this proof,
we let $z_i\coloneqq(x_{0_i},y_i)$,
$\hat{\ell}(z_i)\coloneqq\ell^{\text{trunc}}(z_i;\hat{s})$
and
$\ell^{\circ}(z_i)\coloneqq\ell^{\text{trunc}}(z_i;s^\circ)$.
For simplicity,
we use $\kappa=1/t_0$ for the case in \cref{thm:Main_1} and $\kappa=\log(1/t_0)$ for the case in \cref{thm:Main_2}.

\begin{itemize}

\item 
\textbf{Step A: Rewrite the true risk.}

To derive the upper-bound of the true risk,
we introduce a different set of i.i.d samples $\{x_{0,i}^\prime,y_i^\prime\}_{i=1}^{n}$ independent of the training data drawn from the same distribution.

This allows us to rewrite the true risk as:
\begin{align}
\label{eqn:ghost_true_risk}
{\calR_m}(\hat{s})-{\calR_m}(s^{\circ})=\calL(\hat{s})-\calL(s^{\circ})=\E_{\{x_{i}^{\prime},y_{i}^{\prime}\}_{i=1}^{n}}\left[\frac{1}{n}\sum_{i=1}^{n}\left(\ell(x_{i}^{\prime},y_{i}^{\prime},\hat{s})-\ell(x_{i}^{\prime},y_{i}^{\prime},s^{\circ})\right)\right].
\end{align}

With \eqref{eqn:ghost_true_risk},
we rewrite the difference function:
\begin{align}
\label{phi_2_first}
\Delta_n(\hat{s},s^\circ)
=
\abs{\frac{1}{n}\E_{\{z_{i},z_{i}^{\prime}\}_{i=1}^{n}}\left[\sum_{i=1}^{n}\left(\left(\hat{\ell}(z_{i})-\ell^{\circ}(z_{i})\right) - \left(\hat{\ell}(z_{i}^{\prime})-\ell^{\circ}(z_{i}^{\prime})\right)\right)\right]}.
\end{align}

\item 
\textbf{Step B: Introduce the $\epsilon_c$-covering.}

Before further decomposing \eqref{phi_2_first},
we introduce three definitions.

\begin{itemize}
    \item $\omega_{J}(z)\coloneqq\ell_{J}(z)-\ell^{\circ}(z)$ and $\hat{\omega}(z)\coloneqq\hat{\ell}(z)-\ell^{\circ}(z)$.

    \item $\Omega\coloneqq\max\limits_{1\leq J\leq\calN}\abs{\sum\limits_{i=1}^{n}\frac{\omega_J(z_{i})-\omega_{J}(z_{i}^{\prime})}{h_{J}}}$.

    \item $h_{J}\coloneqq\max\{\calA,\sqrt{\E_{z^{\prime}}[\ell_{J}(z^{\prime})-\ell^{\circ}(z^{\prime})]}\}$ with constant $\calA$ to be chosen later.
\end{itemize}

With $h_{j}$, $\omega_{j}$ and $\Omega$,
we start bounding \eqref{phi_2_first} by writing
\begin{align}
\label{eqn:Delta_first_decomp}
\Delta_n(\hat{s},s^\circ)
= &~\abs{\frac{1}{n}\E_{\{z_{i},z_{i}^{\prime}\}_{i=1}^{n}}\left[\sum_{i=1}^{n}\left(\left(\hat{\ell}(z_{i})-\ell^{\circ}(z_{i})\right) - \left(\hat{\ell}(z_{i}^{\prime})-\ell^{\circ}(z_{i}^{\prime})\right)\right)\right]}
\nonumber\\
\annot{By Replacing $\hat{\ell}$ with $\ell_J$}
\leq &~ 
\abs{\frac{1}{n}\E_{\{z_{i},z_{i}^{\prime}\}_{i=1}^{n}}\left[\sum_{i=1}^{n}\left(\omega_{J}(z_{i})-\omega_{J}(z_{i}^{\prime})\right)\right]}+2\epsilon_{c}
\nonumber\\
\annot{By introducing $\Omega$ and $h_J$}
\leq &~
\frac{1}{n}\E_{\{z_{i},z_{i}^{\prime}\}_{i=1}^{n}}[h_{J}\Omega] +2\epsilon_{c} 
\nonumber\\
\annot{By Cauchy-Schwarz inequality }
\leq& ~ 
\frac{1}{n}\sqrt{\E_{\{z_{i},z_{i}^{\prime}\}_{i=1}^{n}}[h_{J}^{2}]\E_{\{z_{i},z_{i}^{\prime}\}_{i=1}^{n}}[\Omega^{2}]}+2\epsilon_{c} 
\nonumber\\
\annot{By AM-GM inequality}
\leq& ~ 
\frac{1}{n}\left(\frac{n}{2}\E_{\{z_{i},z_{i}^{\prime}\}_{i=1}^{n}}[h_{J}^{2}] + \frac{1}{2n}\E_{\{z_{i},z_{i}^{\prime}\}_{i=1}^{n}}[\Omega^{2}]\right)+2\epsilon_{c}
\nonumber\\
=& ~ 
\underbrace{\frac{1}{2}\E_{\{z_{i},z_{i}^{\prime}\}_{i=1}^{n}}[h_{J}^{2}]}_{\text{(I)}} 
+
\underbrace{\frac{1}{2n^{2}}\E_{\{z_{i},z_{i}^{\prime}\}_{i=1}^{n}}[\Omega^{2}]}_{\text{(II)}}+2\epsilon_{c}.
\end{align}

\begin{itemize}
\item \textbf{Step B.1: Bounding (I).}

By the definition of $h_{J}$,
    \begin{align}
    \label{eq:phi_2_A}
    \E_{\{z_{i},z_{i}^{\prime}\}_{i=1}^{n}}[h_{J}^{2}]
    \leq&~
    \calA^{2} 
    +
    \E_{\{z_{i},z_{i}^{\prime}\}_{i=1}^{n}}\left[\E_{z^{\prime}}[\omega_{J}^{2}(z)]\right]
    \nonumber\\
    \leq&~
    \calA^{2} + \E_{z^{\prime}}[\hat{\omega}^{2}(z^{\prime})]+ 2\epsilon_{c}
    \nonumber\\
    =&~
    \calA^{2} + \E_{\{z_i\}_{i=1}^{n}}
    \left[\calR_m^{\text{trunc}}(\hat{s})\right] + 2\epsilon_{c}.
    \end{align}

\item \textbf{Step B.2: Bounding (II).}

By \cref{lemma:error_trunc_risk},
we have $\abs{\ell(z;s_W)}\lesssim\kappa$,
and by the definition of $\Omega^{2}$,
we write
    \begin{align*}
    \E_{\{z_{i},z_{i}^{\prime}\}_{i=1}^{n}}\left[\sum\limits_{i=1}^{n}\left(\frac{\omega_{J}(z_{i})-\omega_{J}(z_{i}^{\prime})}{h_{J}}\right)^{2}\right]
    \leq&~
    \sum\limits_{i=1}^{n}\E_{\{z_{i},z_{i}^{\prime}\}_{i=1}^{n}}\left[\left(\frac{\omega_{J}(z_{i})}{h_{J}}\right)^{2}+\left(\frac{\omega_{J}(z_{J}^{\prime})}{h_{J}}\right)^{2}\right] \annot{By the independence between $z_i$ and $z_i^\prime$}
    \\
    \leq&~
    \kappa\sum\limits_{i=1}^{n}\E_{\{z_{i},z_{i}^{\prime}\}_{i=1}^{n}}\left[\frac{\omega_{J}^{2}(z_{i})}{h_{J}} + \frac{\omega_{J}^{2}(z_{i}^{\prime})}{h_{J}}\right]
    \\
    \leq&~ 2n\kappa.
    \end{align*}

From the following two facts 
\begin{itemize}
    \item (1) $\abs{\frac{\omega_{J}(z_{i})-\omega_{J}(z_{i}^{\prime})}{h_{J}}}\leq{\kappa}/{\calA}$ 

    \item (2) $\sum\limits_{i=1}^{n}\frac{\omega_{J}(z_{i})-\omega_{J}(z_{i}^{\prime})}{h_{J}}$ is centered
\end{itemize}
we further write 
\begin{align*}
P\left(\left(\sum\limits_{i=1}^{n}\frac{\omega_{J}(z_{i})-\omega_{J}(z_{i}^{\prime})}{h_{J}}\right)^{2}\geq\omega\right)
=
&~
2P\left(\left(\sum\limits_{i=1}^{n}\frac{\omega_{J}(z_{i})-\omega_{j}(z_{i}^{\prime})}{h_{j}}\right)\geq\sqrt{\omega}\right)
\annot{By Bernstein’s inequality}
\leq 
2\exp\left(-\frac{\omega/2}{\kappa\left(2n+\frac{\sqrt{\omega}}{3\calA}\right)}\right),
\end{align*}
for any $J$ and $\omega\geq0$.
Therefore,
we have
\begin{align*}
P\left(\Omega^{2}\geq\omega\right)
\leq
\sum\limits_{J=1}^{\calN}P\left(\left(\sum\limits_{i=1}^{n}\frac{\omega_{J}(z_{i})-\omega_{J}(z_{i}^{\prime})}{h_{J}}\right)^{2}\geq\omega\right)
\leq
2\calN\exp\left(-\frac{\omega/2}{\kappa\left(2n+\frac{\sqrt{\omega}}{3\calA}\right)}\right).
\end{align*}

For some $\omega_0>0$,
we bound $\Omega^{2}$ by 
\begin{align*}
\E_{\{z_{i},z_{i}^{n}\}_{i=1}^{n}}\left[\Omega^{2}\right]=
\annot{By integral identity}& ~ \int_{0}^{\omega_{0}}P\left(\Omega^{2}\geq\omega\right)\dd\omega + \int_{\omega_{0}}^{\infty}P\left(\Omega^{2}\geq\omega\right)\dd\omega,\\
\leq& ~ 
\omega_{0} + 
\int_{\omega_{0}}^{\infty}2\calN\exp\left(-\frac{\omega/2}{\kappa\left(2n+\frac{\sqrt{\omega}}{3\calA}\right)}\right)\dd\omega,\\
\leq& ~ 
\omega_{0} +
2\calN\int_{\omega_{0}}^{\infty}\left\{\exp\left(-\frac{\omega}{8n\kappa}\right) + \exp\left(-\frac{3\calA\sqrt{\omega}}{4\kappa}\right)\right\}\dd\omega,\\
\leq& ~ 
\omega_{0} + 2\calN\left\{8n\kappa\exp\left(-\frac{\omega_{0}}{8n\kappa}\right) + \left(\frac{8\kappa\sqrt{\omega_{0}}}{3\calA} + \frac{32\kappa}{9\calA^{2}}\right)\exp\left(-\frac{3\calA\sqrt{\omega_{0}}}{4\kappa}\right)\right\}.
\end{align*}

Taking $\calA=\sqrt{\omega_{0}}/6n$ and $\omega_{0}=8n\kappa\log{\calN}$,
we have
\begin{align}
\label{eqn:Omega}
\E_{\{z_{i},z_{i}^{n}\}_{i=1}^{n}}[\Omega^{2}]\leq n\kappa\log{\calN}.
\end{align}

\end{itemize}

\item \textbf{Step C: Altogether.}

Combining 
\eqref{eq:phi_2_A} and \eqref{eqn:Omega},
we obtain:
\begin{align*}
\Delta_n(\hat{s},s^\circ)
\leq& ~ 
\frac{1}{2}\E_{\{z_{i},z_{i}^{\prime}\}_{i=1}^{n}}[h_{J}^{2}]
+
{\frac{1}{2n^{2}}\E_{\{z_{i},z_{i}^{\prime}\}_{i=1}^{n}}[\Omega^{2}]} 
+
2\epsilon_{c}\\
\lesssim &~
\frac{1}{2}\E_{\{z_i\}_{i=1}^{n}}\left[\calR_m^{\text{trunc}}(\hat{s})\right] + \frac{\kappa}{2n}\log{\calN} + \frac{7}{2}\epsilon_c.
\end{align*}
\end{itemize}

Recall \cref{def:covering_number_omega} and multiply the above inequality with $2$,
we have 
\begin{align*}
\E_{\{z_i\}_{i=1}^{n}}\left[\calR_m^{\text{trunc}\hat{s}}\right]
\lesssim
2\E_{\{z_i\}_{i=1}^{n}}\left[\hat{\calR}_m^{\text{trunc}}(\hat{s})\right] + \frac{\kappa}{n}\log{\calN} + 7\epsilon_c.
\end{align*}

Therefore,
\begin{align*}
\Delta_n(\hat{s},s^\circ)
\lesssim&~
\E_{\{z_i\}_{i=1}^{n}}\left[\hat{\calR}_m^{\text{trunc}}(\hat{s})\right] +\frac{\kappa}{n}\log{\calN} + 7\epsilon_c
\annot{By \cref{lemma:error_trunc_risk}}\\
\lesssim&~
\E_{\{x_i,y_i\}_{i=1}^{n}}
\left[\hat{\calR}_m(\hat{s})\right] + \kappa\left(R_{\calT}\exp(-C_{2}R_{\calT}^{2}) + \frac{1}{n}\log{\calN}\right) + 7\epsilon_c,
\end{align*}
This completes the proof.
\end{proof}

\subsection{Proof of \texorpdfstring{\cref{thm:main_risk_bounds}}{}}
\label{app:proof_main_risk_bounds}

\begin{proof}[Proof of \cref{thm:main_risk_bounds}]

For simplicity,
we use $\kappa=1/t_0$ for the case in \cref{thm:Main_1} and $\kappa=\log(1/t_0)$ for the case in \cref{thm:Main_2}.
The proof proceeds through the following three steps.

\begin{itemize}

    \item 
    \textbf{Step A: Decompose the mixed risk.}

We denote the ground truth by $s^{\circ}(x,y,t)=\nabla\log{p_{t}(x|y)}$, 
and if $y=\emptyset$ we set $s^{\circ}(x,y,t)=p_{t}(x)$.

Recall from \cref{def:empirical_risk} and \cref{lemma:bound_on_trunc_ghost_vs_real},
by introducing a different set of i.i.ds samples $\{x_{i}^{\prime},y_{i}^{\prime}\}_{i=1}^{n}$ from the initial data distribution $P_{0}(x,y)$ independent of the training samples,
we rewrite the mixed risk:
\begin{align*}
{\calR_m}(\hat{s})
=
\E_{\{x_{i}^{\prime},y_{i}^{\prime}\}_{i=1}^{n}}\left[\frac{1}{n}\sum_{i=1}^{n}\left(\ell(x_{i}^{\prime},y_{i}^{\prime},\hat{s})-\ell(x_{i}^{\prime},y_{i}^{\prime},s^{\circ})\right)\right]
=
\E_{\{x_{i}^{\prime},y_{i}^{\prime}\}_{i=1}^{n}}\left[\hat{\calR}^\prime_{m}(\hat{s})\right],
\end{align*}
where we use $\hat{\calR}_m^\prime(\hat{s})$ to denote the empirical risk of the score estimator $\hat{s}$ trained from i.i.d samples $\{x_i^\prime,y_i^\prime\}_{i=1}^{n}$ .

This allows us to do the decomposition of $\E_{\{x_{i},y_{i}\}_{i=1}^{n}}[{\calR_m}(\hat{s})]$ as follows.
\begin{align*}
\E_{\{x_{i},y_{i}\}_{i=1}^{n}}[{\calR_m}(\hat{s})]=
& ~
\underbrace{
\E_{\{x_{i},y_{i}\}_{i=1}^{n}}
\left[\E_{\{x_{i}^{\prime},y_{i}^{\prime}\}_{i=1}^{n}}
\left[
\hat{\calR}_{m}^{\prime}(\hat{s})
-
\hat{\calR}_{m}^{\prime\,\text{trunc}}(\hat{s})\right]
\right]}
_{\textbf{(I)}}\\
&~ 
+
\underbrace{\E_{\{x_{i},y_{i}\}_{i=1}^{n}}
\left[
\E_{\{x_{i}^{\prime},y_{i}^{\prime}\}_{i=1}^{n}}
\left[
\hat{\calR}_{m}^{\prime\,\text{trunc}}(\hat{s})
-
\hat{\calR}_{m}^{\text{trunc}}(\hat{s})\right]
\right]}
_{\textbf{(II)}}\\
&~ 
+
\underbrace{\E_{\{x_{i},y_{i}\}_{i=1}^{n}}
\left[
\hat{\calR}_{m}^{\text{trunc}}(\hat{s})
-
\hat{\calR}_{m}(\hat{s})
\right]}
_{\textbf{(III)}} 
+ 
\underbrace{\E_{\{x_{i},y_{i}\}_{i=1}^{n}}
\left[
\hat{\calR}_{m}(\hat{s})
\right]}
_{\textbf{(IV)}}
\end{align*}

\item 
\textbf{Step B: Derive the Upper Bound.}

\begin{itemize}
\item 
\textbf{Step B.1: Bound Each Term.}

\begin{itemize}
    \item  
By \cref{lemma:error_trunc_risk},
we have both \textbf{(I)}, \textbf{(III)} $\lesssim\kappa\exp(-C_{2}R_{\calT}^{2})R_{\calT}.$

\item 
By \cref{lemma:bound_on_trunc_ghost_vs_real},
we have \textbf{(II)}
$\lesssim
\textbf{(IV)} + \kappa\left(R_{\calT}\exp(-C_{2}R_{\calT}^{2}) + \frac{1}{n}\log{\calN}\right) + 7\epsilon_c,$

\item 
By the following,
we have \textbf{(IV)}$\leq\min_{s_W\in\calT_R^{\textcolor{blue}{h,s,r}}}{\calR_m}(s)$.
\begin{align*}
\textbf{(IV)}
= 
\E_{\{z_{i}\}_{i=1}^{n}}\left[\hat{\calR}(\hat{s})\right]
\leq  
\E_{\{z_{i}\}_{i=1}^{n}}\left[\hat{{\calR}}_m(s)\right]={\calR_m}(s).
\end{align*}
The inequality holds because $\hat{s}$ is the minimizer of the empirical risk.
\end{itemize}

\item 
\textbf{Step B.2: Combine \textbf{(I)}, \textbf{(II)}, \textbf{(III)}, \textbf{(IV)}.}

Combining these results we obtain
\begin{align}
\label{eqn:risk_upper_bound_1}
\E_{\{x_i,y_i\}_{i=1}^{n}}[{\calR_m}(\hat{s})]
\leq
&~ 2\min\limits_{s_W\in{\calT}_R^{\textcolor{blue}{h,s,r}} }\int_{t_{0}}^{T}\frac{1}{T-t_{0}}\E_{x_{t},y,\tau}\left[\norm{s(x_{t},\tau y,t)-\nabla\log{p_{t}(x_{t}|\tau y)}}_{2}^{2}\right]\dd t
\nonumber\\
&~ +
\calO\left(\frac{\kappa}n\log{\calN}\right) + \calO(\exp(-C_{2}R_{\calT}^{2})\kappa) + \calO\left(\epsilon_{c}\right).
\end{align}

By taking $R_{\calT}=\sqrt{\frac{(C_{\sigma}+2\beta)\log{N}}{C_{2}(d_{x}+d_{y})}}$  along with the result in \cref{lemma:covering_number_truncated_loss},
we further write
\begin{align}
\label{eqn:final_risk}
\E_{\{x_i,y_i\}_{i=1}^{n}}[{\calR_m}(\hat{s})]
\leq
&~2\min_{s\in{\calT}_{R}^{\textcolor{blue}{h,s,r}}}\int_{t_{0}}^{T}\frac{1}{T-t_{0}}\E_{\tau,x_{t},y}\left[\norm{s(x,\tau y,t)-\nabla\log{p_{t}(x|y)}}_{2}^{2}\right]\dd t
\nonumber\\
&~
\calO\left(\frac{\kappa}{n}\log{\calN}\right)
+\calO\left(N^{-\frac{2\beta}{d_{x}+d_{y}}}\right)
+\calO\left(\epsilon_{c}\right).
\end{align}
where we invoke $\kappa\lesssim\frac{1}{t_{0}}=N^{C_{\sigma}}$ to obtain the second term on the RHS.
\end{itemize}

\textbf{Step C: Altogether.}

To apply the previous approximation theorems (\cref{thm:Main_1} and \cref{thm:Main_2}) to the first term on the RHS of \eqref{eqn:risk_upper_bound_1},
we rewrite the expectation as
\begin{align}\label{eq:extend_condi_to_uncondi}
&~\E_{x_{t},y,\tau}\left[\norm{s(x_{t},\tau y,t)-\nabla\log{p_{t}(x_{t}|\tau y)}}_{2}^{2}\right]
\\ 
= &~
\frac{1}{2}\int_{\R^{d_{x}}}\norm{s(x,\emptyset,t)-\nabla\log{p_{t}(x|y)}}_{2}^{2}p_{t}(x)\dd x + 
\frac{1}{2}\E_{y}\left[\int_{\R^{d_{x}}}\norm{s(x,y,t)-\nabla\log{p_{t}(x|y)}}_{2}^{2}p_{t}(x|y)\dd x\right].
\nonumber
\end{align}
Since the marginal distribution $p_{t}(x)$ also satisfies the subgaussian property,
the previous result of the conditional score estimation applies to its unconditional counterpart by removing the label throughout the whole process.

\begin{itemize}

\item 
\textbf{Step C.1: Result under \cref{assumption:conditional_density_function_assumption_1}.}

From \cref{thm:Main_1},
we rewrite \eqref{eqn:final_risk} as
\begin{align*}
\E_{\{z_{i}\}_{i=1}^{n}}[{\calR_m}(\hat{s})]
\lesssim
\underbrace{\calO\left(N^{-\frac{\beta}{d_{x}+d_{y}}}(\log{N})^{d_{x}+\frac{\beta}{2}+1}\right)}_{\textbf{(i)}} + \underbrace{\calO\left(N^{-\frac{2\beta}{d_{x}+d_{y}}}\right)}_{\textbf{(ii)}}+\underbrace{\calO\left(\frac{\kappa}{n}\log{\calN}\right)}_{\textbf{(iii)}} + \underbrace{\calO\left(\epsilon_{c}\right)}_{\textbf{(iv)}}.
\end{align*}

Moreover,
from \cref{lemma:error_trunc_risk} we have $\kappa=\calO(1/t_{0})$
and 
from \cref{lemma:covering_number_truncated_loss} we have
\begin{align*}
\log\mathcal{N}\left(\epsilon_c,\calS(R_{\calT}),\norm{\cdot}_\infty\right)
\lesssim
&~
\frac{\log{n}}{\epsilon_{c}^{2}}N^{\frac{172\beta}{d_{x}+d_{y}}+104C_{\sigma}}(\log{N})^{12d_x+12\beta+2}(R_{\calT})^{2}\\
\coloneqq&~
\frac{\log{n}}{\epsilon_{c}^{2}}N^{\nu_{1}}(\log{N})^{\nu_{2}}(R_{\calT})^{2},
\end{align*}
where \blue{$\nu_{1}=68\beta/(d_x+d_y)+104C_{\sigma}$
and
$\nu_{2}=12d_x+12\beta+2$.}

Taking $N=n^{\frac{1}{\nu_{1}}\frac{d_{x}+d_{y}}{(d_{x}+d_{y}+\beta)}}$ and $\epsilon_{c}=N^{-\frac{1}{4}\frac{\nu_{1}\beta}{(d_{x}+d_{y})}}$ 
renders error
\begin{itemize}
    \item 
    \textbf{(i)} $=\calO\left(\frac{1}{t_{0}}(\log{n})^{d_{x}+\frac{\beta}{2}+1}n^{-\frac{\beta}{\nu_{1}(d_{x}+d_{y}+\beta)}}\right)$ from \eqref{eq:extend_condi_to_uncondi} and \cref{thm:Main_1}

     \item 
     \textbf{(ii)} 
     $=\calO\left(n^{-\frac{2\beta}{\nu_{1}(d_{x}+d_{y}+\beta)}}\right)$
     
     \item 
     \textbf{(iii)} $=\calO\left(\kappa n^{-1}n^{\frac{1}{2}\frac{\beta}{d_{x}+d_{y}+\beta}}(\log{n})n^{\frac{d_{x}+d_{y}}{d_{x}+d_{y}+\beta}}(\log{n})^{\nu_{2}}(\log{n})\right)$

     Rearranging the expression,
     we have
     \textbf{(iii)} $=\calO\left(\frac{1}{t_{0}}n^{-\frac{1}{2}\frac{\beta}{d_{x}+d_{y}+\beta}}(\log{n})^{\nu_{2}+2}\right)$

    \item \textbf{(iv)}
    $=\calO\left(n^{-\frac{1}{4}\frac{\beta}{d_{x}+d_{y}+\beta}}\right)$

The total error is bounded by
\begin{align*}
\E_{\{x_{i},y_{i}\}_{i=1}^{n}}\left[\calR(\hat{s})\right]=
\calO\left(\frac{1}{t_{0}}n^{-\frac{\beta}{\nu_{1}(d_{x}+d_{y}+\beta)}}(\log{n})^{\nu_{2}+2}\right).
\end{align*}
\end{itemize}

\textbf{Step C.2: Result under \cref{assumption:conditional_density_function_assumption_2}.}

With \cref{thm:Main_2},
we further write \eqref{eqn:final_risk} as
\begin{align*}
\E_{\{z_{i}\}_{i=1}^{n}}[{\calR_m}(\hat{s})]
\lesssim
\underbrace{\calO\left(N^{-\frac{2\beta}{d_{x}+d_{y}}}(\log{N})^{\beta+1}\right)}_{\textbf{(i)}} + \underbrace{\calO\left(N^{-\frac{2\beta}{d_{x}+d_{y}}}\right)}_{\textbf{(ii)}}
+
\underbrace{\calO\left(\frac{\kappa}{n}\log{\calN}\right)}_{(\textbf{(iii)}} 
+ \underbrace{\calO\left(\epsilon_{c}\right)}_{\textbf{(iv)}}.
\end{align*}

Moreover,
from \cref{lemma:error_trunc_risk} we have $\kappa=\calO(\log{\frac{1}{t_{0}}})$ 
and from \cref{lemma:covering_number_truncated_loss} 
\begin{align*}
\log\mathcal{N}\left(\epsilon_c,\calS(R_{\calT}),\norm{\cdot}_\infty\right)
=
\frac{\log{n}}{\epsilon_{c}^{2}}N^{\nu_{3}}(\log{N})^{10}(R_{\calT})^{2}.
\end{align*}
where  \blue{$\nu_{3}={\frac{4(12\beta d_{x}+31\beta d+ 6\beta)}{d(d_x+d_y)} + \frac{12(12C_{\alpha}d_x + 25C_{\alpha}\cdot d + 6C_{\alpha})}{d}}
+72C_{\sigma}$.}

Taking $N=n^{\frac{(d_{x}+d_{y})}{\nu_{3}(d_{x}+d_{y}+2\beta)}}$ and $\epsilon_{c}=N^{-\frac{1}{4}\frac{\nu_{3}\beta}{(d_{x}+d_{y})}}$ 
renders error
\begin{itemize}
    \item \textbf{(i)} $=\calO\left(\log{\frac{1}{t_{0}}}(\log{n})^{\beta+1}n^{-\frac{1}{\nu_{3}}\frac{2\beta}{(d_{x}+d_{y}+2\beta)}}\right)$ from \eqref{eq:extend_condi_to_uncondi} and \cref{thm:Main_1}

     \item \textbf{(ii)} $=\calO\left(n^{-\frac{2\beta}{\nu_{3}(d_{x}+d_{y}+2\beta}})\right)$
     
     \item \textbf{(iii)} $=\calO\left(\frac{\kappa}{n}n^{\frac{1}{2}\frac{\beta}{d_{x}+d_{y}+2\beta}}(\log{n})n^{\frac{d_{x}+d_{y}}{d_{x}+d_{y}+2\beta}}(\log{n})^{10}(\log{n})\right)$
    
    Rearranging the expression we have
    \textbf{(iii)} $=\calO\left(\log{\frac{1}{t_{0}}}n^{-\frac{3}{2}\frac{\beta}{d_{x}+d_{y}+2\beta}}(\log{n})^{12}\right)$ 

    \item
    \textbf{(iv)} $=\calO\left(n^{-\frac{1}{4}\frac{\beta}{d_{x}+d_{y}+2\beta}}\right)$
\end{itemize}
The total error is bounded by
\begin{align*}
\E_{\{x_{i},y_{i}\}_{i=1}^{n}}\left[\calR(\hat{s})\right]=
\calO\left(\log{\frac{1}{t_{0}}}n^{-\frac{1}{\nu_{3}}\frac{\beta}{d_{x}+d_{y}+2\beta}}(\log{n})^{{\max(12,\beta+1)}}\right).
\end{align*}

\end{itemize}
\end{itemize}
This completes the proof.
\end{proof}

\clearpage

\subsection{Dominance Transition between \texorpdfstring{$N$}{} and \texorpdfstring{$\log N$}{} for All Norm Bounds under \texorpdfstring{\cref{assumption:conditional_density_function_assumption_1}}{}}
\label{sec:low_dim_result}

Here we show that there is a sharp transition between the dominance of $N$ and $\log N$ in all norm bounds for using transformers to approximate score function under \cref{assumption:conditional_density_function_assumption_1} (in \cref{thm:Main_1}). 

We remark that this sharp transition necessitates separate analyses for the low-dimensional region  ($d_x\ll n$) in \cref{cor:low_dim_score_est,cor:low_dim_distribution_est}.

\begin{lemma}[Dominance Transition between $N$ and $\log N$ for All Norm Bounds]\label{lem:N_logN_transition}
Let $d_x$ be the feature dimension of the data. 
Let $N$ be the discretization resolution of the locally diffused polynomial defined in \cref{clipping_integral} and \cref{remark:N_resolution}. 
Under \cref{assumption:conditional_density_function_assumption_1}, $d_x = \Theta\left(\frac{\log N}{\log \log N}\right)$ divides the dependence of $N$ and $\log N$ into two regions for the required norm bounds on attention weights $W_K, W_Q, W_O, W_1, W_2$ in score approximation using transformer networks (\cref{thm:Main_1}):
\begin{itemize}
    \item \textbf{High-Dimensional Region}: If $d_x = \Omega\left(\frac{\log N}{\log \log N}\right)$, $N$ dominates over $\log N$.
    \item \textbf{Mild and Low-Dimensional Region}: If $d_x = o\left(\frac{\log N}{\log \log N}\right)$, $\log N$ dominates over $N$.
\end{itemize}
\end{lemma}

\begin{proof}[Proof of \cref{lem:N_logN_transition}]
Recall the required parameter norm bounds for approximating score function with transformer networks from \textbf{Step C} of \cref{lemma:Score_Approx_Trans}.
We provide a comprehensive summary of all parameter bounds involving terms dependent on $N$ and $\log{N}$ from each respective operation.

\begin{itemize}
\item \textbf{Bound on $W_Q$ and $W_K$.}
\begin{itemize}
\item 
\textbf{For $\epsilon_{f_1}$:}
    \begin{align*}
    & ~ \norm{W_{Q}}_{2},\norm{W_{K}}_{2}, \norm{W_{Q}}_{2,\infty},\norm{W_{K}}_{2,\infty}
    =
    \calO\left(N^{(9\beta+6C_{\sigma})\cdot\frac{2dL+4d+1}{d}}\cdot(\log{N})^{-3(d_x+\beta)\cdot\frac{2dL+4d+1}{d}}\right).
    \end{align*}
Since $d_x=dL$, $N$ and $\log N$ balance at 
\begin{align*}
N^{\calO(d_x)}=(\log{N})^{\calO(d_x^2)},
\end{align*}
and hence
\begin{align*}
d_x =\calO\left( \frac{\log N}{\log \log N}\right).
\end{align*}

\item 
\textbf{For $\epsilon_{f_2}$:}
    \begin{align*}
    & ~ \norm{W_{Q}}_{2},\norm{W_{K}}_{2}, \norm{W_{Q}}_{2,\infty},\norm{W_{K}}_{2,\infty}
    =
    \calO\left(N^{(3\beta+2C_{\sigma})\cdot\frac{2dL+4d+1}{d}}\cdot(\log{N})^{-(d_x+\beta)\cdot\frac{2dL+4d+1}{d}}\right).
    \end{align*}

Since $d_x=dL$, $N$ and $\log N$ balance at 
\begin{align*}
N^{\calO(d_x)}=(\log{N})^{\calO(d_x)},
\end{align*}
and hence
\begin{align*}
d_x =\calO\left( \frac{\log N}{\log \log N}\right).
\end{align*}

\item \textbf{For $\epsilon_{\text{rec},1}$ and $\epsilon_{\text{rec},2}$:}
\begin{align*}
& ~ \norm{W_{Q}}_{2},\norm{W_{K}}_{2}
,\norm{W_{Q}}_{2,\infty},
\norm{W_{K}}_{2,\infty}
=
\calO\left(N^{(9\beta+6C_{\sigma})}(\log{N})^{-3(d_x+\beta)}\right).
\end{align*}
$N$ and $\log N$ balance at 
\begin{align*}
N^{\calO(1)}= (\log{N})^{\calO({d_x})},
\end{align*}
and hence 
\begin{align*}
d_x = \calO\left(\frac{\log N}{\log \log N}\right).
\end{align*}

\item 
\textbf{For $\epsilon_{\sigma,1}$:}
\begin{align*}
\norm{W_{Q}}_{2},\norm{W_{K}}_{2},\norm{W_{Q}}_{2,\infty},\norm{W_{Q}}_{2,\infty}
=
\calO\left(N^{(27\beta+18C_{\sigma})}(\log{N})^{-9(d_x+\beta)}\right).
\end{align*}
$N$ and $\log N$ balance at 
\begin{align*}
N^{\calO(1)}= (\log{N})^{\calO(d_x)},
\end{align*}
and hence 
\begin{align*}
d_x = \calO\left(\frac{\log N}{\log \log N}\right).
\end{align*}

\item \textbf{For $\epsilon_{\sigma,3}$:}
\begin{align*}
    \norm{W_{Q}}_{2},\norm{W_{K}}_{2},\norm{W_{Q}}_{2,\infty},\norm{W_{Q}}_{2,\infty}
    =
    \calO\left(N^{(21\beta+15C_{\sigma}}(\log{N})^{-6(d_x+\beta)}\right).
    \end{align*}
$N$ and $\log N$ balance at 
\begin{align*}
N^{\calO(1)}=(\log{N})^{\calO(d_x)},
\end{align*}
and hence 
\begin{align*}
d_x = \calO\left(\frac{\log N}{\log \log N}\right).
\end{align*}

\end{itemize}

\item 
\textbf{Bound on $W_O$.}

\begin{itemize}
\item 
\textbf{For $\epsilon_{f_1}$}
\begin{align*}
    & ~ 
    \norm{W_O}_{2},    \norm{W_O}_{2,\infty}
    =\calO\left(N^{-\frac{(9\beta+6C_{\sigma})}{d}}(\log{N})^{\frac{3(d_x+\beta)}{d}}\right).
\end{align*}

$N$ and $\log N$ balance at 
\begin{align*}
N^{\calO(1)}=(\log{N})^{\calO(d_x)},
\end{align*}
and hence 
\begin{align*}
d_x = \calO\left(\frac{\log N}{\log \log N}\right).
\end{align*}

\item 
\textbf{For $\epsilon_{f_2}$}
\begin{align*}
    & ~ 
    \norm{W_O}_{2},    \norm{W_O}_{2,\infty}
    =\calO\left(N^{-\frac{(3\beta+2C_{\sigma})}{d}}(\log{N})^{\frac{(d_x+\beta)}{d}}\right).
\end{align*}

$N$ and $\log N$ balance at 
\begin{align*}
N^{\calO(1)}=(\log{N})^{\calO(d_x)},
\end{align*}
and hence 
\begin{align*}
d_x = \calO\left(\frac{\log N}{\log \log N}\right).
\end{align*}

\item 
\textbf{For $\epsilon_{\text{rec},1}$ and $\epsilon_{\text{rec},2}$:}
\begin{align*}
& ~ 
\norm{W_O}_{2}, \norm{W_O}_{2,\infty}
=\calO\left(N^{-(3\beta+6C_{\sigma})}(\log{N})^{d_x+\beta}\right).
\end{align*}
$N$ and $\log N$ balance at 
\begin{align*}
N^{\calO(1)}=(\log{N})^{\calO(d_x)},
\end{align*}
and hence 
\begin{align*}
d_x = \calO\left(\frac{\log N}{\log \log N}\right).
\end{align*}

\item 
\textbf{For $\epsilon_{\sigma_1}$:}
   \begin{align*}
    & ~ 
    \norm{W_O}_{2}, \norm{W_O}_{2,\infty}
    =\calO\left(dN^{-(9\beta+6C_{\sigma})}(\log{N})^{3(d_x+\beta)}\right).
    \end{align*}
    $N$ and $\log N$ balance at 
    \begin{align*}
    N^{\calO(1)}=(\log{N})^{\calO(d_x)},
    \end{align*}
    and hence 
    \begin{align*}
    d_x = \calO\left(\frac{\log N}{\log \log N}\right).
    \end{align*}

\item 
\textbf{For $\epsilon_{\sigma_2}$}:
    \begin{align*}
    & ~ 
    \norm{W_O}_{2}, \norm{W_O}_{2,\infty}
    =\calO\left(dN^{-(7\beta+5C_{\sigma})}(\log{N})^{2(d_x+\beta)}\right).
     \end{align*}
    $N$ and $\log N$ balance at 
    \begin{align*}
    N^{\calO(1)}=(\log{N})^{\calO(d_x)},
    \end{align*}
    and hence 
    \begin{align*}
    d_x = \calO\left(\frac{\log N}{\log \log N}\right).
    \end{align*}
\end{itemize}

\item \textbf{Bound on $W_1$.}
\begin{itemize}
\item 
\textbf{For $\epsilon_{f_1}$:}
    \begin{align*}
    & ~ \norm{W_1}_{2}, \norm{W_{1}}_{2,\infty}
    =\calO\left(N^{\frac{(9\beta+6C_{\sigma})}{d}}(\log{N})^{-\frac{3(d_x+\beta)}{d}}\cdot(\log{N})\right).
    \end{align*}
    $N$ and $\log N$ balance at 
    \begin{align*}
    N^{o(1)}=(\log{N})^{\calO(d_x)},
    \end{align*}    
    and hence 
    \begin{align*}
    d_x = \calO\left(\frac{\log N}{\log \log N}\right).
    \end{align*}

\item 
\textbf{For $\epsilon_{f_2}$:}
 \begin{align*}
    & ~ 
    \norm{W_O}_{2},    \norm{W_O}_{2,\infty}
    =\calO\left(N^{-\frac{(3\beta+2C_{\sigma})}{d}}(\log{N})^{\frac{(d_x+\beta)}{d}}\right).
\end{align*}
    $N$ and $\log N$ balance at 
    \begin{align*}
    N^{\calO(1)}=(\log{N})^{\calO(d_x)},
    \end{align*}    
    and hence 
    \begin{align*}
    d_x = \calO\left(\frac{\log N}{\log \log N}\right).
    \end{align*}
    
\item 
\textbf{For $\epsilon_{\text{mult},1}$:}
\begin{align*}
    & ~ \norm{W_1}_{2},\norm{W_{1}}_{2,\infty}
    =
\calO\left(N^{(4\beta+C_{\sigma})}(\log{N})^{-\frac{1}{2}(d_x+\beta)}\right).
    \end{align*}
    $N$ and $\log N$ balance at 
    \begin{align*}
    N^{\calO(1)}=(\log{N})^{\calO(d_x)},
    \end{align*}
    and hence 
    \begin{align*}
    d_x = \calO\left(\frac{\log N}{\log \log N}\right).
    \end{align*}

\item 
\textbf{For $\epsilon_{\text{rec},1}$, $\epsilon_{\text{rec},2}$:}
  \begin{align*}
    & ~ \norm{W_1}_{2}, \norm{W_{1}}_{2,\infty}
    =
    \calO\left(N^{(6\beta+4C_{\sigma})}(\log{N})^{-2(d_x+\beta)}\right).
    \end{align*}
    $N$ and $\log N$ balance at 
    \begin{align*}
    N^{\calO(1)}=(\log{N})^{\calO(d_x)},
    \end{align*}
    and hence 
    \begin{align*}
    d_x = \calO\left(\frac{\log N}{\log \log N}\right).
    \end{align*}

\item 
\textbf{For $\epsilon_{\sigma_1}$:}
\begin{align*}
    & ~ \norm{W_1}_{2}, \norm{W_{1}}_{2,\infty}
    =
    \calO\left(N^{(9\beta+6C_{\sigma})}(\log{N})^{-3(d_x+\beta)}\cdot\log{N}\right).
    \end{align*}
    $N$ and $\log N$ balance at 
    \begin{align*}
    N^{\calO(1)}=(\log{N})^{\calO(d_x)},
    \end{align*}
    and hence 
    \begin{align*}
    d_x = \calO\left(\frac{\log N}{\log \log N}\right).
    \end{align*}

\item 
\textbf{For $\epsilon_{\sigma_2}$:}
\begin{align*}
    & ~ \norm{W_1}_{2}, \norm{W_{1}}_{2,\infty}
    =
    \calO\left(N^{(7\beta+5C_{\sigma})}(\log{N})^{-2(d_x+\beta)}\cdot\log{N}\right).
    \end{align*}
    $N$ and $\log N$ balance at 
    \begin{align*}
    N^{\calO(1)}=(\log{N})^{\calO(d_x)},
    \end{align*}
    and hence 
    \begin{align*}
    d_x = \calO\left(\frac{\log N}{\log \log N}\right).
    \end{align*}
\end{itemize}

\item \textbf{Bound on $W_2$.}
\begin{itemize}
    \item \textbf{For $\epsilon_{f_1}$ and $\epsilon_{f_2}$:}
    \begin{align*}
    & ~ \norm{W_2}_{2}, \norm{W_{2}}_{2,\infty}
    =
    \calO\left(N^{\frac{(9\beta+6C_{\sigma})}{d}}(\log{N})^{-3\frac{(d_x+\beta)}{d}}\right).
    \end{align*}
    $N$ and $\log N$ balance at 
    \begin{align*}
    N^{\calO(1)}=(\log{N})^{\calO(d_x)},
    \end{align*}
    and hence 
    \begin{align*}
    d_x = \calO\left(\frac{\log N}{\log \log N}\right).
    \end{align*}

    \item \textbf{For $\epsilon_{f_2}$ and $\epsilon_{f_2}$:}

    \begin{align*}
    & ~ \norm{W_2}_{2}, \norm{W_{2}}_{2,\infty}
    =
    \calO\left(N^{\frac{(3\beta+2C_{\sigma})}{d}}(\log{N})^{-\frac{(d_x+\beta)}{d}}\right).
    \end{align*}

    $N$ and $\log N$ balance at 
    \begin{align*}
    N^{\calO(1)}=(\log{N})^{\calO(d_x)},
    \end{align*}
    and hence 
    \begin{align*}
    d_x = \calO\left(\frac{\log N}{\log \log N}\right).
    \end{align*}

\item 
\textbf{For $\epsilon_{\text{rec},1}$ and $\epsilon_{\text{rec},2}$:}
    \begin{align*}
    & ~ \norm{W_2}_{2}, \norm{W_2}_{2,\infty}
    =
    \calO\left(N^{(3\beta+2C_{\sigma})}(\log{N})^{-(d_x+\beta)}\right).
    \end{align*}
    $N$ and $\log N$ balance at 
    \begin{align*}
    N^{\calO(1)}=(\log{N})^{\calO(d_x)},
    \end{align*}
    and hence 
    \begin{align*}
    d_x = \calO\left(\frac{\log N}{\log \log N}\right).
    \end{align*}

\textbf{For $\epsilon_{\sigma_1}$:}
\begin{align*}
    & ~ \norm{W_1}_{2}, \norm{W_{1}}_{2,\infty}
    =
    \calO\left(N^{(9\beta+6C_{\sigma})}(\log{N})^{-3(d_x+\beta)}\right).
    \end{align*}
    $N$ and $\log N$ balance at 
    \begin{align*}
    N^{\calO(1)}=(\log{N})^{\calO(d_x)},
    \end{align*}
    and hence 
    \begin{align*}
    d_x = \calO\left(\frac{\log N}{\log \log N}\right).
    \end{align*}

\textbf{For $\epsilon_{\sigma_2}$:}
   \begin{align*}
    & ~ \norm{W_1}_{2}, \norm{W_{1}}_{2,\infty}
    =
    \calO\left(N^{(7\beta+5C_{\sigma})}(\log{N})^{-2(d_x+\beta)}\right).
    \end{align*}
    $N$ and $\log N$ balance at 
    \begin{align*}
    N^{\calO(1)}=(\log{N})^{\calO(d_x)},
    \end{align*}
    and hence 
    \begin{align*}
    d_x = \calO\left(\frac{\log N}{\log \log N}\right).
    \end{align*}
\end{itemize}

\end{itemize}
This completes the proof.
\end{proof}

\subsection{Proof of \texorpdfstring{\cref{cor:low_dim_score_est}}{}}
\label{proof:cor:low_dim_score_est}

By brute force, we know $N=\calO(n^{d_x^{\kappa}})$
with\footnote{The options of $\kappa$ values are from the hindsight. 
One must compute all norm bounds to identify the available values} $\kappa=-2, 1$ under \cref{assumption:conditional_density_function_assumption_1}.
This indicates the positive proportionality between the sample size $n$ and the resolution $N$.

By \cref{lem:N_logN_transition}, we conclude:
\begin{itemize}
    \item High-Dimension: $d_x=\Omega(\frac{\log N}{\log\log{N}})$,
    and $\kappa=1$.

    \item  Mild and Low-Dimensional Region: $d_x=o(\frac{\log N}{\log\log{N}})$ and $\kappa=-2$. 
\end{itemize}

\paragraph{Low-Dimension Approximation Result.}
For $d_x=o\left(\log{N}/(\log\log{N})\right)$, since the dominant term in the norm bounds differs (\cref{lem:N_logN_transition}), we obtain a distinct score approximation result from  \cref{thm:Main_1}:

\begin{theorem}[Conditional Score Approximation under \cref{assumption:conditional_density_function_assumption_1} and $d_x=o\left(\log{N}/(\log\log{N})\right)$]
\label{thm:Main_1_low_dim}
Assume \cref{assumption:conditional_density_function_assumption_1} and $d_x=o\left(\log{N}/(\log\log{N})\right)$. 
For any precision parameter $0 < \epsilon < 1$ and smoothness parameter $\beta > 0$, let $\epsilon \le \calO(N^{-\beta})$ for some $N \in \mathbb{N}$.
For some positive constants $C_{\alpha},C_{\sigma}>0$, for any $y \in [0,1]^{d_{y}}$ and $t \in [N^{-C_{\sigma}}, C_{\alpha} \log N]$, there exists a $\calT_{\text{score}}(x,y,t)\in\calT_R^{h,s,r}$ such that the conditional score approximation satisfies 
\setlength{\abovedisplayskip}{0pt}
\setlength{\belowdisplayskip}{0pt}
\begin{align*}
\int_{\R^{d_{x}}} \| \calT_{\text{score}}(x,y,t) - \nabla \log p_{t}(x|y) \|_{2}^{2} \cdot p_{t}(x|y) \, \dd x = \calO\left( \frac{B^{2}}{\sigma_{t}^{2}} \cdot N^{-\frac{\beta}{d_{x} + d_{y}}} \cdot (\log N)^{d_{x} + \frac{\beta}{2} + 1} \right).
\end{align*}
Notably, for $\epsilon=\calO(N^{-\beta})$, the approximation error has the upper bound 
$\tilde{\calO}( \epsilon^{1/(d_{x} + d_{y})}/\sigma_{t}^{2} )$.

The parameter bounds for the transformer network class are as follows:

\begin{align*}
& ~ \norm{W_{Q}}_{2},\norm{W_{K}}_{2}, \norm{W_{Q}}_{2,\infty},\norm{W_{K}}_{2,\infty}\\
&~=
\calO\left(N^{\frac{9\beta(2d_x+4d+1)}{d(d_x+d_y)} + \frac{6C_{\sigma}(2d_x+4d+1)}{d}}\cdot(\log{N})^{-3(d_x+\beta)\cdot\frac{2dL+4d+1}{d}}\right);\\
&
\norm{W_{V}}_{2}=\calO(\sqrt{d}); \norm{W_{V}}_{2,\infty}=\calO (d); \norm{W_{O}}_{2},\norm{W_{O}}_{2,\infty}=\calO\left(N^{-\frac{\beta}{d_x+d_y}}\right);\\ 
& 
\norm{W_1}_{2}, \norm{W_1}_{2,\infty}
=
\calO\left(N^{\frac{9\beta}{d_x+d_y} + 6C_{\sigma}}(\log{N})^{-2(d_x+\beta)+1}\right);
\\
& ~ 
\norm{W_2}_{2}, \norm{W_2}_{2,\infty}
=
\calO\left(N^{\frac{9\beta}{d_x+d_y} + 6C_{\sigma}}(\log{N})^{-2(d_x+\beta)}\right);\\
&
\norm{E^{\top}}_{2,\infty}=\calO\left(d^{\frac{1}{2}}L^{\frac{3}{2}}\right);
C_\calT=\calO\left(\sqrt{\log{N}}/\sigma_{t}^{2}\right).
\end{align*}
\end{theorem}

\begin{proof}[Proof of \cref{thm:Main_1_low_dim}]
We show the proof by the following two steps.
\begin{itemize}
    \item \textbf{Step A: Upper-Bound Selection.}

For $d_x=o\left(\log{N}/(\log\log{N})\right)$,
$N$ dominates the $\log{N}$ term.
We set the parameter based on the order of $N$ when $N$ and $\log{N}$ coexist.
By \textbf{Step C} in the proof of \cref{lemma:Score_Approx_Trans},
we have:
\begin{itemize}
    \item \textbf{Bound on $W_Q$ and $W_K$.}
    
    We set the parameter to the largest upper bound determined by the approximation error $\epsilon_{f_1}$:
    \begin{align*}
    & ~ \norm{W_{Q}}_{2},\norm{W_{K}}_{2}, \norm{W_{Q}}_{2,\infty},\norm{W_{K}}_{2,\infty}
    =
    \calO\left(N^{(9\beta+6C_{\sigma})\cdot\frac{2dL+4d+1}{d}}\cdot(\log{N})^{-3(d_x+\beta)\cdot\frac{2dL+4d+1}{d}}\right).
    \end{align*}

    \item \textbf{Parameter Bound on $W_{O}$ and $W_{V}$.}

     We set the parameter to the largest upper bound determined by the approximation error $\epsilon_{\text{mult},2}$ and 
     $\epsilon_{\text{rec},3}$:
    \begin{align*}
    & ~ \norm{W_O}_{2},\norm{W_O}_{2,\infty}
    =\calO\left(N^{-\beta}\right).
\end{align*}
Note that only $\epsilon_{f_{1}}$ and $\epsilon_{f_{2}}$ involve the reshape operation.
That is,
approximation other than $f_1$ and $f_2$ has $\norm{W_V}_{2}, \norm{W_V}_{2,\infty}=\calO(1)$.
Therefore,
we take $\calO(\sqrt{d})$ and $\calO(d)$ for $\norm{W_{V}}_{2}$ and $\norm{W_{V}}_{2,\infty}$ by \cref{lemma:trans_para_bound} respectively.

\item \textbf{Parameter Bound on $W_1$.}

 We set the parameter to the largest upper bound determined by the approximation error $\epsilon_{\sigma,1}$ and $\epsilon_{\sigma,2}$.
 That is,
 we take $N^{(9\beta+6C_{\sigma})}$ from the former and we take $(\log{N})^{-2(d_x+\beta)}$ from the latter.
\begin{align*}
    & ~ \norm{W_1}_{2}, \norm{W_{1}}_{2,\infty}
    =
    \calO\left(N^{(9\beta+6C_{\sigma})}(\log{N})^{-2(d_x+\beta)}\cdot\log{N}\right).
\end{align*}

\item \textbf{Parameter Bound on $W_2$.}

Following the same argument for $W_1$,
we have
\begin{align*}
    & ~ \norm{W_2}_{2}, \norm{W_2}_{2,\infty}
    =
    \calO\left(N^{(9\beta+6C_{\sigma})}(\log{N})^{-2(d_x+\beta)}\right).
\end{align*}
\end{itemize}

\item \textbf{Step B: Change of Variables.}

Recalling from the last step in the proof of \cref{thm:Main_1} (in \cref{sec:appendix_proof_main1}),
we replace $N$ with $N^{1/(d_x+d_y)}$ and $C_{\sigma}$ with $(d_x+d_y)C_{\sigma}$ to obtain the final approximation result.
Here we perform the same change of variables.
\end{itemize}
This completes the proof.
\end{proof}

We compute the covering number for the function class of truncated loss $\calS(R_{\calT})$ (defined in \cref{def:trunc_loss}) under \cref{assumption:conditional_density_function_assumption_1} in low-dimensional region $d_x=o\left(\log{N}/(\log\log{N})\right)$ .

\begin{lemma}[Covering Number for $\calS(R_{\calT})$]
\label{lemma:covering_number_truncated_loss_low_dim}
Given $\epsilon_{c}>0$ and consider $\norm{x}_{\infty}\leq R_{\calT}$.
With sample $\{x_{i},y_{i}\}_{i=1}^{n}$,
the $\epsilon_{c}$-covering number for $\calS(R_\calT)$ with respect to $\norm{\cdot}_{L_{\infty}}$ under the network configuration in \cref{thm:Main_1} satisfies
\begin{align*}
\log\mathcal{N}\left(\epsilon_c,\calS(R_{\calT}),\norm{\cdot}_2\right)
\lesssim
\frac{\log{n}}{\epsilon_{c}^{2}}N^{\nu_4}(\log{N})^{\nu_5}(R_{\calT})^{2},
\end{align*}
where $\nu_4=144d\beta(L+2)(d_x+2d+1)/(d_x+d_y)+96dC_{\sigma}(L+2)(d_x+2d+1)-8\beta$ and 
$\nu_5=-16d(d_x+\beta)(L+2)(3d_{x}+6d+2)+2$.
\end{lemma}

\begin{proof}[Proof of \cref{lemma:covering_number_truncated_loss_low_dim}]
The proof closely follows \cref{lemma:covering_number_truncated_loss}.
Applying \cref{lemma:covering_number},
we calculate 
\begin{align*}
    &~ \log\mathcal{N}(\epsilon_c,\calT_R^{{h,s,r}},\norm{\cdot}_{2}) \notag\\
    \leq  &~ 
    \frac{\log n}{\epsilon_c^2}\cdot 
    \alpha^{2}
    \Bigg(
    \underbrace{2\left((C_F)^2 C_{OV}^{2,\infty}\right)^{\frac{2}{3}}}_{\textbf{(I)}} 
    +
    \underbrace{(d^{\frac{2}{3}}\left(C_F^{2,\infty}\right)^{\frac{4}{3}}}_{\textbf{(II)}}
    +
    \underbrace{d^{\frac{2}{3}}\left(2(C_F)^2 C_{OV} C_{KQ}^{2,\infty}\right)^{\frac{2}{3}}}_{\textbf{(III)}}
    \Bigg)^3,
\end{align*}
where \textbf{(III)} dominates \textbf{(I)} and \textbf{(II)}.

Plug in the network configuration from \cref{thm:Main_1_low_dim}:
\begin{align*}
& ~ \norm{W_{Q}}_{2},\norm{W_{K}}_{2}, \norm{W_{Q}}_{2,\infty},\norm{W_{K}}_{2,\infty}\\
&~=
\calO\left(N^{\frac{9\beta(2d_x+4d+1)}{d(d_x+d_y)} + \frac{6C_{\sigma}(2d_x+4d+1)}{d}}\cdot(\log{N})^{-3(d_x+\beta)\cdot\frac{2dL+4d+1}{d}}\right);\\
&
\norm{W_{V}}_{2}=\calO(\sqrt{d}); \norm{W_{V}}_{2,\infty}=\calO (d); \norm{W_{O}}_{2},\norm{W_{O}}_{2,\infty}=\calO\left(N^{-\frac{\beta}{d_x+d_y}}\right);\\ 
& 
\norm{W_1}_{2}, \norm{W_1}_{2,\infty}
=
\calO\left(N^{\frac{9\beta}{d_x+d_y} + 6C_{\sigma}}(\log{N})^{-2(d_x+\beta)+1}\right);
\\
& ~ 
\norm{W_2}_{2}, \norm{W_2}_{2,\infty}
=
\calO\left(N^{\frac{9\beta}{d_x+d_y} + 6C_{\sigma}}(\log{N})^{-2(d_x+\beta)}\right);\\
&
\norm{E^{\top}}_{2,\infty}=\calO\left(d^{\frac{1}{2}}L^{\frac{3}{2}}\right);
C_\calT=\calO\left(\sqrt{\log{N}}/\sigma_{t}^{2}\right).
\end{align*}

Note that $W_{K,Q}=W_{Q}W_{K}^{\top}$,
we take $\norm{W_{Q}}_{2,\infty}\cdot\norm{W_{K}}_{2,\infty}$ as the upper bound for $\norm{W_{KQ}}_{2,\infty}$.
Since $W_{Q}$, $W_{K}$ share identical upper-bound,
we calculate $(C_{K}^{2,\infty})^{4}$ for $(C_{K,Q}^{2,\infty})^{2}$.
Similarly we use $\norm{W_{O}}_{2,\infty}\cdot\norm{W_{V}}_{2,\infty}$ as the upper bound for $\norm{W_{OV}}_{2,\infty}$.
Moreover,
we take $C_{F}=\max\{C_{f_1},C_{f_2}\}$.

\begin{itemize}
    \item 
    \textbf{ Bound on $C_{F}^{4}=(C_{f_2})^4$:}
    \begin{align*}
    (C_{f_2})^4\lesssim
    N^{\frac{36\beta}{d_x+d_y} + 24C_{\sigma}}(\log{N})^{-8(d_x+\beta)}.
    \end{align*}

    \item \textbf{Bound on $(C_{K}^{2,\infty})^{4}$:}
    \begin{align*}
    (C_{K}^{2,\infty})^{4}\lesssim
    N^{\frac{36\beta(2d_x+4d+1)}{d(d_x+d_y)} + \frac{24C_{\sigma}(2d_x+4d+1)}{d}}\cdot(\log{N})^{-12(d_x+\beta)\cdot\frac{2dL+4d+1}{d}}.
    \end{align*}
\end{itemize}

The bound on \textbf{(III)} follows:

\begin{align*}
\textbf{(III)}
\lesssim
&~\left((C_{f_2})^{4}(C_{OV})^{2}(C_{KQ}^{2,\infty})^{2}\right)^{\frac{1}{3}}\\
\lesssim&~
\left(\underbrace{N^{\frac{36\beta(2d_x+5d+1)}{d(d_x+d_y)} + \frac{24C_{\sigma}(2d_x+5d+1)}{d}}(\log{N})^{-\frac{(d_x+\beta)(24dL+56d+12)}{d}}}_{(C_{f_2})^4\cdot(C_{K}^{2,\infty})^4}\cdot \underbrace{N^{-2\beta}}_{(C_{OV})^2}\right)^{\frac{1}{3}}.
\end{align*}

Moreover,
$\alpha \coloneqq (C_F)^2 C_{OV} (1+4C_{KQ})(R_{\calT}+C_E)$,
we have:
\begin{align*}
\alpha^{2}
\lesssim&~
(C_{f_{1}})^{4}(C_{OV})^{2}(C_{KQ})^{2}(R_{\calT}+C_{E})^{2}
\lesssim
\textbf{(III)}^3\cdot R_{\calT}^{2}.
\end{align*}

By the \textbf{Step C} in \cref{lemma:covering_number_truncated_loss},
we extend the log-covering number of transformer to the truncated loss $\calS(R_{\calT})$ with $\norm{x}_\infty\leq R_{\calT}$ by replacing $\epsilon_c$ with $\epsilon_c/\log{N}$.

Combining \textbf{(III)} and $\alpha^{2}$ for network configuration in \cref{thm:Main_2},
we obtain:
\begin{align*}
\log\mathcal{N}\left(\epsilon_c,\calS(R_{\calT}),\norm{\cdot}_2\right)
\lesssim&~
N^{\frac{72\beta(2d_x+5d+1)}{d(d_x+d_y)} + \frac{48C_{\sigma}(2d_x+5d+1)}{d} - 4\beta}
(\log{N})^{-\frac{8(d_x+\beta)(6dL+14d+3)}{d}+2}\cdot(R_{\calT})^2\\
\coloneqq&~
\frac{\log{n}}{\epsilon_{c}^{2}}N^{\nu_4}(\log{N})^{\nu_5}(R_{\calT})^{2},
\end{align*}
where 
$\nu_4=\frac{72\beta(2d_x+5d+1)}{d(d_x+d_y)} + \frac{48C_{\sigma}(2d_x+5d+1)}{d} - 4\beta$ 
and 
$\nu_5=-\frac{8(d_x+\beta)(6dL+14d+3)}{d}+2$.

This completes the proof.
\end{proof}

\begin{proof}[Proof of \cref{cor:low_dim_score_est}]
The proof closely follows the high-dimensional result where $d_x=\Omega(\log{N}/(\log\log{N}))$ in \cref{app:proof_main_risk_bounds}.
The only distinction lies in the  covering number with transformer network (defined in \cref{def:covering_number}),
characterized by $\nu_i$ with $i\in[5]$.
Therefore,
we replace $\nu_1, \nu_2$ in \cref{thm:main_risk_bounds} with $\nu_4$ and $\nu_5$.

Specifically,
for score estimation under \cref{assumption:conditional_density_function_assumption_1},
by taking $N=n^{\frac{1}{\nu_{4}}\cdot \frac{d_{x}+d_{y}}{\beta+d_{x}+d_{y}}}$, $t_0=N^{-C_{\sigma}}<1$ and $T=C_{\alpha}\log{n}$, it holds
\begin{align*}
\E_{\{x_{i},y_{i}\}_{i=1}^{n}}\left[\calR(\hat{s})\right]
=&~
\calO\left(\frac{1}{t_{0}}n^{-\frac{1}{\nu_{4}}\cdot \frac{\beta}{d_{x}+d_{y}+\beta}}(\log{n})^{\nu_{5}+2}\right)\\
=&~\calO\left(\frac{1}{t_{0}}n^{-\frac{1}{\nu_{4}}\cdot \frac{\beta}{d_{x}+d_{y}+\beta}}\right),
\annot{$n$ term surpasses $\log n$ term}
\end{align*}
$\nu_4=\frac{72\beta(2d_x+5d+1)}{d(d_x+d_y)} + \frac{48C_{\sigma}(2d_x+5d+1)}{d} - 4\beta$ 
and 
$\nu_5=-\frac{8(d_x+\beta)(6dL+14d+3)}{d}+2$.

This completes the proof.
\end{proof}

\clearpage

\subsection{Auxiliary Lemmas for \texorpdfstring{\cref{thm:distribution_TV_bound}}{}.}
\label{sec:aux_thm:tv_bound}

We give the following two lemmas serving as the key components in the proof of \cref{thm:distribution_TV_bound}.

\begin{lemma}[Proposition D.1 of \citet{oko2023diffusion}, Lemma D.4 of \citet{fu2024unveil} and also \citet{chen2022sampling}]
\label{lemma:Girsanov}
Consider probability distribution $p_{0}$ and two stochastic processes $h=\{h_{t}\}_{t\in[0,T]}$ and $h^{\prime}=\{h_{t}^{\prime}\}_{t\in[0,T]}$ that satisfy the following SDE respectively
\begin{align*}
&dh_{t} = b(h_{t},t)\dd t + \dd W_{t}\quad h_{0}\sim p_{0}\\
&dh_{t}^{\prime} = b^{\prime}(h_{t}^{\prime},t)\dd t + \dd W_{t}\quad h_{0}^{\prime}\sim p_{0}.
\end{align*}
Plus denote the distribution of the two processes at time $t$ as $p_{t}$ and $p_{t}^{\prime}$.
Then suppose
\begin{align}
\label{eq:replace_novikov}
\int_{x}p_{t}(x)\norm{(b-b^{\prime})(x,t)}\dd x\leq C
\end{align}
holds for any $t\in[0,T]$,
then we have
\begin{align*}
\text{KL}(p_{T}\mid\mid p_{T}^{\prime})=\int_{0}^{T}\frac{1}{2}\int_{x}p_{t}(x)\norm{(b-b^{\prime})(x,t)}\dd x\dd t
\end{align*}
\end{lemma}
The bound for KL divergence stems from Girsanov’s Theorem,
with the extension to the case where the Novikov’s condition is replaced with \eqref{eq:replace_novikov} by \citet{chen2022sampling}.
Moreover,
we need the following lemma to bound to total variation.
\begin{lemma}[Lemma D.5 of \citet{fu2024unveil}]
\label{lemma:tv_early_stop}
Assume \cref{assumption:conditional_density_function_assumption_1} or \cref{assumption:conditional_density_function_assumption_2}.
For any $y\in[0,1]^{d_{y}}$ we have
\begin{align*}
\text{TV}\left(P_{0}(\cdot|y),P_{t_{0}}(\cdot|y)\right)=
\calO\left(\sqrt{t_{0}}\log^{\frac{d_{x}+1}{2}}\left(\frac{1}{t_{0}}\right)\right).
\end{align*}
\end{lemma}

With the above lemmas and discussion,
we begin the proof of \cref{thm:distribution_TV_bound}.

\subsection{Main Proof of \texorpdfstring{\cref{thm:distribution_TV_bound}}{}}
\label{app:proof_distribution_TV_bound}

\begin{proof}[Proof of \cref{thm:distribution_TV_bound}]

Given label $y$,
we let $\hat{P}_{t_{0}}(\cdot|y)$ denote the data distribution with early-stopped time $t_{0}$ generated by the reverse process with the score estimator from transformer network class.

The decomposition of the total variation between the processes driven by the ground truth and the score estimator follows
\begin{align*}
\text{TV}\left(P(\cdot|y),\hat{P}_{t_{0}}(\cdot|y)\right)
\lesssim
\underbrace{\text{TV}\left(P(\cdot|y),P_{t_{0}} (\cdot|y)\right)}_{\textbf{(I)}} +
\underbrace{\text{TV}\left(P_{t_{0}}(\cdot|y),\tilde{P}_{t_{0}} (\cdot|y)\right)}_{\textbf{(II)}} + 
\underbrace{\text{TV}\left(\tilde{P}_{t_{0}}(\cdot|y),\hat{P}_{t_{0}}(\cdot|y)\right)}_{\textbf{(III)}}
\end{align*}

\begin{itemize}
\item 
\textbf{Step A: Derive the Upper Bound}

\begin{itemize}

\item 
\textbf{Step A.1: Bounding $\textbf{(I)}$.}

From \cref{lemma:tv_early_stop} we have  $\text{TV}\left(P(\cdot|y),\tilde{P}_{t_{0}} (\cdot|y)\right)=\calO\left(\sqrt{t_{0}}\log^{\frac{d_{x}+1}{2}}\left(\frac{1}{t_{0}}\right)\right)$.

\item 
\textbf{Step A.2: Bounding \textbf{(II)}.} 

We use the following process that represents the reverse process starting with standard Gaussian.
\begin{align*}
\dd \tilde{X}_{t}^{\leftarrow} = \left[\frac{1}{2}\dd\tilde{X}_{t}^{\leftarrow} + \nabla\log{p_{T-t}(\tilde{X}_{t}^{\leftarrow}|y)}\right]\dd t + \dd\bar{W}_{t}\quad \tilde{X}_{0}^{\leftarrow}\sim N(0,I_{d_{x}}).
\end{align*}
The distribution of $\tilde{X}_{t}^{\leftarrow}$ conditioned on the label $y$ is denoted by $\tilde{P}_{T-t}(\cdot|y)$.

Next,
by Data Processing Inequality and Pinsker’s Inequality \citep[Lemma 2]{canonne2022short}  we have 
\begin{align}
\label{eq:pinsker}
\text{TV}\left(P_{t_{0}}(\cdot|y),\tilde{P}_{t_{0}}(\cdot|y)\right)
\lesssim&\sqrt{\text{KL}(P_{t_{0}}(\cdot|y)\mid\mid\tilde{P}_{t_{0}}(\cdot|y))}
\nonumber\\
\lesssim&
\sqrt{\text{KL}(P_{T}(\cdot|y)\mid\mid N(0,I_{d_{x}}))}
\nonumber\\
\lesssim&
\sqrt{\text{KL}(P(\cdot|y)\mid\mid N(0,I_{d_{x}}))}\exp(-T).
\end{align}

Therefore for \textbf{(II)},
from \eqref{eq:pinsker} we have
\begin{align*}
\text{TV}\left(P_{t_{0}}(\cdot|y),\tilde{P}_{t_{0}} (\cdot|y)\right)\lesssim
&\sqrt{\text{KL}(P(\cdot|y)\mid\mid N(0,I_{d_{x}}))}\exp(-T)\\
\lesssim&\exp(-T)
\end{align*}

\item 
\textbf{Step A.3: Bounding \textbf{(III)}.}

From \eqref{eq:pinsker} and \cref{lemma:Girsanov},
we have
\begin{align*}
\text{TV}\left(\tilde{P}_{t_{0}}(\cdot|y),\hat{P}_{t_{0}}(\cdot|y)\right)\lesssim\sqrt{\int_{t_{0}}^{T}\frac{1}{2}\int_{x}p_{t}(x|y)\norm{\hat{s}(x,y,t)-\nabla\log{p_{t}(x|y)}}^{2}\dd x\dd t}.
\end{align*}
\end{itemize}

\item 
\textbf{Step B: Altogether.}

Combining $\textbf{(I)}$ $\textbf{(II)}$ and $\textbf{(III)}$,
we take the expectation to the total variation with respect to $y$
\begin{align*}
&\E_{y}\left[\text{TV}\left(P(\cdot|y),\hat{P}_{t_{0}}(\cdot|y)\right)\right]\\
\lesssim&
\sqrt{t_{0}}\log^{\frac{d_{x}+1}{2}}\left(\frac{1}{t_{0}}\right)+\exp(-T)+\sqrt{\int_{t_{0}}^{T}\frac{1}{2}\int_{x}p_{t}(x|y)\norm{\hat{s}(x,y,t)-\nabla\log{p_{t}(x|y)}}^{2}\dd x\dd t}
\annot{By Jensen’s inequality}\\
\lesssim&\sqrt{t_{0}}\log^{\frac{d_{x}+1}{2}}\left(\frac{1}{t_{0}}\right) + \exp(-T) + \sqrt{\frac{T}{2}\calR(\hat{s})}.
\end{align*}
Lastly,
take the expectation with respect to the sample $\{x_{i},y_{i}\}_{i=1}^{n}$ and take $T=C_{\alpha}\log{n}$ we have
\begin{align*}
&\E_{\{x_{i},y_{i}\}_{i=1}^{n}}\left[\E_{y}\left[\text{TV}\left(P(\cdot|y),\hat{P}_{t_{0}}(\cdot|y)\right)\right]\right]\\
\lesssim&\sqrt{t_{0}}\log^{\frac{d_{x}+1}{2}}\left(\frac{1}{t_{0}}\right)
+
n^{-C_{\alpha}}+\sqrt{\log{n}}\E_{\{x_{i},y_{i}\}_{i=1}^{n}}\left[\sqrt{\calR(\hat{s})}\right]
\annot{By Jenson's Inequality}\\
\lesssim&
\underbrace{\sqrt{t_{0}}\log^{\frac{d_{x}+1}{2}}\left(\frac{1}{t_{0}}\right)}_{\textbf{(i)}}
+
n^{-C_{\alpha}}
+
\underbrace{\sqrt{\log{n}}\sqrt{\E_{\{x_{i},y_{i}\}_{i=1}^{n}}\left[\calR(\hat{s})\right]}}_{\textbf{(ii)}}
\end{align*}

\begin{itemize}

\item 
\textbf{Step B.1: Result under \cref{assumption:conditional_density_function_assumption_1}.}

We apply \cref{thm:main_risk_bounds} and
setting $C_{\alpha}=\frac{2\beta}{d_{x}+d_{y}+2\beta}$ and 
$t_{0}=n^{-{\beta}/{(d_{x}+d_{y}+\beta)}}$,
we further write the above expression into
\begin{align*}
&~\E_{\{x_{i},y_{i}\}_{i=1}^{n}}\left[\E_{y}\left[\text{TV}\left(P(\cdot|y),\hat{P}_{t_{0}}(\cdot|y)\right)\right]\right]\\
\lesssim&~
\underbrace{n^{-\frac{\beta}{2(d_{x}+d_{y}+\beta)}}(\log{n})^{(\frac{d_{x}+1}{2})}}_{\textbf{(i)}}
+ 
n^{-\frac{2\beta}{d_{x}+d_{y}+2\beta}} 
+ 
\underbrace{(\log{n})^{\frac{1}{2}}\left(\frac{1}{t_{0}}n^{-\frac{\beta}{\nu_{1}(d_{x}+d_{y}+\beta)}}(\log{n})^{\nu_{2}+2}\right)^{\frac{1}{2}}}_{\textbf{(ii)}}
\end{align*}
Therefore,
under \cref{assumption:conditional_density_function_assumption_1} 
we have
\begin{align*}
\E_{\{x_{i},y_{i}\}_{i=1}^{n}}\left[\E_{y}\left[\text{TV}\left(P(\cdot|y),\hat{P}_{t_{0}}(\cdot|y)\right)\right]\right]=
\calO\left(n^{-\frac{\beta}{2(\nu_{1}-1)(d_{x}+d_{y}+\beta)}}(\log{n})^{\frac{\nu_{2}}{2}+\frac{3}{2}}\right)
\end{align*}

\item 
\textbf{Step B.2: Result under \cref{assumption:conditional_density_function_assumption_2}.}

We apply \cref{thm:main_risk_bounds} and set $t_{0}=n^{-\frac{4\beta}{d_{x}+d_{y}+2\beta}-1}$.
Note that we have
\begin{align*}
\sqrt{t_{0}}\left(\log{\frac{1}{t_{0}}}\right)^{\frac{d_{x}+1}{2}}\lesssim n^{-\frac{2\beta}{d_{x}+d_{y}+2\beta}}.
\end{align*}

We further write
\begin{align*}
&~\E_{\{x_{i},y_{i}\}_{i=1}^{n}}\left[\E_{y}\left[\text{TV}\left(P(\cdot|y),\hat{P}_{t_{0}}(\cdot|y)\right)\right]\right]\\
\lesssim&~
\underbrace{n^{-\frac{2\beta}{d_{x}+d_{y}+2\beta}}}_{\textbf{(i)}}
+ 
n^{-\frac{2\beta}{d_{x}+d_{y}+2\beta}} 
+ 
\underbrace{(\log{n})^{\frac{1}{2}}\left(\log{\frac{1}{t_{0}}}n^{-\frac{1}{\nu_{3}}\frac{\beta}{d_{x}+d_{y}+2\beta}}(\log{n})^{{\max(10,\beta+1})}\right)^{\frac{1}{2}}}_{\textbf{(ii)}}.
\end{align*}
Therefore we have
\begin{align*}
\E_{\{x_{i},y_{i}\}_{i=1}^{n}}\left[\E_{y}\left[\text{TV}\left(P(\cdot|y),\hat{P}_{t_{0}}(\cdot|y)\right)\right]\right]=\calO\left(n^{-\frac{1}{2\nu_{3}}\frac{\beta}{d_{x}+d_{y}+2\beta}}(\log{n})^{\max(6,(\beta+3)/2)}\right)
\end{align*}
\end{itemize}
\end{itemize}
This completes the proof.
\end{proof}

\subsection{Proof of \texorpdfstring{\cref{cor:low_dim_distribution_est}}{}}
\label{proof:cor:low_dim_distribution_est}

\begin{proof}[Proof of \cref{cor:low_dim_distribution_est}]
The proof closely follows the high-dimensional result where $d_x=\Omega(\log{N}/(\log\log{N}))$ in \cref{app:proof_main_risk_bounds}.
The only distinction lies in the  covering number with transformer network (defined in \cref{def:covering_number}),
characterized by $\nu_i$ with $i\in[5]$.
Therefore,
we replace $\nu_1, \nu_2$ in \cref{thm:distribution_TV_bound} with $\nu_4$ and $\nu_5$.
This completes the proof.
\end{proof}

\clearpage
\def\arxivfont{\rm}
\bibliographystyle{plainnat}

\bibliography{refs}

\end{document}